\documentclass[10pt, lettersize,journal]{IEEEtran}
\usepackage{amsmath,amsfonts}
\usepackage{algorithmic}
\usepackage{algorithm}
\usepackage{array}
\usepackage[font=scriptsize,labelfont={scriptsize,bf}]{subfig}

\usepackage{textcomp}
\usepackage{stfloats}
\usepackage{url}
\usepackage{hyperref}

\usepackage{verbatim}
\usepackage{graphicx}
\usepackage{cite}
\IEEEoverridecommandlockouts
\usepackage{times}
\usepackage{epsfig}
\usepackage{amssymb}
\usepackage{xcolor}
\begin{document}

%\title{Attention Enhanced Residual Refining Generative Adversarial Network for Automatic Building Detection in Remotely Sensed Images }
\title{Uncertainty, Edge, and Reverse-Attention Guided Generative Adversarial Network for Automatic Building Detection in Remotely Sensed Images}

\author{Somrita Chattopadhyay and Avinash C. Kak \thanks{The authors are with the School of Electrical and Computer Engineering at Purdue University, West Lafayette, Indiana 47907, USA (e-mail:
chattops@purdue.edu; kak@purdue.edu). }%~\IEEEmembership{Staff,~IEEE,
}
        % <-this % stops a space
%\thanks{This paper was produced by the IEEE Publication Technology Group. They are in Piscataway, NJ.}% <-this % stops a space
%\thanks{Manuscript received April 19, 2021; revised August 16, 2021.}}



\maketitle

\begin{abstract}
Despite recent advances in deep-learning based semantic
segmentation, automatic building detection from remotely
sensed imagery is still a challenging problem owing to large
variability in the appearance of buildings across the
globe. The errors occur mostly around the boundaries of the
building footprints, in shadow areas, and when detecting
buildings whose exterior surfaces have reflectivity
properties that are very similar to those of the surrounding
regions. To overcome these problems, we propose a generative
adversarial network based segmentation framework with
\textbf{uncertainty attention unit} and \textbf{refinement
  module} embedded in the generator. The refinement module,
composed of edge and reverse attention units, is designed to
refine the predicted building map. The edge attention
enhances the boundary features to estimate building
boundaries with greater precision, and the reverse attention
allows the network to explore the features missing in the
previously estimated regions. The uncertainty attention unit
assists the network in resolving uncertainties in
classification.  As a measure of the power of our approach,
as of December 4, 2021, it ranks at the second place on
DeepGlobe's public leaderboard despite the fact that main
focus of our approach --- refinement of the building edges
--- does not align exactly with the metrics used for
leaderboard rankings.  Our overall F1-score on DeepGlobe's
challenging dataset is 0.745.  We also report improvements
on the previous-best results for the challenging INRIA
Validation Dataset for which our network achieves an overall
IoU of 81.28\% and an overall accuracy of 97.03\%.  Along
the same lines, for the official INRIA Test Dataset, our
network scores 77.86\% and 96.41\% in overall IoU and
accuracy.  We have also improved upon the previous best
results on two other datasets: For the WHU Building Dataset,
our network achieves 92.27\% IoU, 96.73\% precision, 95.24\%
recall and 95.98\% F1-score. And, finally, for the Massachusetts
Buildings Dataset, our network achieves 96.19\% relaxed IoU
score and 98.03\% relaxed F1-score over the previous best
scores of 91.55\% and 96.78\% respectively, and in terms of
non-relaxed F1 and IoU scores, our network outperforms the
previous best scores by 2.77\% and 3.89\% respectively.
\end{abstract}

\begin{IEEEkeywords}
Semantic Segmentation, Attention, Deep Learning, Generative Adversarial Networks
\end{IEEEkeywords}

\section{Introduction}
\label{sec:intro}

While a great deal of progress has already been made in the
automatic detection of building footprints in aerial and
satellite imagery, several challenges still remain.  Most of
these can be attributed to the high variability in how the
buildings show up in such images in different parts of the
world, by the effect of shadows on the sensed data, and by
the presence of occlusions caused by nearby tall structures
and high vegetation.  Problems are also caused by the fact
that the reflectivity signatures of several types of
building materials are close to those for the materials that
are commonly used for the construction of roads and parking
lots.

With regard to the performance of the deep-learning based
methods for discriminating between the buildings and the
background, the commonly used metrics used for evaluating
the algorithms only ensure that the bulk of the building
footprints is extracted.  The metrics do not enforce the
requirement of contiguity of the pixels that belong to the
same building \cite{MnihThesis,saito2015building,
  chen2017deeplab, khalel2018automatic,
  iglovikov2018ternausnetv2, maggiori2016convolutional}.
This has led some researchers to formulate post-processing
steps like the Conditional Random Fields (CRFs)
\cite{wallach2004conditional, zhu2020building} during
inference for invoking spatial contiguity in the output
label maps.

Even more importantly, the semantic-segmentation metrics for
identifying the buildings are silent about the quality of
the boundaries of the pixel blobs
\cite{MnihThesis,chen2017deeplab, cheng2020panoptic,
  sebastian2020adversarial, pan2019building,
  wang2020automatic}.  Since the number of pixels at the
perimeter of a convex shape is roughly proportional to the
square-root of the pixels in the interior, incorrectly
labeling even a tiny fraction of the overall building pixels
may correspond to an exaggerated effect on the quality of
the boundary.

These problems related to enforcing the spatial contiguity
constraint and to ensuring the quality of the building
boundaries only become worse in the presence of confounding
factors such as shadows, the similarity between the
reflectivity properties of the building exteriors and their
surroundings, etc.

We address these challenges in a new generative adversarial
network (GAN) \cite{goodfellow2014generative} for segmenting
building footprints from high-resolution remotely sensed
images. We adopt an adversarial training strategy to enforce
long-range spatial label contiguity, {\em without adding any
  complexity to the trained model during inference}.  In our
adversarial network, the discriminator is designed to
correctly distinguish between the predicted labels and the
ground-truth labels and is trained by optimizing a
multi-scale L1 loss \cite{xue2018segan}. The generator, an
encoder-decoder framework with embedded \emph{uncertainty
  attention} and \emph{refinement modules}, is trained to
predict one-channel binary maps with pixel-wise labels for
building and non-building classes.

Our network incorporates several novel ideas, such as the
Uncertainty Attention Unit that is introduced at each data
abstraction level between the concatenation of the encoder
feature map with the decoder feature map.  This unit focuses
on those feature regions where the network has not shown
confidence during its previous predictions. That is likely
to happen at the boundaries of the building shapes, in
shadow areas, and in those regions of an image where the
building pixel signatures are too close to the background
pixel signatures.

\begin{figure*}[h]
\centering
\includegraphics[width=\textwidth]{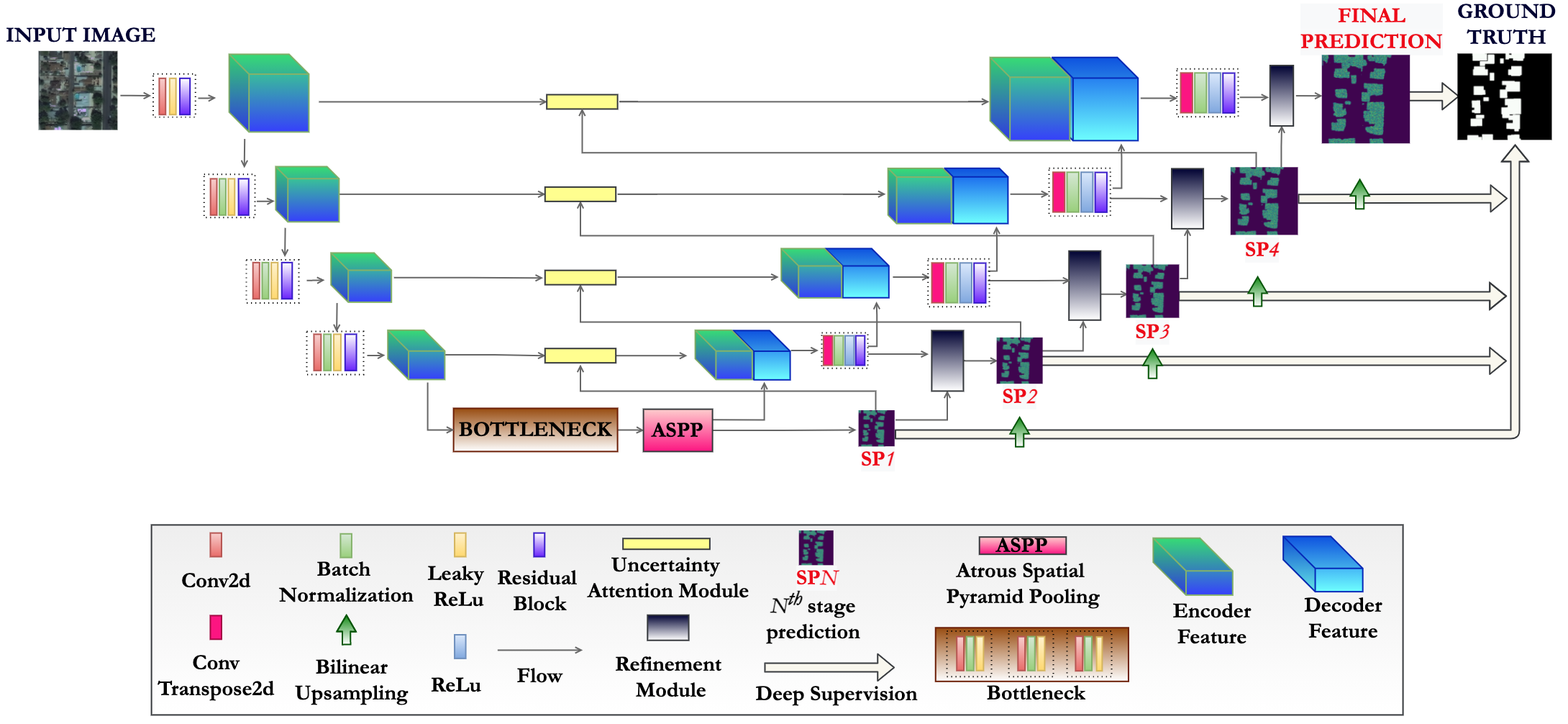}
\caption{\centering Segmentation Framework}
\label{fig:gen}
\end{figure*}

Another novel aspect of our network is the Refinement Module
that consists of a \emph{Reverse Attention Unit} and an
\emph{Edge Attention Unit}.  This module is introduced after
each stage in the decoder to gradually refine the prediction
maps.  Starting with the bottleneck layer of the
encoder-decoder network and using an Atrous Spatial Pyramid
Pooling (ASPP) \cite{chen2017deeplab} layer, the network
first predicts a coarse prediction map that is rich in
semantic information but lacks fine detail
(Figure~\ref{fig:gen}). The coarse prediction map is then
gradually refined by adding residual predictions obtained
from the two attention units in each stage of decoding. The
Edge Attention Unit amplifies the boundary features, and,
thus, helps the network to learn precise boundaries of the
buildings. And the Reverse Attention Unit allows the network
to explore the regions that were previously classified as
non-building, which enables the network to discover the
missing building pixels in the previously estimated
results. 

In addition to the adversarial loss, we also use deep
supervision (shown as thick arrows in Figure~\ref{fig:gen}) in
our architecture for efficient back propagation of the
gradients through the deep network structure. By deep
supervision, we refer to the losses computed for each
intermediate prediction map. These losses are added to the
final layer's loss. Deep supervision guides the intermediate
prediction maps to become more directly predictive of the
final labels.  We compute weighted dice loss and shape loss
for the final prediction map as well as for each
intermediate prediction map.

The power of our approach is best illustrated by its ranking
at number 2 in the ``DeepGlobe Building Extraction
Challenge'' at the following website:\footnote{Our entry is
  under the username `chattops' with the upload date November
  30, 2021.  As mentioned earlier in the Introduction, the metrics
  used in all such competitions only measure the extent of
  the bulk extraction of the pixels corresponding to the
  building footprints. {\em In other words, these metrics
    do not directly address the main focus of our paper,
    which is on improving the boundaries of the extracted
    shapes and the contiguity of the pixel blobs that are
    recognized as the building pixels.}  Nonetheless, it is
  noteworthy that improving the boundary and the pixel
  contiguity properties also improves the traditional
  metrics for building segmentation.}

\begin{center}
  \url{https://competitions.codalab.org/competitions/18544#results}
\end{center}

\begin{figure}
    \centering
    \subfloat[Input Image]{{\includegraphics[width=4cm]{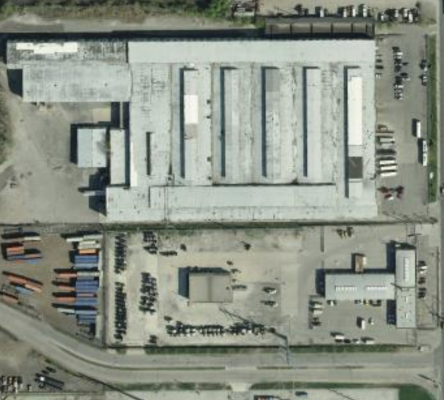} }}
    %\quad
    \subfloat[GAN-SCA \cite{pan2019building}]{{\includegraphics[width=4cm]{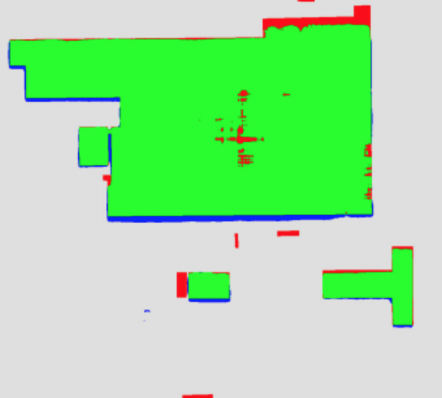} }}
    \quad
    \subfloat[\centering Our baseline network with no attention units]{{\includegraphics[width=3.9cm]{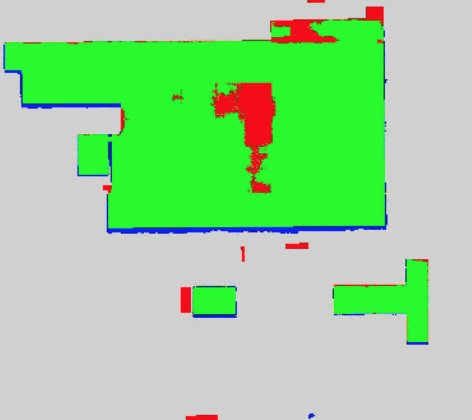} }}
    \subfloat[ \centering Our network with attention units]{{\includegraphics[width=4cm]{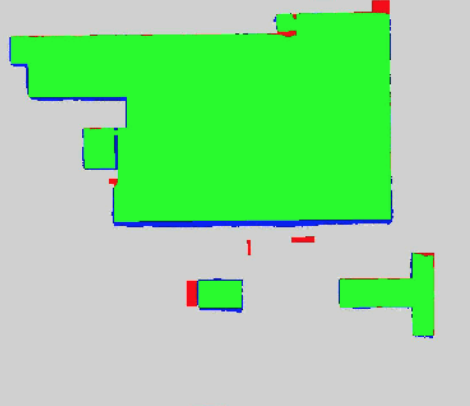} }}
    \caption{Comparing segmentation results using our approach and another state-of-the-art approach (GAN-SCA) on an image patch over Chicago from the INRIA Dataset.  Green: True positives ; Blue: False Positives; Red: False negatives, Grey: True negatives.}
    \label{fig:compare}
\end{figure}

In the experimental results that we will report in this
paper, the reader will see significant performance
improvements over the previous-best results for four
different datasets, two of which are known to be challenging
(DeepGlobe and INRIA), and two others that are older but very
well known in semantic segmentation research (WHU and the
Massachusetts Buildings Dataset).  While our performance
numbers presented in the Results section speak for
themselves, the reader may also like to see a visual example
of the improvements in the quality of the building
prediction maps produced by our framework.
Figure~\ref{fig:compare} shows a typical example.
%In addition to generating accurate semantic labels of
%building footprints from aerial imagery, 
Additionally, our results on the INRIA Aerial Image Labeling
Dataset \cite{maggiori2017dataset} demonstrate that
our proposed network can be generalized to detect buildings
in different cities across the world without being directly
trained on each of them.

The rest of the paper is organized as follows. In
Section~\ref{sec:related}, we review current
state-of-the-art building segmentation algorithms and
explain distinctive features of our proposed algorithm in
relation to the past literature. Section~\ref{sec:method}
gives a detailed description of our network architecture and
its various components. We explain our training strategy and
the loss functions used in Section~\ref{sec:training}. In
Section~\ref{sec:data}, we describe the datasets we have
used for our experiments. Subsequently, in
Section~\ref{sec:exp}, we provide detailed description of
our experimental setup. In Section~\ref{sec:results}, we
compare the performance of our approach with other
state-of-the-art methods. We conduct a detailed discussion about our results and present an ablation study involving various components of our
network in Section~\ref{sec:discussionandablation}. Finally, we conclude and summarize the paper in
Section~\ref{sec:conclusion}.

\section{Related Works}
\label{sec:related}

The past decade of research in image segmentation methods
has witnessed the deep learning based approaches
\cite{ronneberger2015u, tao2020hierarchical, yuan2019object,
  cheng2020panoptic, li2020improving, chen2018encoder,
  zhao2017pyramid, he2017mask, lin2017refinenet,
  wu2019fastfcn, fu2019dual,yu2015multi} outperforming the
more traditional approaches
\cite{zhang2006overview,lindeberg1997segmentation,al2010image,
  khokher2013image, muthukrishnan2011edge,
  angelina2012image, kaganami2009region, yambal2013image,
  dehariya2010clustering, beucher1992watershed}. Inspired by
the success of the deep learning based methods, more
recently the researchers have focused on developing neural
network based frameworks for detecting building footprints
from high-resolution remotely sensed images
\cite{lin2019esfnet, marcu2018multi, marcu2017object,
  iglovikov2018ternausnetv2, chen2021dr, shao2020brrnet,
  liao2021joint, ran2021building, liao2020learning,
  ma2020building, liu2020arc, wu2021hrlinknet,
  golovanov2018building, li2018semantic}.

Mnih was the first to use a CNN to carry out patch-based
segmentation in aerial images \cite{MnihThesis}. Saito et
al. in \cite{saito2015building} also used a patch based CNN
for road and building detection from aerial images. However,
the patch-based methods suffer from the problem of limited
receptive fields and large computational overhead, and
require post-processing steps \cite{wallach2004conditional}
to refine the segmentation results. The patch-based
approaches were soon surpassed by pixel-based methods
\cite{maggiori2016convolutional, khalel2018automatic} that
applied state-of-the-art neural network models, like the
hierarchical fully convolutional network (FCN) and stacked
U-Nets, to perform pixel-wise prediction of building
footprints in aerial images. However, these approaches do
not fully utilize the structural and contextual information
of the ground objects that can help to distinguish the
buildings from their heterogeneous backgrounds.

The shortcomings of the current state-of-the-art in deep
learning based methods are being addressed by several
ongoing research efforts \cite{Hamaguchi_2018_CVPR_Workshops, hu2019light, liao2021joint, bischke2019multi, Zhao_2018_CVPR_Workshops, ji2018fully, mou2019relation, zhang2020local}. The work reported in
\cite{Hamaguchi_2018_CVPR_Workshops} addresses the problems
caused by large variations in the building sizes in satellite
imagery.  On the other hand, the works reported in
\cite{hu2019light, bischke2019multi, Zhao_2018_CVPR_Workshops, he2021boundary, zhu2021ed, jung2021boundary} deal with the preservation of
the sharpness of the building boundaries.  There are also the
works reported in \cite{ji2018fully, wang2020automatic} that
attempt to detect buildings even when only a part of a
building is visible.

In order to leverage large-scale contextual information
and extract critical cues for identifying building pixels in
the presence of complex background and when there is
occlusion, researchers have proposed methods to capture
local and long-range spatial dependencies among the ground
entities in the aerial scene \cite{mou2019relation, zhang2020local}. Several researchers are also using tranformers \cite{chen2021self}, attention modules \cite{wang2020automatic, sebastian2020contextual, zhang2020semantic, zhou2020robust} and multi-scale information \cite{ran2021building, ma2020building, zhu2020building, liu2020arc, guo2021scale, wei2019toward, liu2019building} for this purpose. Recently, multi-view satellite images \cite{comandur2021semantic, Weir_2019_ICCV} are also being used to perform semantic segmentation of points on ground. 

GANs \cite{goodfellow2014generative} are also gaining popularity in solving semantic segmentation problems. In
GAN-based approaches to building detection \cite{sebastian2020adversarial, pan2019building, li2018building, abdollahi2020building}, the generator is
basically a segmentation network that aims to produce
building label maps that cannot be distinguished from the
ground-truth ones by the discriminator. By training
the segmentation and the discriminator networks
alternatively, the likelihood associated with the joint
distribution of all the labels that are possible at the
different pixel locations can be maximized as a whole, which
amounts to enforcing long-range spatial dependency among the
labels.  Using this logic, in
\cite{sebastian2020adversarial}, Sebastian et
al. illustrated how the use of adversarial learning can
improve the performance of the existing benchmark semantic
segmentation networks \cite{zhao2017pyramid,
  chen2017deeplab}.  Along roughly the same lines, Li et
al. adopted adversarial training in \cite{li2018building} to
detect buildings in overhead images where the segmentation
network is a fully convolutional DenseNet model and the
discriminator an autoencoder. In
\cite{abdollahi2020building}, the authors used a SegNet
model with a bidirectional convolutional LSTM as the
segmentation network. 

The work presented in this paper comes closest to the approach adopted in
\cite{pan2019building} in which the authors have proposed a
GAN with spatial and channel attention mechanisms to detect
buildings in high-resolution aerial imagery. In this
contribution, the spatial and the channel attention
mechanisms are embedded in the segmentation architecture to
selectively enhance important features on the basis of their
spatial information in the different channels. In contrast with \cite{pan2019building}, our framework focuses the attention units where they are needed the most --- these would be the pixels where the predictions are being made with low probabilities.

Despite the successes of the previous contributions
mentioned in this section, the predicted building label maps
are still found lacking with regard to the overall quality
of building segmentations.  At the pixel level, we still
have misclassifications at a higher rate at those locations
where the classification accuracy is most important --- at
and in the vicinity of the boundaries of the buildings and
where there are shadows and obscurations.  Furthermore, the
methods that have been proposed to date tend to be locale
specific. That is, they do not generalize straightforwardly
to the different geographies around the world without
further training. In this paper, we aim to overcome these
shortcomings with the help of uncertainty and refinement
modules that we embed in the segmentation network of our
adversarial framework. We show empirically that our model
outperforms the state-of-the models on well-known publicly
available datasets \cite{MnihThesis, maggiori2017dataset,  ji2018fully, van2018spacenet, DeepGlobe18}.

\section{Proposed Architecture}
\label{sec:method}
In this section, we describe our proposed attention-enhanced
generative adversarial network for detecting building
footprints in remotely sensed images. The framework is
composed of two parts: an attention-enhanced segmentation
network ($\mathcal{S}$) and a critic network
($\mathcal{C}$). Our segmentation network, attention units
and critic network are described in details in
Sections~\ref{sec:seg}, \ref{sec:attn} and \ref{sec:critic}
respectively.

\subsection{Segmentation Network}
\label{sec:seg}

Our segmentation framework ($\mathcal{S}$), illustrated in
Figure~\ref{fig:gen}, is a fully convolutional encoder-decoder
network that takes in a 3-channel remotely sensed image and generates
a 1-channel prediction map in which each pixel value
indicates that pixel's probability of belonging to the
building class.

$\mathcal{S}$ uses four strided convolutional (Conv) layers
for \emph{encoding} the input images. The kernel size is set
to 7 for the first two layers and 5 for the next two.  The
stride is set to 2 in all the layers.  The output of the
encoder is a feature map at $1/16$-th the spatial resolution
of the input images. The number of channels goes up by a
factor of 2 in each layer.

The feature maps thus produced at the bottleneck layer of
the network are processed by an ASPP module
\cite{chen2017deeplab} to capture the global contextual
information for more accurate pixel-wise predictions. The
ASPP module consists of a $1\times1$ Conv layer, three
$3\times3$ Conv layers with dilation rates of 2, 4, and 6,
and a global context layer incorporating average pooling and
bilinear interpolation.  The resulting feature maps from the
five layers of ASPP are concatenated and passed through
another $3\times3$ Conv layer, where they form the output of
the ASPP module that is fed directly into the decoder. In
addition to that, we pass the feature maps from the ASPP
module through a $1\times1$ Conv layer to produce the
top-most prediction map that is low in resolution but rich
in semantic information.

The \emph{decoder} uses kernels with increasingly larger
receptive fields (7,9 and 11) in order to enlarge the
representational scope of each pixel. Each layer of the
decoder uses a transpose convolution (ConvTranspose2d) to
up-sample the incoming feature map while halving the number
of feature channels.

Residual blocks are added after every downsampling and
upsampling layer. Each residual block consists of a $1\times1$
Conv, followed by a $3\times3$ Conv and then another $1\times1$ Conv. Skip
connections are used in a similar fashion as that of the
U-Net \cite{ronneberger2015u} to concatenate the
corresponding layers of the encoder and the decoder.  As
shown by the yellow boxes in Figure~\ref{fig:gen}, an
\emph{Uncertainty Attention Module} is used for this concatenation at
each abstraction level in network.  This allows the network
to focus on the features in those regions where the network
has not shown confidence in the predictions made at the
lower abstraction level.  Detailed description of this
module is presented in Section~\ref{sec:uncertainty}.

Batch normalization is used after each convolutional layer
except the first layer of the encoder. After each batch
normalization, Leaky ReLU with a leak slope of 0.2 is used
in all downsampling blocks, and a regular ReLU has been used
for all the upsampling layers.

We also apply a \emph{Refinement Module} consisting of a
\emph{Reverse Attention Unit} and an \emph{Edge Attention
  Unit} in each stage of the decoder. This module is used to
learn residual predictions after every stage of decoding and
gradually refine the prediction map estimated in the
previous stage until the final prediction map is
obtained. Details of this module are provided in
Section~\ref{sec:side}.

\subsection{Attention in Segmentation Network}
\label{sec:attn}
\subsubsection{Refinement Module}
\label{sec:side}

\begin{figure}[ht] 
  \label{ fig7} 
  \begin{minipage}[b]{\linewidth}
    \centering
    \includegraphics[width=0.9\linewidth]{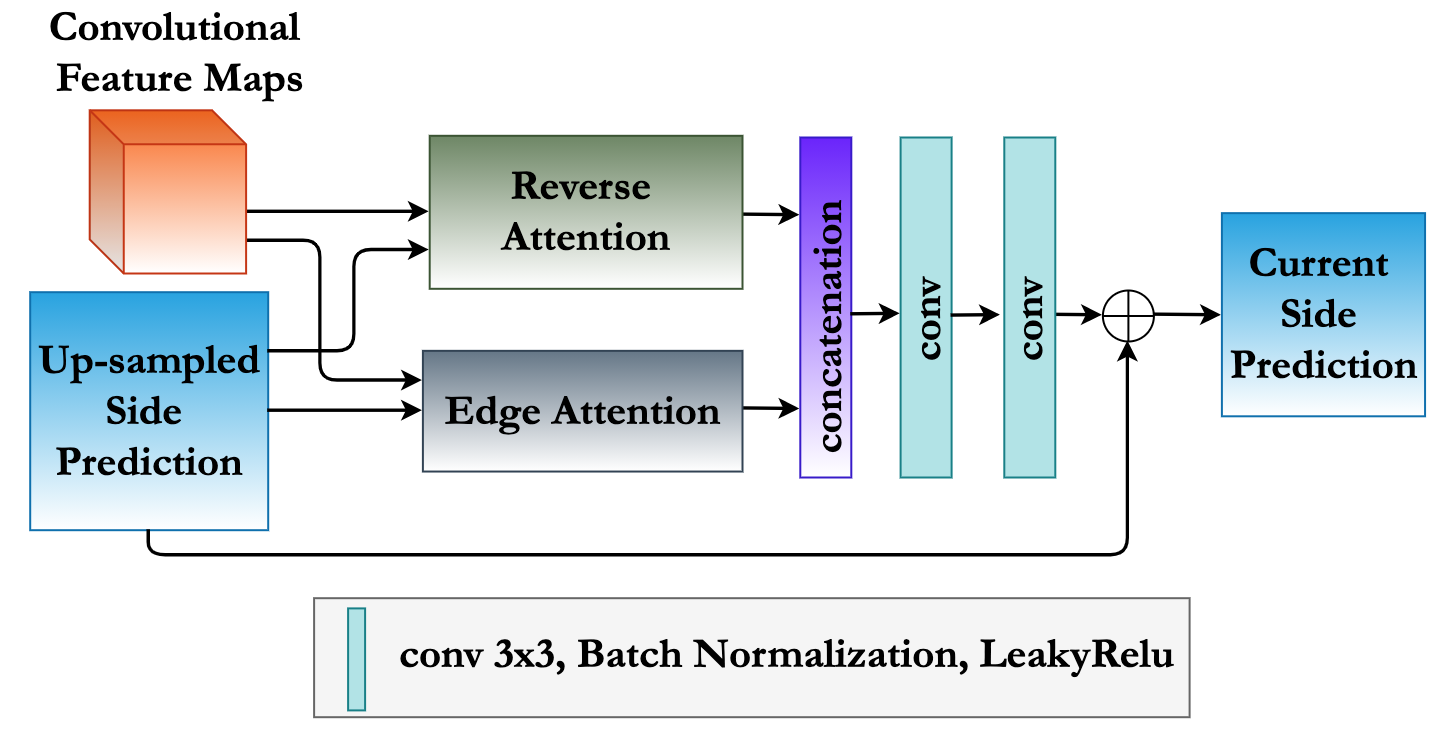}
    \caption{\centering Refinement Module (RM)}% $\oplus$ denotes element-wise addition.}
    \label{fig:side} 
  \end{minipage}
  \\
  \\
  \begin{minipage}[b]{\linewidth}
    \centering
    \includegraphics[width=0.9\linewidth]{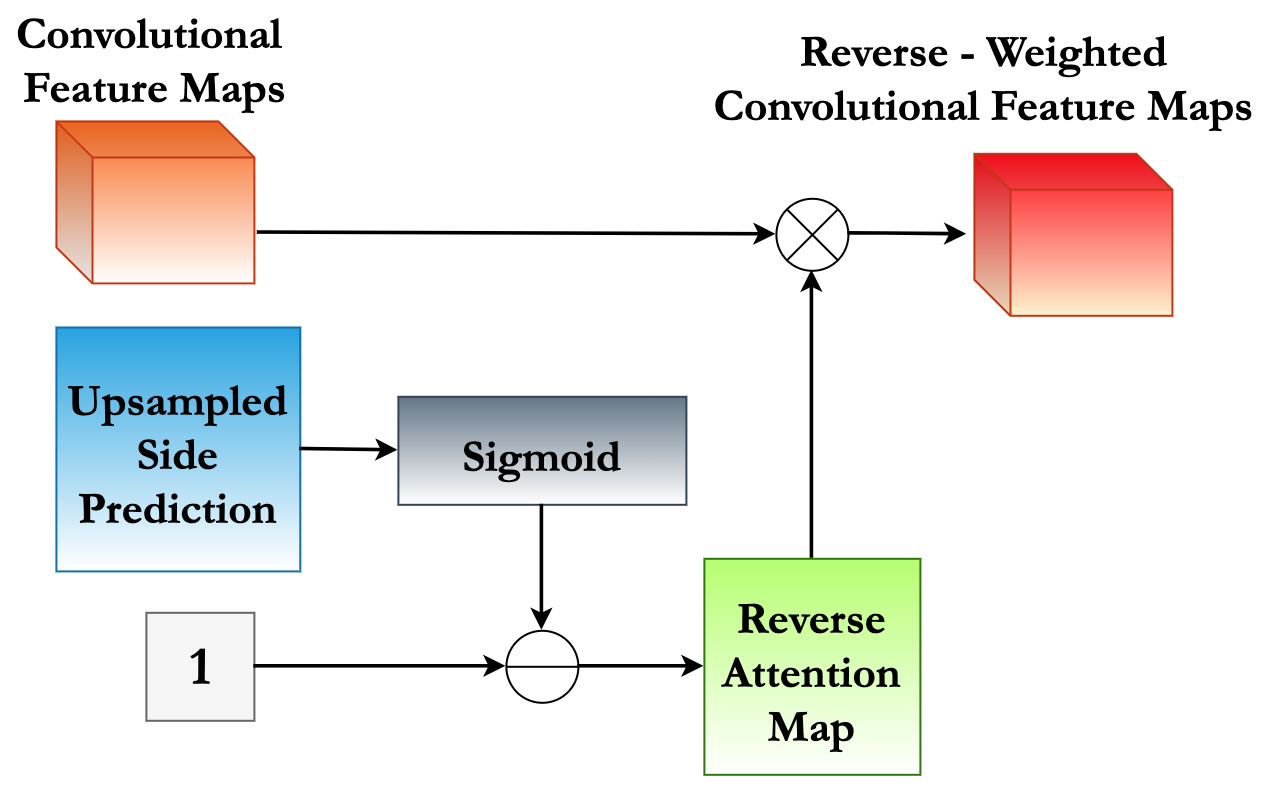} 
    \caption{\centering Reverse Attention Unit (RAU)} %$\otimes$ and $\ominus$ denote element-wise multiplication and subtraction respectively.}
    \label{fig:reverse} 
  \end{minipage} 
  \\
  \\
  \begin{minipage}[b]{\linewidth}
    \centering
    \includegraphics[width=0.9\linewidth]{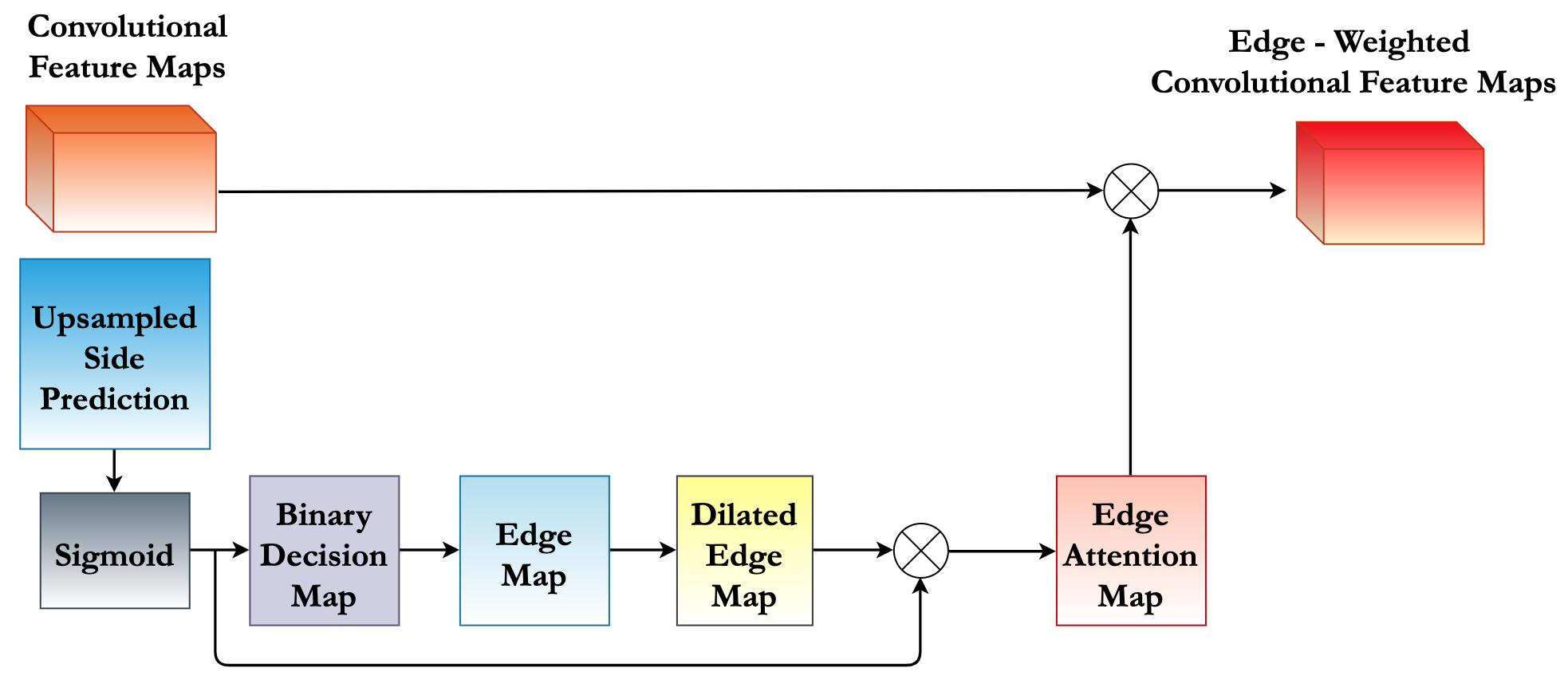} 
    \caption{\centering Edge Attention Unit (EAU)}
    \label{fig:edge}
  \end{minipage}
\end{figure}

In general, given a deep network for image segmentation, the
high-level feature maps extracted in layers closer to the
final output will contain accurate localization information
about the objects in the image, but will be lacking in fine
detail regarding those objects.  On the other hand, the
layers closer to the input will be rich in fine detail but
with unreliable estimates of where exactly the object is
located.  The purpose of the Refinement Module is to fuse
the fine detail from the lower-indexed layers with the
spatial features in the higher-indexed layers with the
expectation that such a fusion would lead to a segmentation
mask that is rich in fine details and that, at the same
time, exhibits high accuracy with regard to object
localization.

\begin{figure*}
    \centering
    \subfloat{{\includegraphics[width=2.4cm]{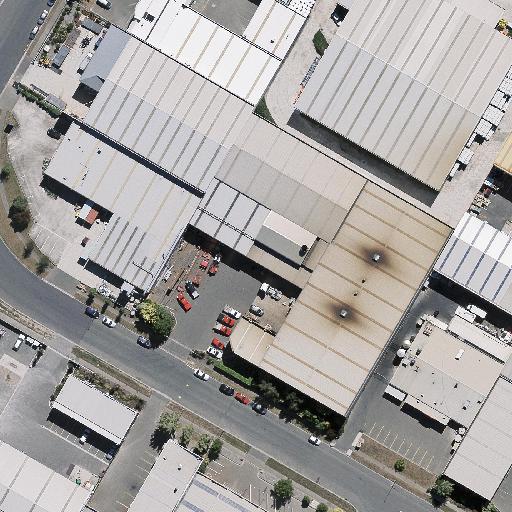} }}
    \subfloat{{\includegraphics[width=2.4cm]{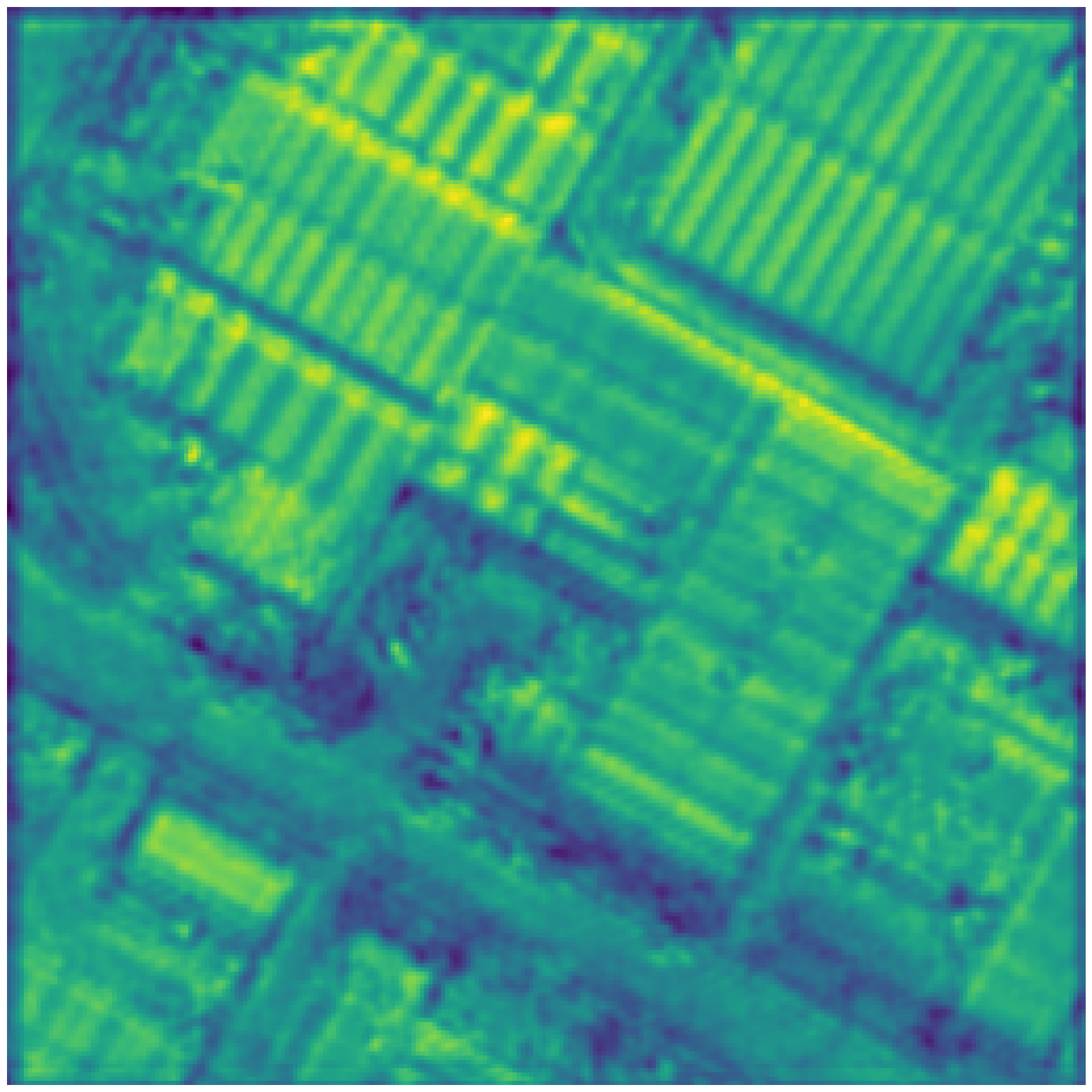} }}
    \subfloat{{\includegraphics[width=2.4cm]{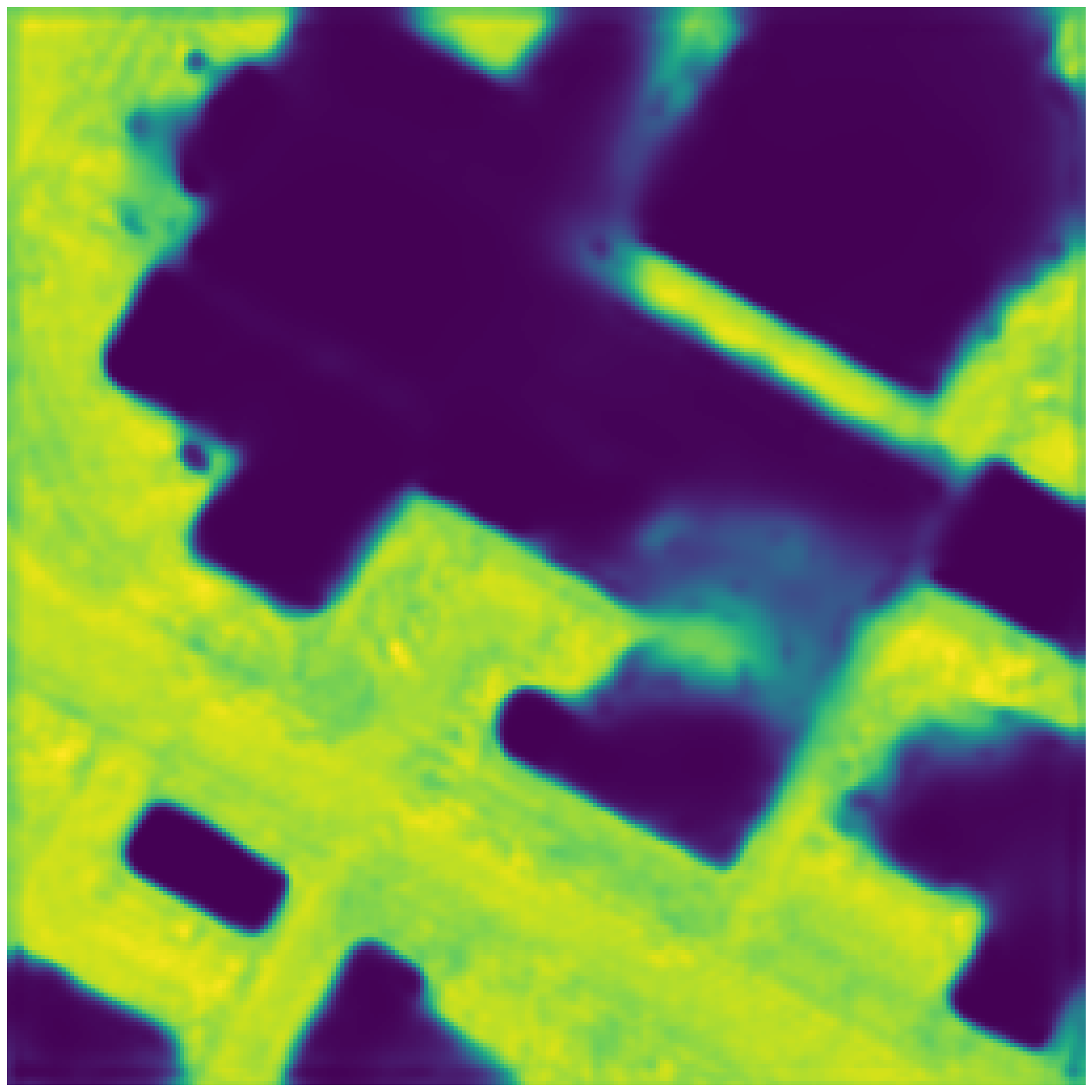} }}
    \subfloat{{\includegraphics[width=2.4cm]{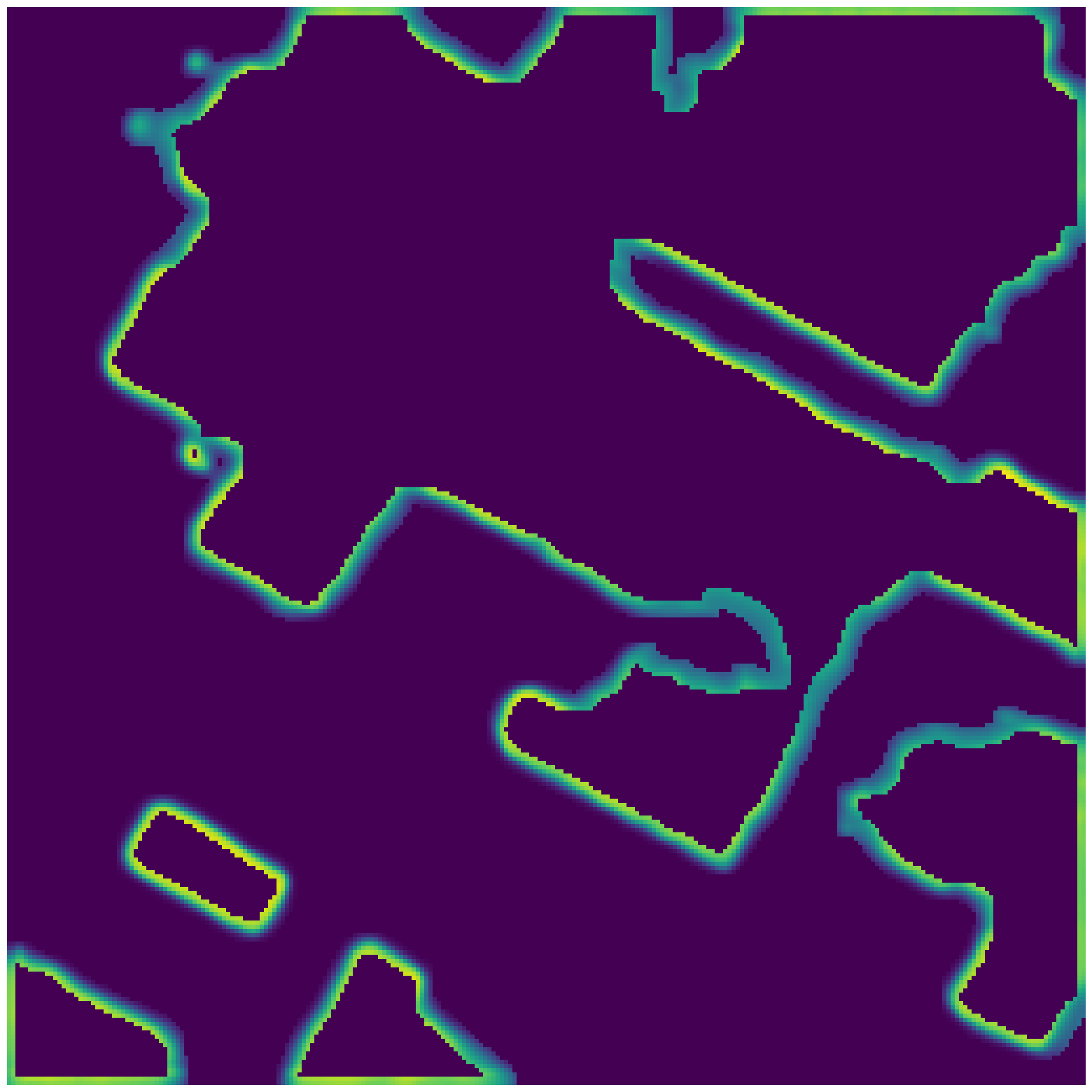}}}
    \subfloat{{\includegraphics[width=2.4cm]{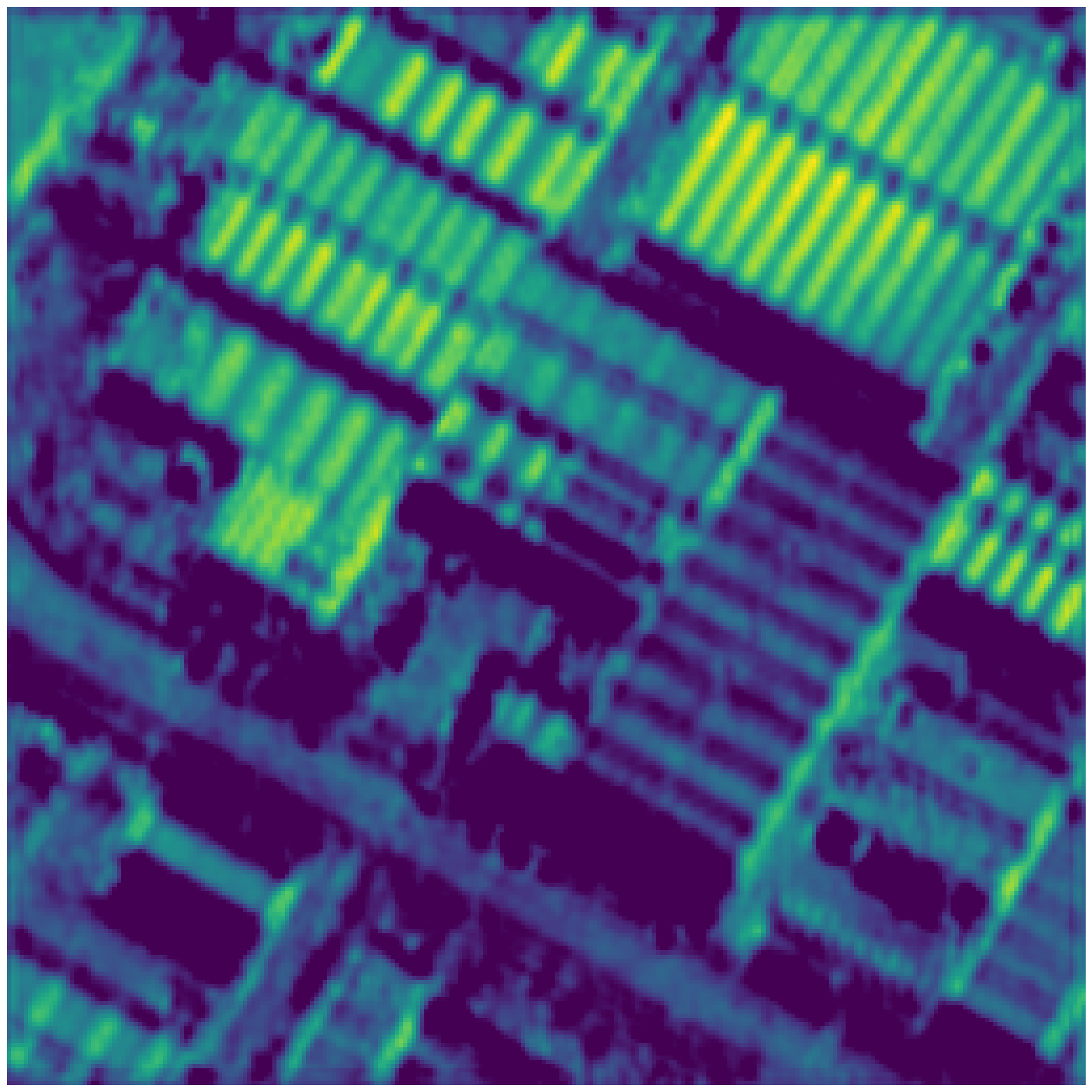} }}
    \subfloat{{\includegraphics[width=2.4cm]{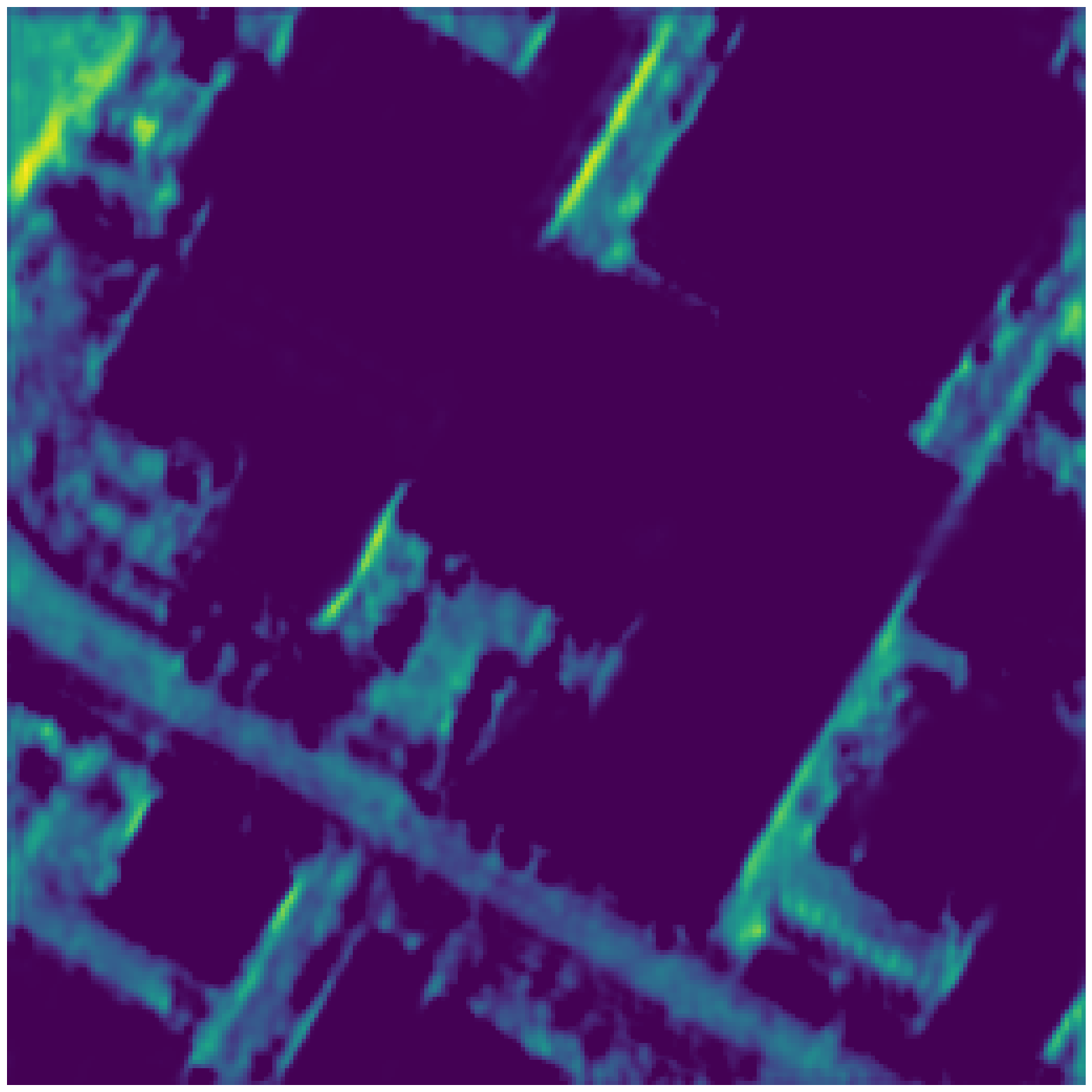} }}
    \subfloat{{\includegraphics[width=2.4cm]{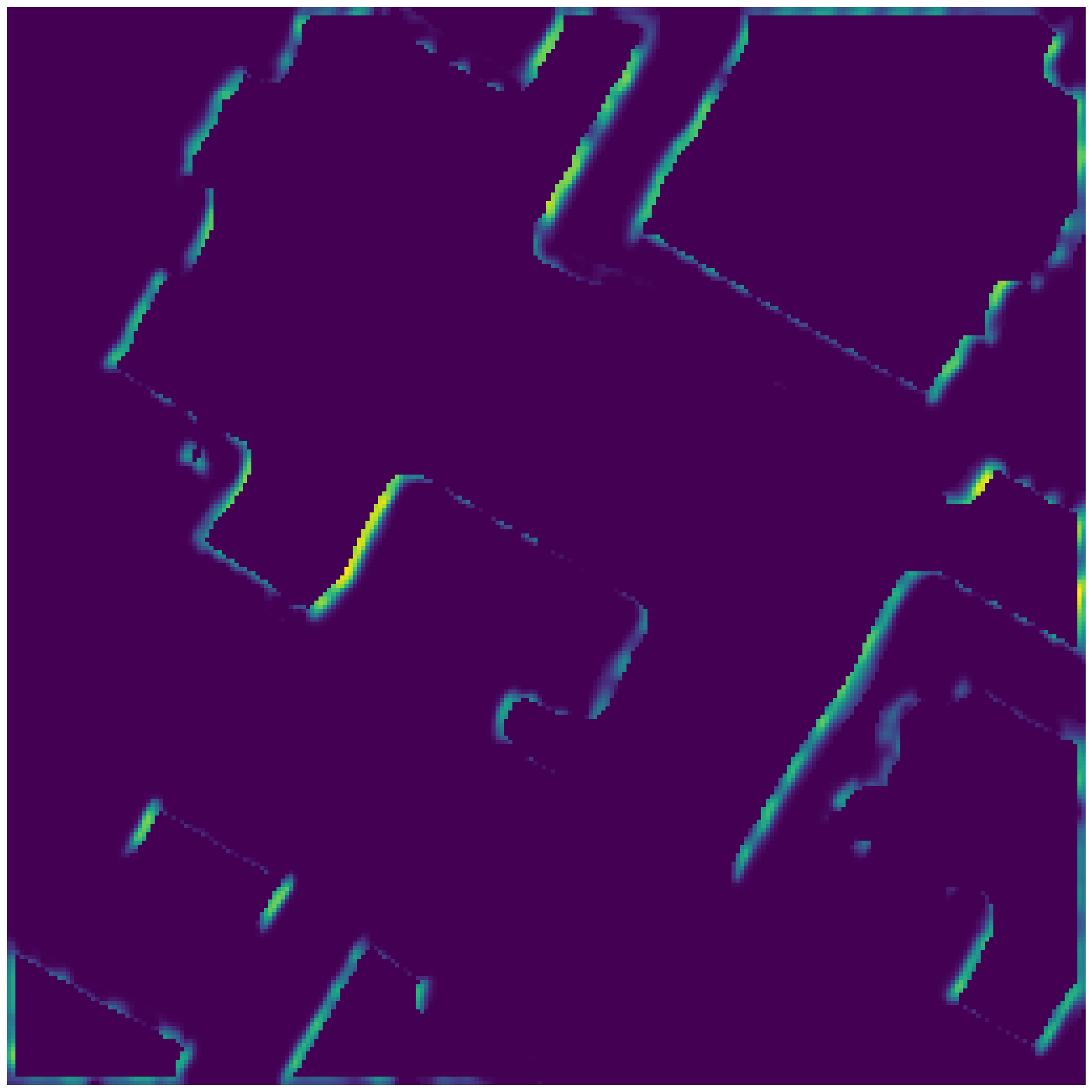}}}
    \hfill
    \subfloat{{\includegraphics[width=2.4cm]{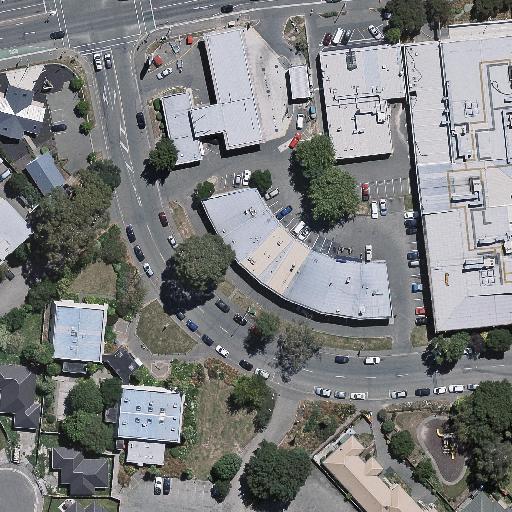} }}
    \subfloat{{\includegraphics[width=2.4cm]{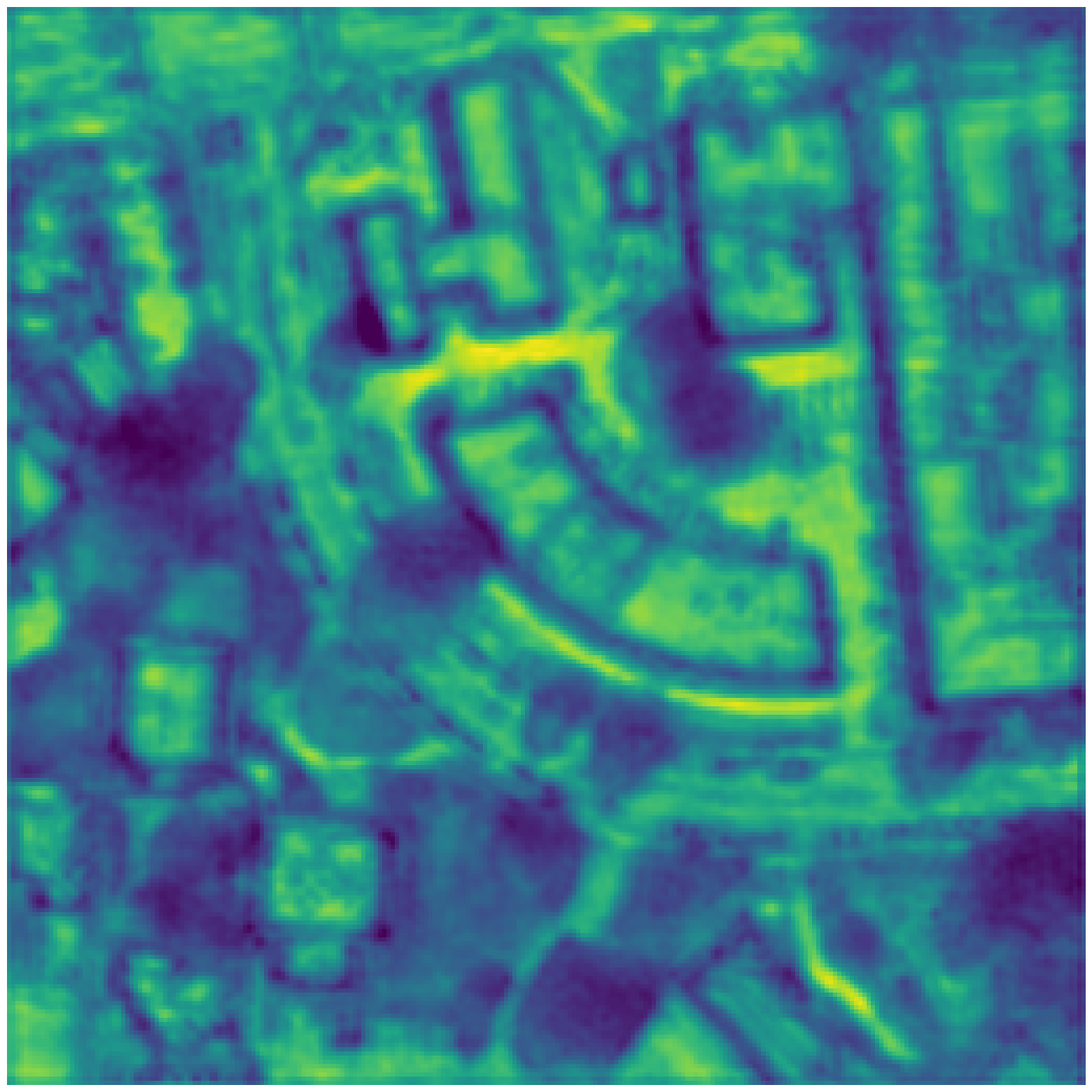} }}
    \subfloat{{\includegraphics[width=2.4cm]{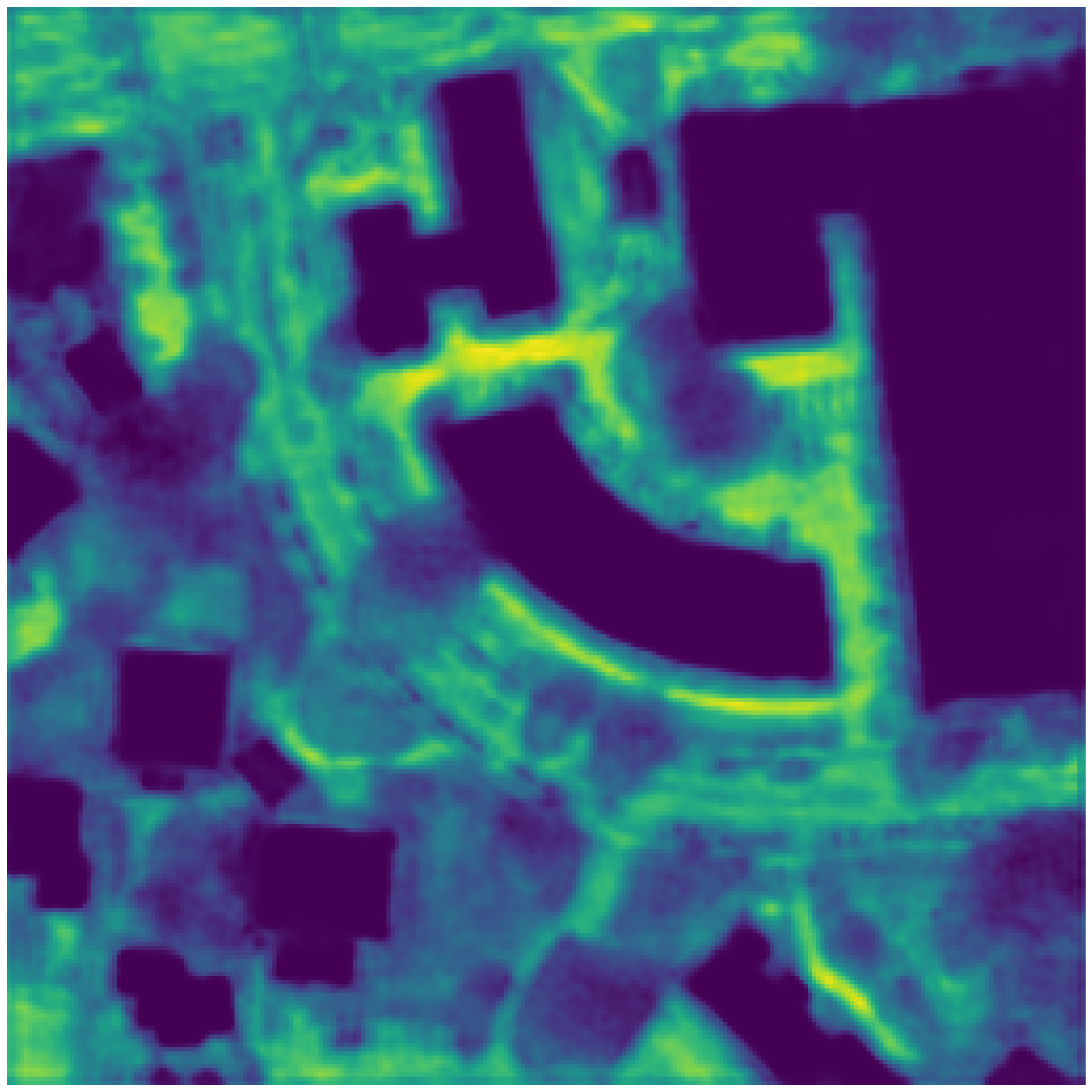} }}
    \subfloat{{\includegraphics[width=2.4cm]{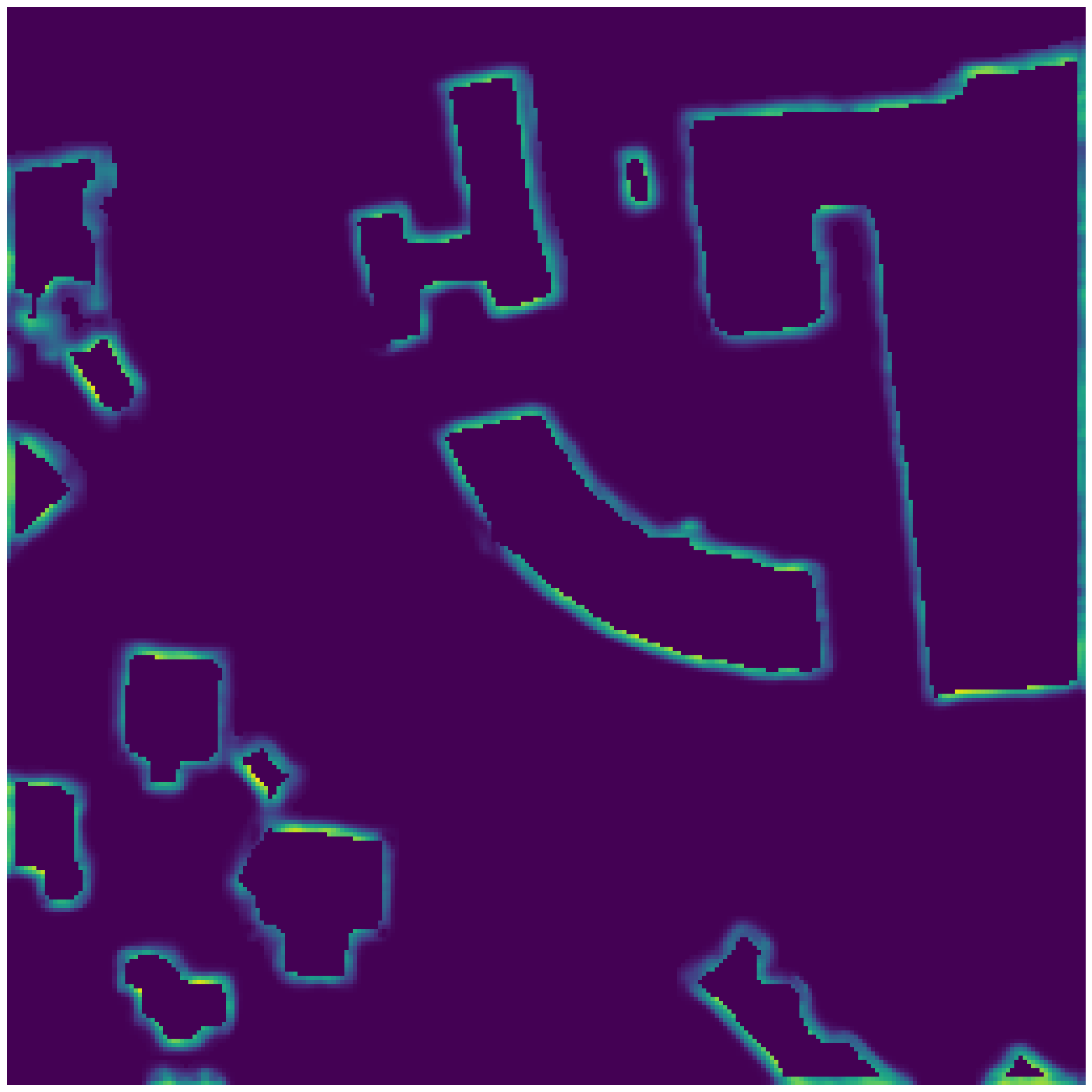}}}
    \subfloat{{\includegraphics[width=2.4cm]{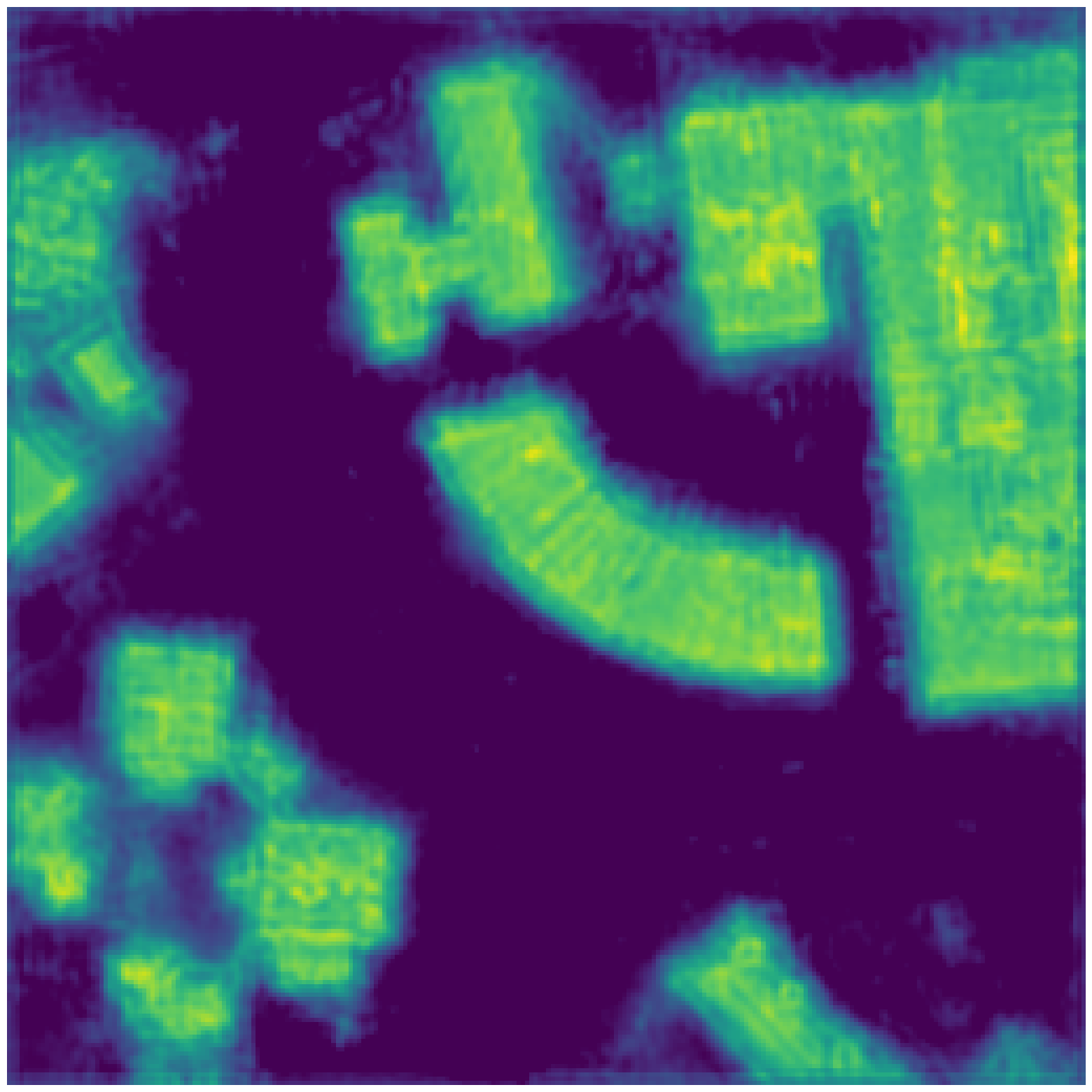} }}
    \subfloat{{\includegraphics[width=2.4cm]{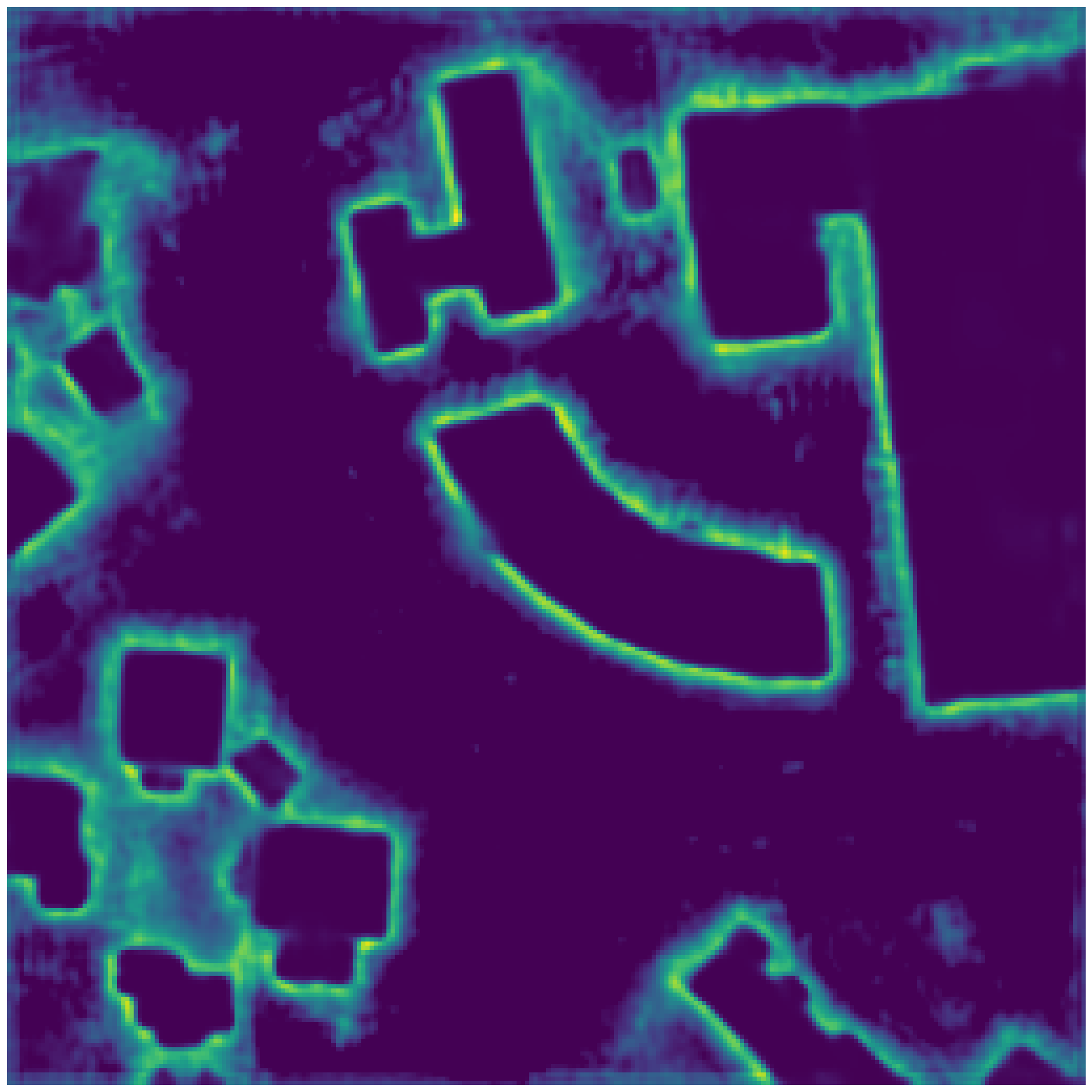} }}
    \subfloat{{\includegraphics[width=2.4cm]{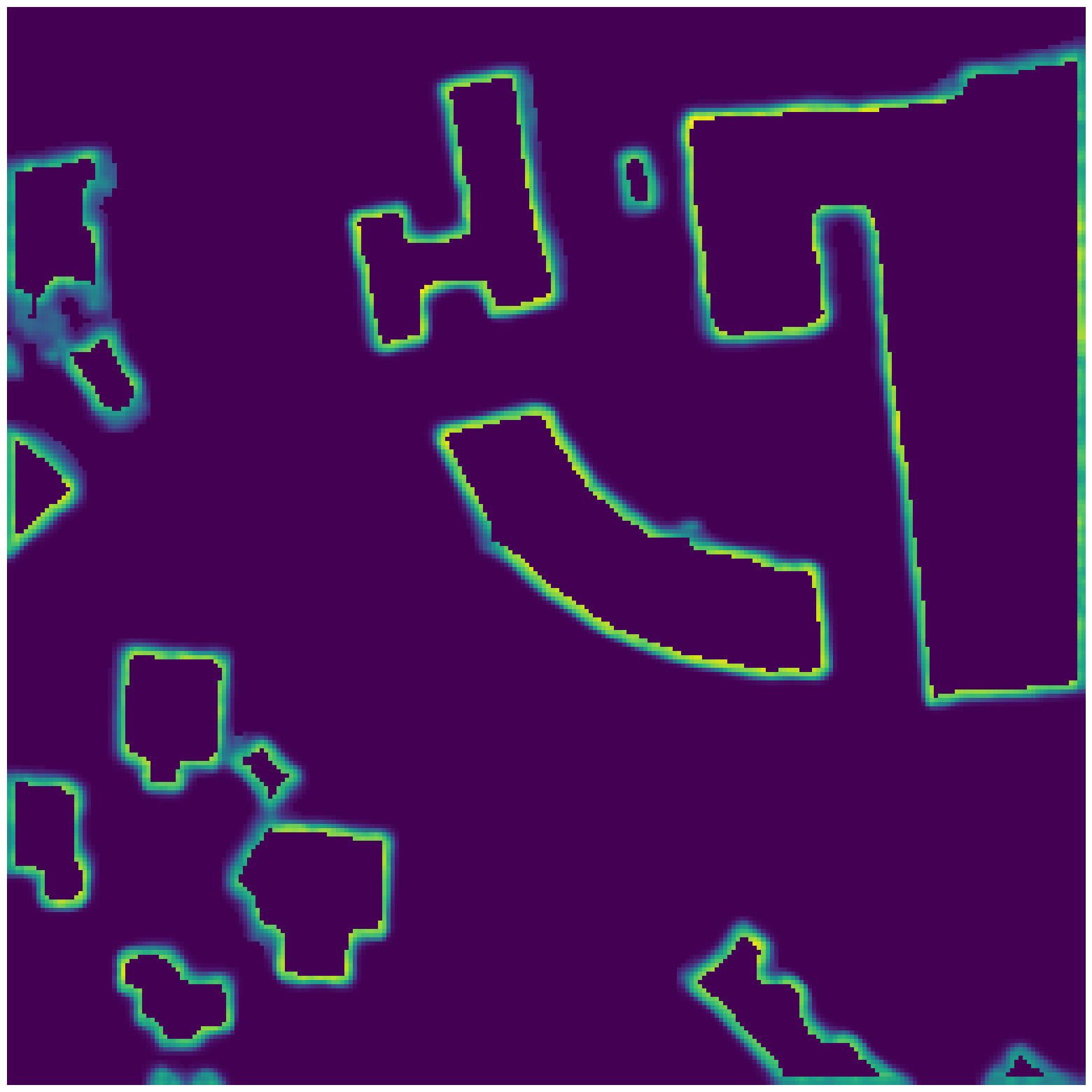}}}
    \hfill
    \subfloat{{\includegraphics[width=2.4cm]{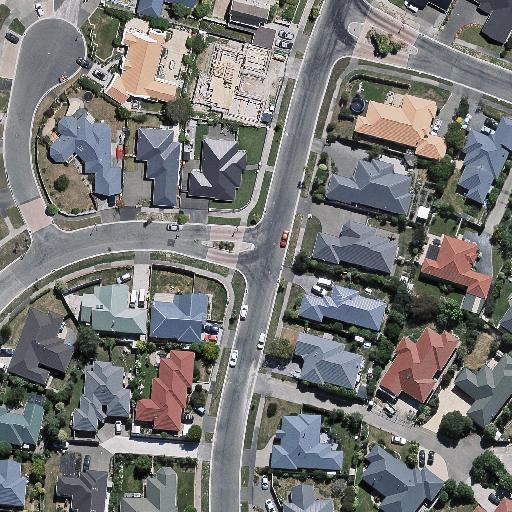} }}
    \subfloat{{\includegraphics[width=2.4cm]{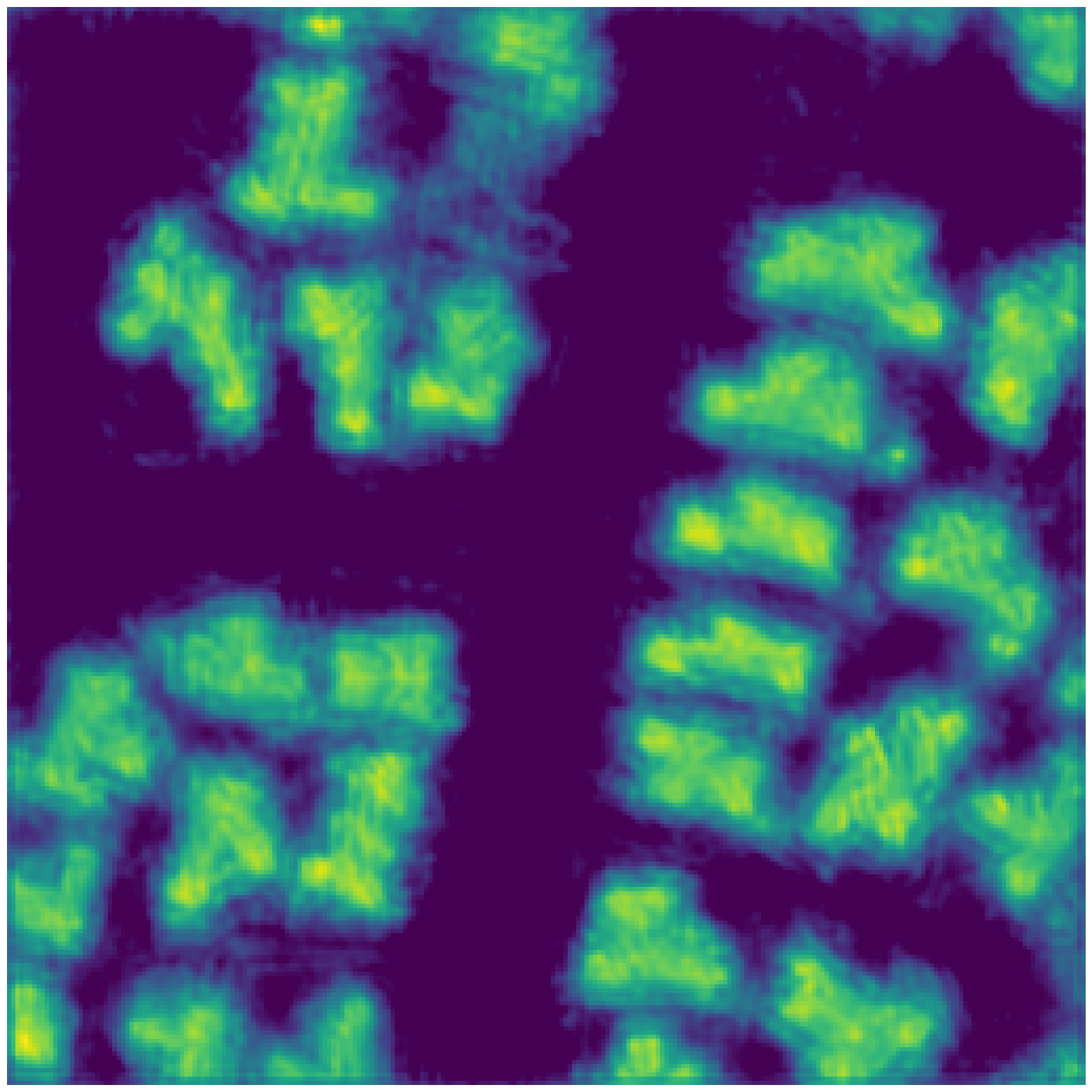} }}
    \subfloat{{\includegraphics[width=2.4cm]{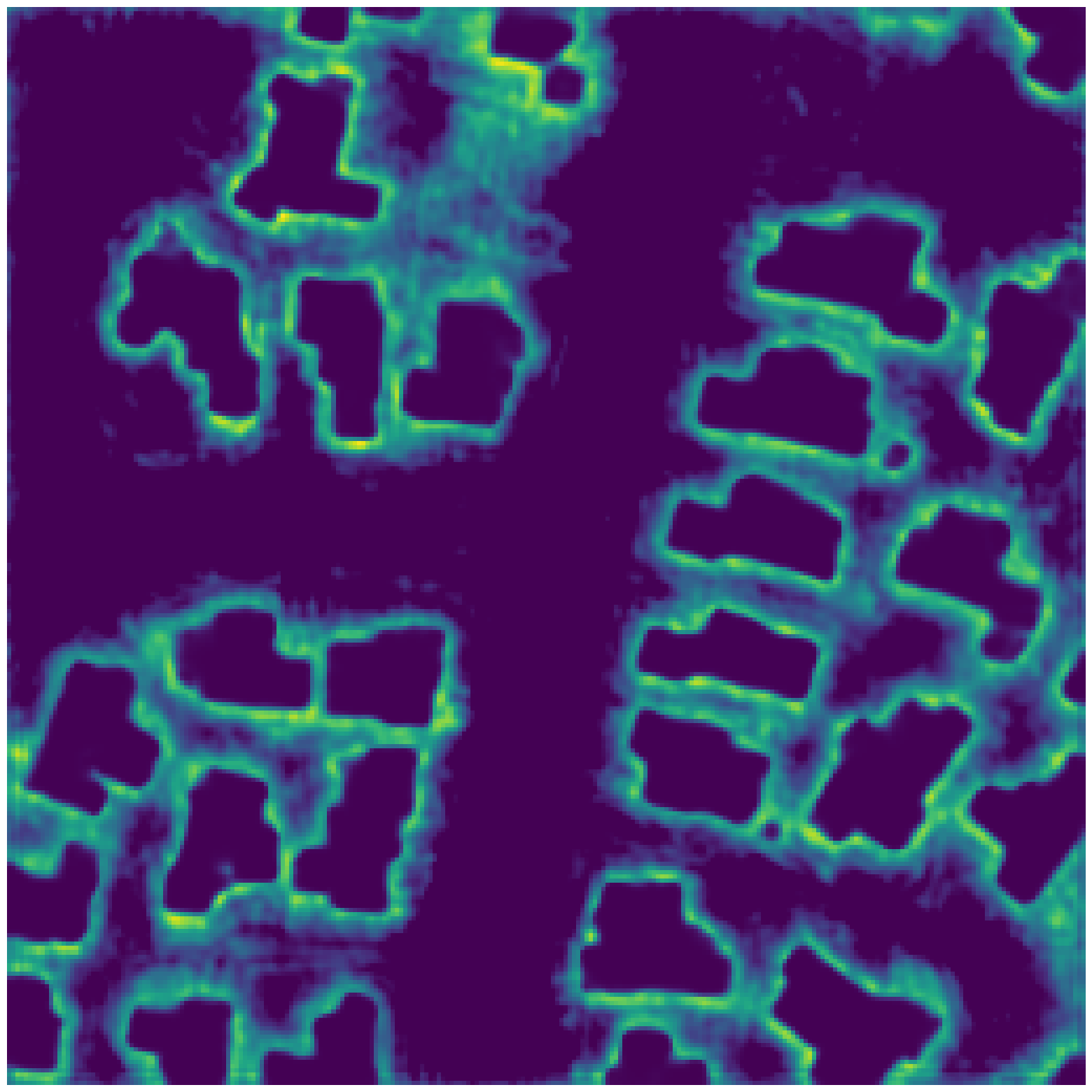} }}
    \subfloat{{\includegraphics[width=2.4cm]{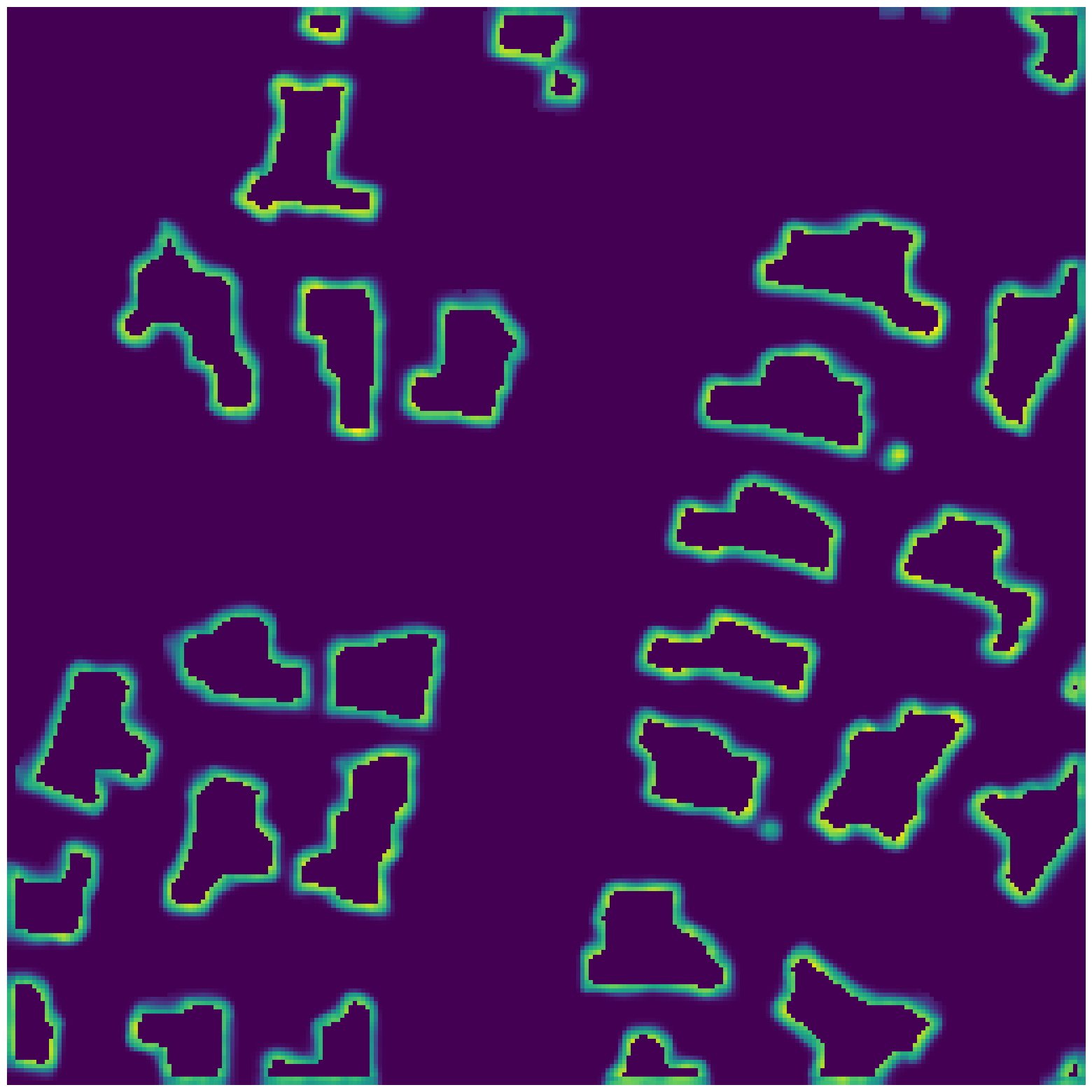}}}
    \subfloat{{\includegraphics[width=2.4cm]{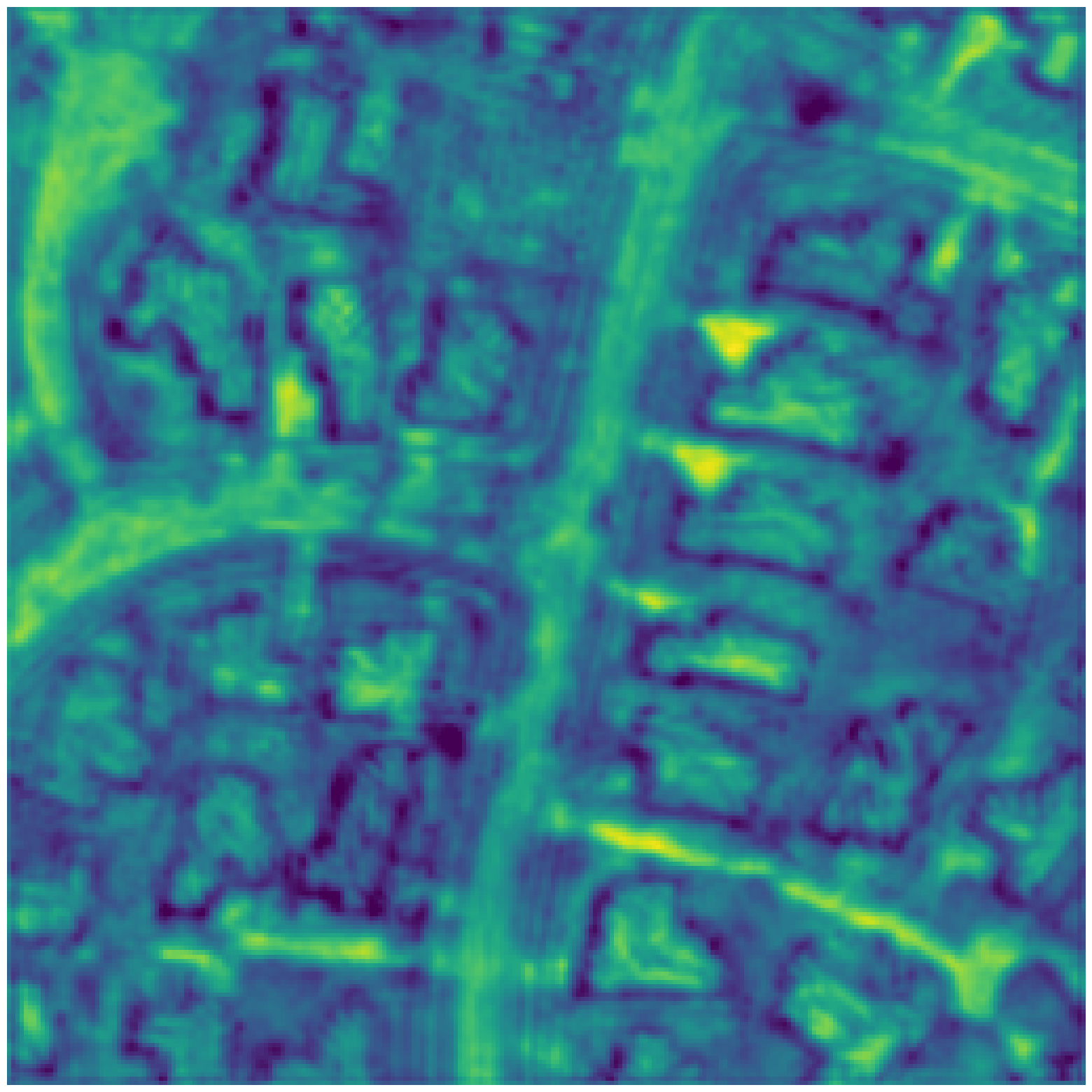} }}
    \subfloat{{\includegraphics[width=2.4cm]{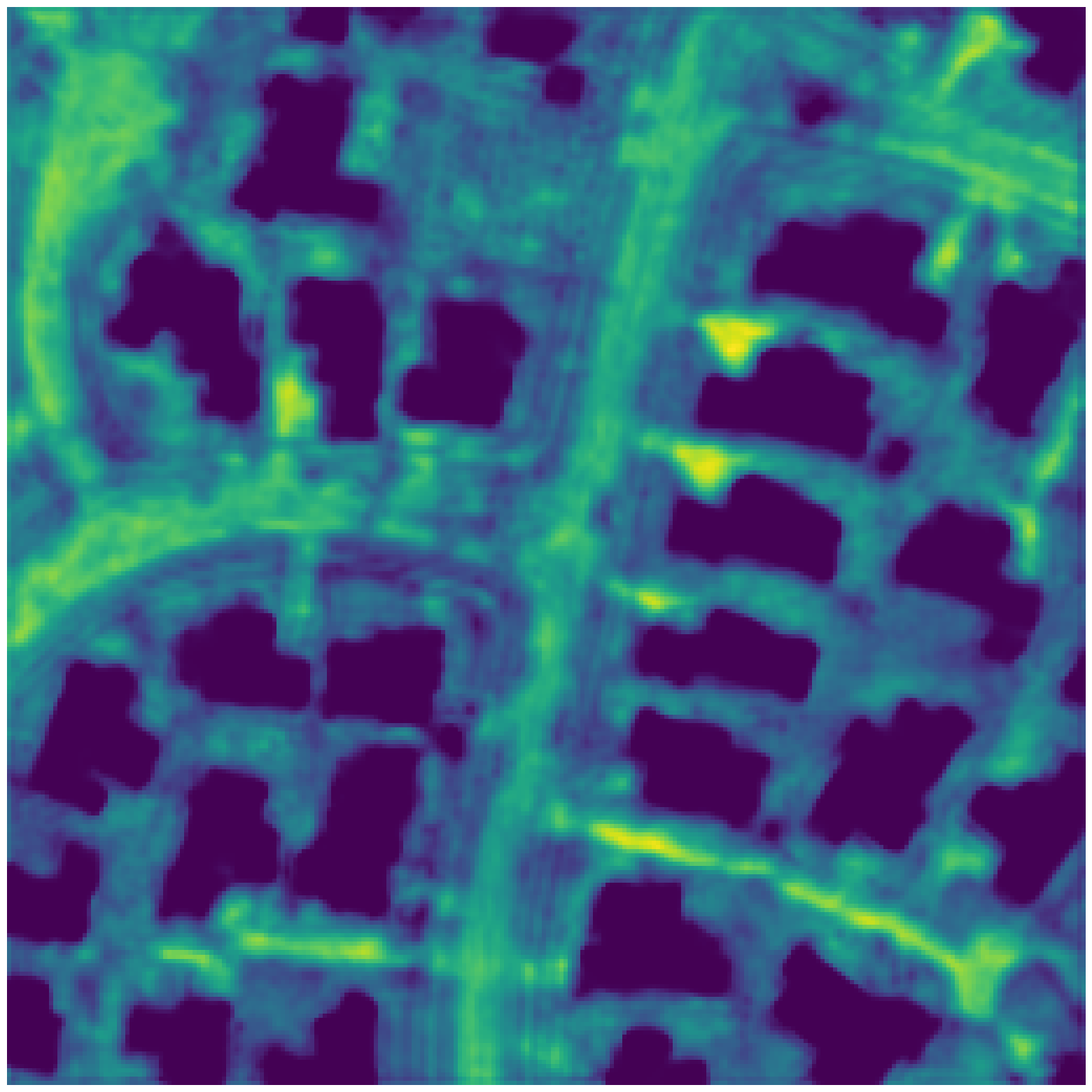} }}
    \subfloat{{\includegraphics[width=2.4cm]{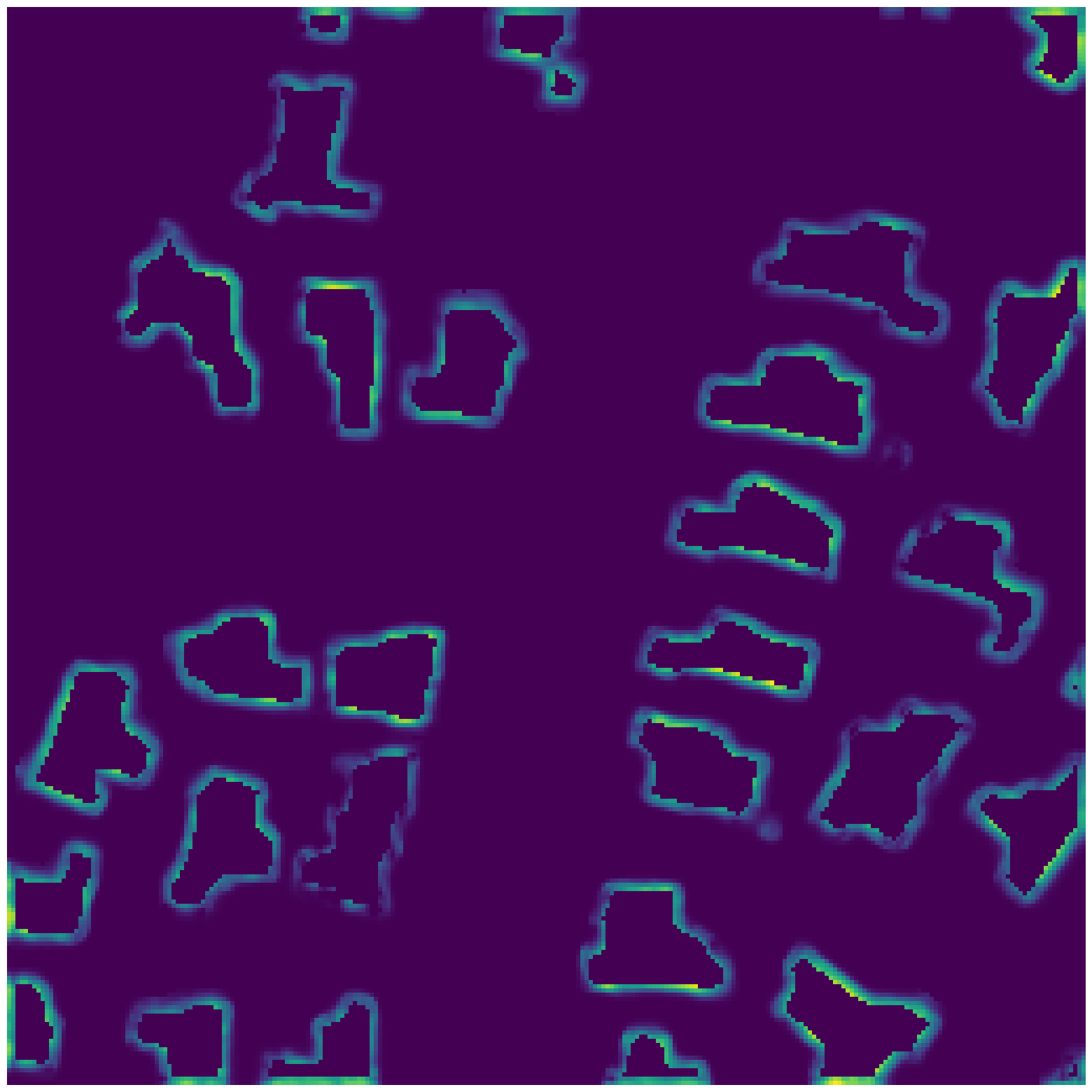}}}
    \hfill
    \subfloat{{\includegraphics[width=2.4cm]{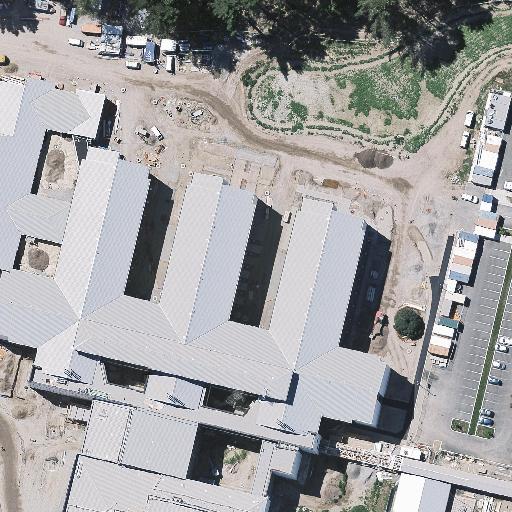} }}
    \subfloat{{\includegraphics[width=2.4cm]{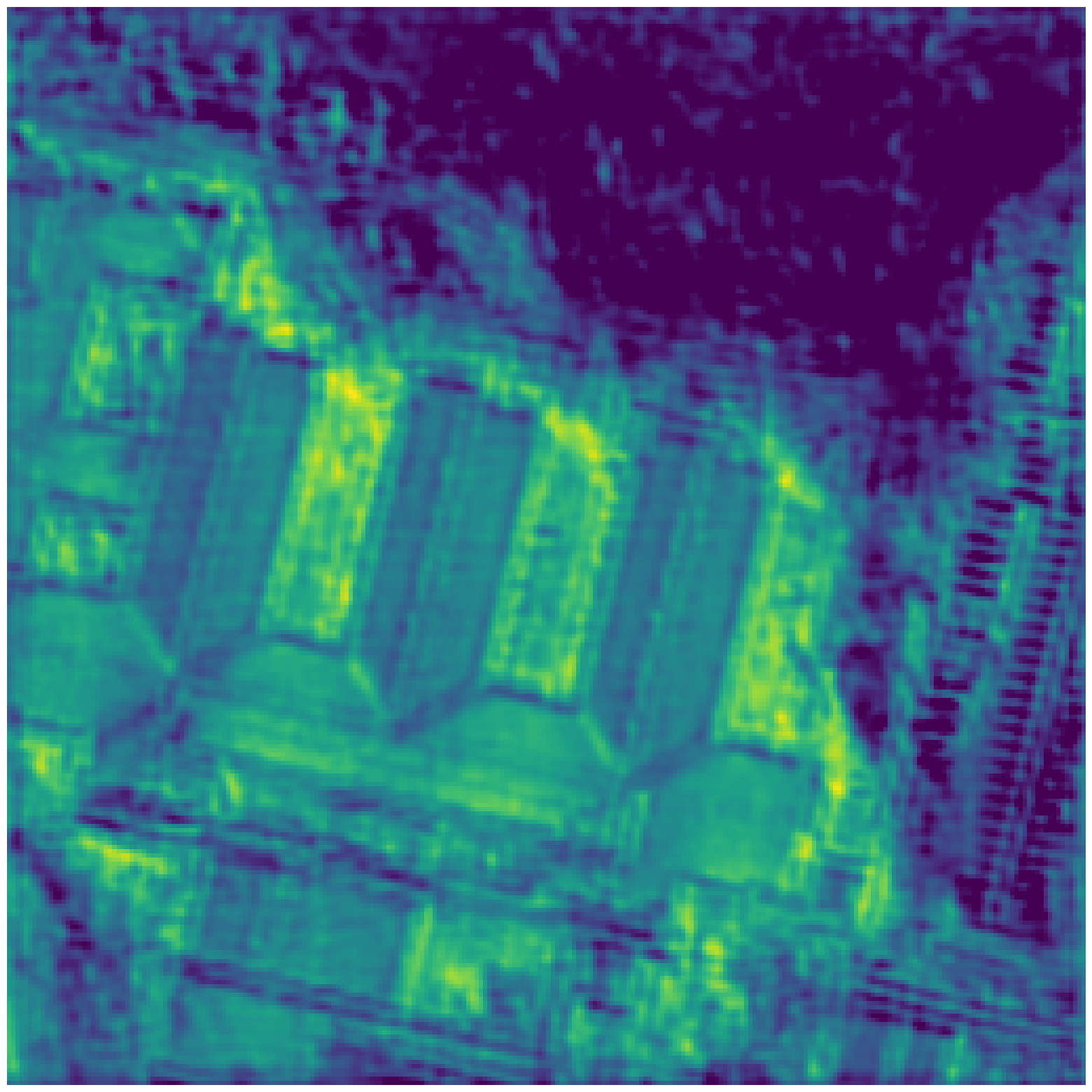} }}
    \subfloat{{\includegraphics[width=2.4cm]{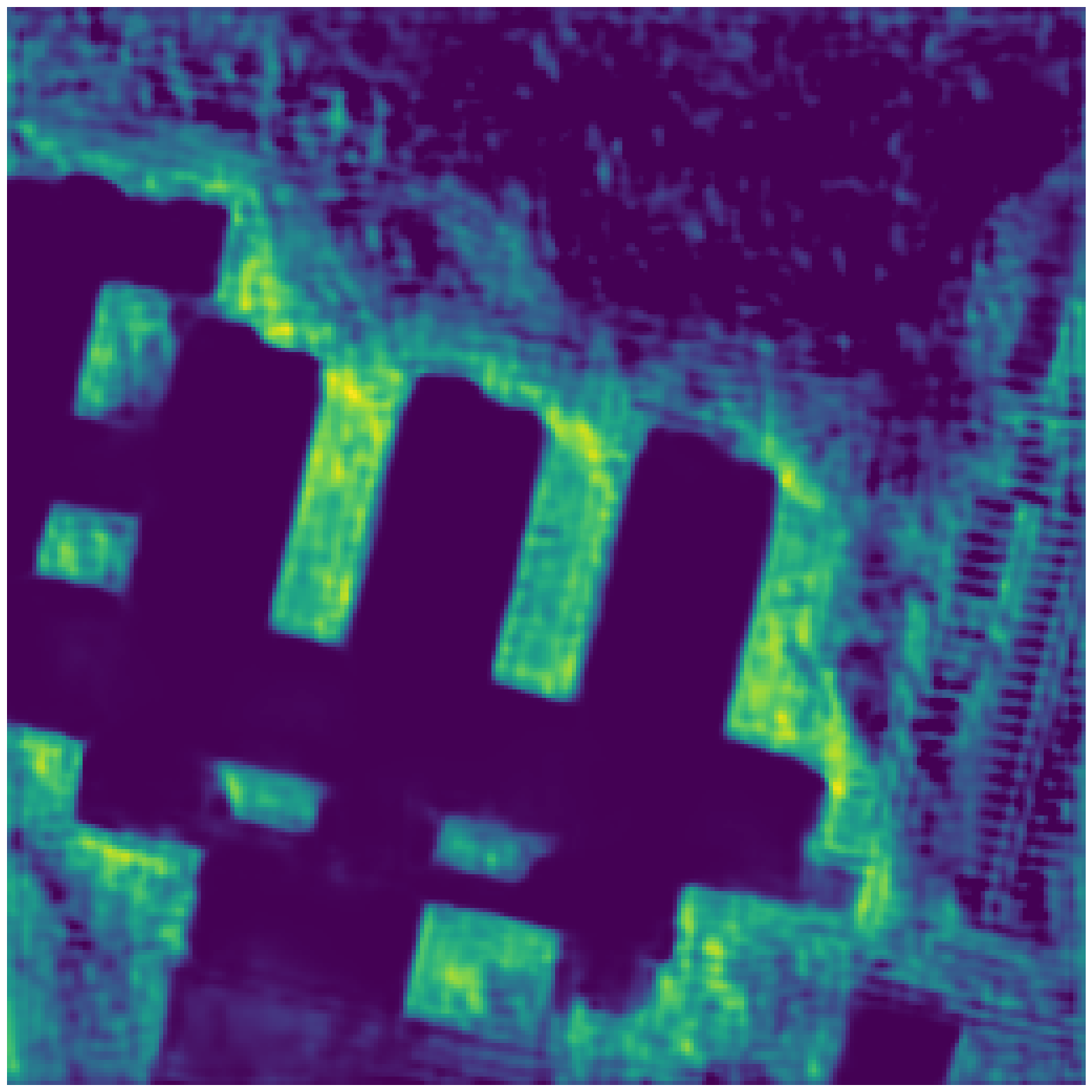} }}
    \subfloat{{\includegraphics[width=2.4cm]{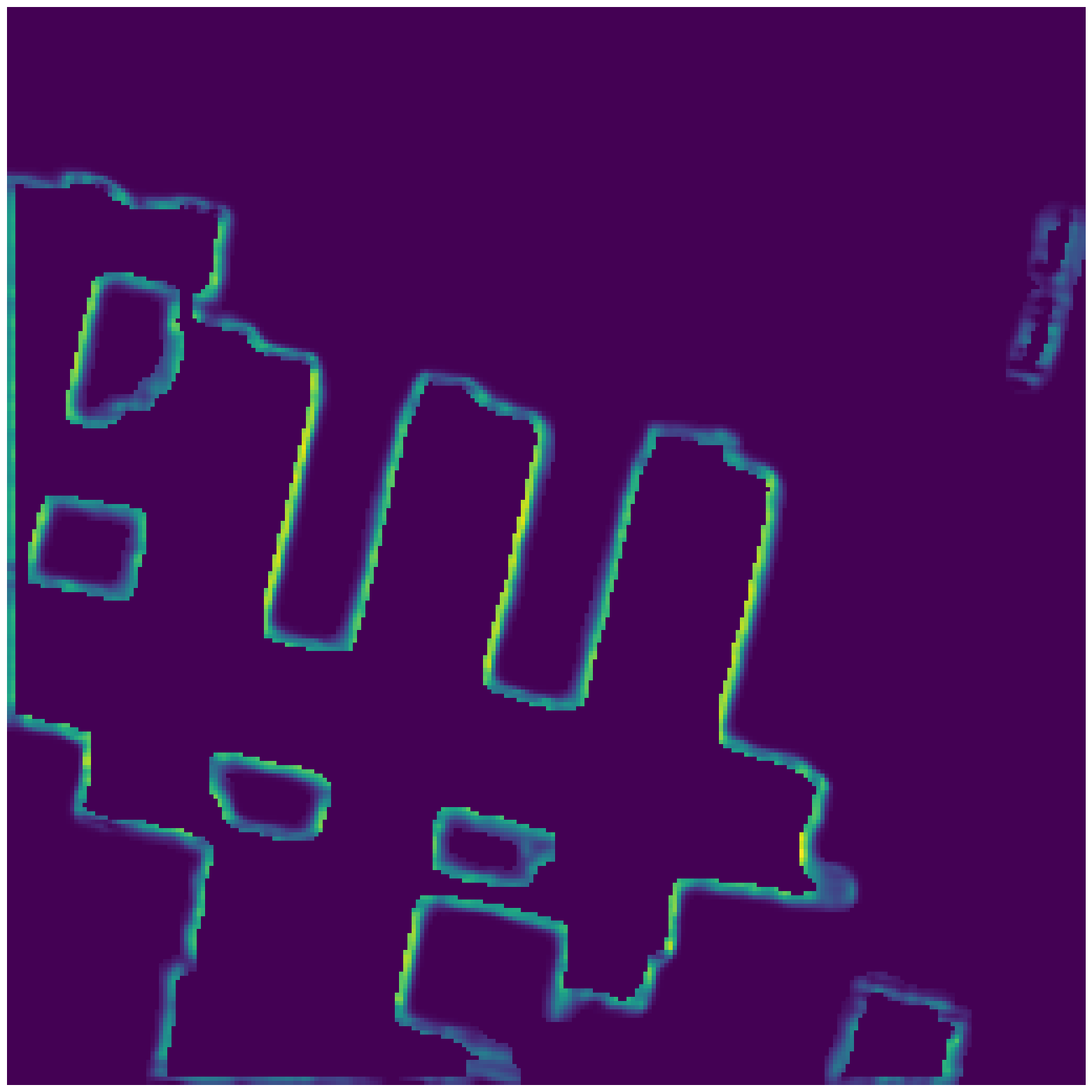}}}
    \subfloat{{\includegraphics[width=2.4cm]{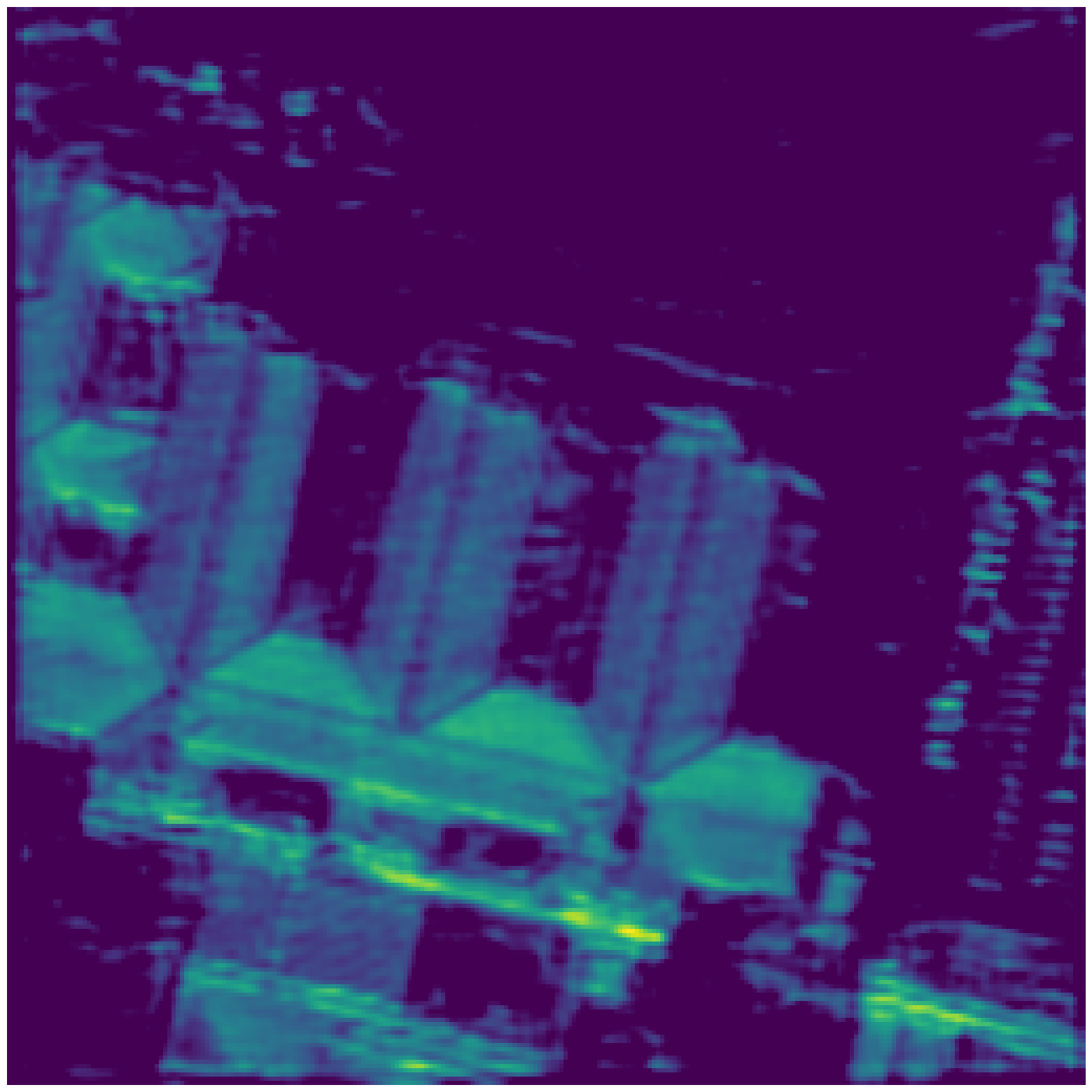} }}
    \subfloat{{\includegraphics[width=2.4cm]{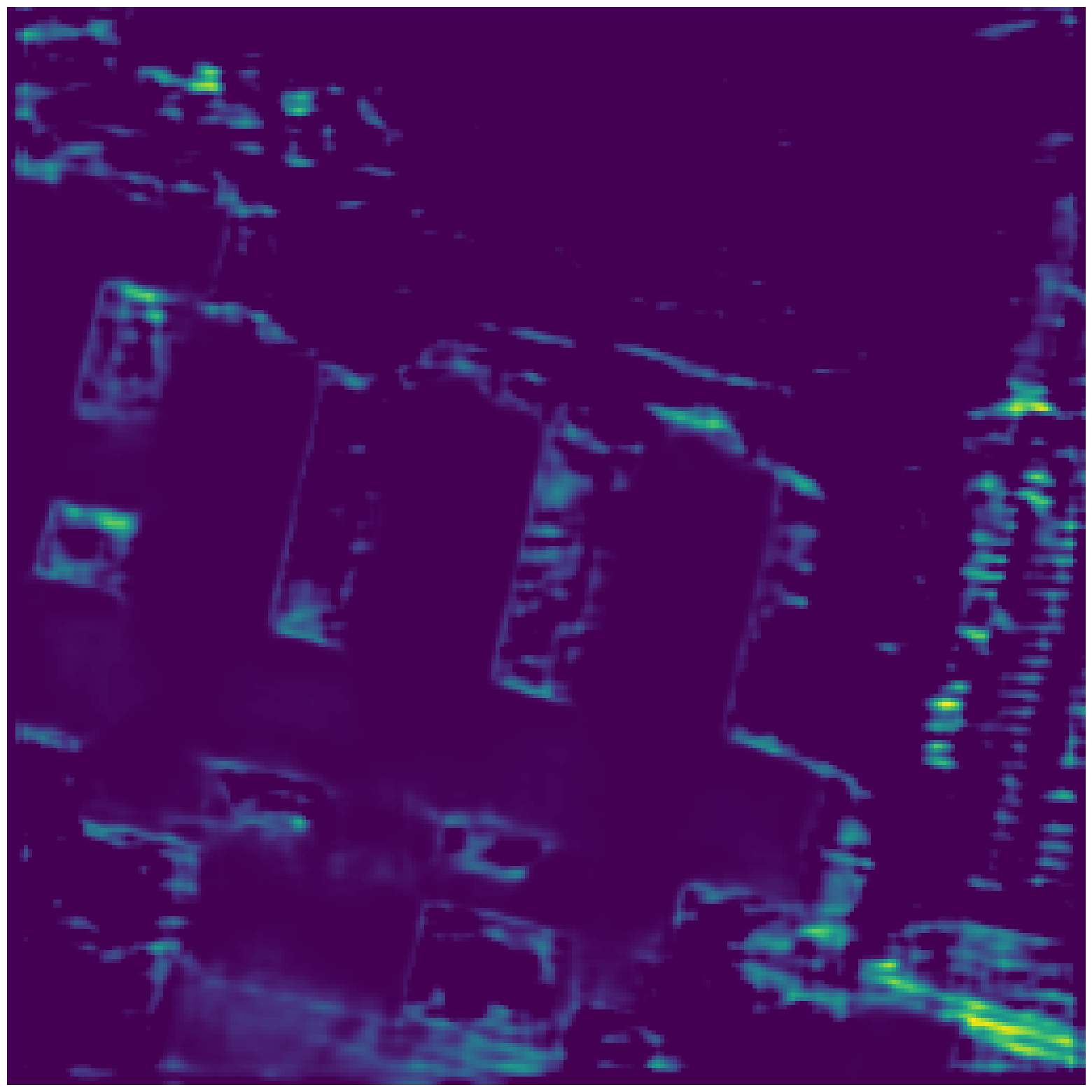} }}
    \subfloat{{\includegraphics[width=2.4cm]{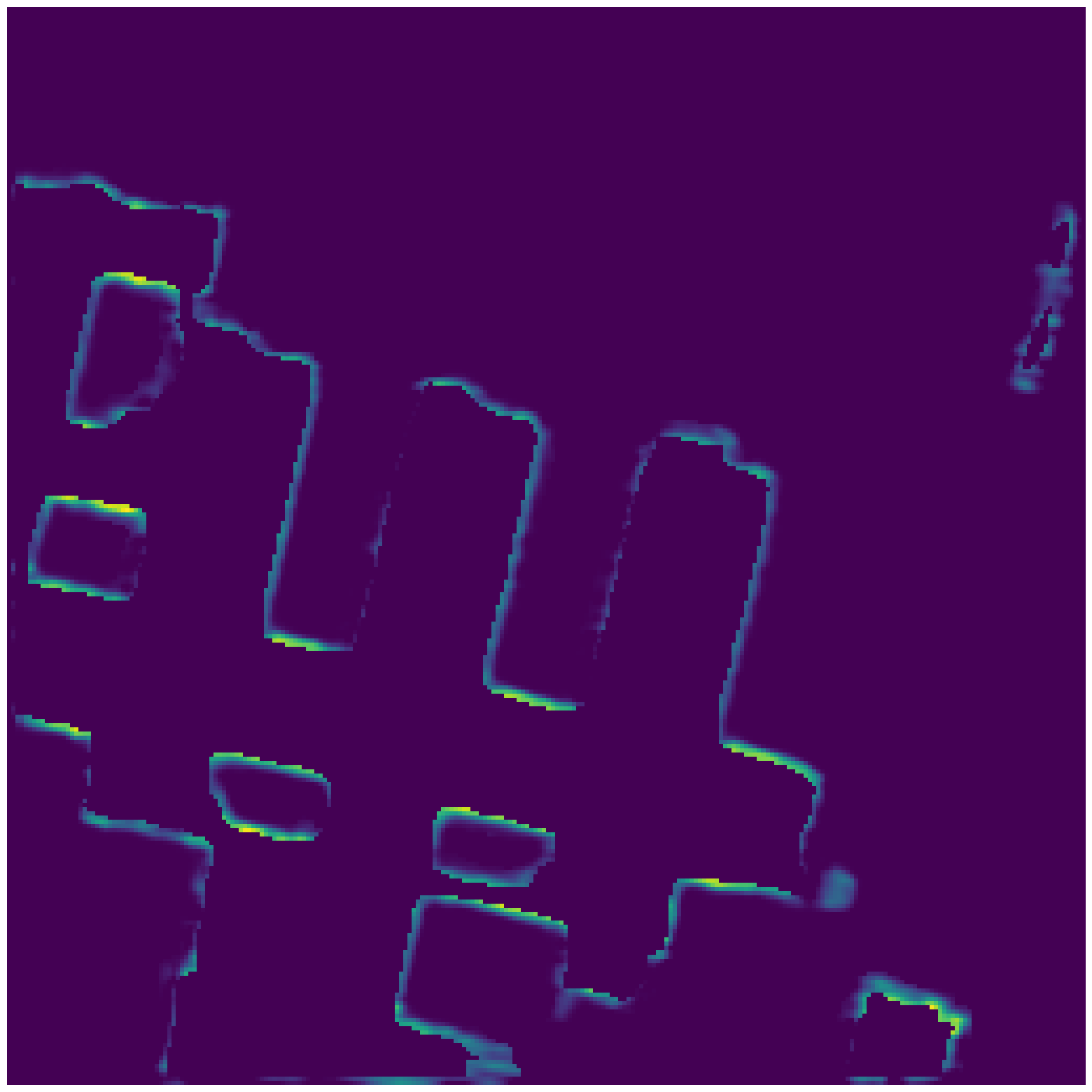}}}
    \hfill
    \caption{\centering Visualization of the decoder feature maps before and after applying reverse and edge attention. Column 1: Input image. Column 2, 5: Decoded Convolutional Features {\em  without} any attention. Column 3, 6: Decoded Convolutional Features {\em  with} Reverse Attention. Column 4, 7: Decoded Convolutional Features {\em  with} Edge Attention.}
    \label{fig:rm_feats}
\end{figure*}

Such a fusion in our framework is carried out by the
\emph{Refinement Module} that is used in each stage of the
decoder for refining the prediction map gradually by
recovering the fine details lost during encoding.  This
module does its work through two attention units: {\em
  Reverse Attention Unit} (RAU) and {\em Edge Attention
  Unit} (EAU).  Through residual learning, both these units
seek to improve the quality of the predictions made in the
previous decoder level on the basis of the finer image
detail captured during the current decoder level.  What's
important here is the fact that both these actions are meant
to be carried out in those regions of an image where the
accuracy of semantic segmentation is likely to be poor ---
e.g. in the vicinity of building boundaries, as can be seen in Figure~\ref{fig:rm_feats}.

For example, starting with the bottleneck, the encoded
features extracted from the ASPP module predict the top-most
prediction map that is at low resolution but rich in
semantic information.  The decoder starts with this coarse
prediction map and looks back at it in the next layer of the
decoder where additional image detail is available for
improving the prediction probabilities that were put out by
ASPP and for improving the edge detail associated with the
predictions. The former is accomplished by RAU and the
latter by EAU.  
%It is important to mention that such
While similar techniques have been used in the past to
improve the output of semantic
segmentation \cite{sun2019reverse, sun2020lightweight} and
object detection \cite{chen2018reverse},
%However, to the best of our knowledge, 
we believe that ours is the first contribution that
incorporates these ideas for a reliable extraction of
building footprints in aerial and satellite imagery.
%in the area of remote sensing which explores such a refinement 
%technique for building segmentation.

As shown in Figure~\ref{fig:side}, the Refinement Module
concatenates the feature maps that are produced by RAU and
EAU. The concatenated feature maps are then passed through two  $3\times3$ Conv layers, and the
output of the Refinement Module is then added to the
upsampled upper-layer prediction to obtain a finer
lower-level prediction, as shown in the figure. The circle with a plus sign inside it in the figure
means an element-wise addition of the two inputs. Details
regarding the two attention units are presented in
the next two subsections.
%Sections~\ref{sec:reverse} and ~\ref{sec:edge} .

\paragraph{Reverse Attention}
\label{sec:reverse}

The idea of reverse attention is to reconsider the
predictions coming out of a lower-indexed layer in the
decoder in light of the spatial details available at the
current layer. This amounts to a backward look in the
decoder chain and justifies the name of this attention unit.

Figure~\ref{fig:reverse} illustrates how the reverse
attention mechanism works. The RAU takes two inputs: (1) the upsampled version of the building prediction map produced by the previous decoder layer; and (2) the finer detailed Conv features copied over from the encoder side after they have been processed by the decoder logic in the current layer. As should be evident from the
data flow arrows in Figure~\ref{fig:gen}, the Reverse
Attention Unit (RAU) guides the network to use the fine
detail in the current layer of the decoder and reevaluate
the building predictions coming out of the lower layer.  We
refer to these reassessed predictions as {\em Reverse
  Attention Map}.  At the $n^{th}$ layer, the Reverse
Attention Map is generated as follows:
\begin{equation}
   A_{R}^{n} = 1 - Sigmoid(U(P_{n-1}))
\end{equation}
where $P_{n-1}$ is the building prediction map produced by
the $(n-1)^{th}$ layer and $U(P_{n-1})$ is its upsampled
version that can be understood directly in the $n^{th}$
layer. 

There is a very important reason for the subtraction in the
equation shown above: As one would expect, the building
detection probabilities are poor near the building edges and
that's exactly where we want to direct RAU's firepower,
hence the reversal of the probabilities in the equation
shown above.  As it turns out, this is another reason for
``Reversal'' in the name of this attention unit.

We now define a {\em Reverse--Weighted Feature Map},
$F^n_{R}$, for the $n^{th}$ layer:
\begin{equation}
    F_{R}^n = A_{R}^{n} \otimes F_n
\end{equation}
where the symbol $\otimes$ denotes element-wise
multiplication, and $F_n$ represents the convolutional
feature maps of the $n^{th}$ layer.

\paragraph{Edge Attention}
\label{sec:edge}

The purpose of the edge attention is to improve the quality
of the boundary edges of the building predictions made by
the previous layer of the decoder using the additional image
detail available in the current layer.

Essential to the logic of what improves the boundary edges
is the notion of contour extraction. At each layer on the
decoder side, we want to extract the contours in the fine
detail provided by the encoder side in order to improve the
edges in the building prediction map yielded by the lower
layer.  Note that there is a significant difference between
just detecting the edge pixels and identifying the
contours. Whereas the former could yield just a disconnected
set of pixels on the object edges, the latter is more likely
to yield a set of connected boundary points --- even when
using just contour fragment (as opposed to, say, closed
contours).  On account of the need to make these
calculations GPU compatible, at the moment the notion of
contour extraction is carried out by applying the Sobel edge
detector \cite{gao2010improved} to a building prediction map
followed by a p-pixel dilation of the edge pixels identified
in order to connect what would otherwise be disconnected
pixels.

As shown in Figure~\ref{fig:edge}, the Edge Attention Unit
(EAU) takes two inputs: 1) the upsampled version of the
building prediction map produced by the previous decoder
layer; and 2) the finer detailed convolutional features copied
over from the encoder side after they have been processed by
the decoder logic in the current layer.  The output of EAU
consists of an {\em edge-weighted feature map}.  If $n$
denotes the index for the current layer in the decoder, the
building prediction map produced by the previous layer,
denoted $P_{n-1}$, is first upsampled using bilinear
interpolation to get $U(P_{n-1})$, which is then used to
generate a {\em binary decision map}, $B_{E}^{n}$, for the
current layer as follows:
\begin{equation}
    B_{E}^{n} = 
    \begin{cases}
      1 & \text{if $Sigmoid(U(P_{n-1})) \geq$ 0.5}\\
      0 & \text{otherwise}\\
      \end{cases}
\end{equation}

\noindent Subsequently, the Sobel edge detector is applied
to the binary decision map in order to detect edge fragments
in the predicted binary map.  As shown in
Figure~\ref{fig:edge}, the next step is to dilate the edge
fragments produced by Sobel so that they become $p$-pixels
wide.  The edge dilation step connects what could otherwise
be disjoint edge fragments.  Typically, we dilate the edge
pixels by a kernel of size $7 \times 7$ to get a {\em
  dilated edge map}, $D_{E}^{n}$, which leads to the {\em
  edge attention map} as defined by:
\begin{equation}
    A_{E}^{n} = Sigmoid(U(P_{n-1})) \otimes D_{E}^{n}
\end{equation}

\noindent The edge attention map could be thought of as a
boundary confidence map. This confidence map is then
multiplied with the $n^{th}$ layer feature map to obtain
the edge-weighted features, $F_{E}^n$ as shown below:
\begin{equation}
    F_{E}^n = A_{E}^{n} \otimes F_n
\end{equation}
where $F_n$ is the $n^{th}$ layer feature map. 
  
\subsubsection{Uncertainty Attention}
\label{sec:uncertainty}

In general, a classical encoder-decoder network does not
provide for feature selection when fusing together the
high-level features going through decoder with the low-level
features being copied over from the encoder side through the
skip connections.  A manifestation of this phenomenon is
over-segmentation in the final output of the network that is
caused by indiscriminately fusing the low-level features from the encoder with the high-level features in the decoder.

\begin{figure}[h]
    \centering
    \includegraphics[width=0.48\textwidth]{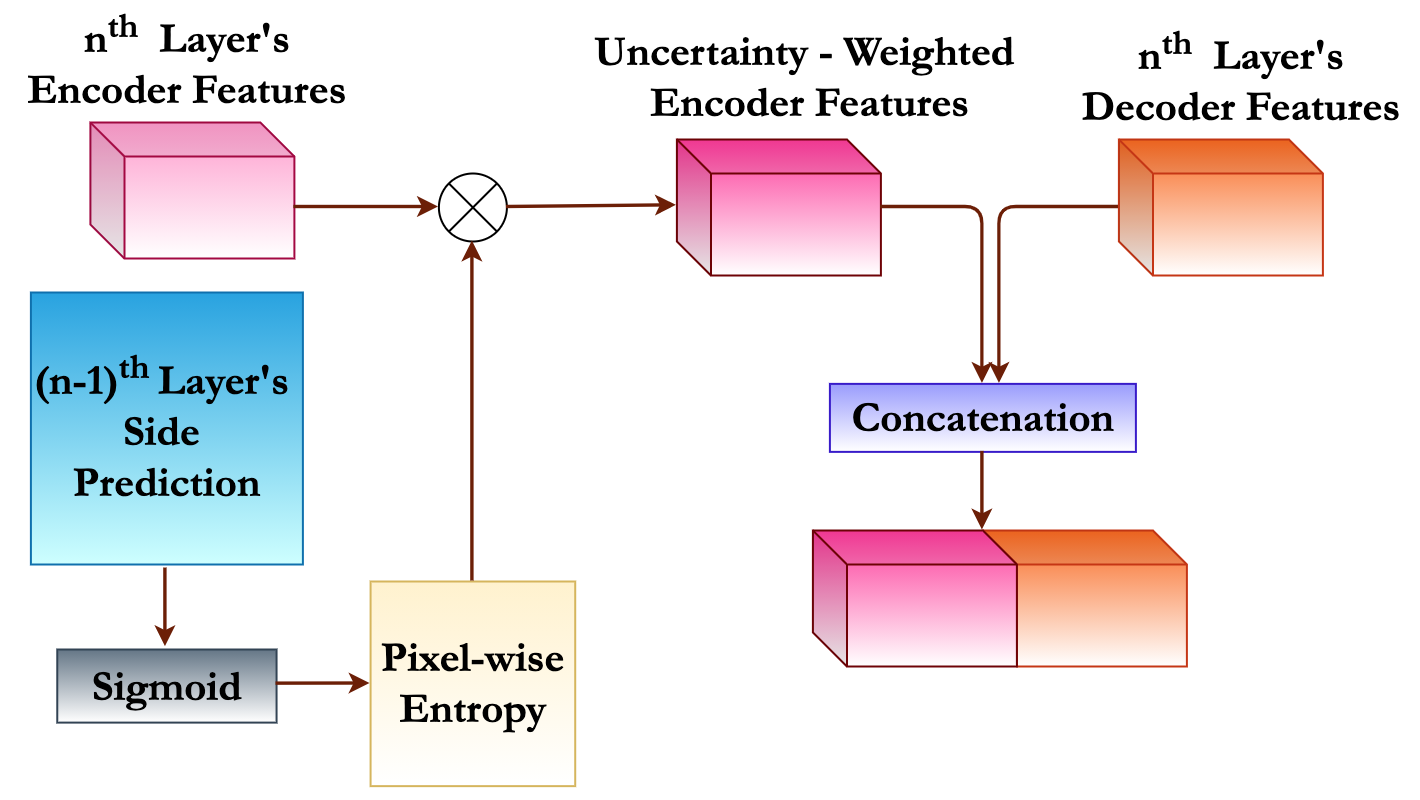}
    \caption{\centering Uncertainty Attention Module (UAM)}
    \label{fig:uncertainty}
\end{figure}

\begin{figure*}
    \centering
    \subfloat{{\includegraphics[width=2.9cm]{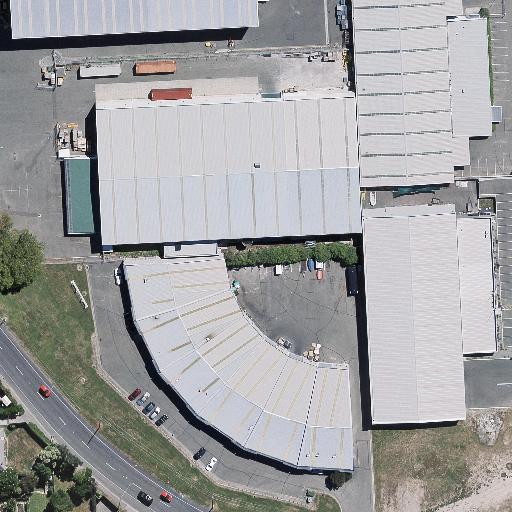} }}
    \subfloat{{\includegraphics[width=2.9cm]{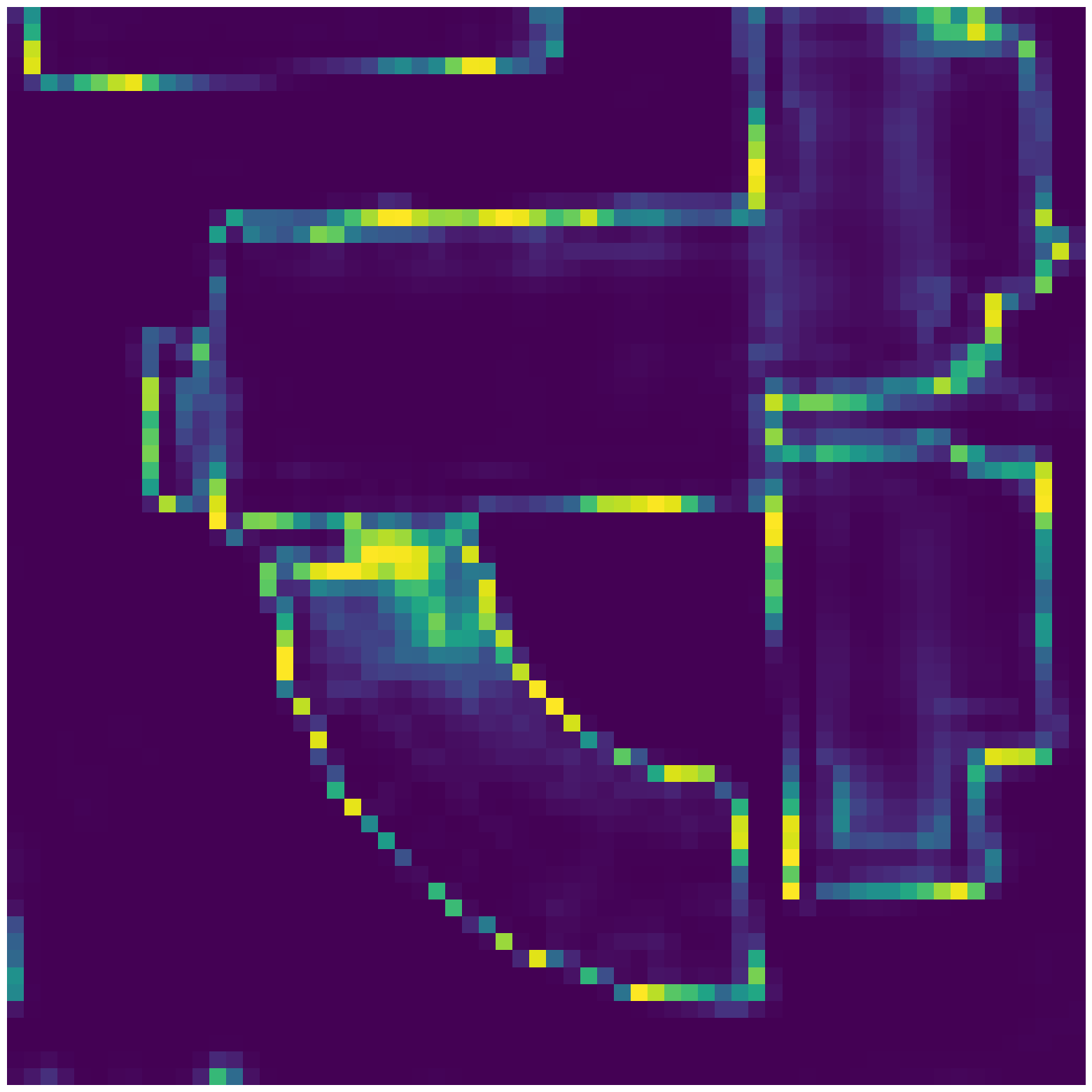} }}
    \subfloat{{\includegraphics[width=2.9cm]{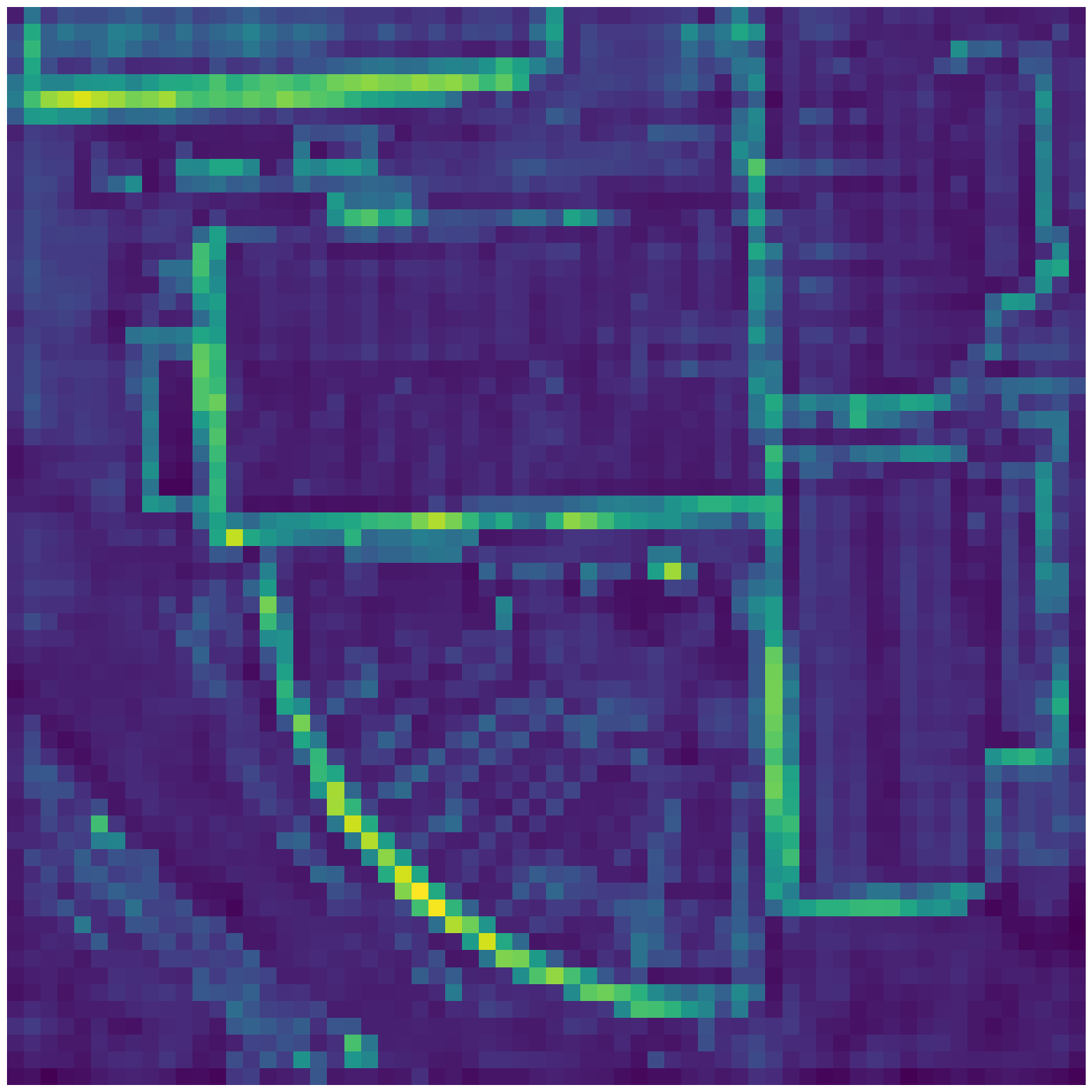} }}
    \subfloat{{\includegraphics[width=2.9cm]{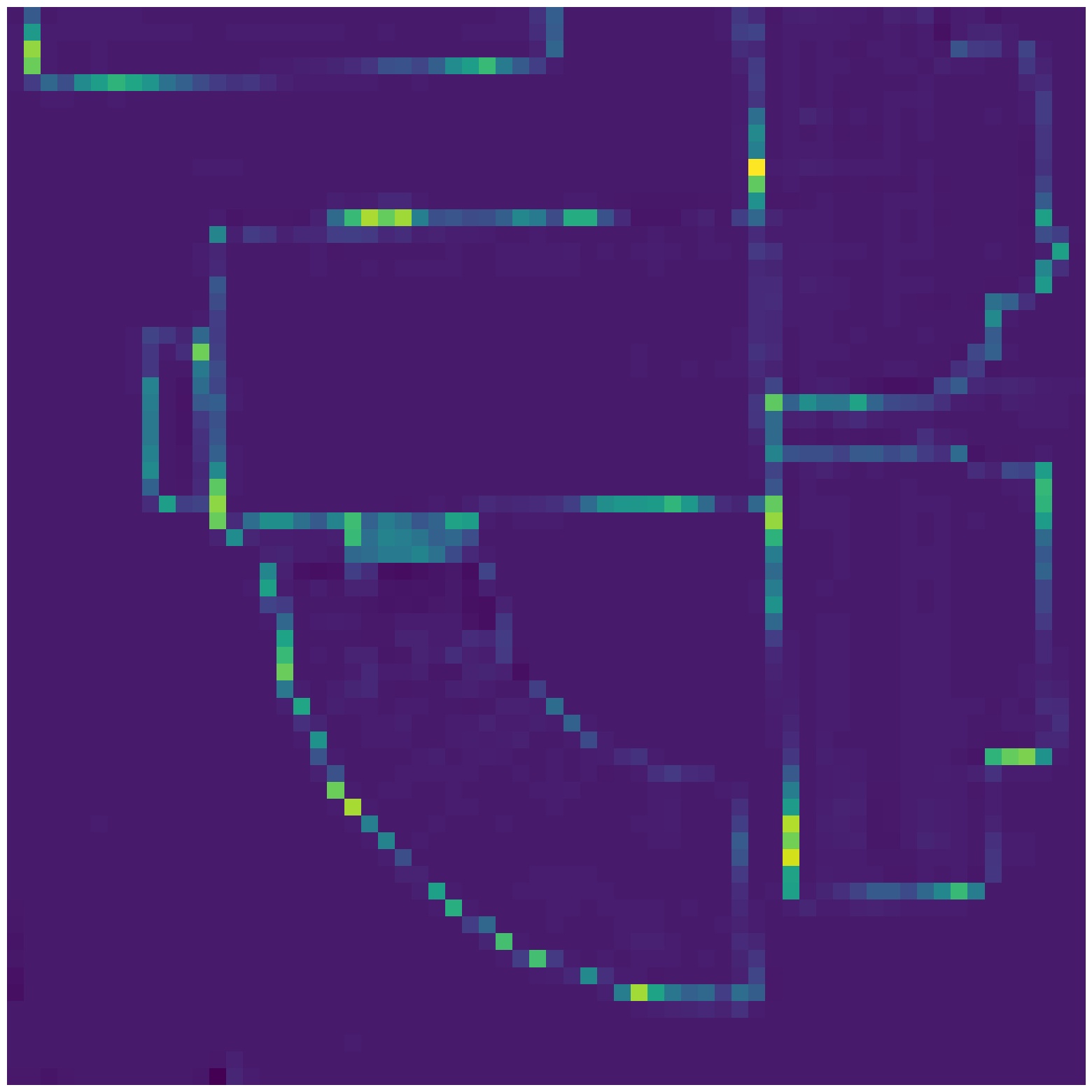}}}
    \subfloat{{\includegraphics[width=2.9cm]{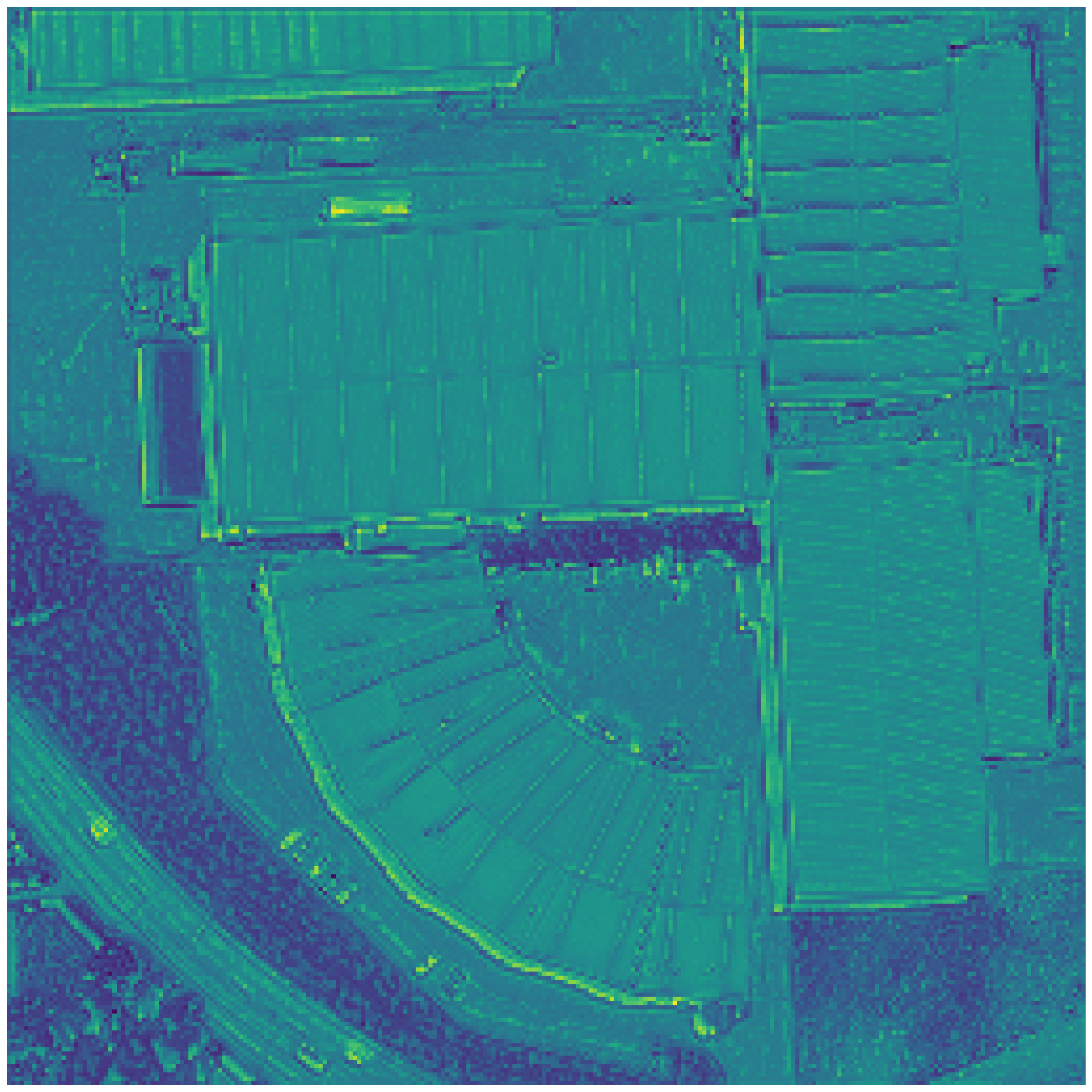} }}
    \subfloat{{\includegraphics[width=2.9cm]{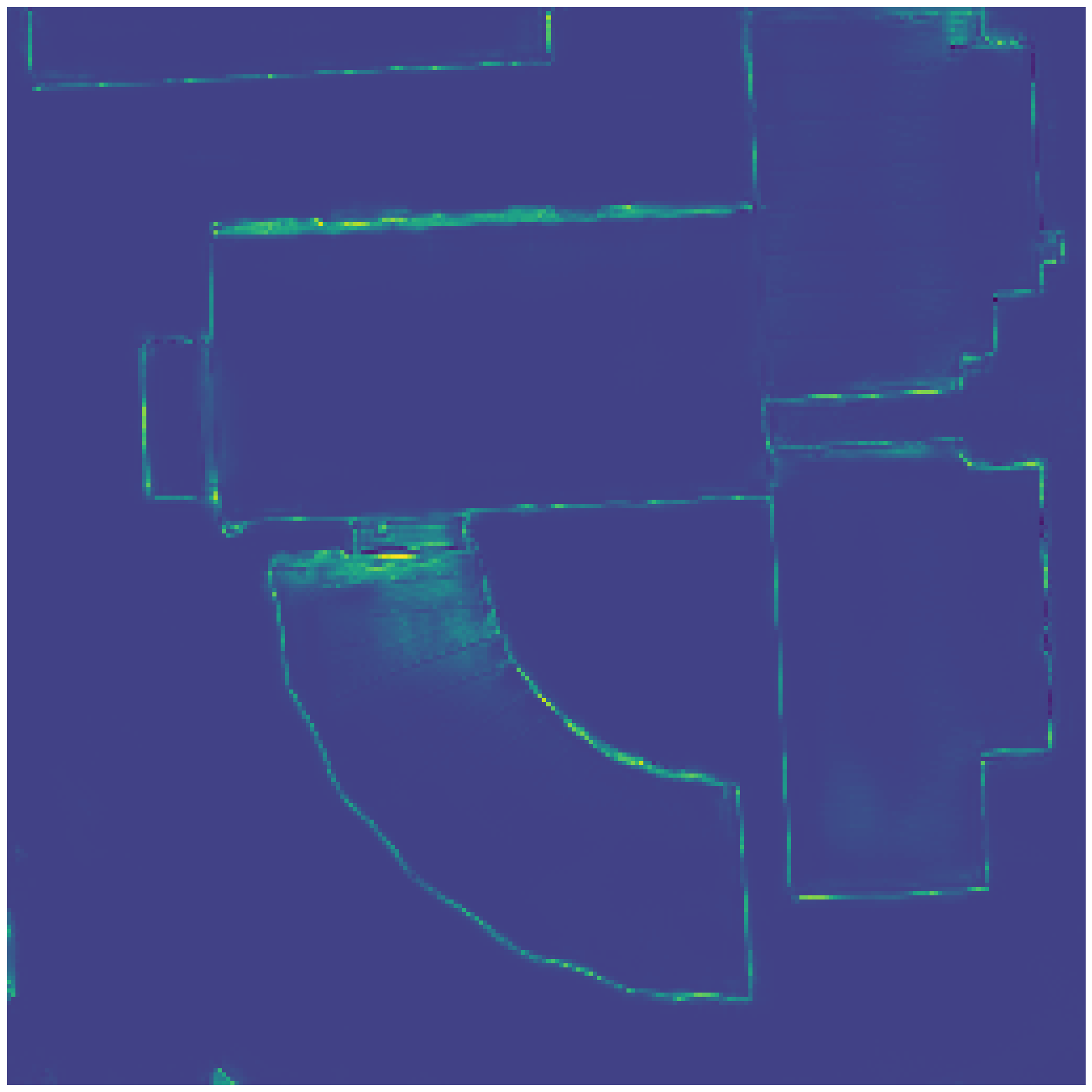}}}
    \hfill
    \subfloat{{\includegraphics[width=2.9cm]{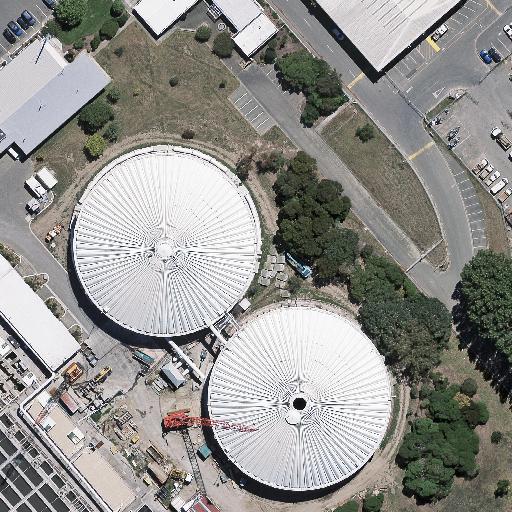} }}
    \subfloat{{\includegraphics[width=2.9cm]{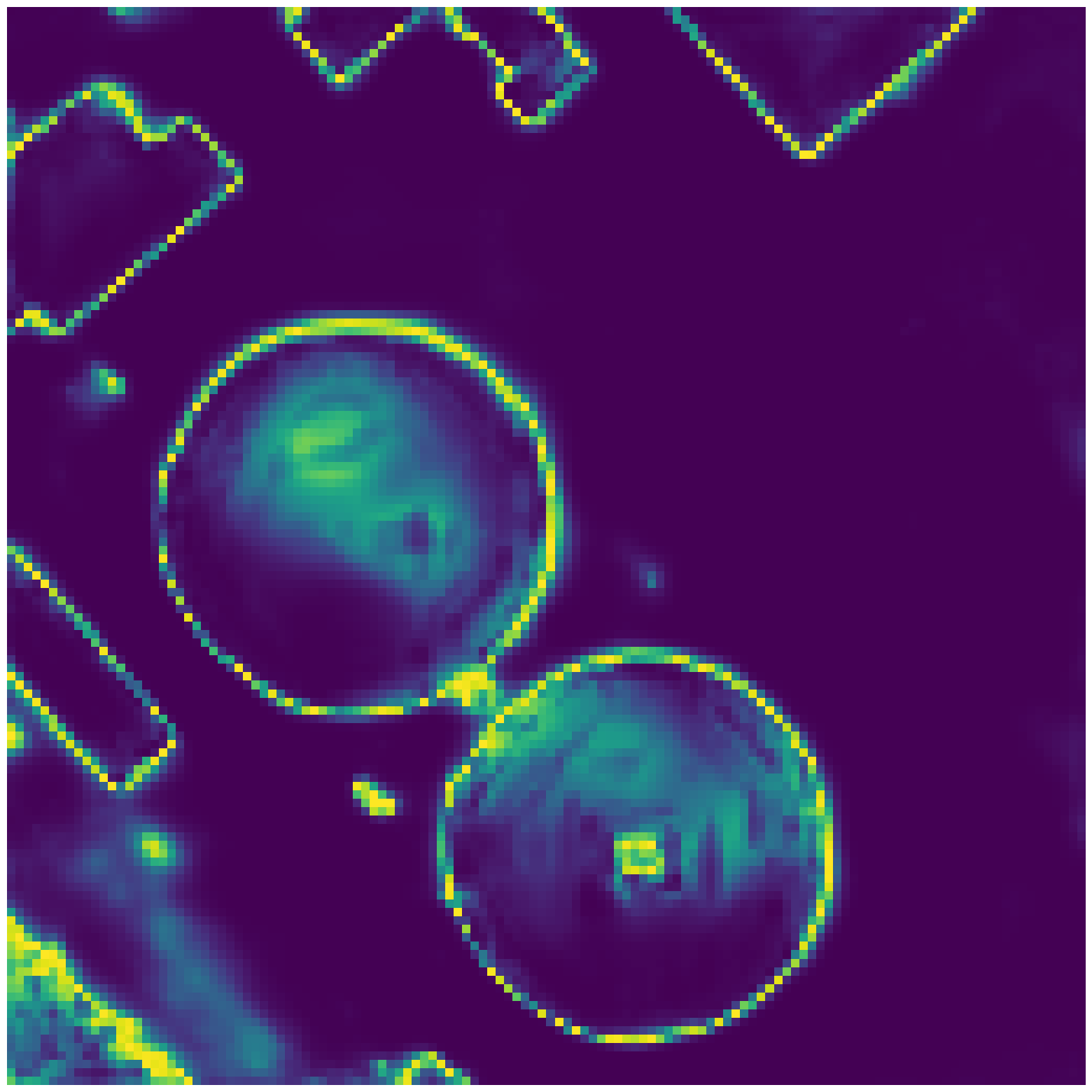} }}
    \subfloat{{\includegraphics[width=2.9cm]{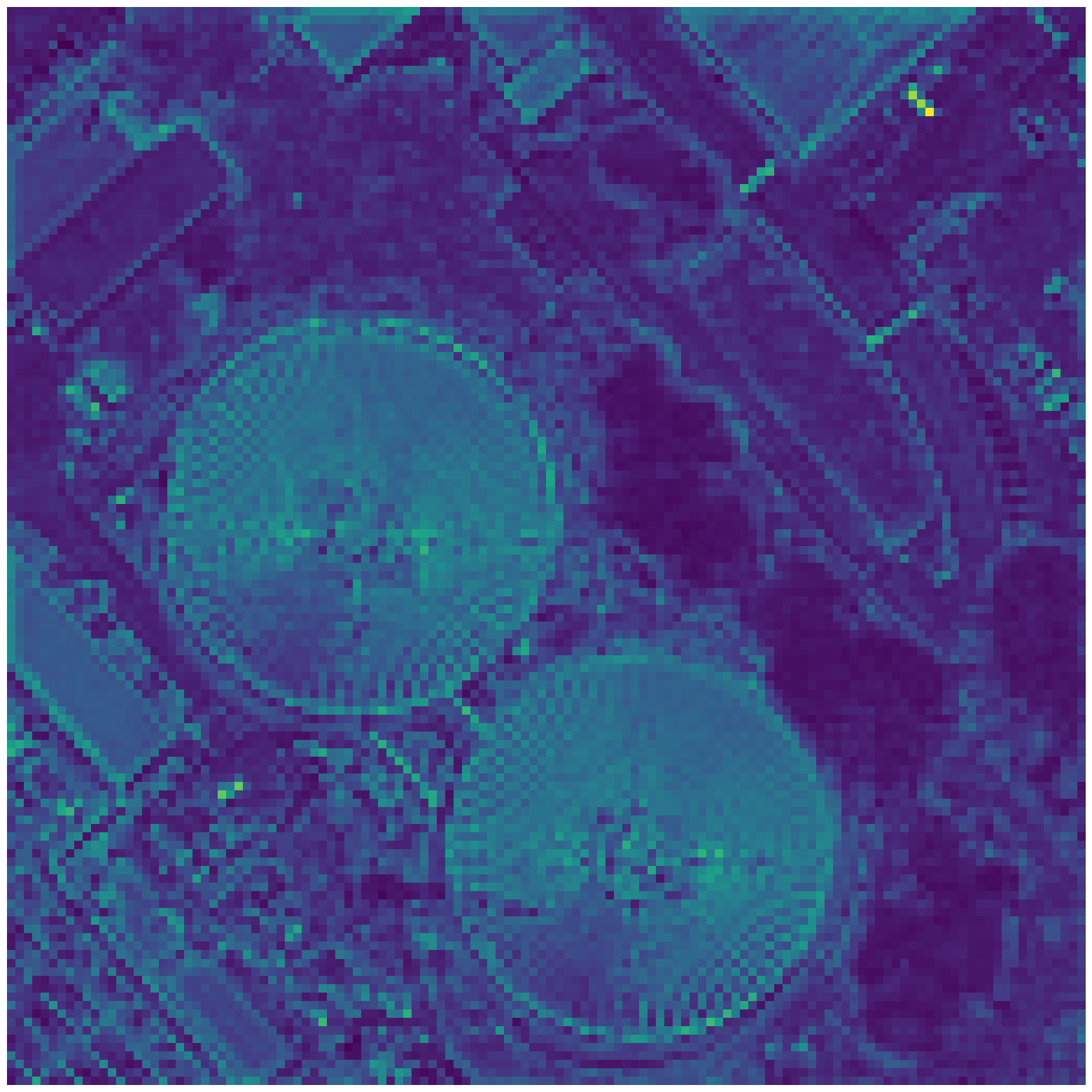} }}
    \subfloat{{\includegraphics[width=2.9cm]{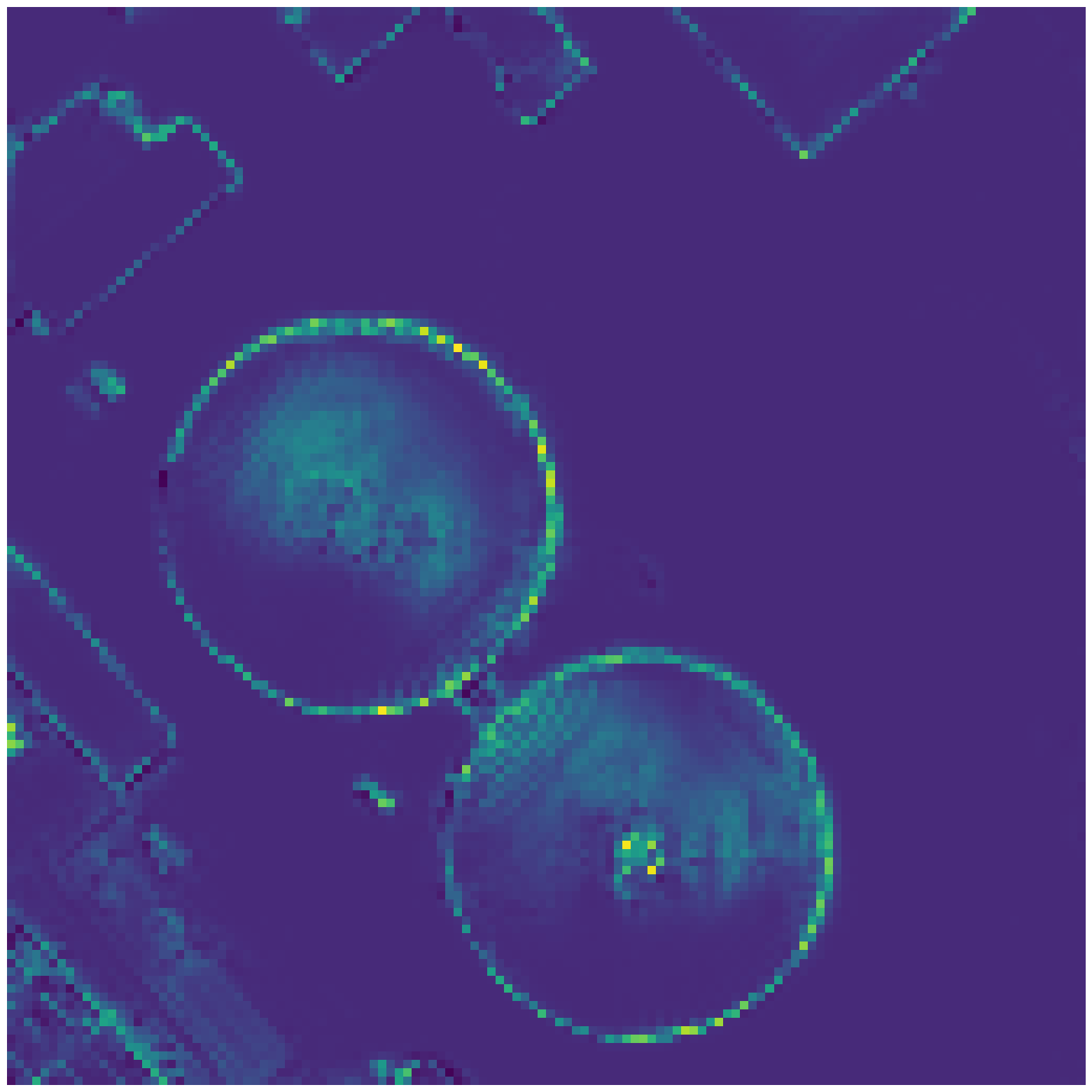}}}
     \subfloat{{\includegraphics[width=2.9cm]{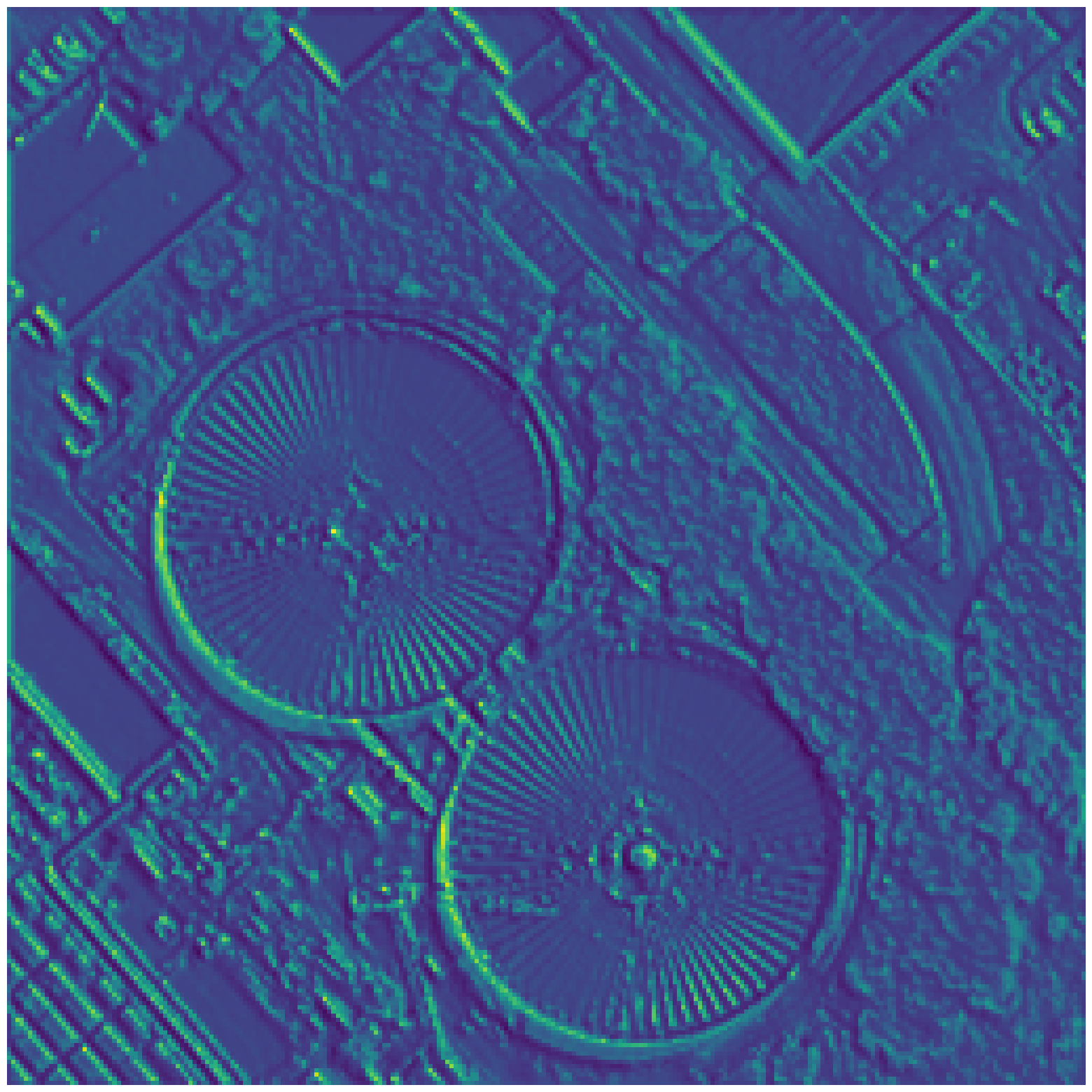} }}
    \subfloat{{\includegraphics[width=2.9cm]{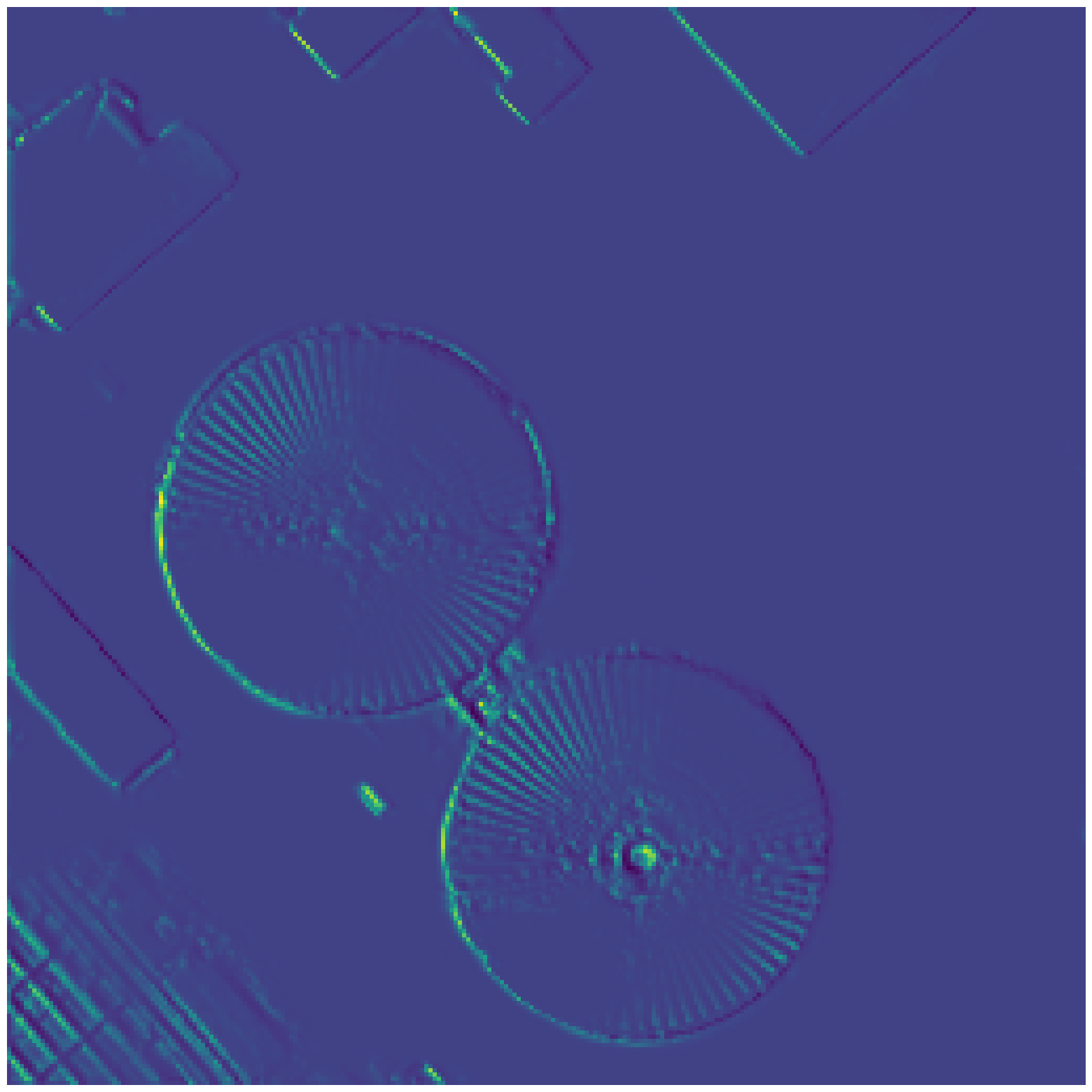}}}
    \hfill
    \subfloat{{\includegraphics[width=2.9cm]{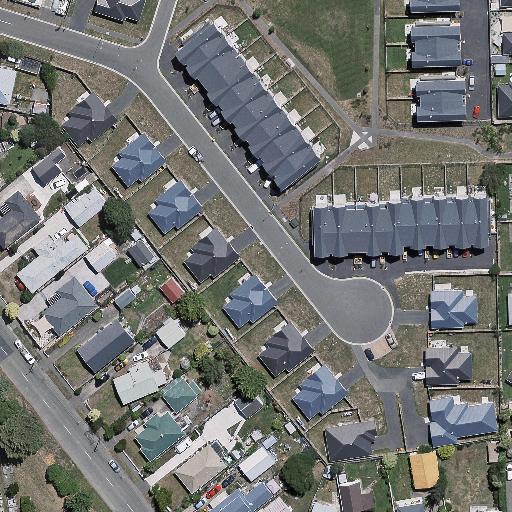} }}
    \subfloat{{\includegraphics[width=2.9cm]{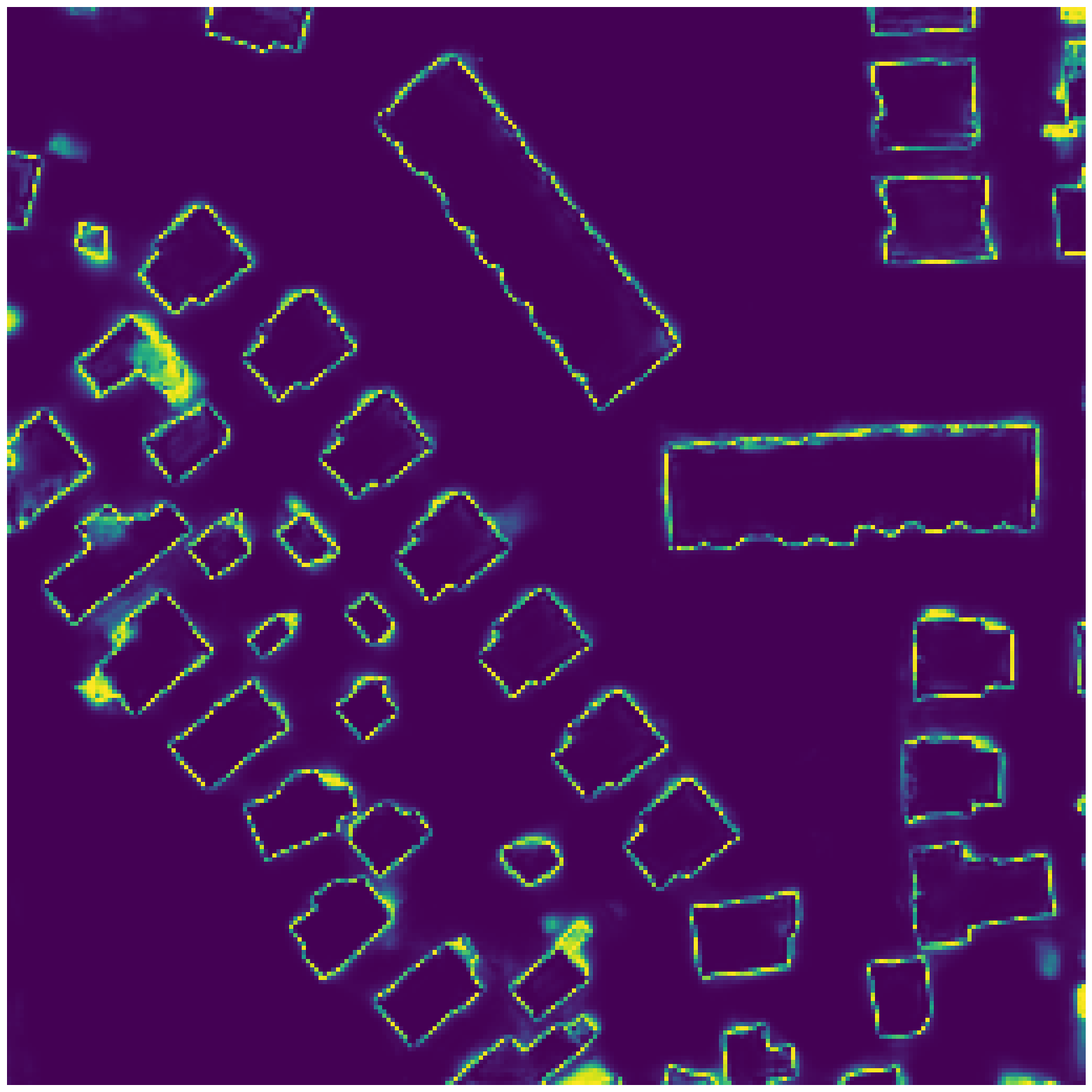} }}
    \subfloat{{\includegraphics[width=2.9cm]{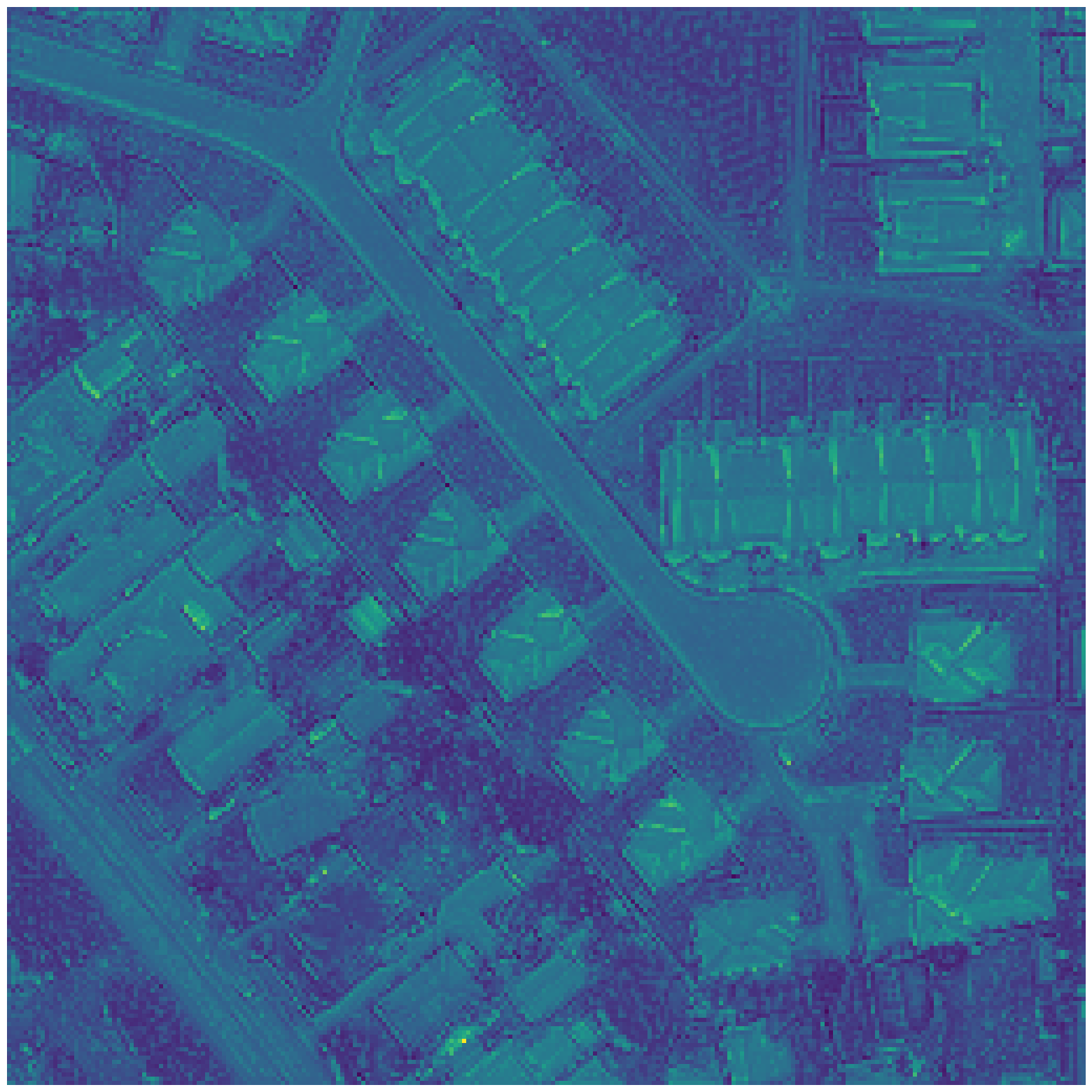} }}
    \subfloat{{\includegraphics[width=2.9cm]{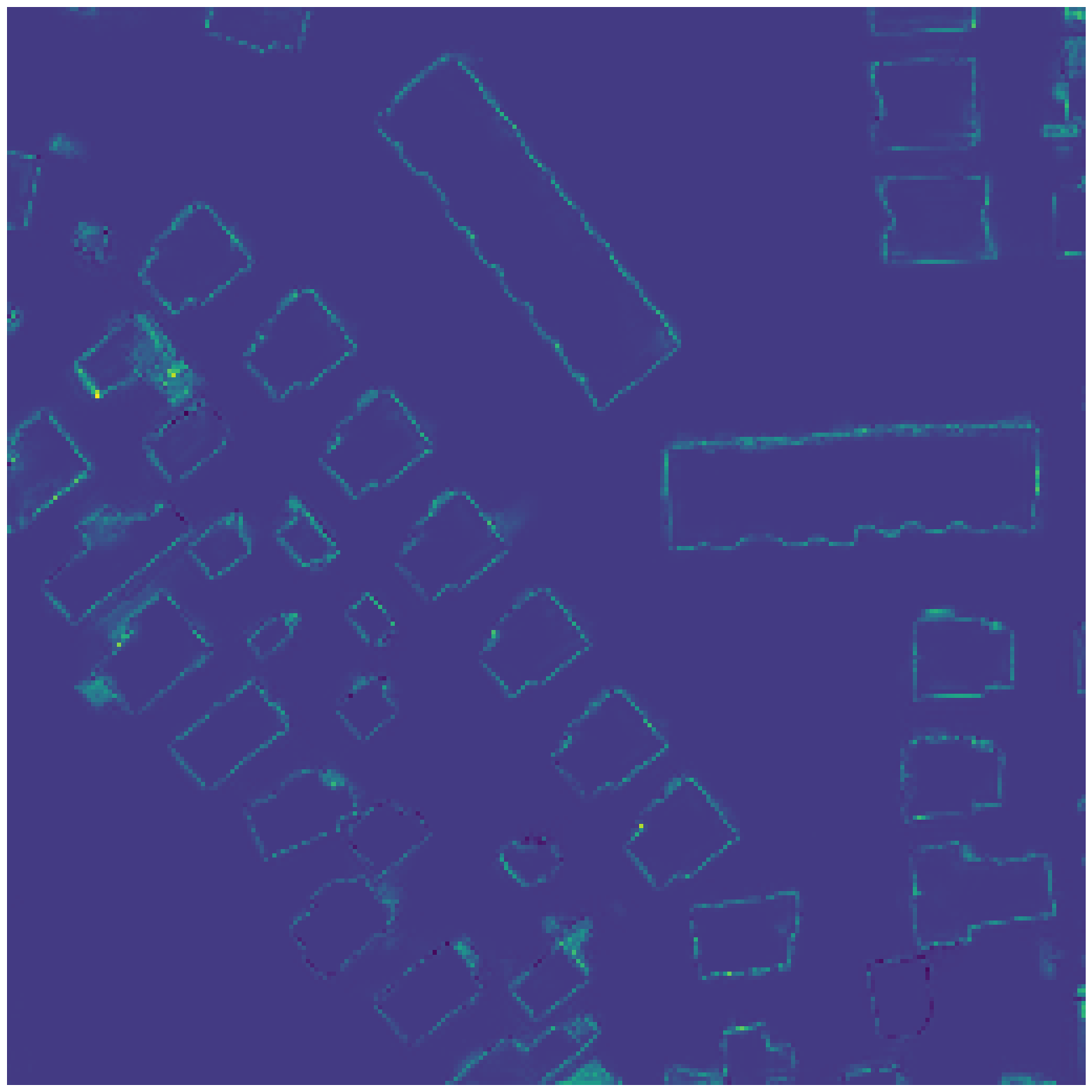}}}
     \subfloat{{\includegraphics[width=2.9cm]{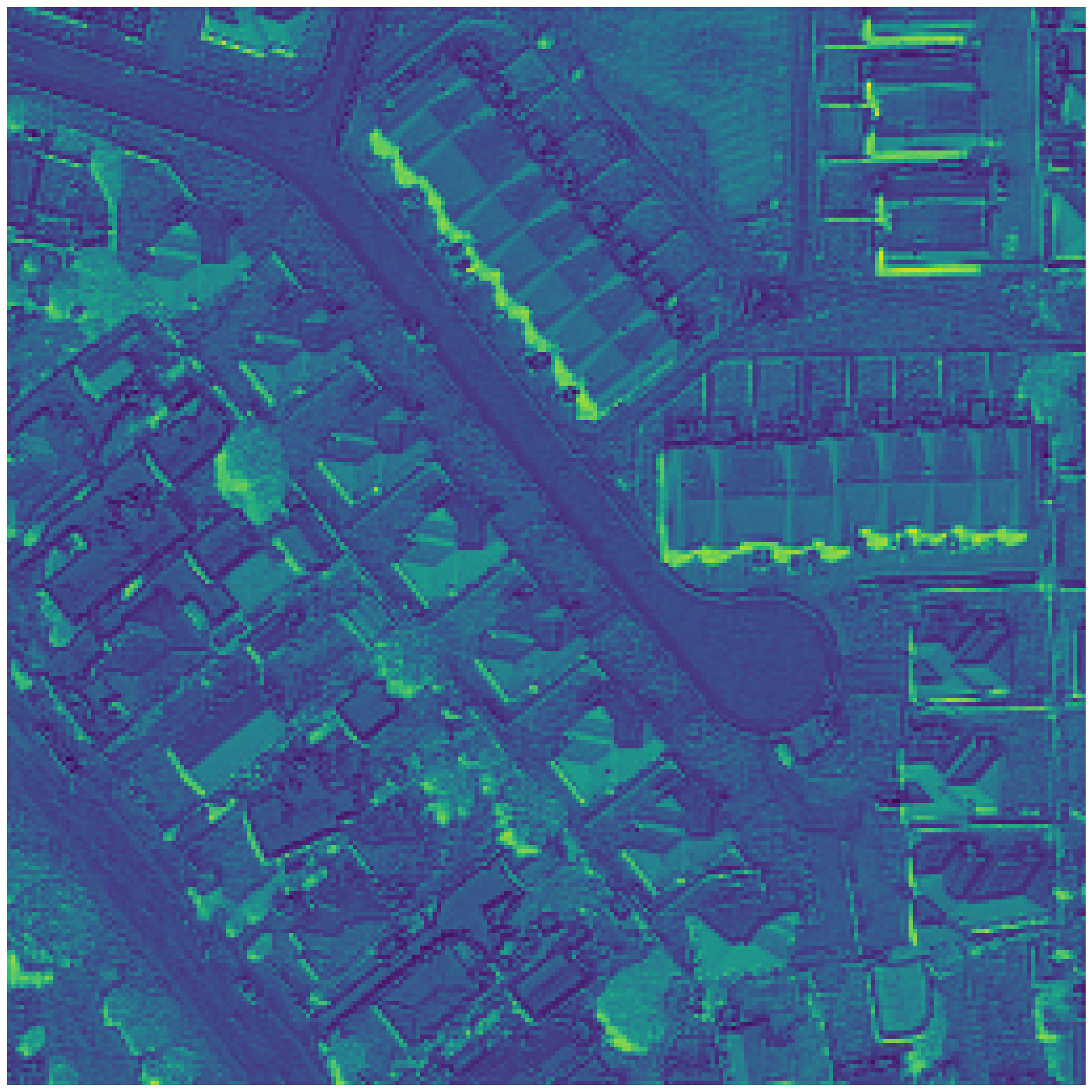} }}
    \subfloat{{\includegraphics[width=2.9cm]{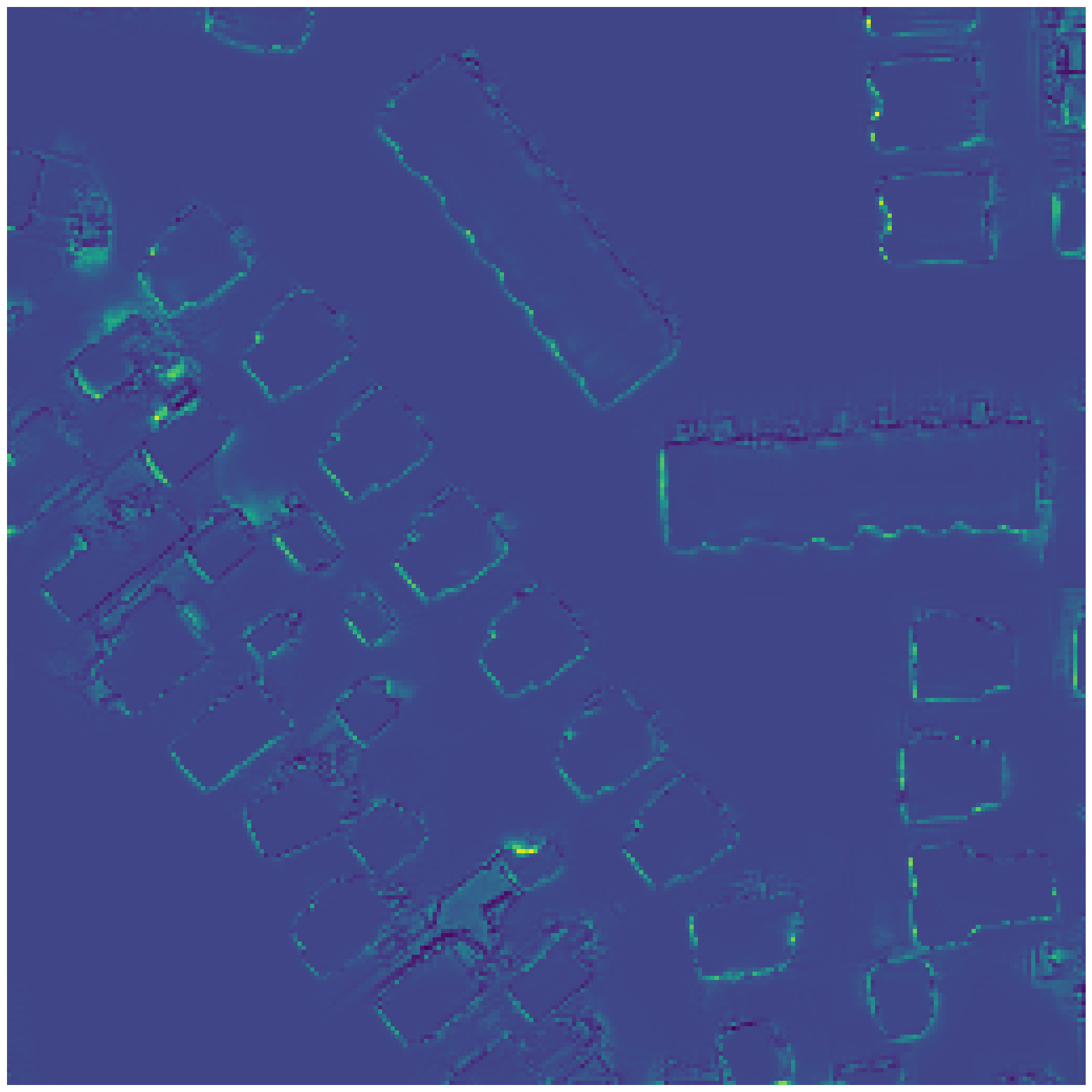}}}
    \hfill
    \subfloat{{\includegraphics[width=2.9cm]{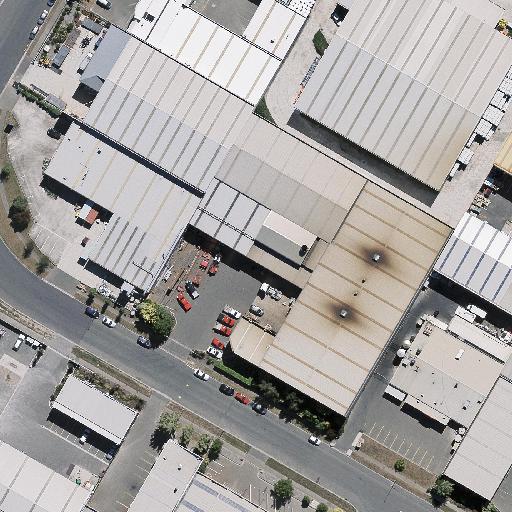} }}
    \subfloat{{\includegraphics[width=2.9cm]{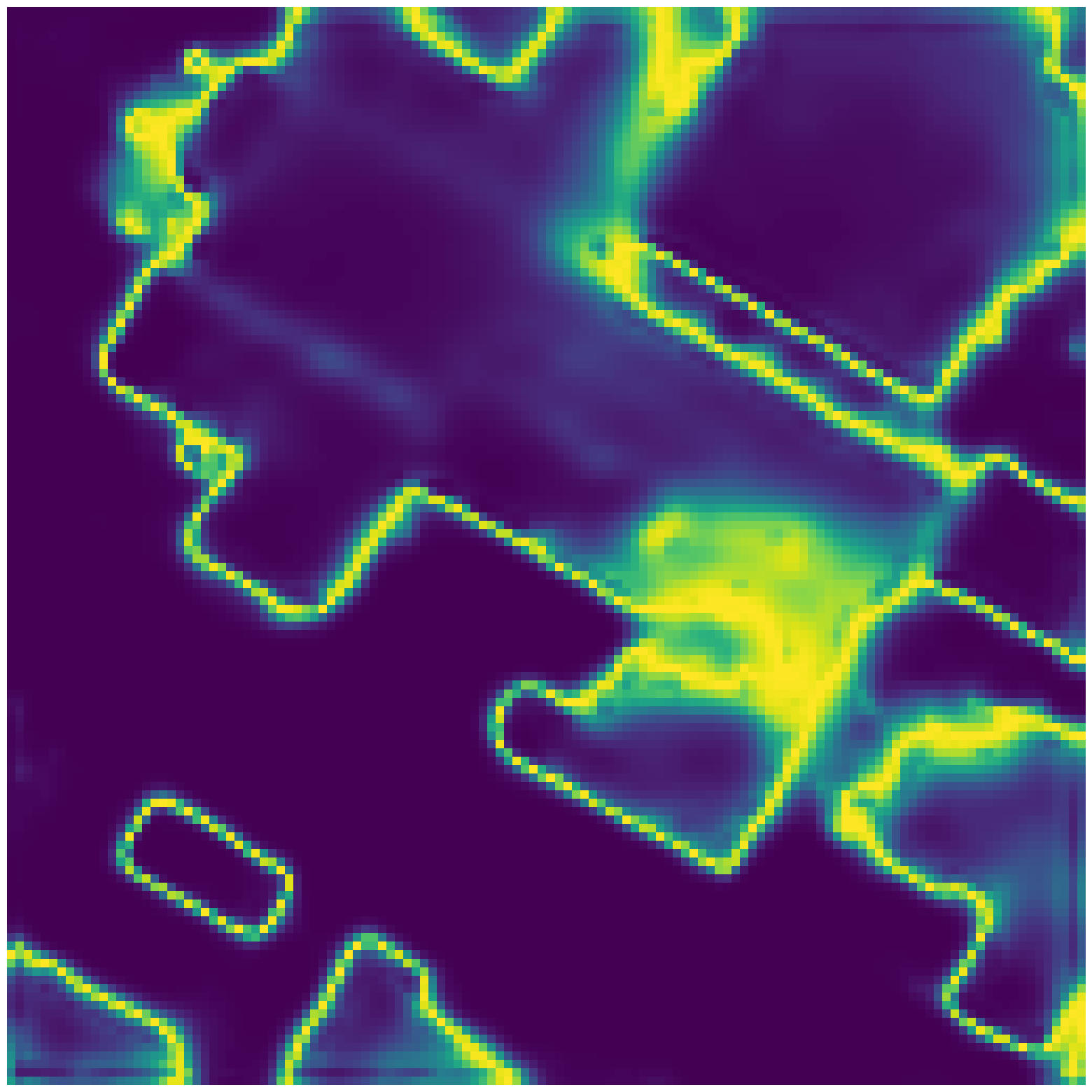} }}
    \subfloat{{\includegraphics[width=2.9cm]{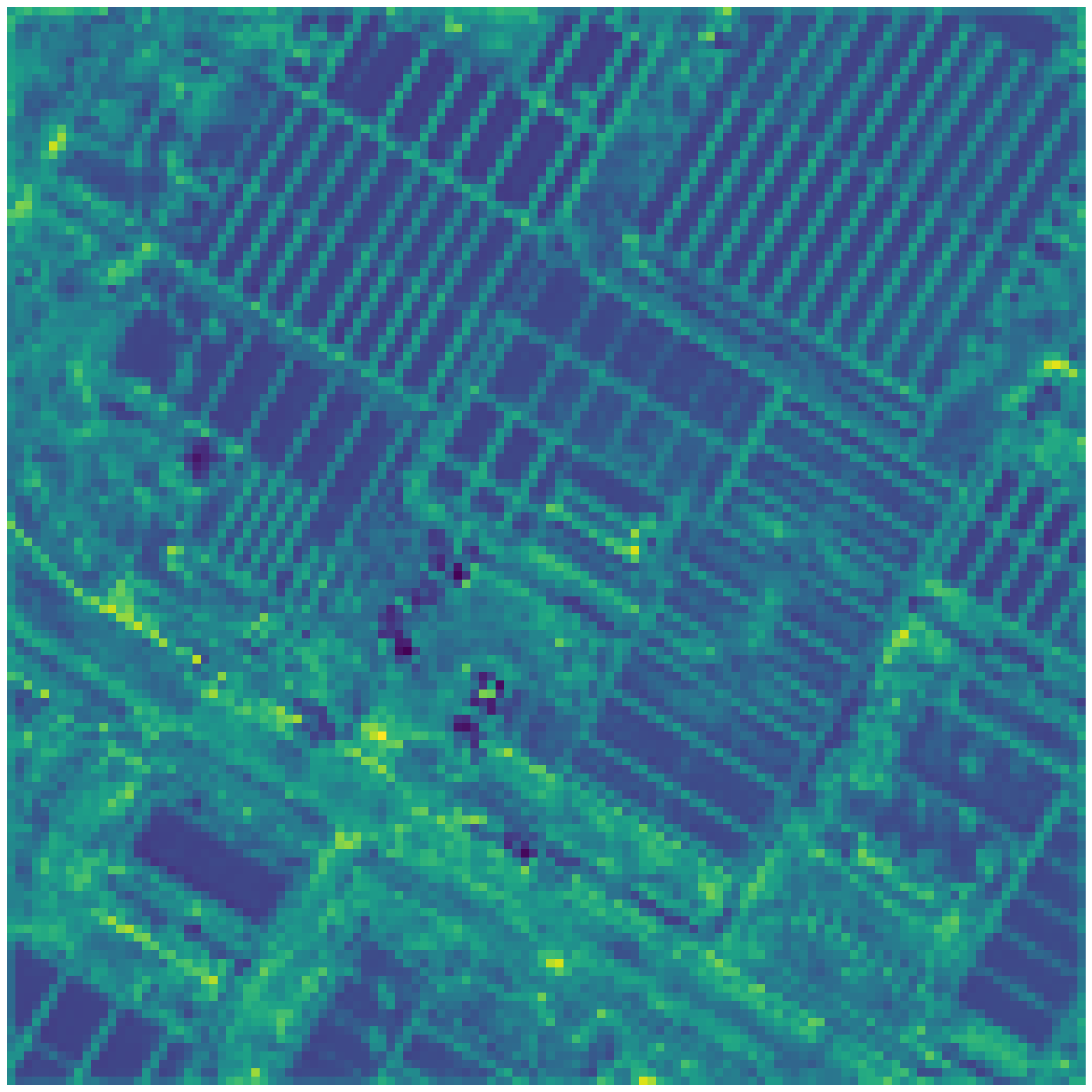} }}
    \subfloat{{\includegraphics[width=2.9cm]{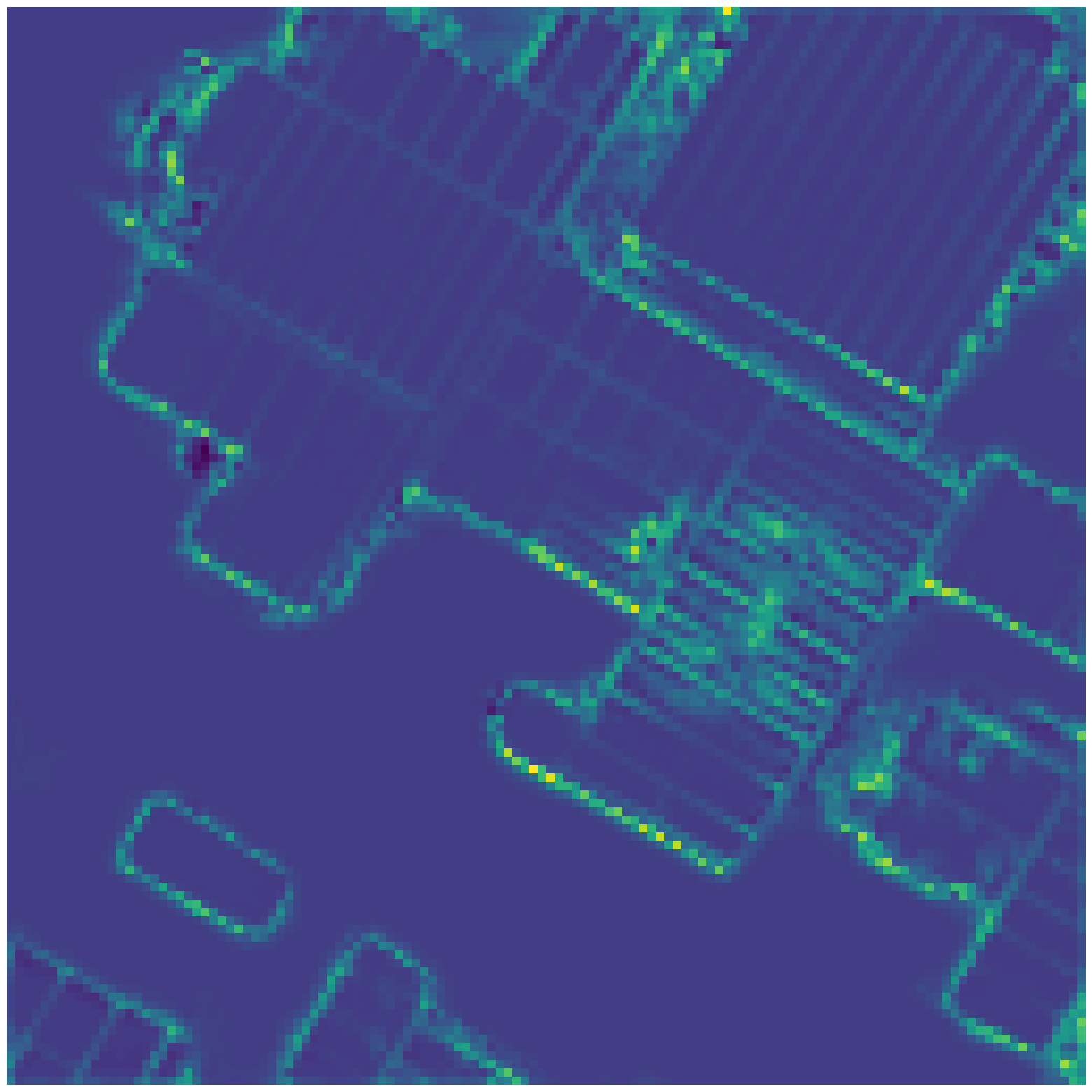}}}
     \subfloat{{\includegraphics[width=2.9cm]{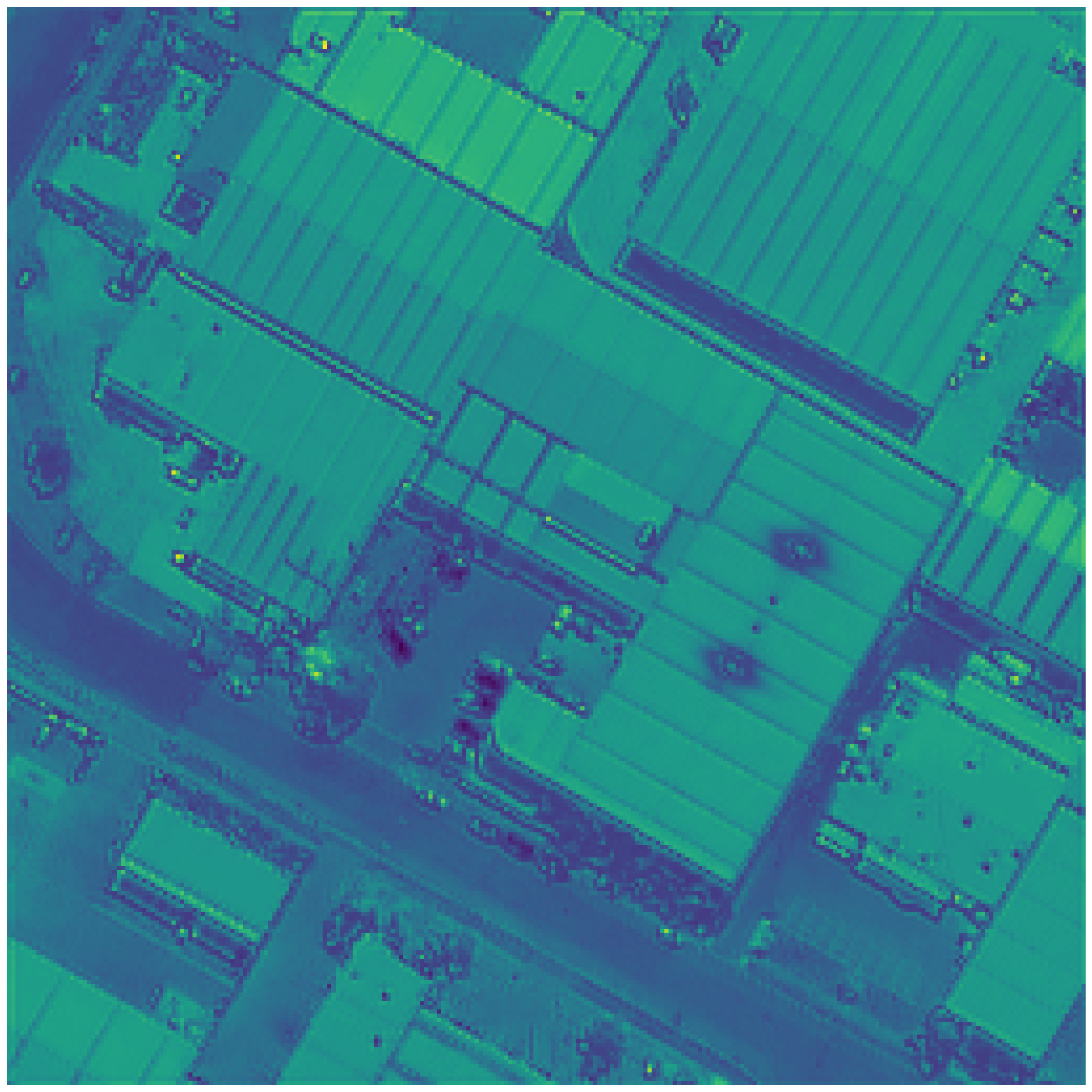} }}
    \subfloat{{\includegraphics[width=2.9cm]{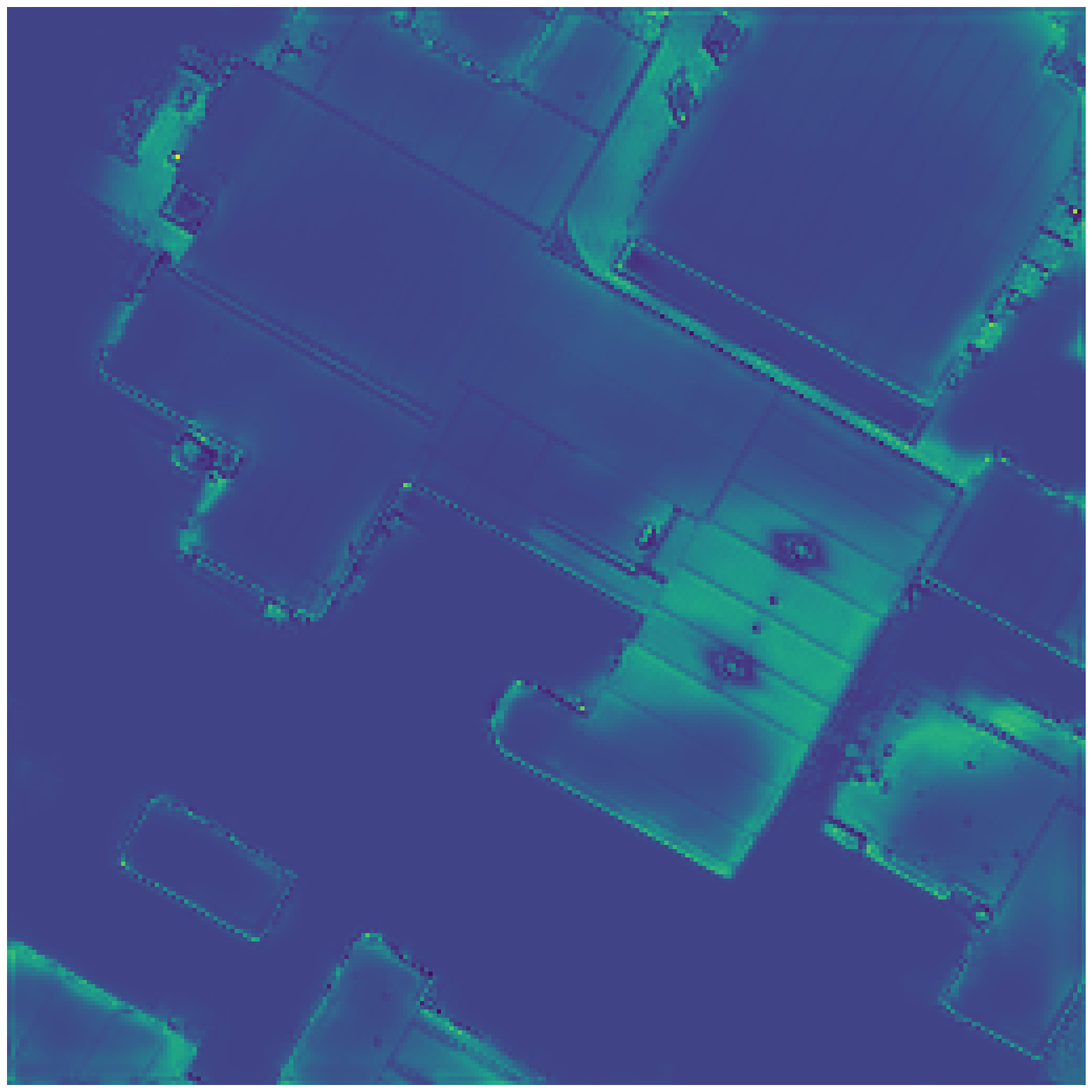}}}
    \hfill
    \caption{\centering Visualization of the encoder feature maps before and after applying uncertainty attention. Column 1: Input image. Column 2: Uncertainty Attention Map. Columns 3, 5: Encoder Features {\em  without} Uncertainty Attention. Column 4, 6: Corresponding Encoder Features {\em with} Uncertainty Attention.}
    \label{fig:ua_feats}
\end{figure*}

To mitigate such over-segmentations, we introduce an
{\em Uncertainty Attention Module} in every
encoder-to-decoder skip connection, as shown by the yellow
boxes in the middle of the `U' in Figure~\ref{fig:gen}.
The purpose of these attention units is to mediate the level
of inclusion for the encoder-generated low-level features
when they are copied over to the decoder side.  More
specifically, we want the Uncertainty Attention Module to
use the low-level detail made available by the encoder only
in those regions of a prediction map where the degree of
uncertainty exceeds a threshold.  Experience with such
architectures tells us that we can expect the uncertainty to
be relatively large in the vicinity of the object boundaries
in the input images, as can be seen in Figure~\ref{fig:ua_feats}.

That raises the question of how to measure the degree of
uncertainty associated with the predictions on the decoder
side.  As it turns out, that's an easy thing to do by
measuring the entropy associated with the building
predictions in the different levels of decoder.  We compute
pixel-wise entropy in a prediction map to produce the
uncertainty attention map at each level of our network as
follows:
\begin{equation}
  E(i) = - p_i \: \log (p_i) - (1 - p_i) \: \log (1-p_i)
\end{equation}

\noindent where $p_i$ denotes the probability of the
$i^{th}$ pixel belonging to the \emph{building class}. This
uncertainty attention map is then element-wise multiplied
with the low-level feature maps in that specific layer to
create an uncertainty--weighted low-level feature map, as
shown in Figure~\ref{fig:uncertainty}.

Recent research \cite{pang2019towards} has shown that
concatenating shallow encoder features with deep decoder
features can adversely affect the predictions if the
semantic gap between the features is large. And, it stands
to reason that introducing uncertainty attention prior to
concatenation has the possibility of amplifying this problem
by injecting ``noisy'' encoder features in those regions of
a building prediction map where the probabilities are low.
We guard against such corruption of the prediction maps by
using deep supervision (shown by thick arrows in Figure~\ref{fig:gen}) that
forces the intermediate feature maps to be discriminative at
all levels of the docoder. Deep supervision \cite{wang2015training, lee2015deeply, xie2015holistically, dou20173d} allows for more
direct backpropagation of loss to the hidden layers of the
network.

\subsection{Critic Network}
\label{sec:critic}

We now present the details regarding the critic network
($\mathcal{C}$) in our framework. The network for
$\mathcal{C}$ is essentially the same as the encoder in
$\mathcal{S}$ \emph{minus the residual blocks}. Our
experiments have shown that adding the residual blocks in
$\mathcal{C}$ increase the parameter space of the model
without any significant improvement in the performance of
the critic. 

\begin{figure}[h]
\centering
\includegraphics[width=0.48\textwidth]{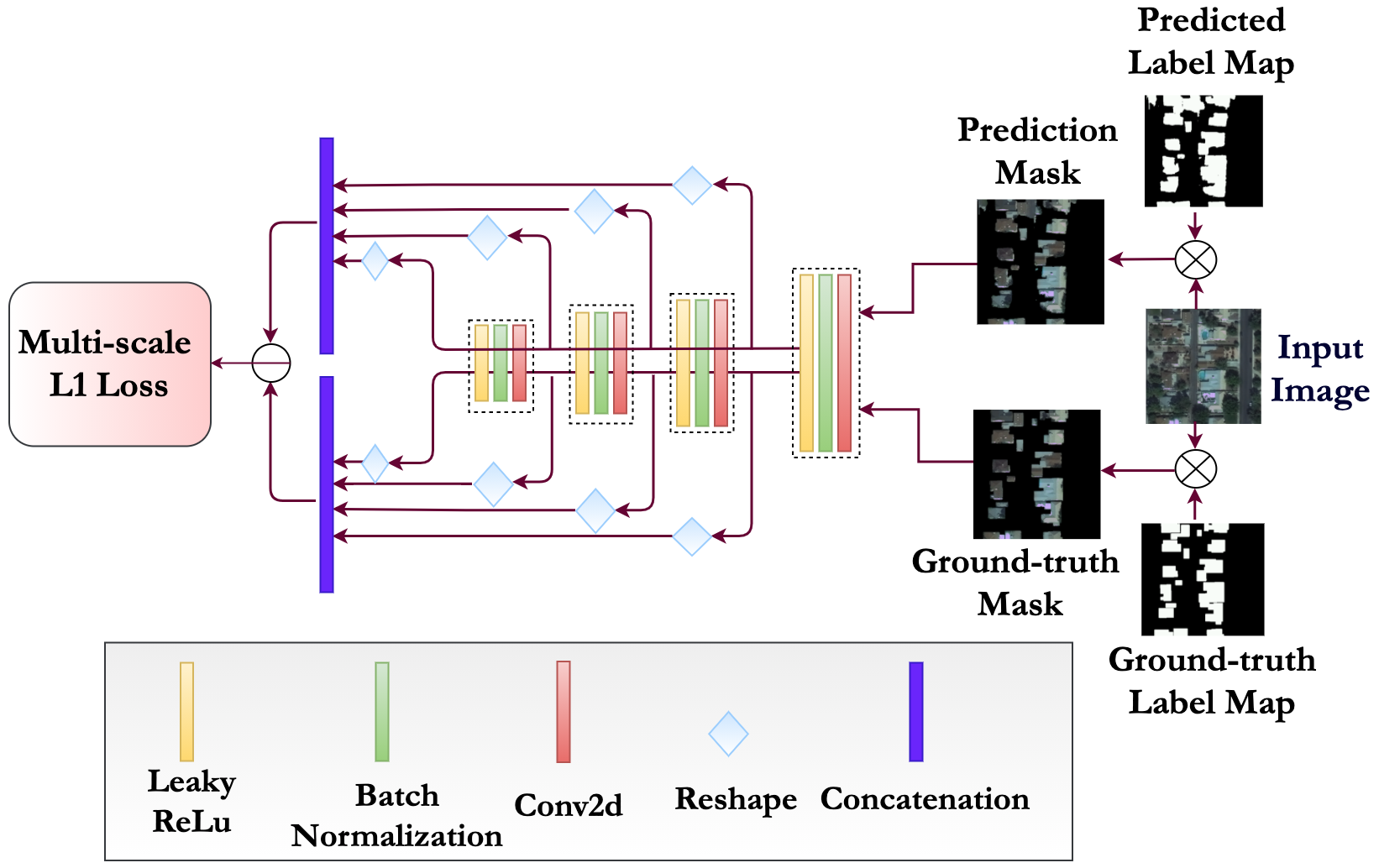}
\caption{\label{fig:critic}Critic Framework}.
\end{figure}

$\mathcal{C}$ is supplied with two inputs: (a) 3-channel
remotely sensed images masked by the corresponding ground-truth
building labels; and (b) 3-channel remotely sensed images masked by
the building labels generated by $\mathcal{S}$. These masks
(predicted and the ground-truth) are created by element-wise
multiplication of the one-channel label maps with the
original RGB images, as shown in Figure~\ref{fig:critic}. $\mathcal{C}$ extracts features from the
predicted mask as well as the ground-truth mask at multiple
scales, reshapes these multi-scale features into
one-dimensional vectors and concatenate them
together. Finally, $\mathcal{C}$ seeks to maximize the
difference between the vectors created from the true
instances and the predicted instances.

\section{Training Strategy}
\label{sec:training}

The generator i.e. the segmentor ($\mathcal{S}$) and the
critic ($\mathcal{C}$) in our proposed architecture are
trained alternatively in an adversarial
fashion. $\mathcal{S}$ tries to predict an accurate label
map for the buildings present in the input image such that
$\mathcal{C}$ cannot distinguish between the predicted map
and the ground-truth map, whereas $\mathcal{C}$ aims to
discriminate the predicted maps from the ground-truth
maps. To train the network in an adversarial fashion, we
calculate the multi-scale ${L}_1$ loss, as explained in
Section \ref{sec:l1loss}, using the hierarchical features
extracted from the multiple layers of $\mathcal{C}$. This
multi-scale ${L}_1$ loss, proposed by Xue et al. in
\cite{xue2018segan}, enables the network to capture the long
and short range spatial relations between the pixels. First,
we train $\mathcal{C}$ keeping the parameters of
$\mathcal{S}$ fixed and try to minimize the negative of
${L}_1$ loss. Next, we keep the parameters of $\mathcal{C}$
fixed and train $\mathcal{S}$ minimizing the same ${L}_1$
loss. Moreover, we incorporate extra supervision in the form
of weighted dice and shape losses to stabilize the training
of $\mathcal{S}$ and boost its performance.

\subsection{Adversarial Loss: Multi-scale ${L}_1$ Loss}
\label{sec:l1loss}
We define our adversarial loss function ${L}_1$ as:
\begin{multline}
    \label{eq:l1loss}
     {L}_1 = 
    \frac{1}{N} \sum_{i=1}^{N} l_{mae} (f_C(x_i \circ S(x_i)), f_C(x_i \circ y_i))
\end{multline}
where $N$ is the batch size and $x_i$ is the $i^{th}$ image
in a batch.  The notation $S(x_i)$ stands for the output
label map of $\mathcal{S}$, and $y_i$ is the corresponding
ground-truth label map. The notation $x_i \circ S(x_i)$
stands for the original input sample masked by predicted map
and $x_i \circ y_i$ is the input image masked by the
ground-truth label map. 
%That brings us to the explanation of
%$f_C(x)$ and $l_{mae}$ in the $L_1$ loss.
The notation $f_C(x)$ stands for the features extracted from
the image $x$ in multiple layers of $\mathcal{C}$ and
$l_{mae}$ stands for the Mean Absolute Error (MAE) defined as:
\begin{equation}
\label{eq:l11loss}
    l_{mae} (f_C(x), f_C(y))  = \frac{1}{L} \sum_{k=1}^{L}\| (f^{k}_C(x) - f^{k}_C(y))\|_1
\end{equation}
where $f^{k}_C(x)$ is the feature map extracted from the
image $x$ at the $k^{th}$ layer of $\mathcal{C}$, the
subscript $mae$ stands for ``mean absolute error'', `$L$' is
the number of layers in $\mathcal{C}$, and $\|.\|_1$
represents $\ell1$ norm.

\subsection{Joint Dice and Shape Loss}
\label{sec:joint}

The overall loss function used also includes dice and shape
losses for stabilizing the training of $\mathcal{S}$ and for
boosting its performance.  We have observed that only using
adversarial loss leads to unstable training of the GAN.  The
\emph{dice} part of the loss, shown below in
Eq. (\ref{eq:dice}), optimizes the dice similarity
coefficient (DSC) and the {\em shape} part of the same,
shown in Eq. (\ref{eq:shape}), minimizes the Hausdorff
Distance (HD) \cite{karimi2019reducing} between the ground-truth and prediction.

Here is the formula used for the dice loss:
%\small
\begin{equation}
\label{eq:dice}
\begin{aligned}
    L_{dice} = &1 - \left[\alpha_1  \frac{2*\sum_{i}^N p_i g_i}{\sum_{i}^N p_i^2 + \sum_{i}^N g_i^2} \right. \\ & \left . + \alpha_2  \frac{2*\sum_{i}^N (1-p_i) (1-g_i)}{\sum_{i}^N (1-p_i)^2 + \sum_{i}^N (1-g_i)^2} \right]
\end{aligned}
\end{equation}

where $\alpha_1 + \alpha_2 = 1$. $\alpha_1$, $\alpha_2 \geq 0$. $p_i, g_i$ represent, respectively, the $i^{th}$ pixel of the ground-truth and the prediction map. This way, in addition to the contribution from the positive samples, we also ensure contribution from
the negative samples. This becomes particularly useful if an
entire sample is composed of only foreground or only
background class. In our experiments, we set $\alpha_1 =
0.8$.

Regarding the shape loss, it helps the system
keep a check on the shape similarity between the
ground-truth and predicted building labels by minimizing the
HD distance between them. Hausdorff Distance loss aims
to estimate HD from the CNN output probability so as to
learn to reduce HD directly. Specifically, HD can be
estimated by the distance transform of ground-truth and
segmentation. We compute the average shape loss as follows -
\begin{equation}
\label{eq:shape}
L_{HD} = \frac{1}{N} \sum_{i=1}^N \left[(p_i - g_i)^2 (d_{p_i}^2 + d_{g_i}^2)\right] 
\end{equation}
where $d_{p_i}$ and $d_{g_i}$ are the taxicab (i.e. $\ell
1$) distance transforms of the ground-truth and predicted
label maps.

\section{Datasets and Evaluation Metrics}
\label{sec:data}
In this paper, we show results on four publicly available
datasets - Massachusetts Buildings (MB) Dataset \cite{MnihThesis}, INRIA Aerial Image Labeling Dataset
\cite{maggiori2017dataset}, WHU Building Dataset \cite{ji2018fully} and DeepGlobe Building Detection Dataset \cite{van2018spacenet, DeepGlobe18}. These datasets cover different
regions of interest across the world and include diverse
building characteristics. We have used different evaluation
metrics for different datasets in order to carry out a fair
comparison with the other state-of-the-art methods.

\subsection{Massachusetts Buildings Dataset}
\label{sec:massdata}

The Massachusetts Buildings (MB) Dataset \cite{MnihThesis}
consists of 151 high-resolution aerial images of urban and
suburban areas around Boston. Each image is $1500 \times 1500$ 
pixels and covers an area of $2250 \times 2250
m^2$. The dataset is randomly divided into training (137
tiles), validation (4 tiles), and testing (10 tiles)
subsets.

We now elaborate on the metrics that we have used for comparisons. For the Massachusetts Buildings Dataset, we report \textbf{relaxed as well as non-relaxed (i.e. regular) versions of F1-score and IoU score}. We use the \textbf{relaxed version of precision, recall, and F1-score} to calculate the precision-recall breakeven point as in \cite{MnihThesis}. A relaxation factor of $\rho$ was introduced to consider a
building prediction correct if it falls within a radius of
$\rho$ pixels of any ground-truth building pixel. This
relaxation factor is used to provide a realistic performance
measure because the building masks in the Massachusetts Buildings Dataset are not perfectly aligned to the actual
buildings in the images. The formula for the F1-measure is:
\begin{equation}
\label{eq:f1}
    F1 = 2\times \frac{precision \times recall}{precision + recall}
\end{equation}
where
\begin{equation}
\label{eq:precision}
    precision = \frac{tp}{tp + fp}
\end{equation}
\begin{equation}
\label{eq:recall}
    recall = \frac{tp}{tp + fn}
\end{equation}
The relaxed version of precision denotes the fraction of
predicted building pixels that are within a radius of $\rho$
pixels of a ground-truth building pixel, and the relaxed
version of recall represents the fraction of the ground-truth
building pixels that are within a radius of $\rho$ pixels of
a predicted building pixel. To conduct a fair comparision
with previous research \cite{pan2019building, khalel2018automatic}, we set $\rho = 3$.

\subsection{INRIA Aerial Image Labeling Dataset}
\label{sec:inriadata}
This dataset \cite{maggiori2017dataset} features aerial
orthorectified color imagery having a spatial resolution of
0.3m with a coverage of $810 km^2$ and contains publicly
available ground-truth labels for the building footprints in
the training and validation subsets. The images range from densly populated
areas like San Francisco to sparsely populated areas in the
alpine regions of Austria. Thus, the dataset represents
highly contrasting terrains and landforms. {\em Moreover,
  the population centers in the training subset are
  different from those in the testing subset, which makes
  the dataset very appropriate for assessing a network's
  generalization capability.}

The training set contains 180 color image tiles of size
$5000 \times 5000$, covering a surface of $1500 \times 1500
m^2$ each (at a 0.30m resolution). There are 36 tiles for
each of the following regions: Austin, Chicago, Kitsap
County, Western Tyrol and Vienna. Each tile has a
correspinding one-channel label image indicating buildings
(255) and the not-building class. The test set also contains
180 tiles but from different areas: Bellingham (WA),
Bloomington (IN), Innsbruck, San Francisco and Eastern
Tyrol. 

The performance measures used for this dataset are:
\textbf{(a) Intersection over Union (IoU)}: number of pixels
labeled as building in both the prediction and the ground
truth, divided by the number of pixels labeled as pixel in
the prediction or the ground-truth, and, \textbf{(b)
  Accuracy (acc)}: percentage of correctly classified
pixels. The metrics are defined as:
\begin{equation}
\label{eq:iou}
    IoU = \frac{tp}{tp+fp+fn}
\end{equation}
\begin{equation}
\label{eq:acc}
    acc = \frac{tp+tn}{tp+tn+fp+fn}
\end{equation}
where $tp$, $tn$, $fp$ and $fn$ represent the true
positives, true negatives, false positives and false
negatives respectively.

\subsection{WHU Aerial Building Dataset}
\label{sec:whudata}

\begin{figure}[ht]
\centering
\includegraphics[width=0.48\textwidth]{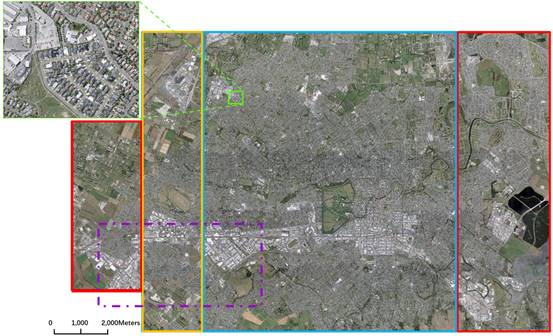}
\caption{\label{fig:whudata} The WHU Aerial Building Dataset in Christchurch, New Zealand. The boxes in blue, yellow and red represent the areas used for creating the training, validation and test sets, respectively.}.
\end{figure}

The WHU Aerial Buiding Dataset \cite{ji2018fully} covers an area of 450 $km^2$ around Christchurch, New Zealand (Figure~\ref{fig:whudata}) and consists more than 187,000 buildings. The original dataset having a ground resolution of 0.075m comes from the New Zealand Land Information services website. Ji et al. \cite{ji2018fully} has downsampled the images to 0.3m resolution and cropped them into 8189 non-overlapping tiles with $512\times512$ pixels. The dataset is divided into three parts --- 4,736 tiles (130,500 buildings) for training, 1,036 tiles (14,500 buildings) for validation and 2,416 tiles (42,000 buildings) for testing. In this paper, we have used the following metrics for evaluating the performance of our proposed method on this dataset -- IoU (Eq.\ref{eq:iou}), Precision (Eq.~\ref{eq:precision}), Recall (Eq.~\ref{eq:recall}) and F1-score (Eq.~\ref{eq:f1}). 

\subsection{DeepGlobe Building Dataset}
\label{sec:spacenetdata}
The DeepGlobe Building Dataset \cite{DeepGlobe18} uses the SpaceNet Building Detection Dataset \cite{van2018spacenet} (Challenge 2 of the SpaceNet Series). This dataset has been used for the DeepGlobe 2018 Satellite Image Understanding Challenge organised as a part of CVPR 2018 Workshops.

The DeepGlobe Dataset for building detection consists of Digital Globe's WorldView-3 satellite images with 30 cm resolution. The dataset covers 4 different areas of interest (AOIs)  with very different landscapes -- Vegas, Paris, Shanghai and Khartoum. The training set has 3851 images for Vegas, 1148 images for Paris, 4582 images for Shanghai and 1012 images for Khartoum. In the test set, there are 1282, 381, 1528 and 336 images for Vegas, Paris, Shanghai and Khartoum respectively. Each image is of size $650 \times 650$ pixels and covers $200 \times 200$ $m^2$ area on the ground. Each region consists of
high-resolution RGB, panchromatic, and 8-channel lower resolution multi-spectral images. In our experiments, we use pansharpened RGB images. Each image comes with its corresponding geojson file with list of polygons as building instances. 

The dataset provides its own evaluation tool to compute F1-score as a performance measure. The F1-score is based on individual building object prediction. Each proposed building is a geospatially defined polygon label representing the footprint of the building. The proposed footprint is considered a ``true positive'' if the intersection over union (IoU) between the  proposed and the ground-truth label is at least 0.5. For each labeled polygon, there can at most one “true positive”. The number of true positives and false positives are counted for all the test images, and the F1-score is computed from this aggregated count.

\section{Experimental Settings and Data Preparation}
\label{sec:exp}
Our entire segmentation pipeline involves the following steps -- image preparation, training our GAN based segmentation model using the training and validation datasets, and, finally applying our trained model to predict building masks for the test images. In this paper, we have shown results on 4 different datasets. Due to the diverse characteristics of the datasets and for performing a fair comparison of our algorithm with other state-of-art methods on those datasets, we preprocess our data differently for each dataset. In this section, we first describe our experimental setup. Then, we give detailed explanation of the data processing strategies that we use for each dataset during training and inference.

\subsection{Experimental Setup}
We have trained our network on four Nvidia GeForce GTX 1080
Ti (11GB) GPUs with images of size $400 \times 400$ and
batch size of 32. We used the Adam stochastic optimizer with
an initial learning rate of 0.0005 and a momentum of 0.9.  A
poly-iter learning rate \cite{mishra2019polynomial} with a
$power$ of 0.9 was used for 200 epochs. The poly-iter
learning rate is calculated as -
\begin{equation}
    lr = lr_0 * \left(1 - \frac{i}{T_i}\right)^ {power}
\end{equation}
where $lr$ is the learning rate in the $i^{th}$ iteration,
$lr_0$ is the initial learning rate and $T_i$ is the total
numbr of iterations. To avoid overfitting, an L$_2$
regularization was applied with a weight decay of 0.0002.

\subsection{Data Augmentation}

During training and inference, we carry out different data
augmentation strategies on all four datasets. During
training, we perform the following data augmentations --
random horizontal flips, random vertical flips, random
rotations, and color jitter.

To improve predictive performance of our algorithm, we apply
a data augmentation technique during inference -- popularly
known as Test Time Augmentation (TTA). Specifically, it
creates multiple augmented copies of each image in the test
set, the model then makes a prediction for each;
subsequently, it returns an ensemble of those
predictions. We perform 5 different transformations on each
test image -- flipping the image horizontally and
vertically, and rotating the image by $90^\circ$,
$180^\circ$ and $270^\circ$. This means we obtain 6
predictions for each image patch. We align these 6
predictions by applying appropriate inverse transformation,
and produce the final prediction for each patch by averaging
these predictions.

\subsection{Creating Training, Test and Validation Datsets} 
\label{sec:datacreate}
The WHU and Massachusetts datasets provide training,
validation and testing subsets.

The DeepGlobe dataset provides training and test subsets. We
randomly divide the training set into 80/20 ratio with 80\%
images in the training dataset and 20\% images in the
validation dataset. This 80/20 subsets are formed such that
the ratios of number of images in each of the 4 AOIs is
maintained in the training and validation sets.

For the INRIA dataset, we take a different approach for
creating the training, validation and test subsets. This
dataset also provides training and testing subsets; however,
the regions covered in the training and testing subsets are
different. The regions in the training subset includes
Austin, Chicago, Kitsap, Vienna and West Tyrol; whereas, the
test subset consists of image patches from Bellingham,
Bloomington, Innsbruck, San Francisco and East Tyrol. It is
evident that this dataset is created with the purpose of
investigating how transferable models trained on one set of
cities to another set of cities are; to fulfill the same
purpose and make our model generalizable to any city in the
world, we adopt a k-fold validation technique for training
our model, and accordingly, we generate our train, test and
validation subsets.

Following the suggestion of the authors of the INRIA dataset
paper \cite{maggiori2017dataset}, we create a dataset of 25
images by taking out the first five tiles of each city from
the training set (e.g., Austin{1-5}). In the original
dataset paper \cite{maggiori2017dataset}, these 25 images
serve as the validation dataset. So, throughout this paper,
we have referred to these 25 images as {\em INRIA Validation
  Dataset}. However, most of the state-of-the-art papers
have regarded these 25 images as the testing subset and
shown inference results on these images. In our paper, we
report the performance of our algorithm on the INRIA
Validation Dataset (Table~\ref{tab:results2}) as well as on
the actual test dataset (Table~\ref{tab:inriatest}).

The rest of the training data now consists of a total of 155
images with 31 images from each region. We split these
images into 5 folds, one for each region. We train an
ensemble of 5 models - each model being trained on 4 regions
and validated on the $5^{th}$ region. Finally, we use an
ensemble of 5 models to do prediction on the test images in
the INRIA dataset. We compute the integral prediction for an
input patch by averaging predictions for each of the models
in the ensemble.

\begin{figure*}[b]
    \centering
    \subfloat{{\includegraphics[width=5.6cm]{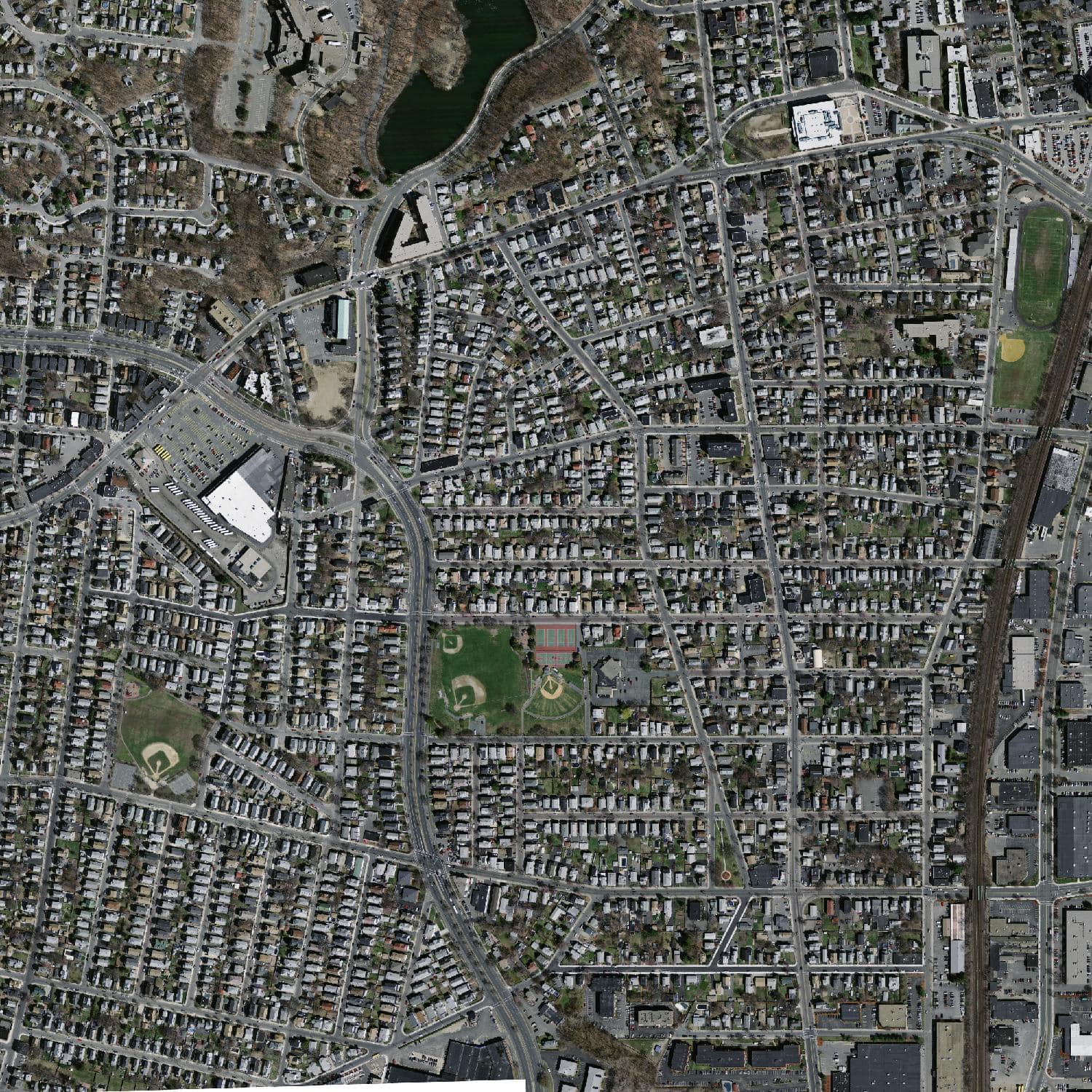} }}
    \subfloat{{\includegraphics[width=5.6cm]{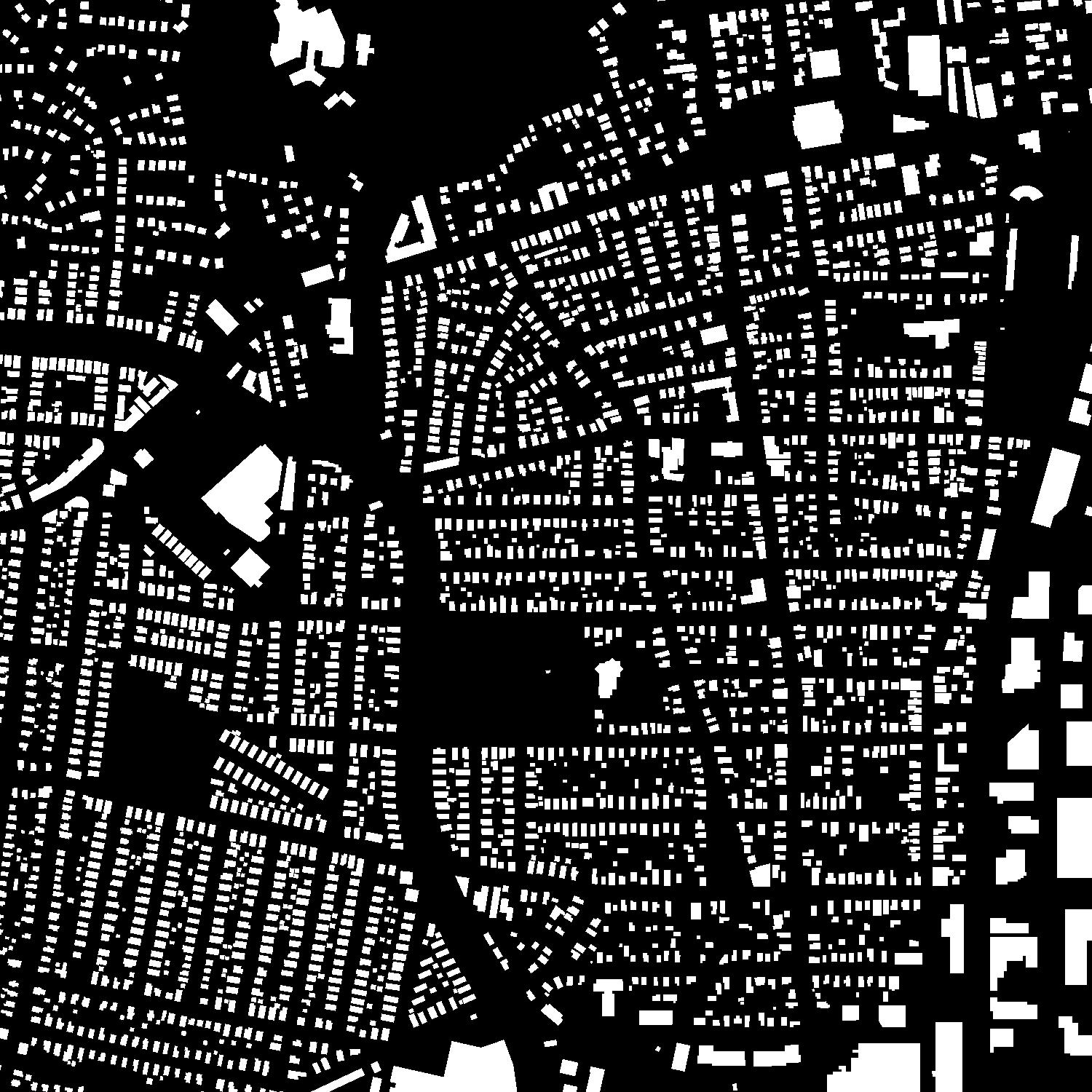} }}
    \subfloat{{\includegraphics[width=5.6cm]{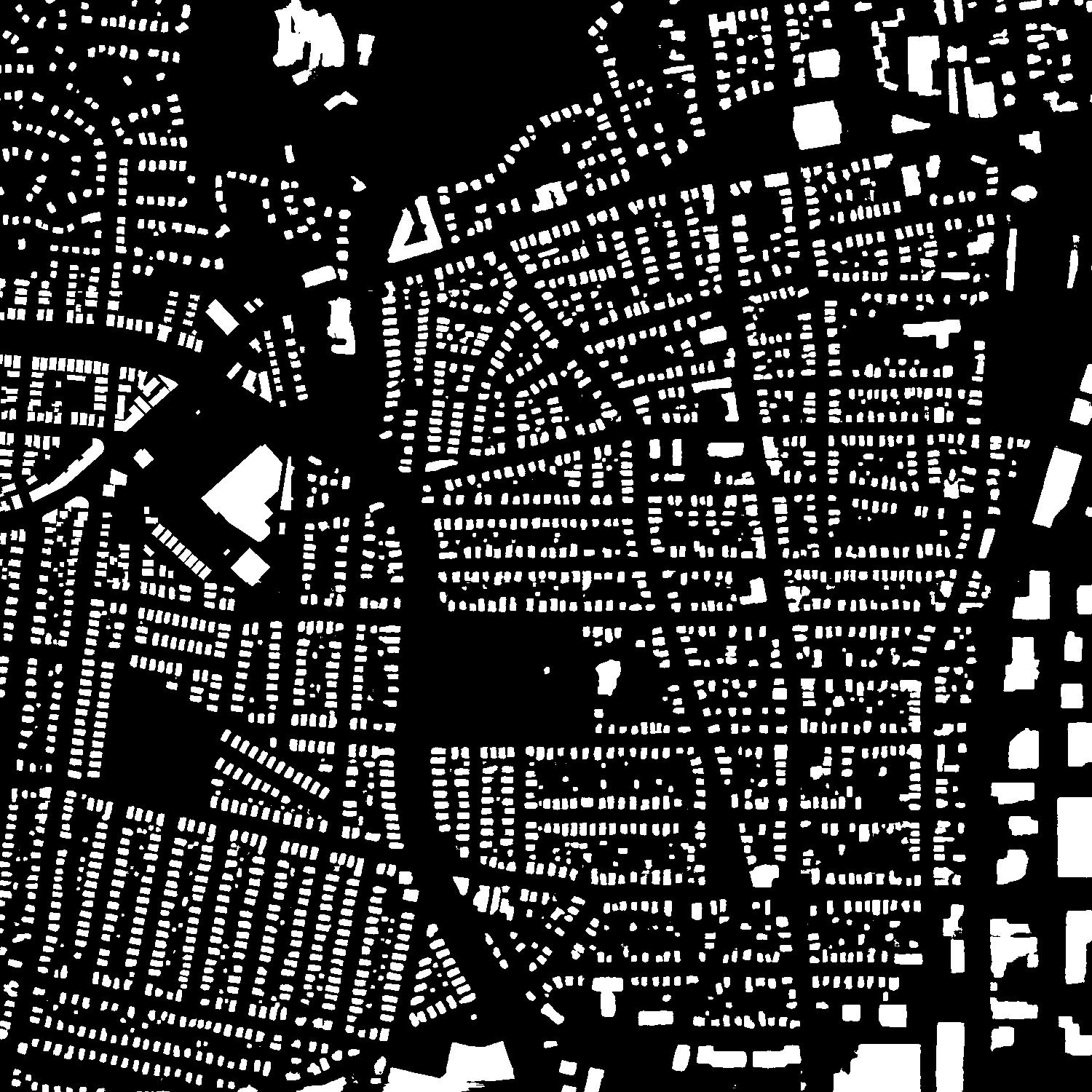} }}
    \hfill
    \subfloat{{\includegraphics[width=5.6cm]{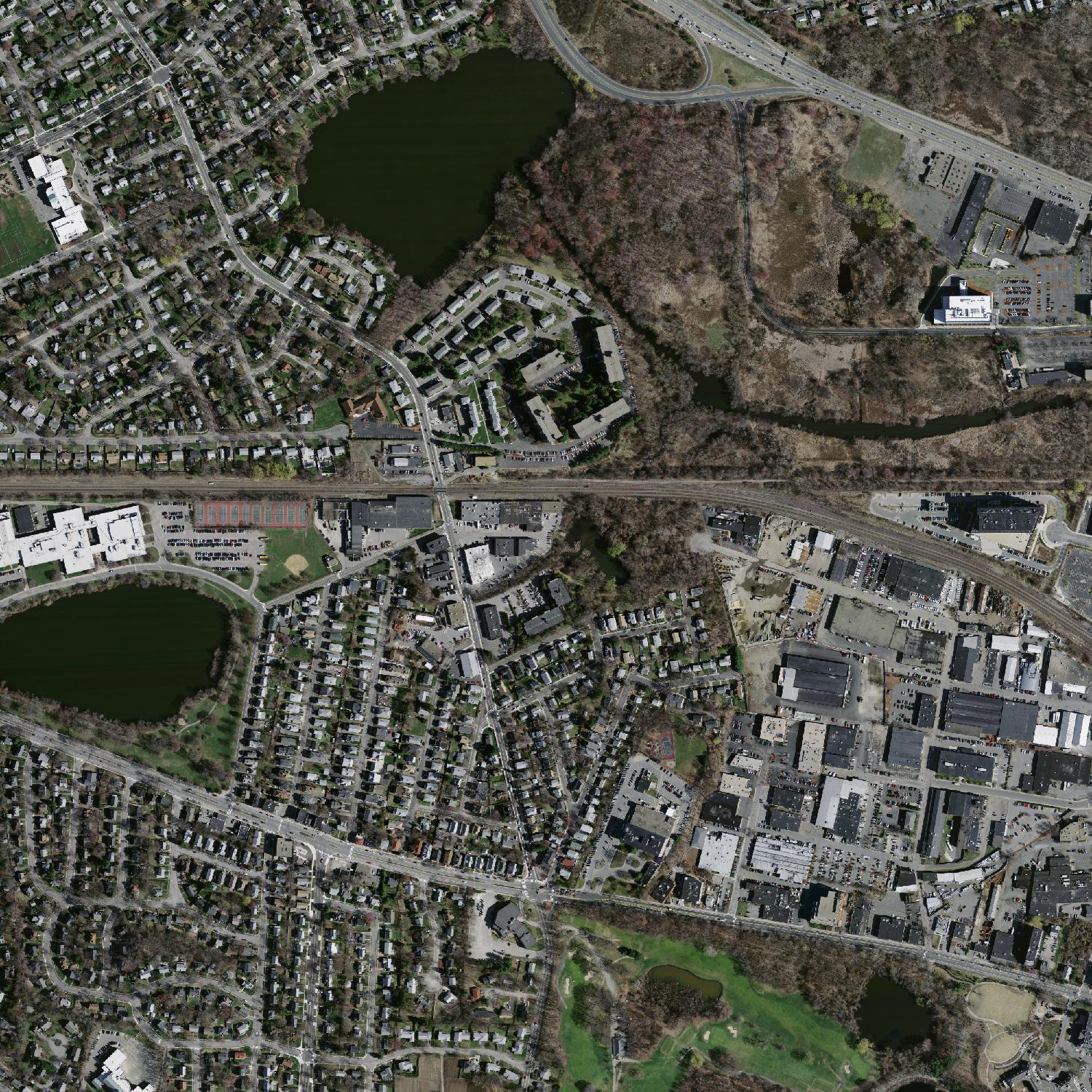} }}
    \subfloat{{\includegraphics[width=5.6cm]{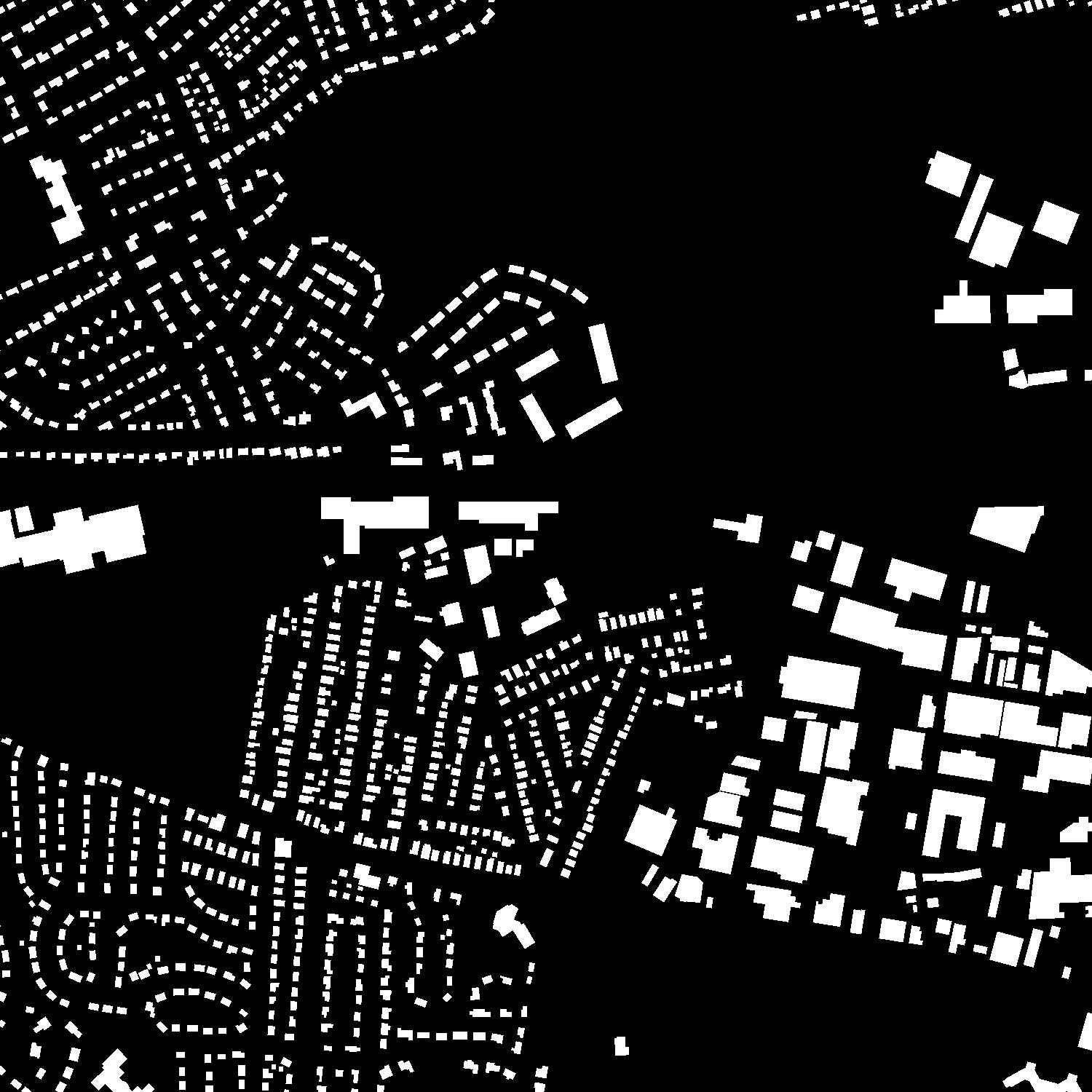} }}
    \subfloat{{\includegraphics[width=5.6cm]{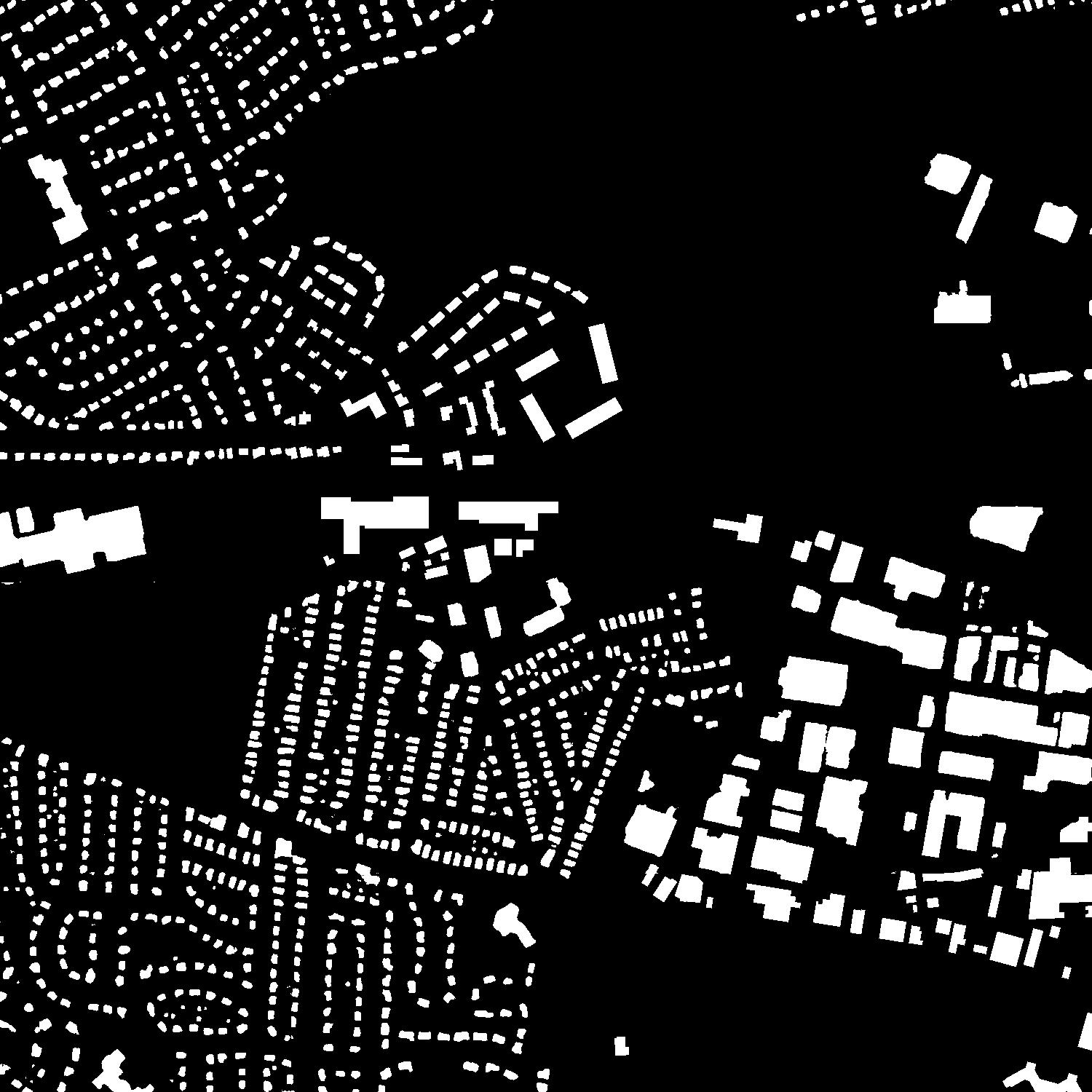} }}
    \caption{\centering Results on the Massachusetts Buildings Dataset. Column 1: Input image. Column 2: Ground-truth Label Map. Column 3: Predicted Label Map.}
    \label{fig:seg_mass}
\end{figure*}

\subsection{Patch Extraction and Prediction Fusion}

During training, we use image patches of size $400 \times 400$. For the INRIA Aerial image Labeling Dataset and the Massachusetts Buildings Dataset, the images provided in the datasets are huge -- $5000 \times 5000$ for the INRIA dataset and $1500 \times 1500$ for the Massachusetts dataset. To fit into the GPU memory, we extract a series of patches, of size $400 \times 400$, from the original RGB input images and the corresponding ground-truth label maps. The patches are extracted with 30\% overlap so that different parts of the images are seen in multiple patches in different locations. The size of the images in the DeepGlobe dataset is $650 \times 650$ and that in the WHU dataset is $512 \times 512$. So instead of creating overlapping patches, for these two datasets, we randomly crop patches of size $400 \times 400$ as a part of the dynamic data augmentation process.

During inference, memory constraint of a 1080Ti GPU limits the maximum image size to be processed by our algorithm to $2000 \times 2000$. We could process whole images from the WHU, Massachusetts and DeepGlobe datasets in one pass. However, to evaluate the performance of our algorithm on the INRIA dataset, we extract patches of size $2000 \times 2000$ with 50\% overlap, perform segmentation on individual patches and merge the predictions of individual patches into an integral prediction for the whole image. Weighted averaging is applied to merge the predictions in overlapping areas. 

\subsection{Post-processing}
Once we have a prediction map for a whole test image, we binarize it to obtain our final building mask. The optimal threshold for binarization is determined by evaluating the respective metrics on the validation images of a specific dataset.

\section{Results}
\label{sec:results}
In this section, we present a comparison of our proposed
framework with some of the state-of-the work building
segmentation approaches.

\subsection{Quantitative Evaluation on the Massachusetts Buildings Dataset}

Table~\ref{tab:results1} presents a relaxed F1-Score
(ref. Section~\ref{sec:massdata}) based comparison between
the different frameworks on the Massachusetts Buildings
Dataset. Our network without TTA achieves a 0.53\%
performance improvement over the previous best performance
\cite{li2018building} with a significantly deeper neural
network using 158 layers. The non-TTA version of our
algorithm outperforms the shallower version of their network
(56 layers) by 0.92\% in terms of relaxed F1-score. With
TTA, we outperform the previous best model by 1.29\%.

Table~\ref{tab:non} 
demonstrates that our proposed method outperforms 
other state-of-the-art approaches by at least 2.77\% and 
3.89\% in terms of non-relaxed F1 and IoU scores respectively.
Figure~\ref{fig:seg_mass} presents our semantic
segmentation result on $1500 \times
1500$ test image patches from the Massachusetts Buildings Dataset. 

In Table~\ref{tab:adv}, we report the relaxed F1 as well as
relaxed IoU scores for our framework and compare the
performance of the framework with some benchmark image
segmentation approaches when adversarial loss is added to
them \cite{sebastian2020adversarial}. Rows 5 and 6 show the
performance of our vanilla generator (no attention) and our
attention-enhanced generator (with attention) networks. It
is clear that the addition of adversarial loss consistently
offers better performance across all the metrics, and our
attention-guided adversarial model performs best among all
the adversarial networks as well.

\bgroup
\def\arraystretch{1.2} \setlength\tabcolsep{0.2cm}
\begin{table}[!ht]
\begin{center}
\resizebox{0.7\linewidth}{!}{%
    \begin{tabular}{ | c | c | c | c | p{1.5cm} |}
    \hline
    Method & \multicolumn{1}{| c |}{Relaxed F1} \\
    \hline
    \hline
    \hline
    Mnih \& Hinton \cite{MnihThesis} & 92.11  \\
    \hline
    Saito et al. \cite{saito2015building} & 92.30 \\
    \hline
    DeepLab v3+ \cite{sebastian2020adversarial} & 92.65 \\
    \hline
    Khalel et al. \cite{khalel2018automatic} & 96.33  \\
    \hline
    MSMT-Stage-1 \cite{marcu2018multi} & 96.04  \\
    \hline
    GAN-SCA \cite{pan2019building} & 96.36  \\
    \hline
    Building-A-Nets (56 layers) \cite{li2018building} & 96.40 \\
    \hline
    Zhang et al. \cite{zhang2020local} & 96.72 \\
    \hline
    Building-A-Nets (158 layers) \cite{li2018building} & 96.78 \\
    \hline
    Our Method (no TTA) & 97.29\\  
    \hline
    Our Method + TTA & \textbf{98.03}\\  
    \hline
    \end{tabular}}
\end{center}
\caption{Relaxed F1-scores of different deep learning based 
networks on the Massachusetts Buildings Dataset. TTA: Test Time Augmentation. The best results are highlighted in bold.}
\label{tab:results1}
\end{table}
\egroup
\bgroup

\bgroup
\def\arraystretch{1.2}\setlength\tabcolsep{0.65cm}
\begin{table}[!ht]
%\small
\begin{center}
\resizebox{0.9\columnwidth}{!}{%
    \begin{tabular}{ | c | c | c | c | c | c | p{1.7cm} |}
    \hline
    Method & \multicolumn{1}{| c |}{F1} & {IoU}\\
    \hline
    \hline
    \hline
    DRNet \cite{chen2021dr} & 79.50 & 66.0  \\ 
    \hline
    GMEDN \cite{ma2020building} & - & 70.39  \\
    \hline
    SRI-Net \cite{liu2019building} & 83.58 & 71.8 \\
    \hline
    ENRU-Net \cite{wang2020automatic} & 84.41 & 73.02  \\
    \hline
    MSCRF \cite{zhu2020building} & 84.75 & 71.19  \\
    \hline
    Chen et al. \cite{chen2021self} & 84.72 & 73.49 \\
    \hline
    DS-Net2 \cite{guo2021scale} & 84.91 & 73.79 \\
    \hline
    DS-Net \cite{liao2020learning} & - & 74.43  \\
    \hline
    BMFR-Net \cite{ran2021building} & 85.14 & 74.12  \\
    \hline
    BRRNet \cite{shao2020brrnet} & 85.36 & 74.46  \\  
    \hline
    Liao et al. \cite{liao2021joint} & 85.39 & 74.51  \\
    \hline
    Zhang et al. \cite{zhang2020local} & 85.49 & -  \\
    \hline
    Our Method (no TTA) & 86.98 & 76.97 \\  
    \hline
    Our Method + TTA & \textbf{87.86} & \textbf{77.41} \\  
    \hline
    \end{tabular}}
\end{center}
\caption{\centering Regular F1 and IoU scores for the state-of-the-art networks on the Massachusetts Buildings Dataset.  TTA: Test Time Augmentation. The best results are highlighted in bold.}
\label{tab:non}
\end{table}
\egroup
\bgroup

\bgroup
\def\arraystretch{1.2}
\setlength\tabcolsep{0.15cm}
\begin{table}[!ht]
\begin{center}
\resizebox{0.9\linewidth}{!}{%
    \begin{tabular}{ | c | c | c | c | c | c | p{1.5cm} |}
    \hline
    Method & \multicolumn{1}{| c |}{Relaxed F1} & {Relaxed IoU}\\
    \hline
    \hline
    \hline
    PSPNet  & 89.52 & 81.2  \\
    \hline
    PSPNet + \emph{adv} & 91.17 & 83.78  \\
    \hline
    FC-DenseNet & 94.33 & 89.27  \\
    \hline
    FC-DenseNet + \emph{adv}  & 95.59 & 91.55 \\
    \hline
    Our vanilla Generator & 94.11 & 91.64  \\  
    \hline
    Our proposed Generator ($\mathcal{S}$)%Vanilla Generator + attention units + deep supervision}
    & 96.82 & 94.79  \\  
    \hline
    Our Method ($\mathcal{S}$ + $\mathcal{C}$) & \textbf{98.03} & \textbf{96.19}  \\  
    \hline
    \end{tabular}}
\end{center}

\caption{\centering Comparison of benchmark image segmentation models
  with adversarial loss on the Massachusetts Buildings Dataset. \emph{adv} represents adversarial loss. The scores of our method reflect the results of our algorithm using TTA. The best results are highlighted in bold.}
\label{tab:adv}
\end{table}
\egroup
\bgroup

\subsection{Quantitative Evaluation on the INRIA Aerial Image Labeling Dataset}

\begin{table*}
\centering
\begin{tabular}{ccccc}
    Austin & 
    \raisebox{-.5\height}{\includegraphics[width=3.6cm]{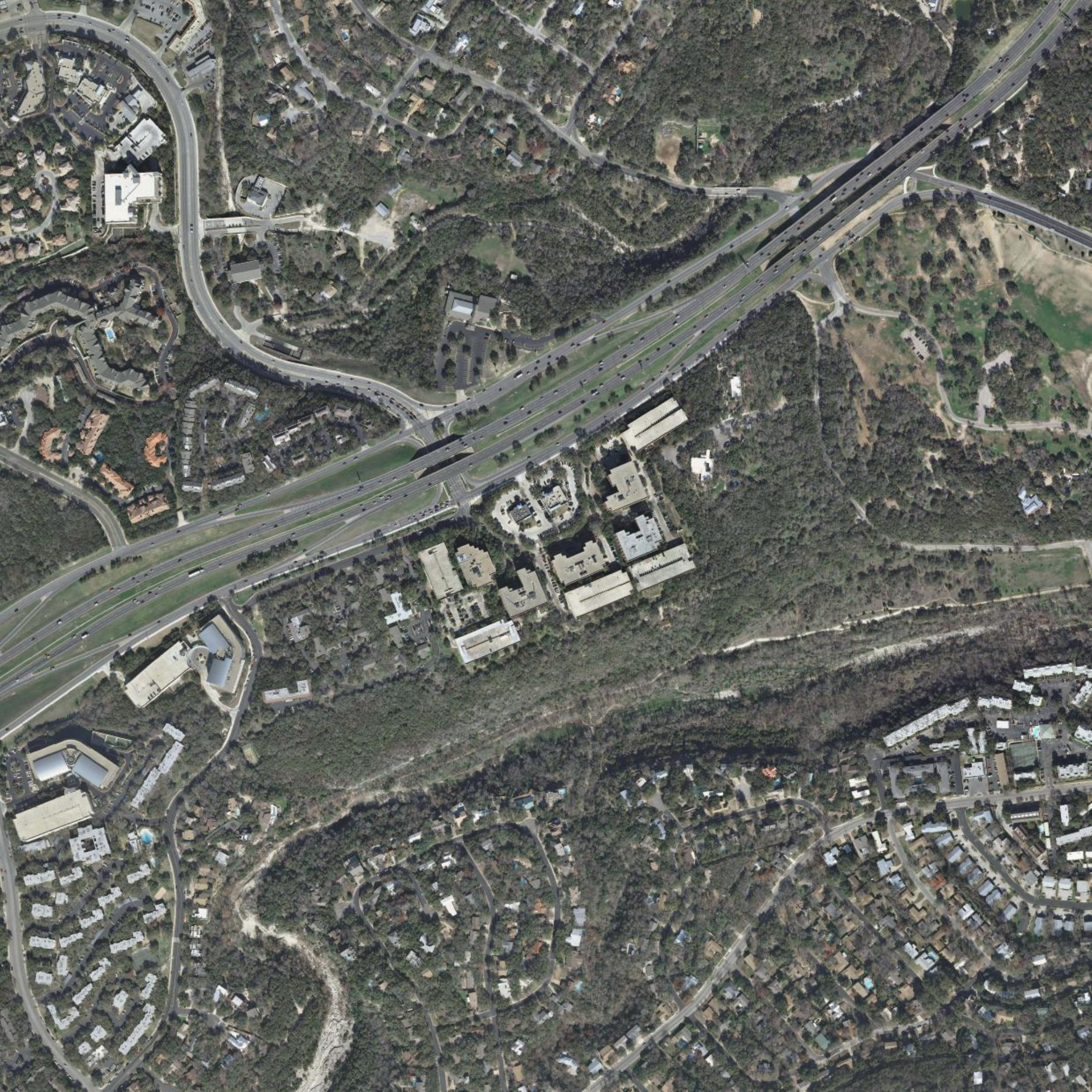}} & 
    \raisebox{-.5\height}{\includegraphics[width=3.6cm]{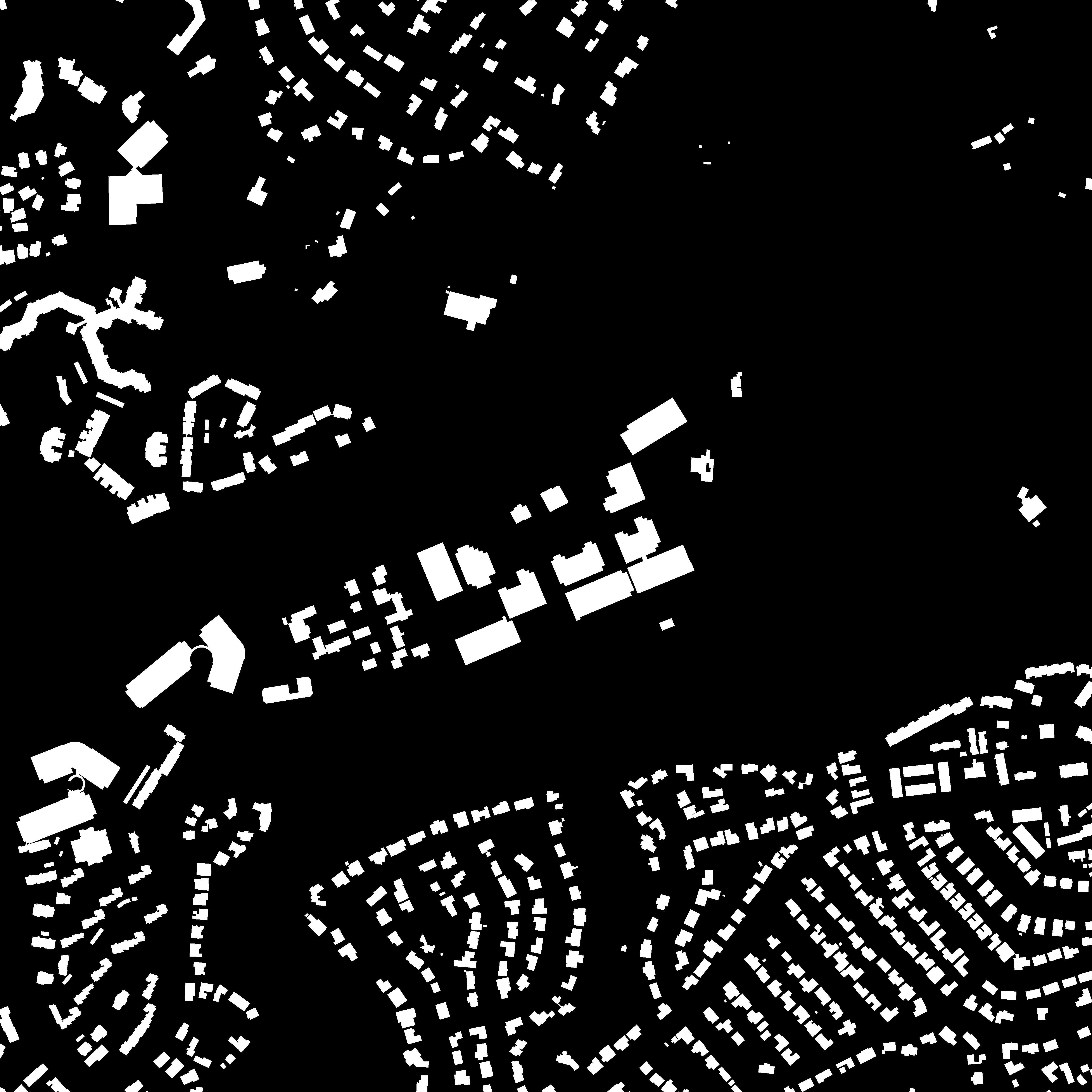}} &
    \raisebox{-.5\height}{\includegraphics[width=3.6cm]{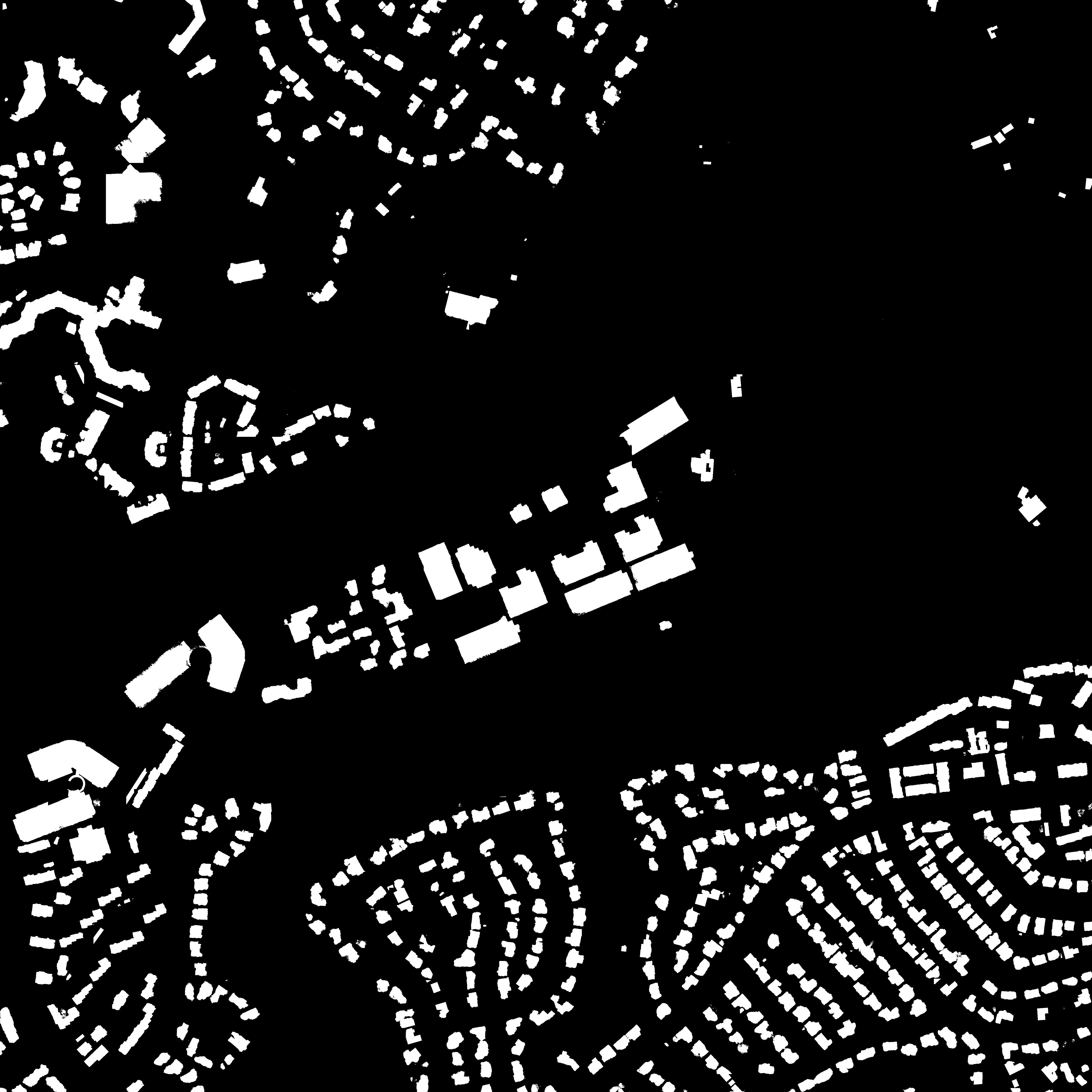}} &
    \raisebox{-.5\height}{\includegraphics[width=3.6cm]{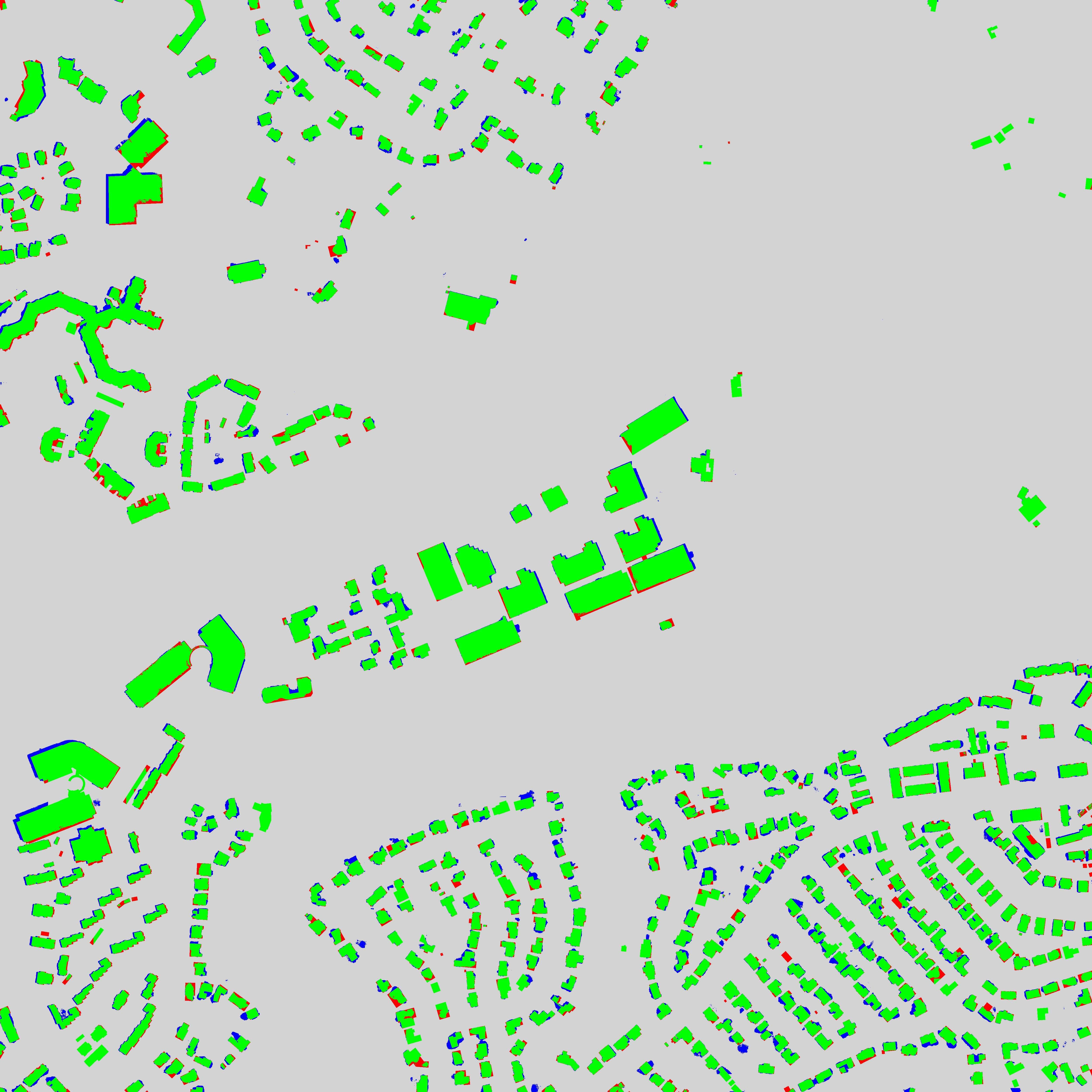}} \\
    \\
    Chicago & 
    \raisebox{-.5\height}{\includegraphics[width=3.6cm]{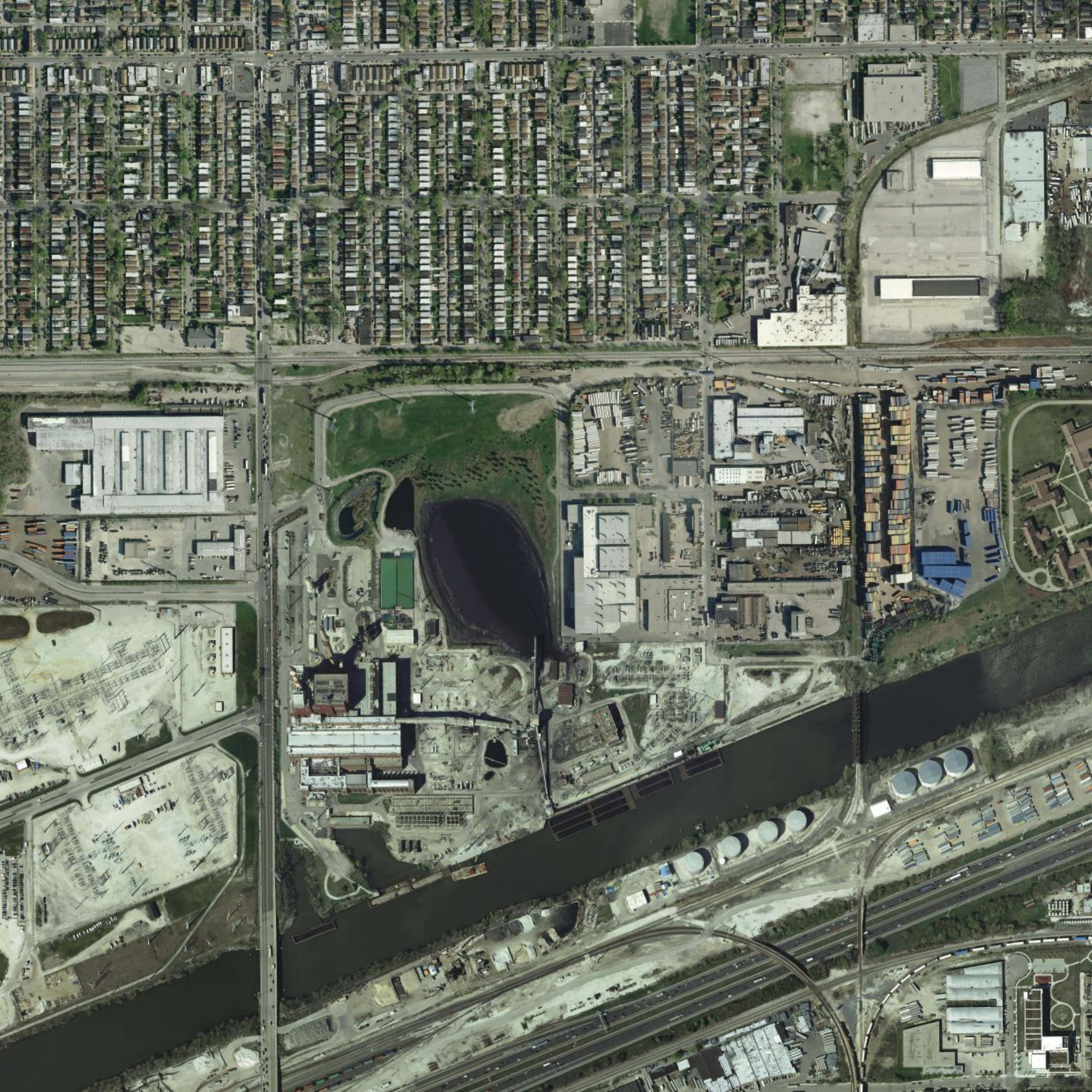}} & 
    \raisebox{-.5\height}{\includegraphics[width=3.6cm]{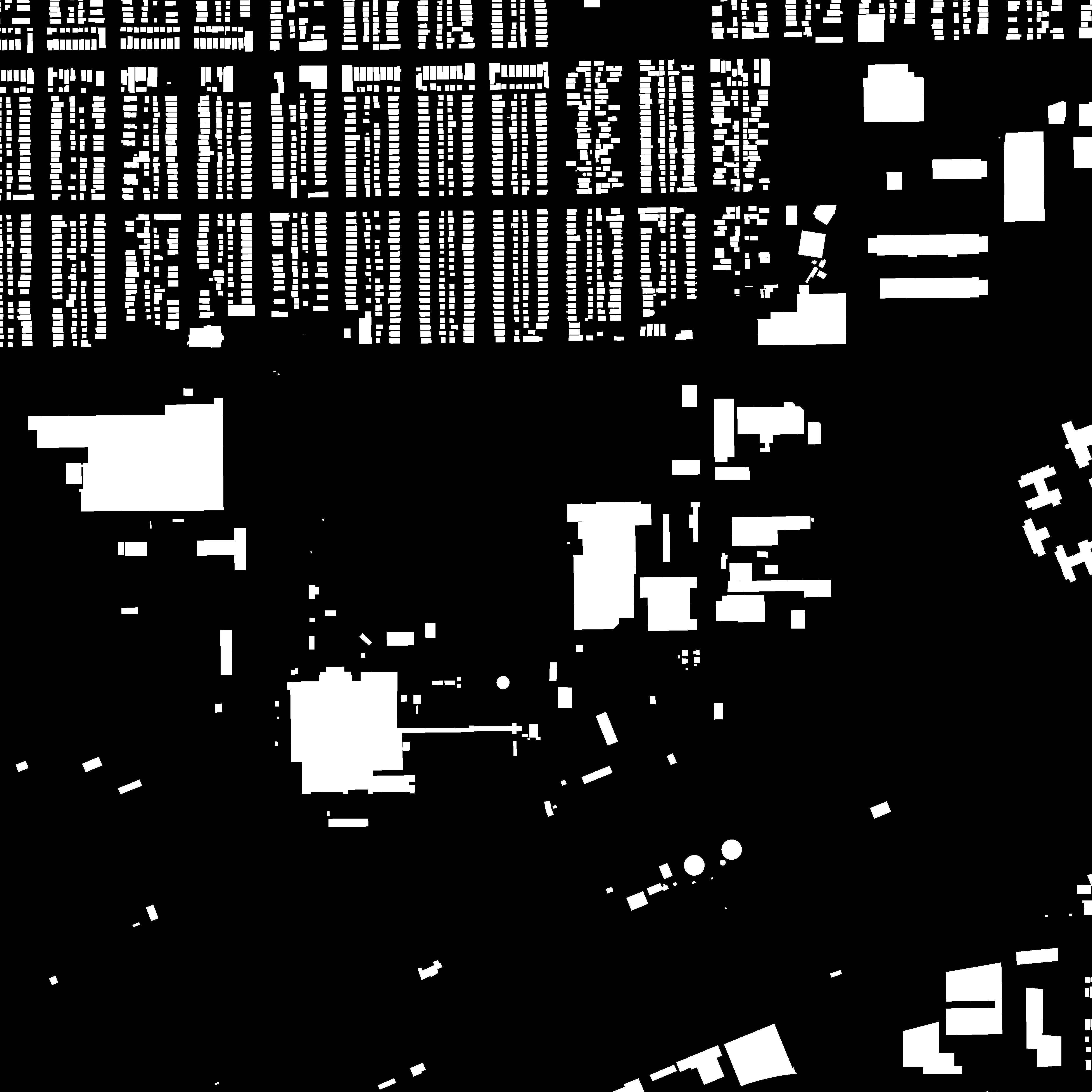}} & 
    \raisebox{-.5\height}{\includegraphics[width=3.6cm]{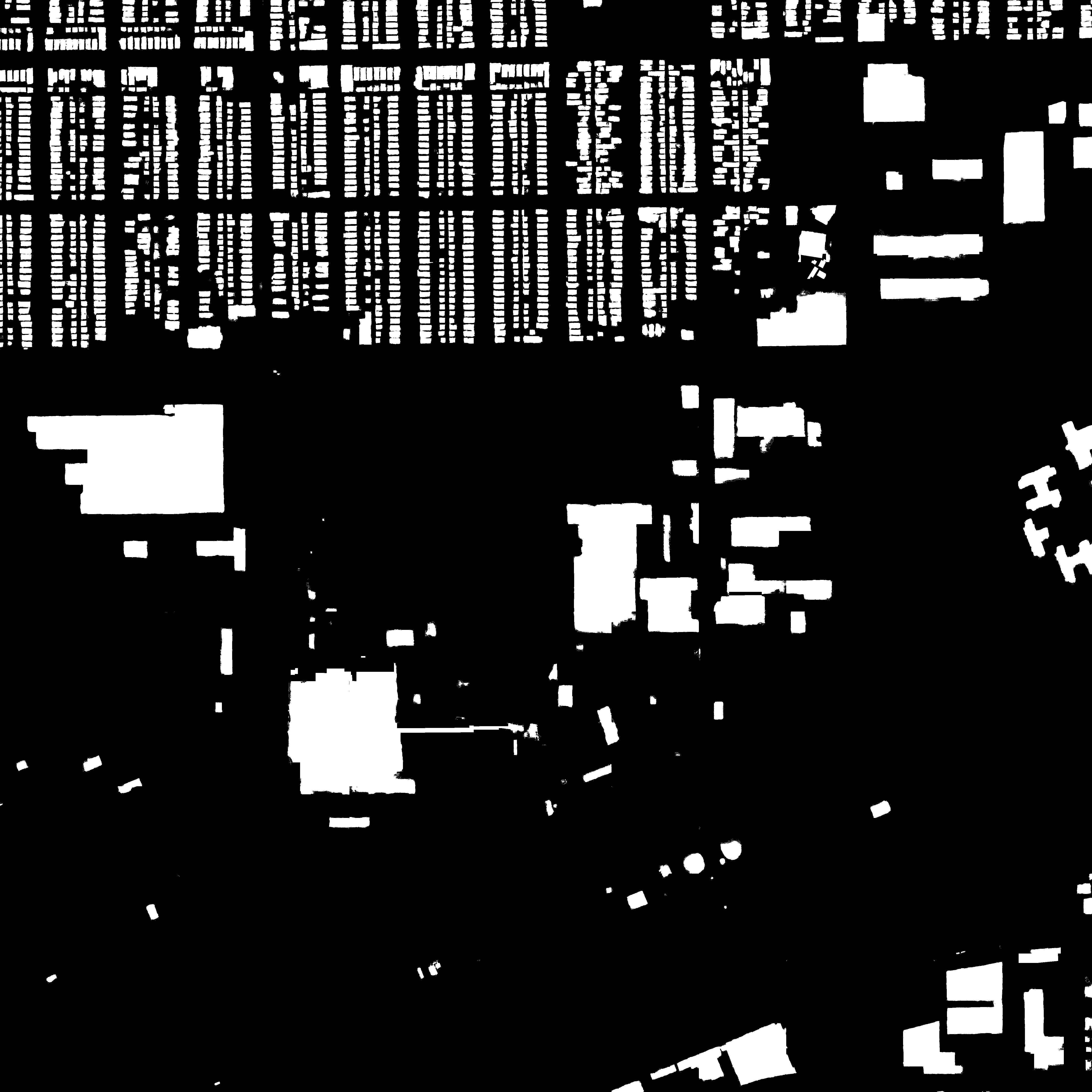}} &
    \raisebox{-.5\height}{\includegraphics[width=3.6cm]{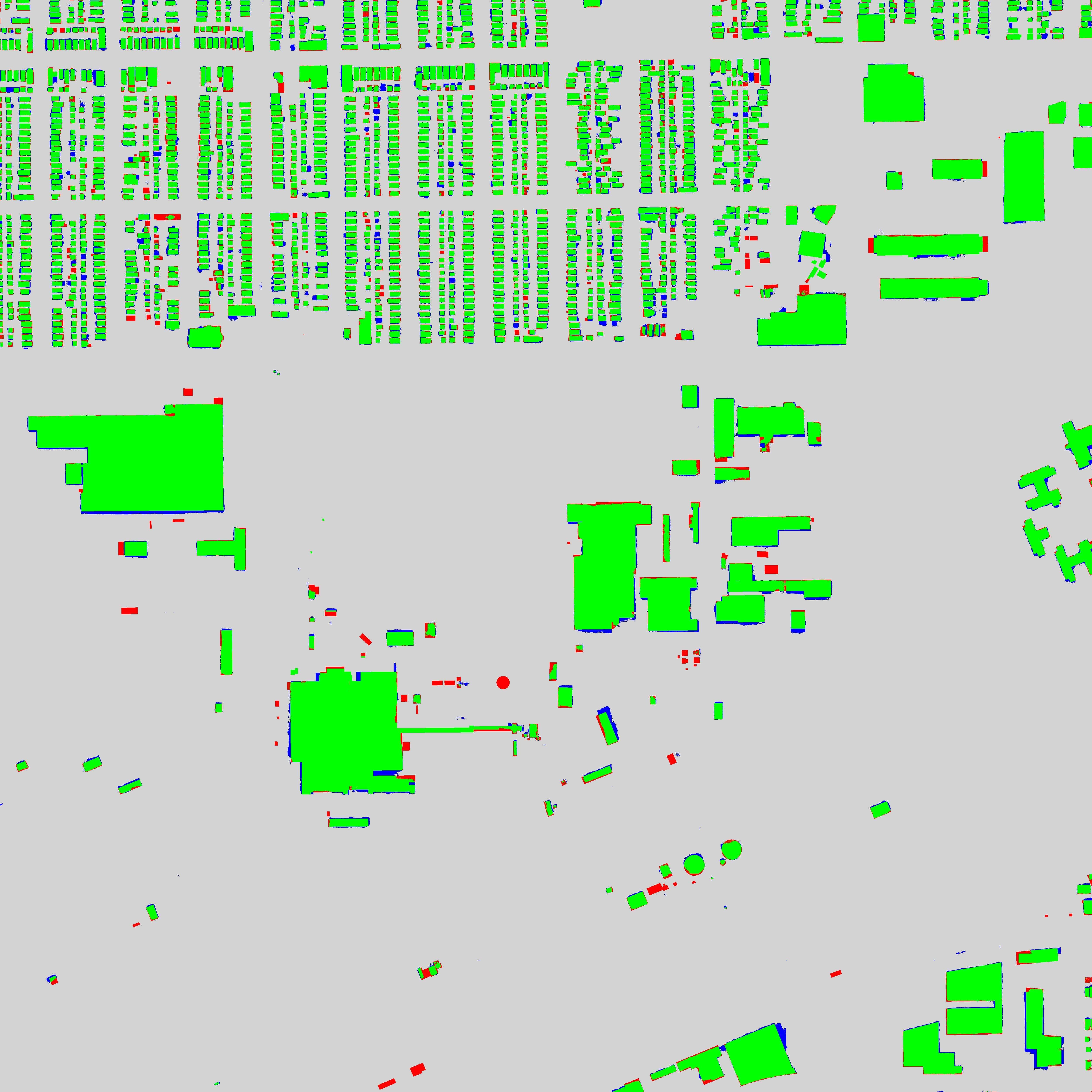}}
    \\
    \\
    Vienna & 
    \raisebox{-.5\height}{\includegraphics[width=3.6cm]{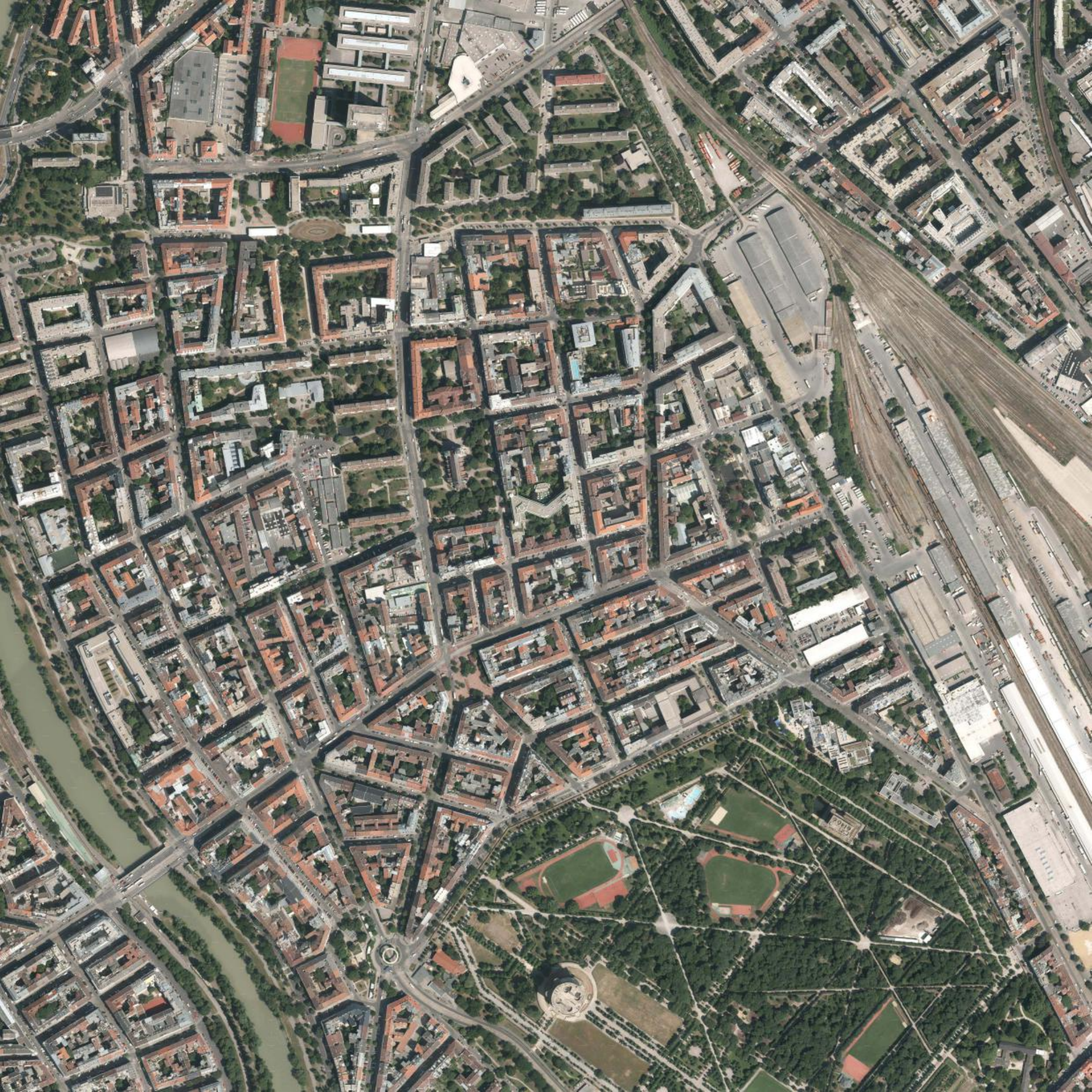}} & 
    \raisebox{-.5\height}{\includegraphics[width=3.6cm]{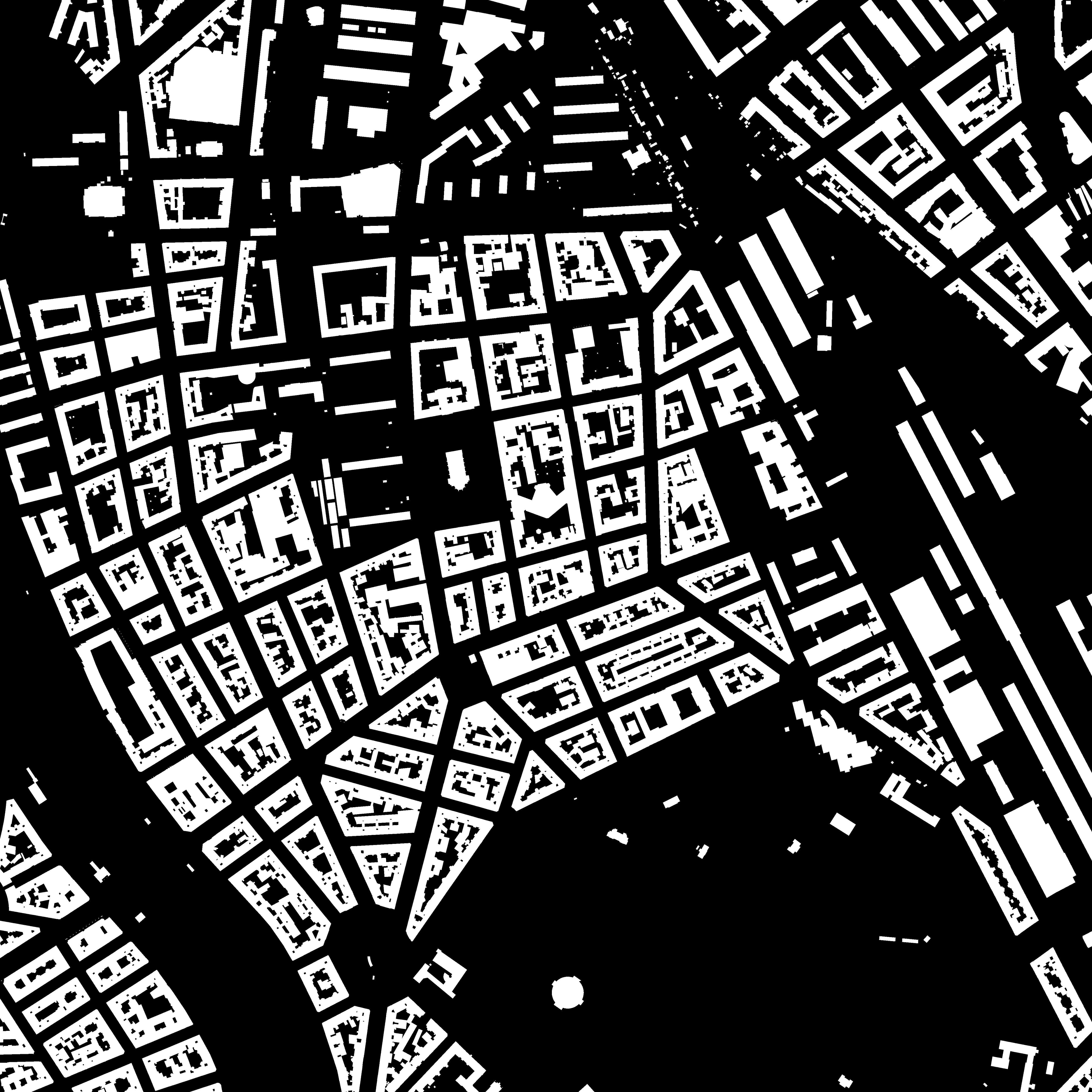}} & 
    \raisebox{-.5\height}{\includegraphics[width=3.6cm]{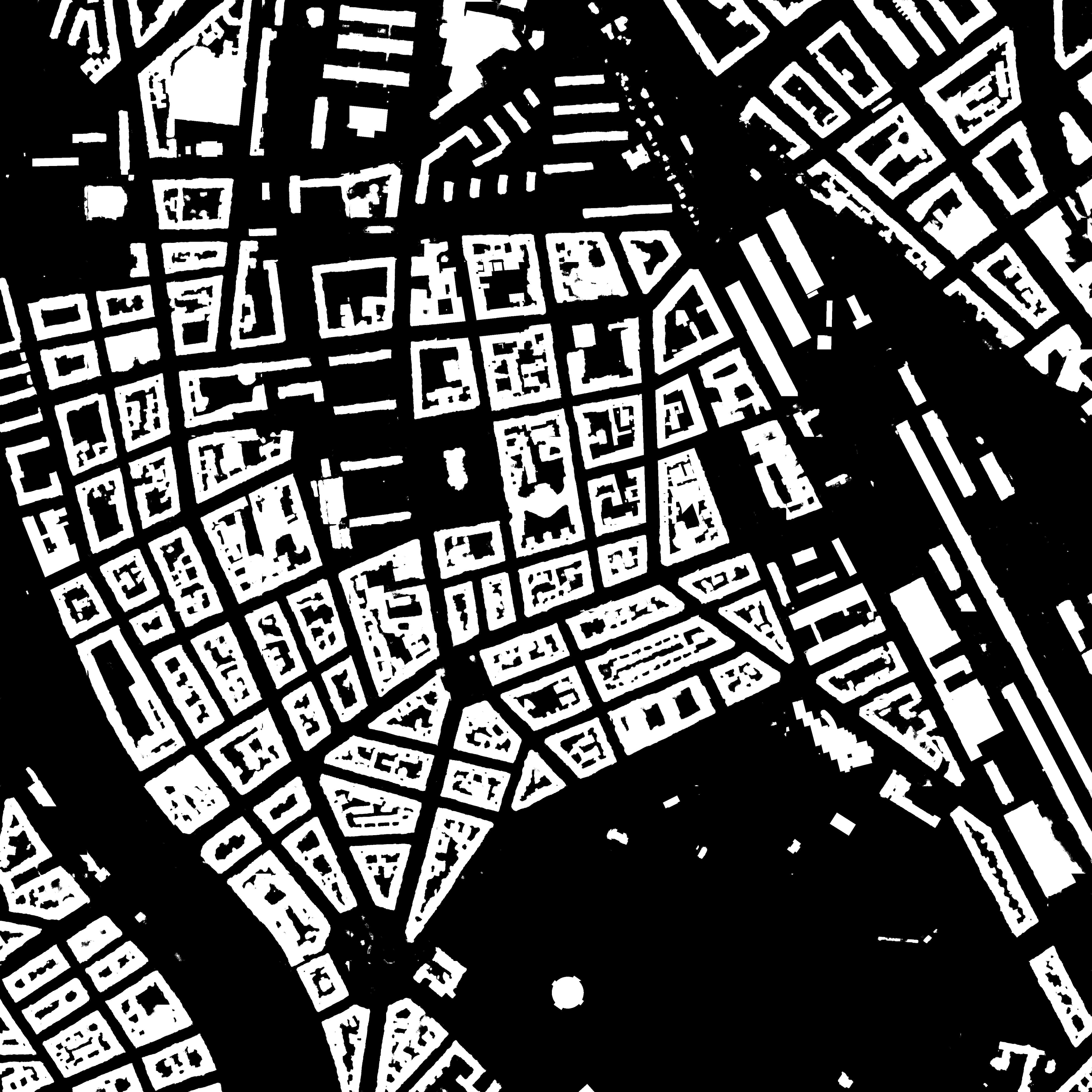}} &
    \raisebox{-.5\height}{\includegraphics[width=3.6cm]{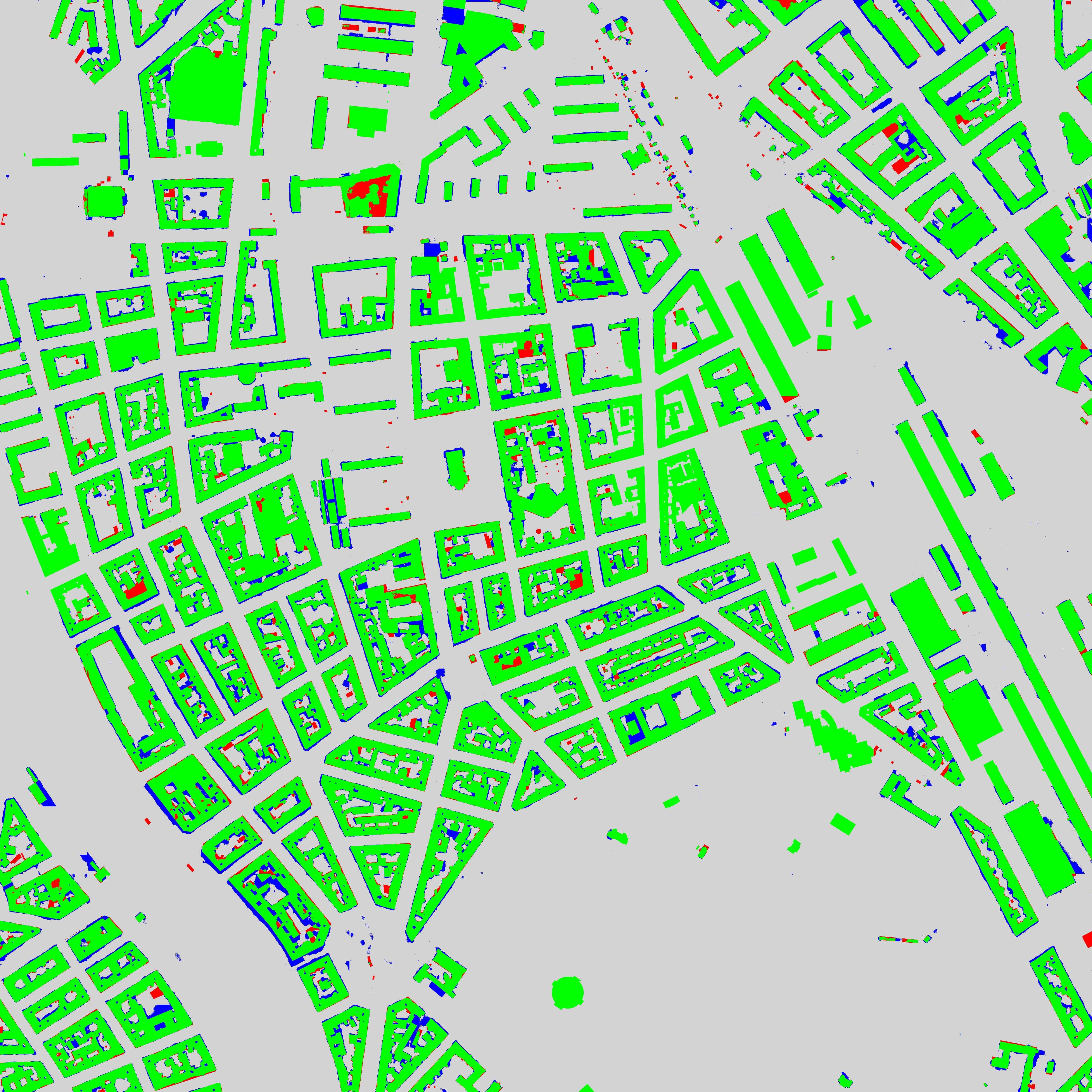}}
    \\
    \\
    Kitsap & 
    \raisebox{-.5\height}{\includegraphics[width=3.6cm]{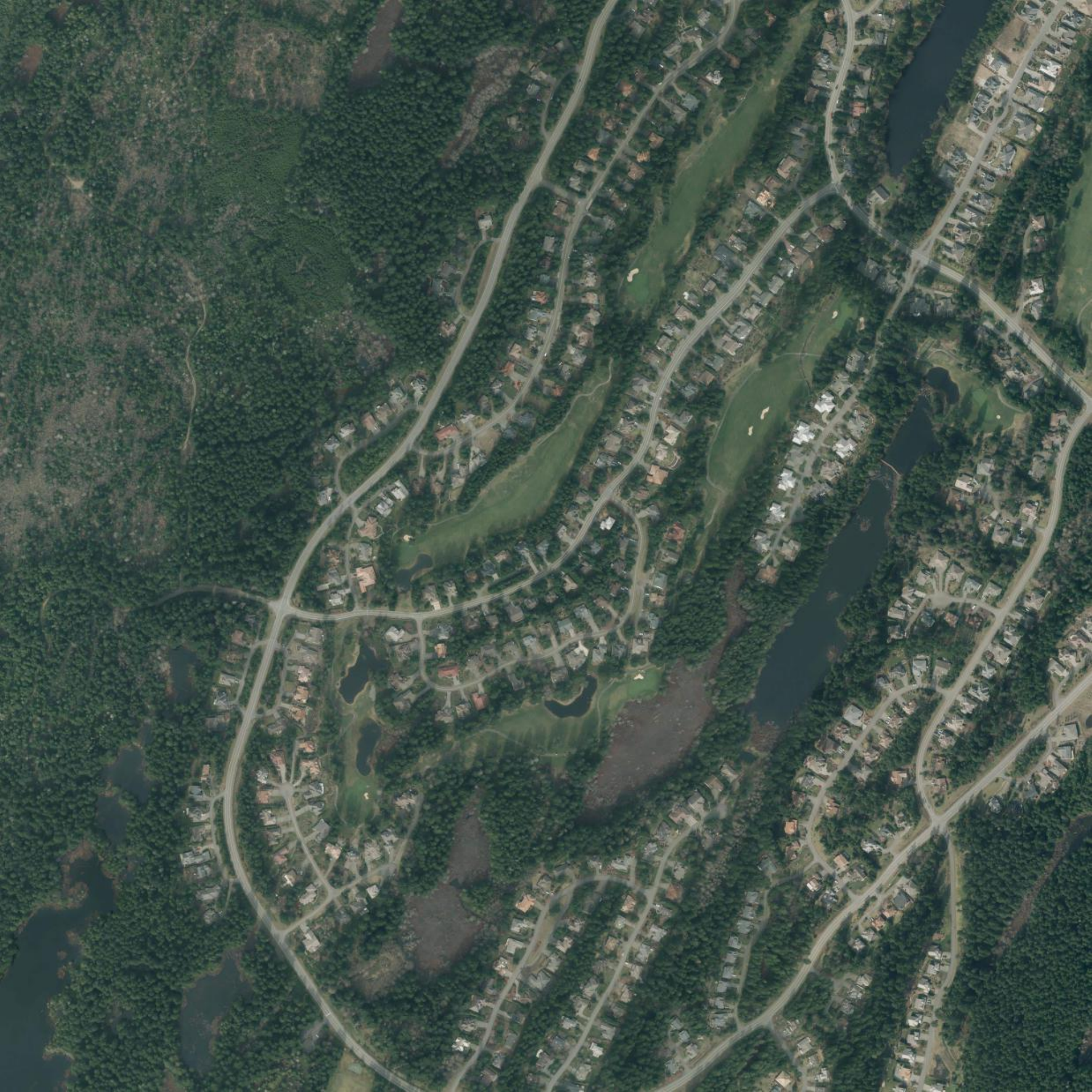}} & 
    \raisebox{-.5\height}{\includegraphics[width=3.6cm]{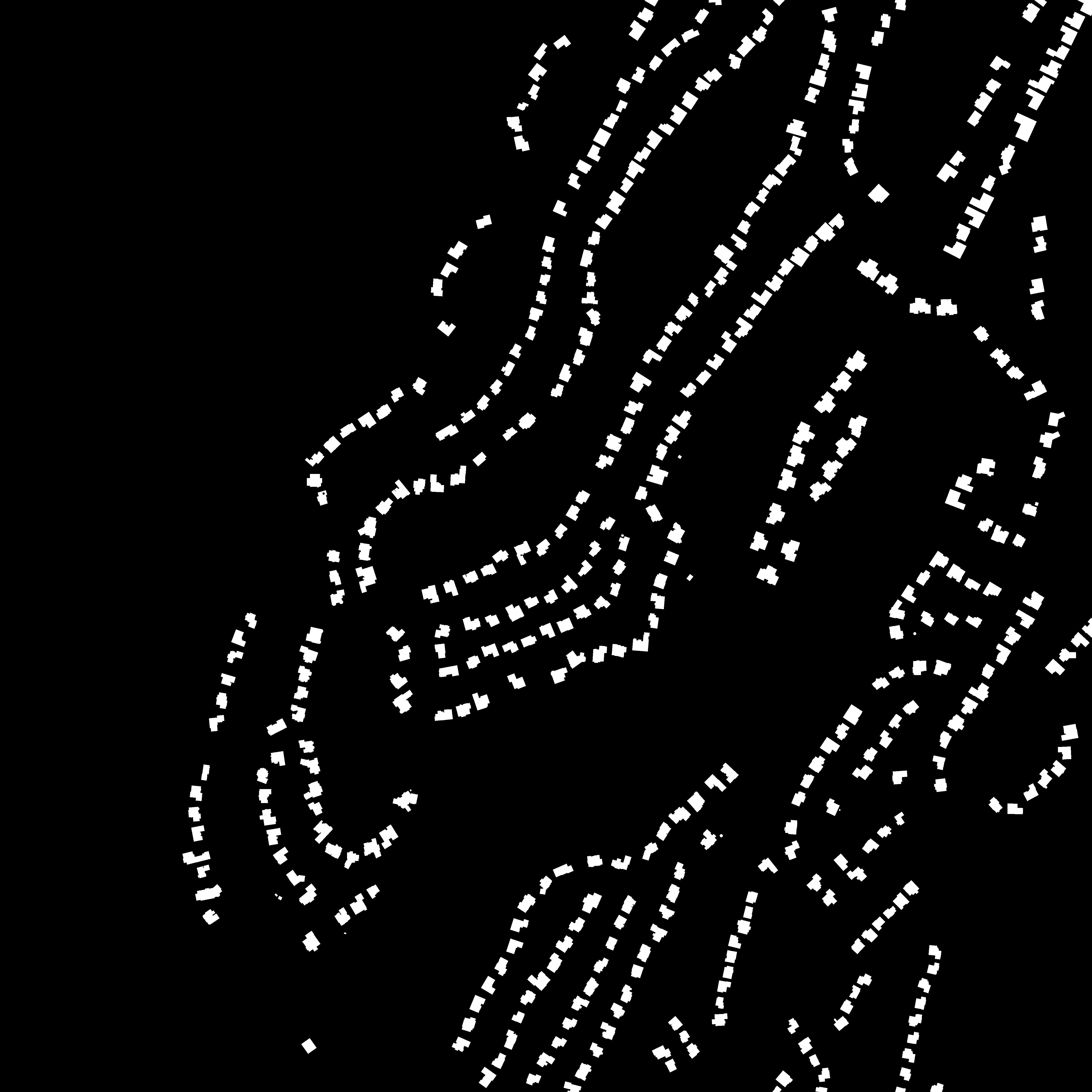}} & 
    \raisebox{-.5\height}{\includegraphics[width=3.6cm]{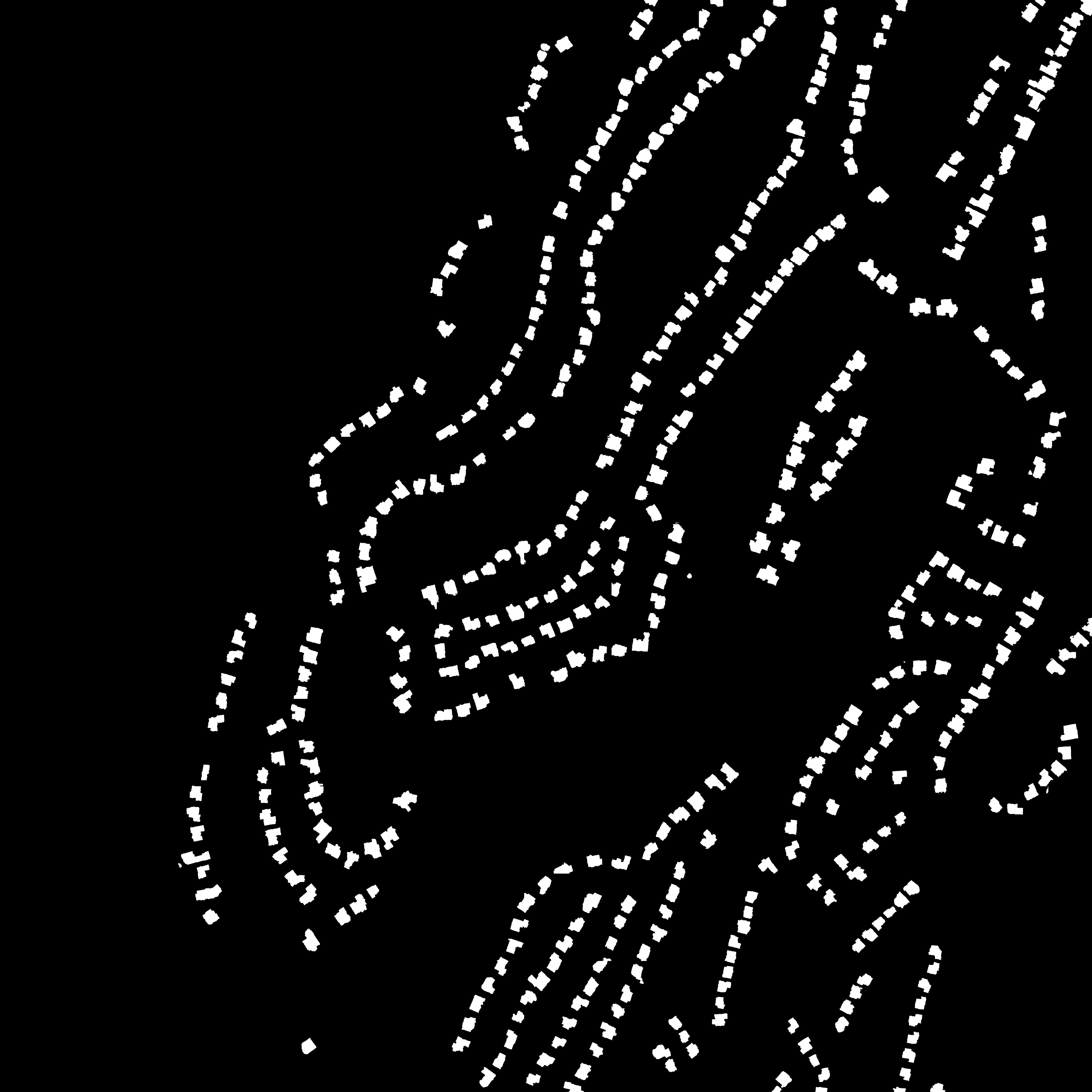}} &
    \raisebox{-.5\height}{\includegraphics[width=3.6cm]{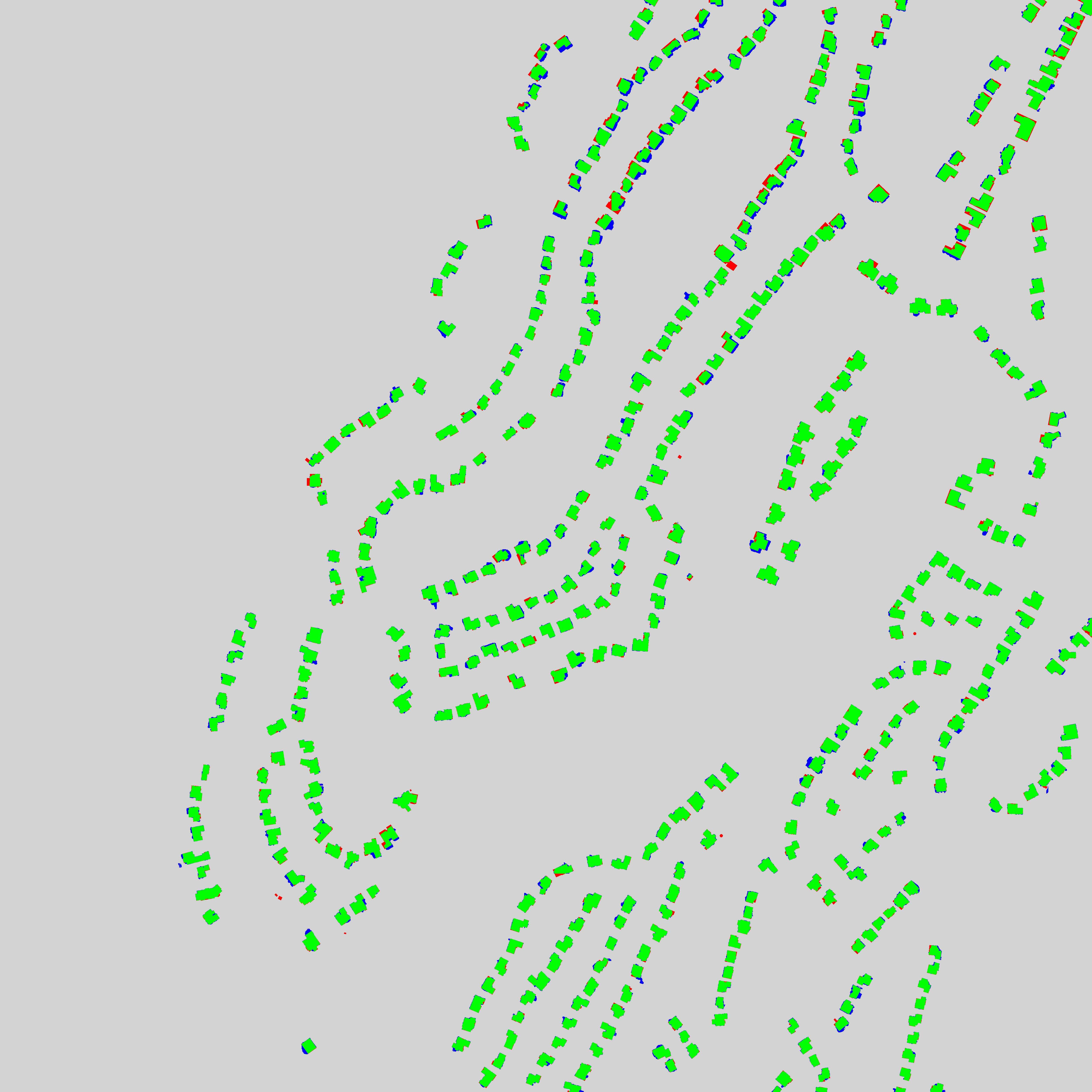}}
    \\
    \\
    West Tyrol & 
    \raisebox{-.5\height}{\includegraphics[width=3.6cm]{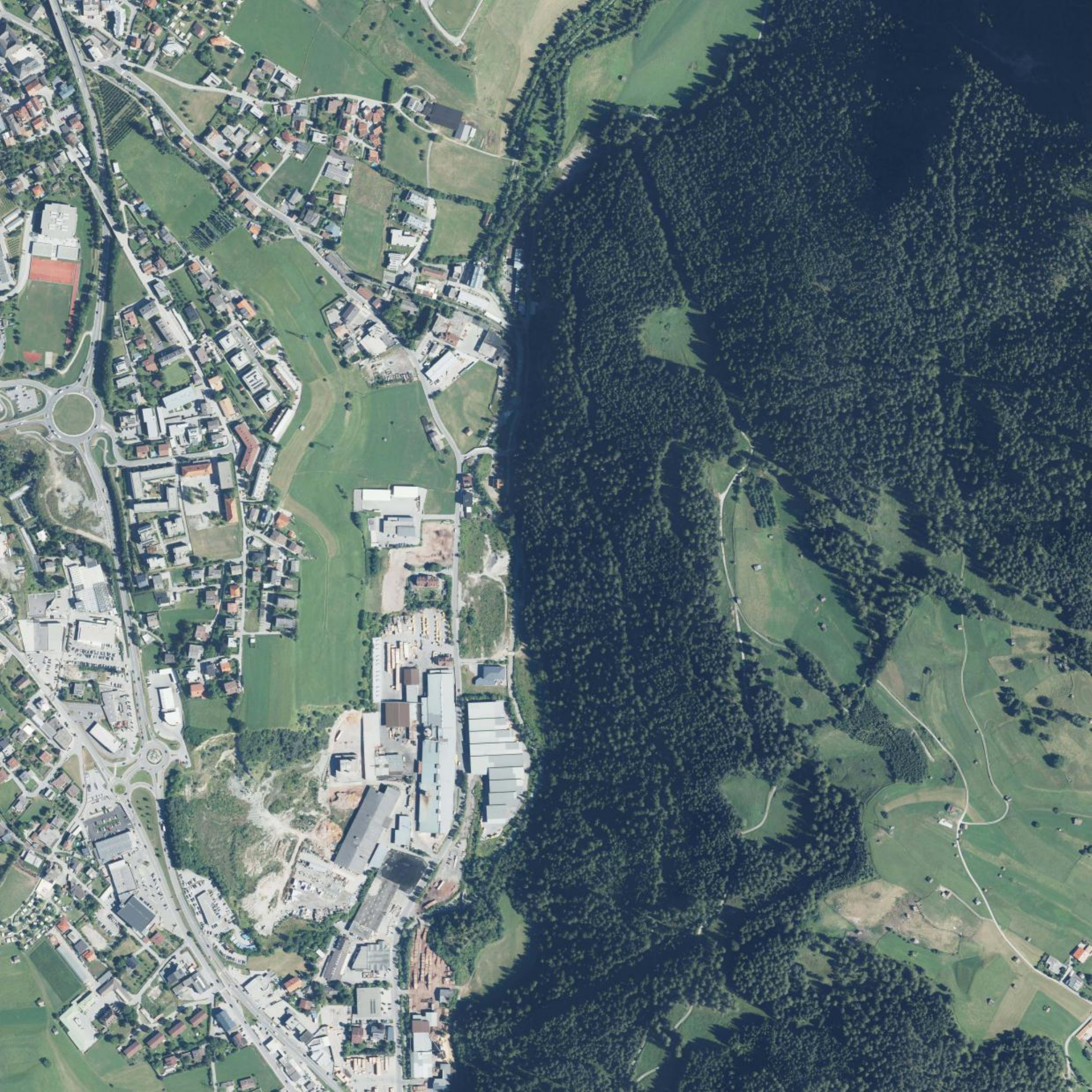}} & 
    \raisebox{-.5\height}{\includegraphics[width=3.6cm]{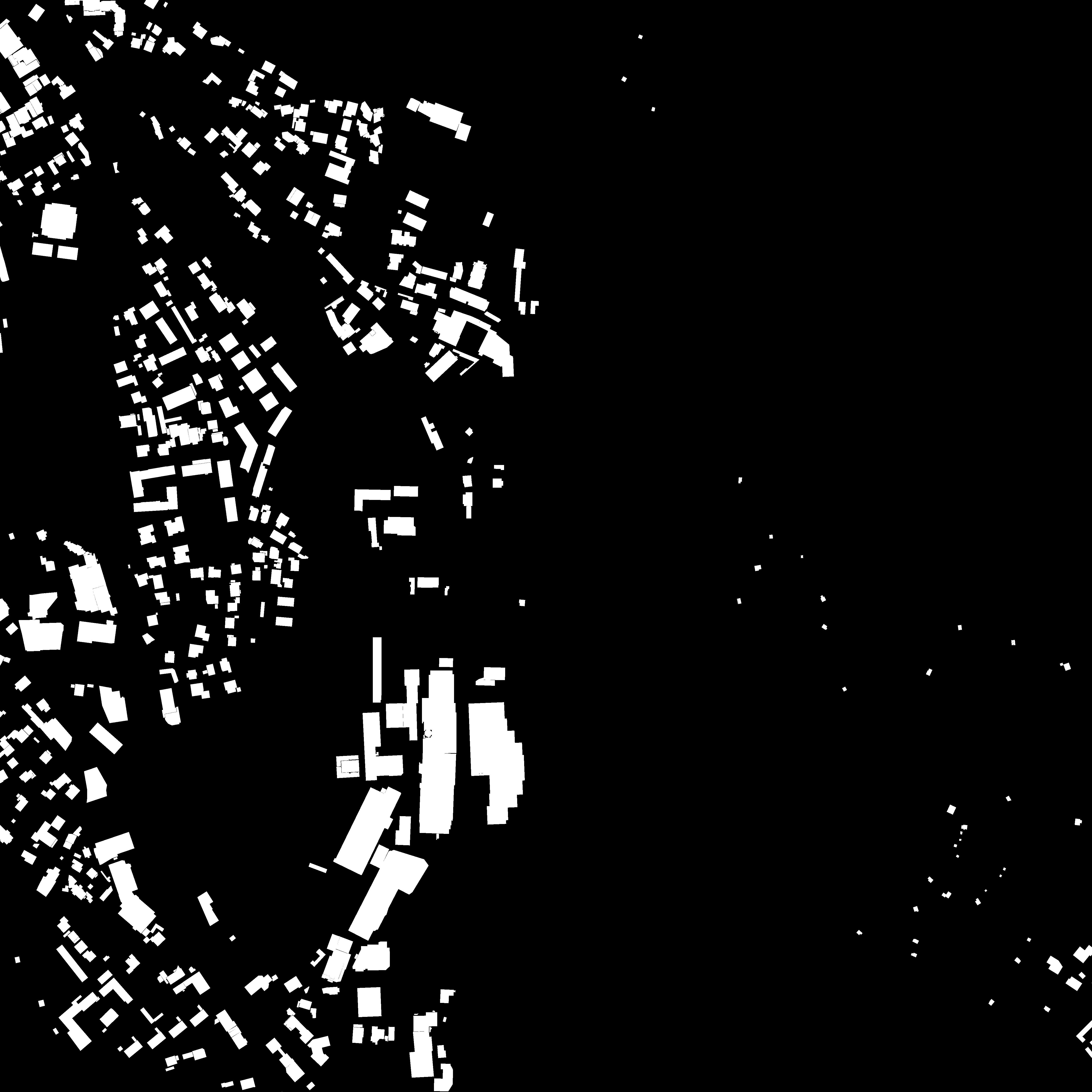}} & 
    \raisebox{-.5\height}{\includegraphics[width=3.6cm]{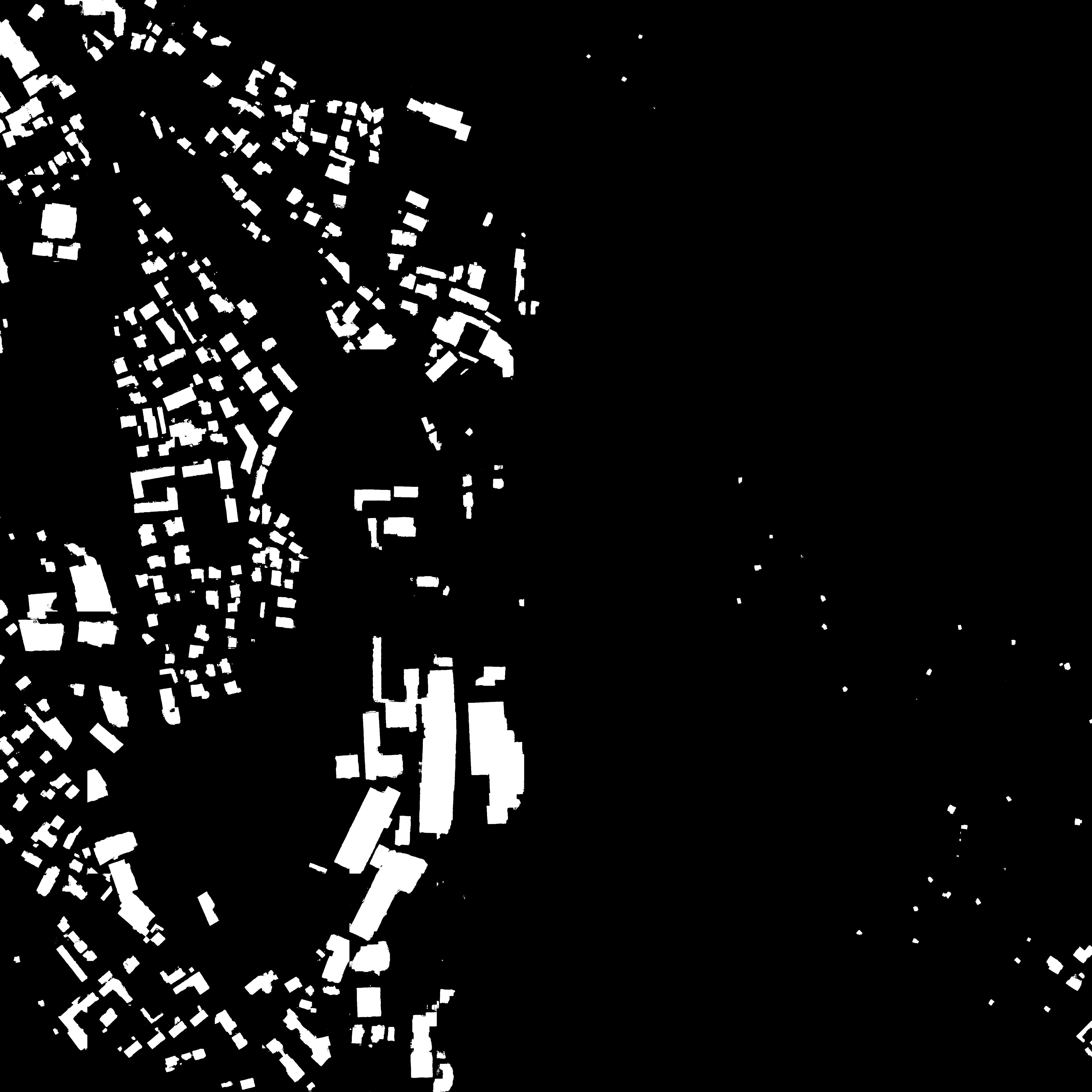}} &
    \raisebox{-.5\height}{\includegraphics[width=3.6cm]{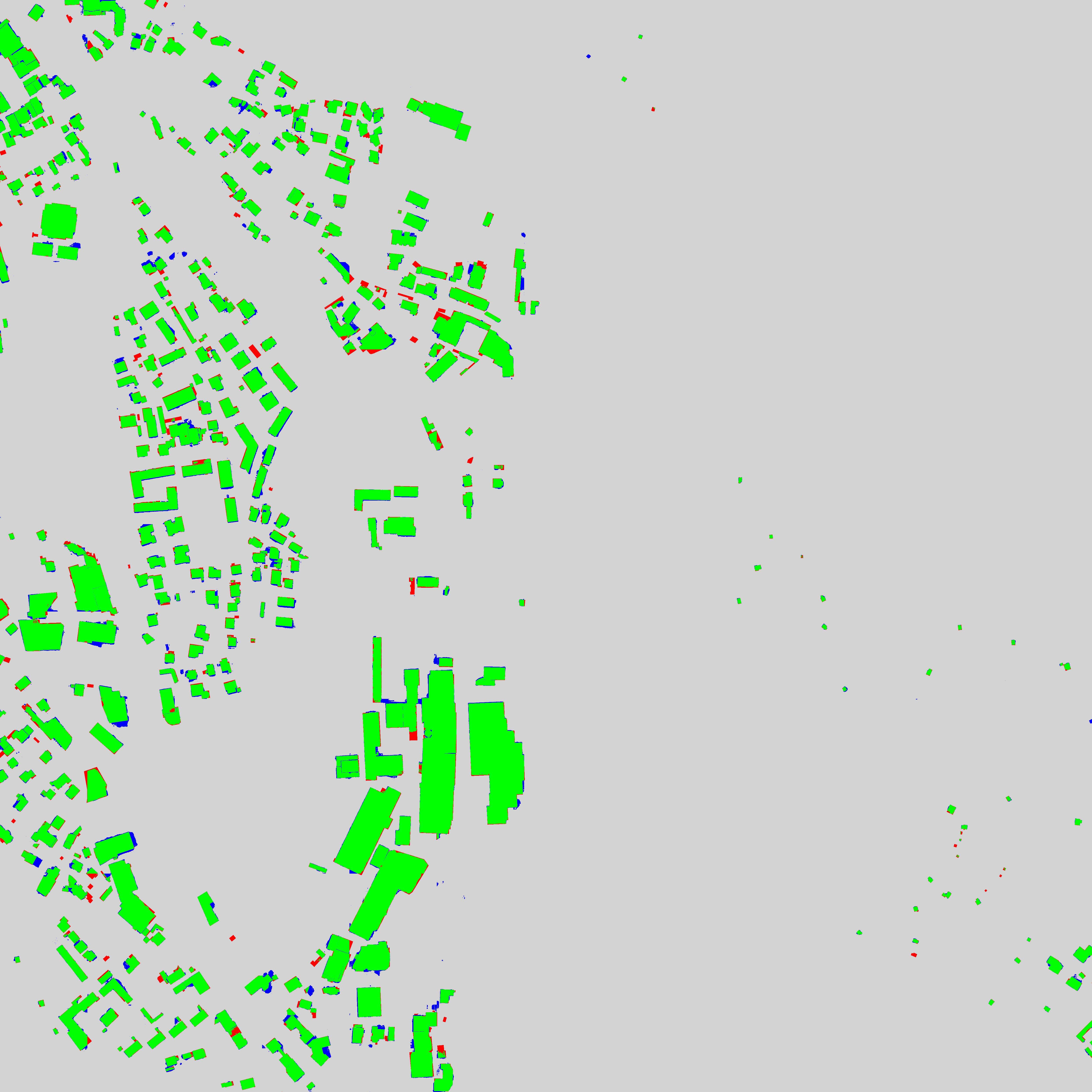}}
\end{tabular}
\captionof{figure}{\centering Results on the INRIA Aerial Image Labeling Validation Dataset. Column 1: Input image. Column 2: Ground-truth Label Map. Column 3: Predicted Label Map. Column 4: Green: True Positives; Blue: False Positives; Red: False Negatives; Grey: True Negatives.}
 \label{fig:seg_inria1}
\end{table*}

\begin{table*}
\centering
\begin{tabular}{ccccc}
    Bellingham & Bloomington & Innsbruck & San Francisco & East Tyrol
    \\
    \includegraphics[width=3.2cm]{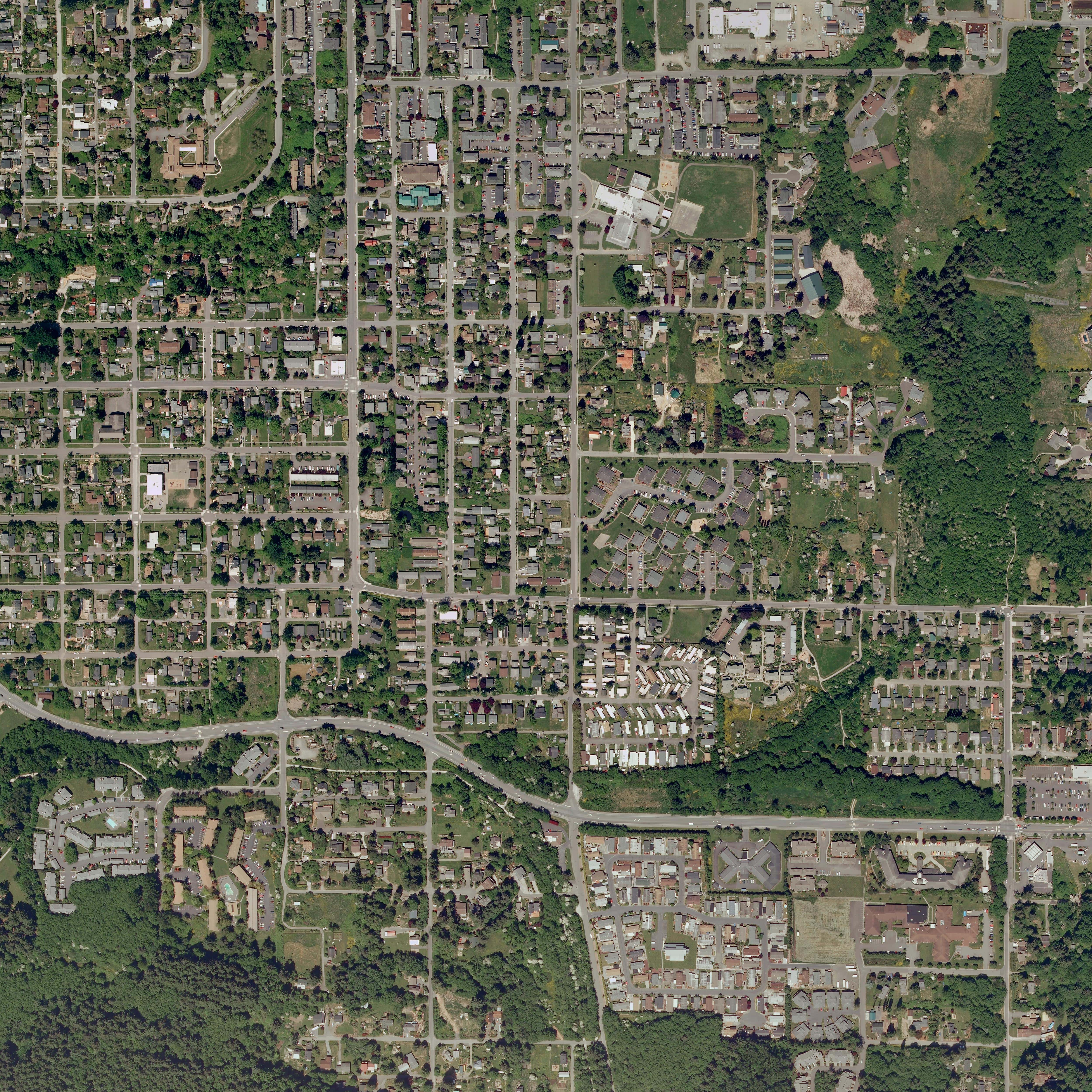} &
    \includegraphics[width=3.2cm]{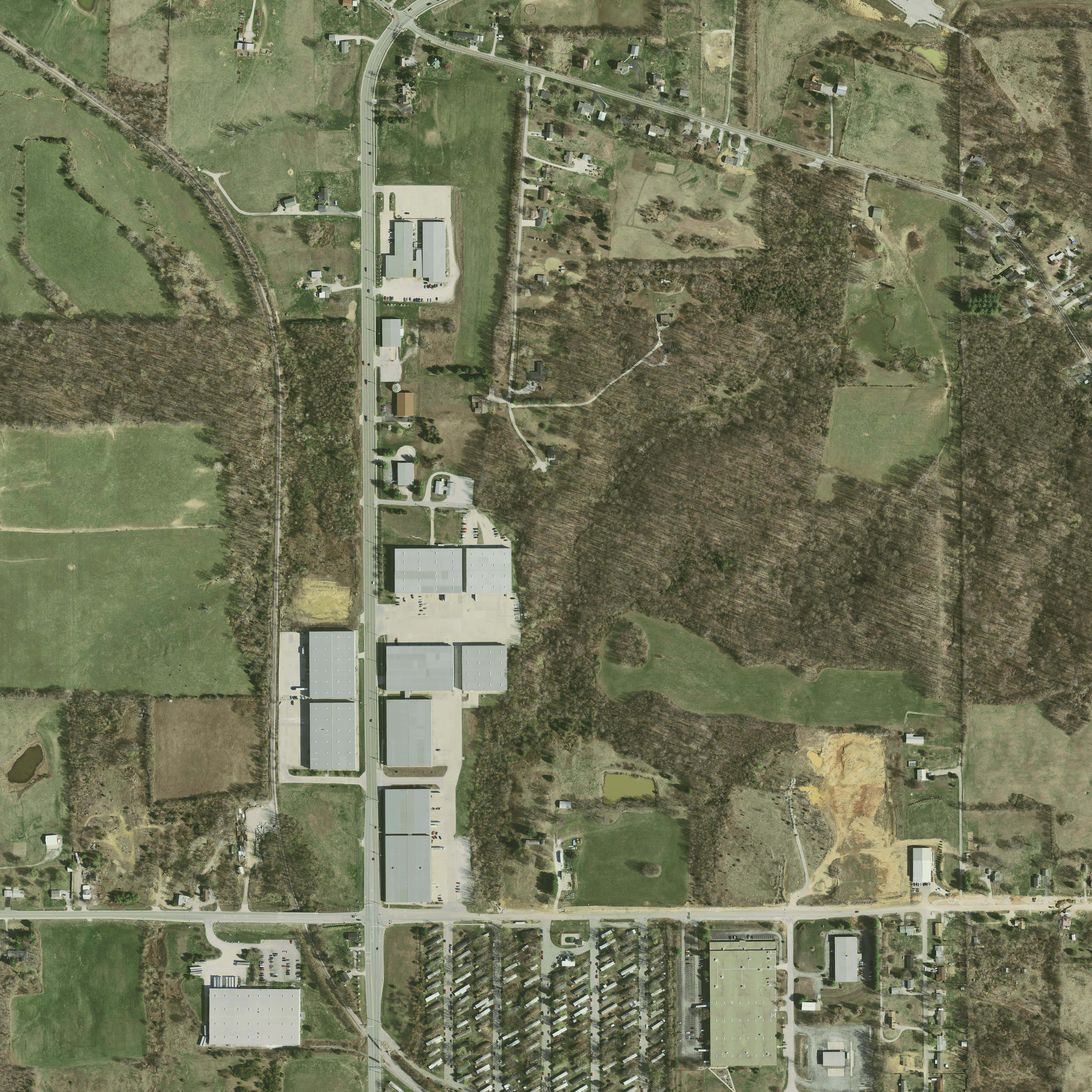} &
    \includegraphics[width=3.2cm]{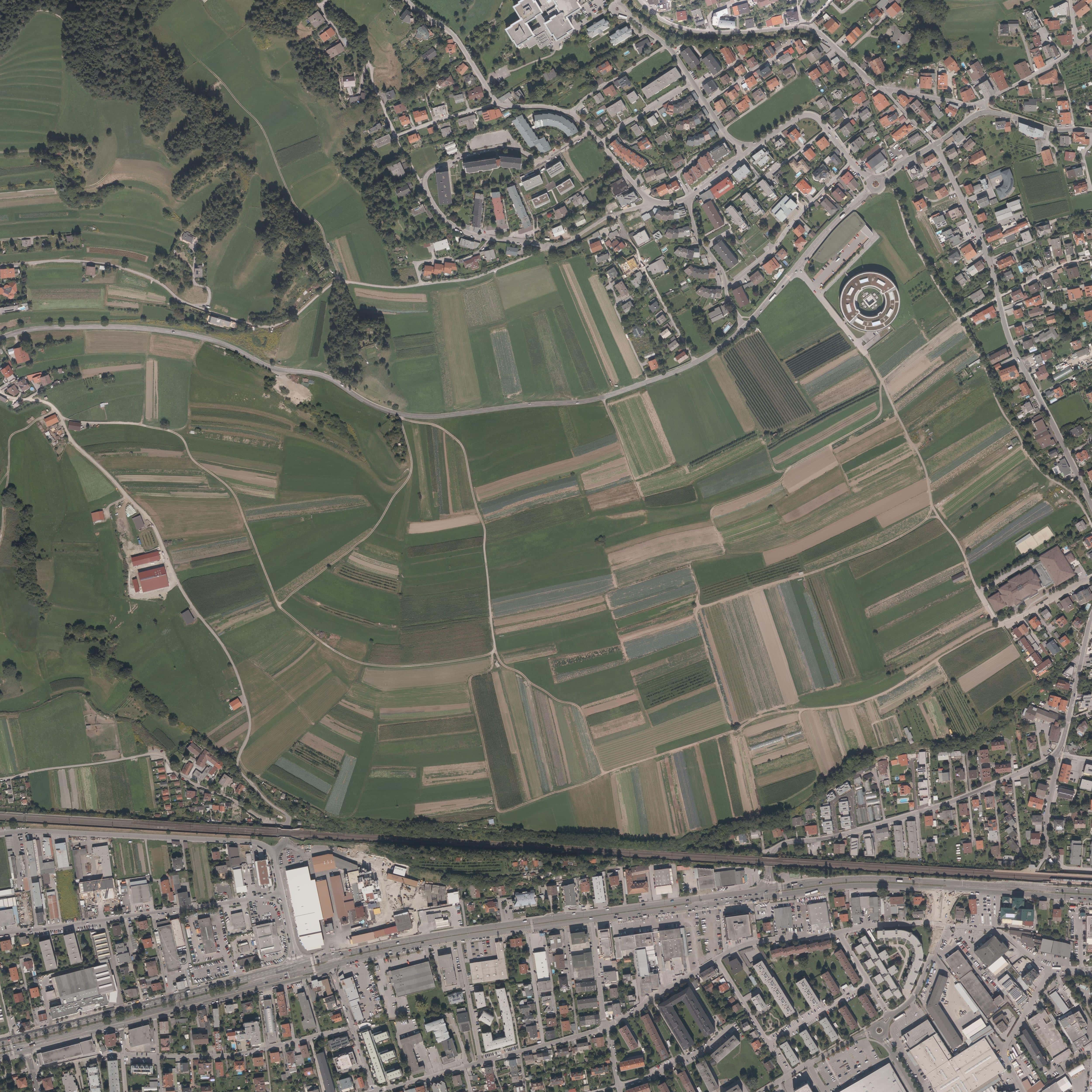} & 
    \includegraphics[width=3.2cm]{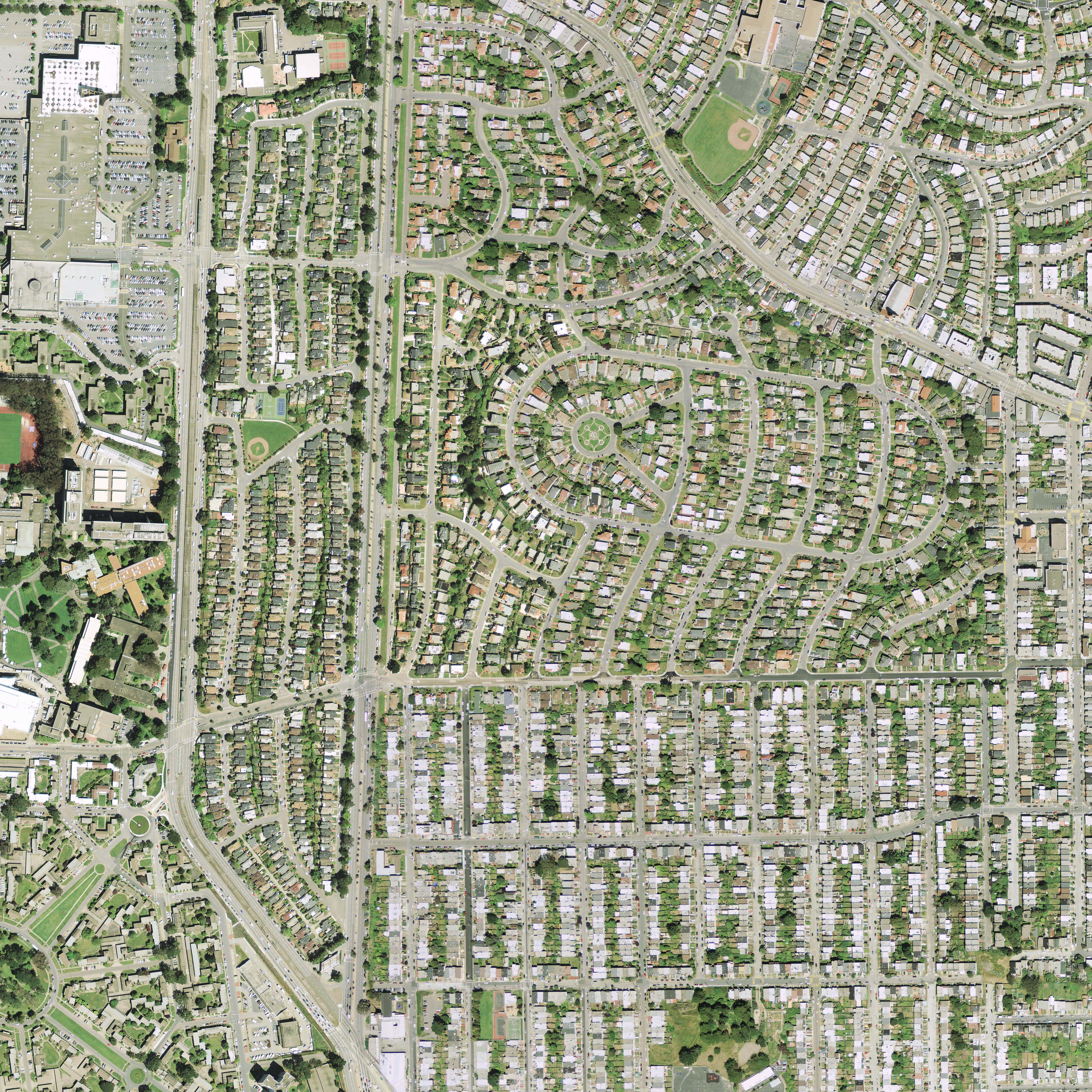} &
    \includegraphics[width=3.2cm]{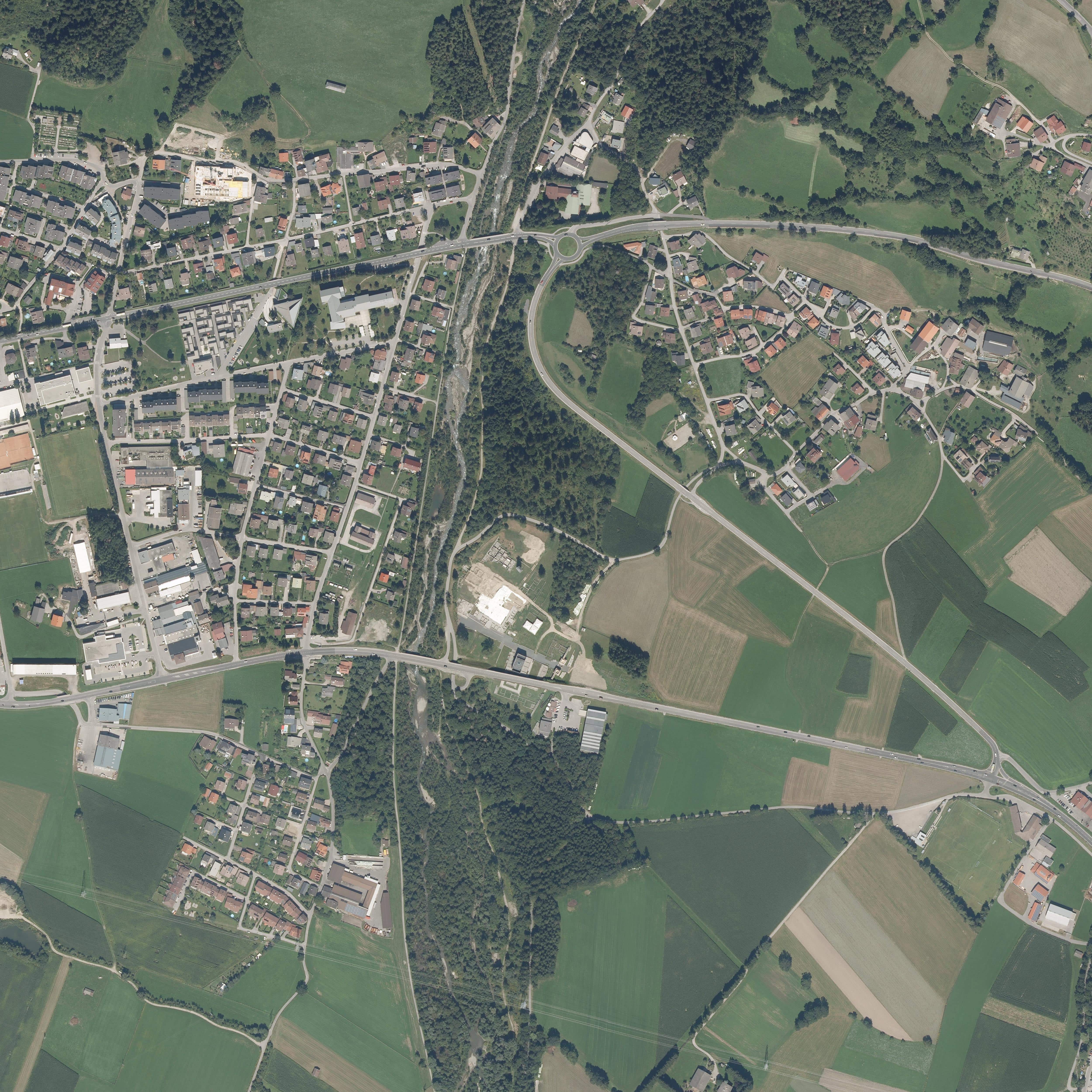}
    \\
    \\
    \includegraphics[width=3.2cm]{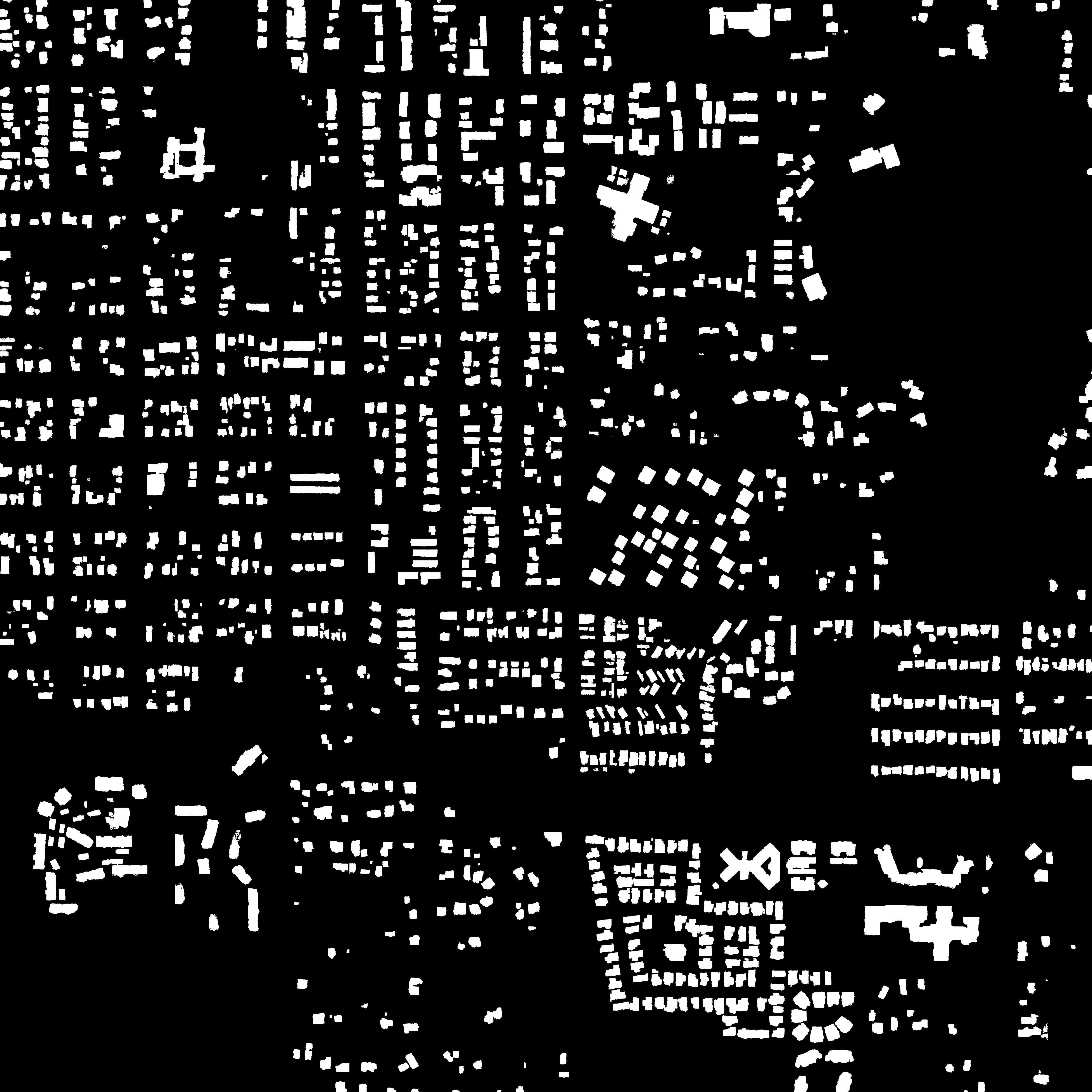} &
    \includegraphics[width=3.2cm]{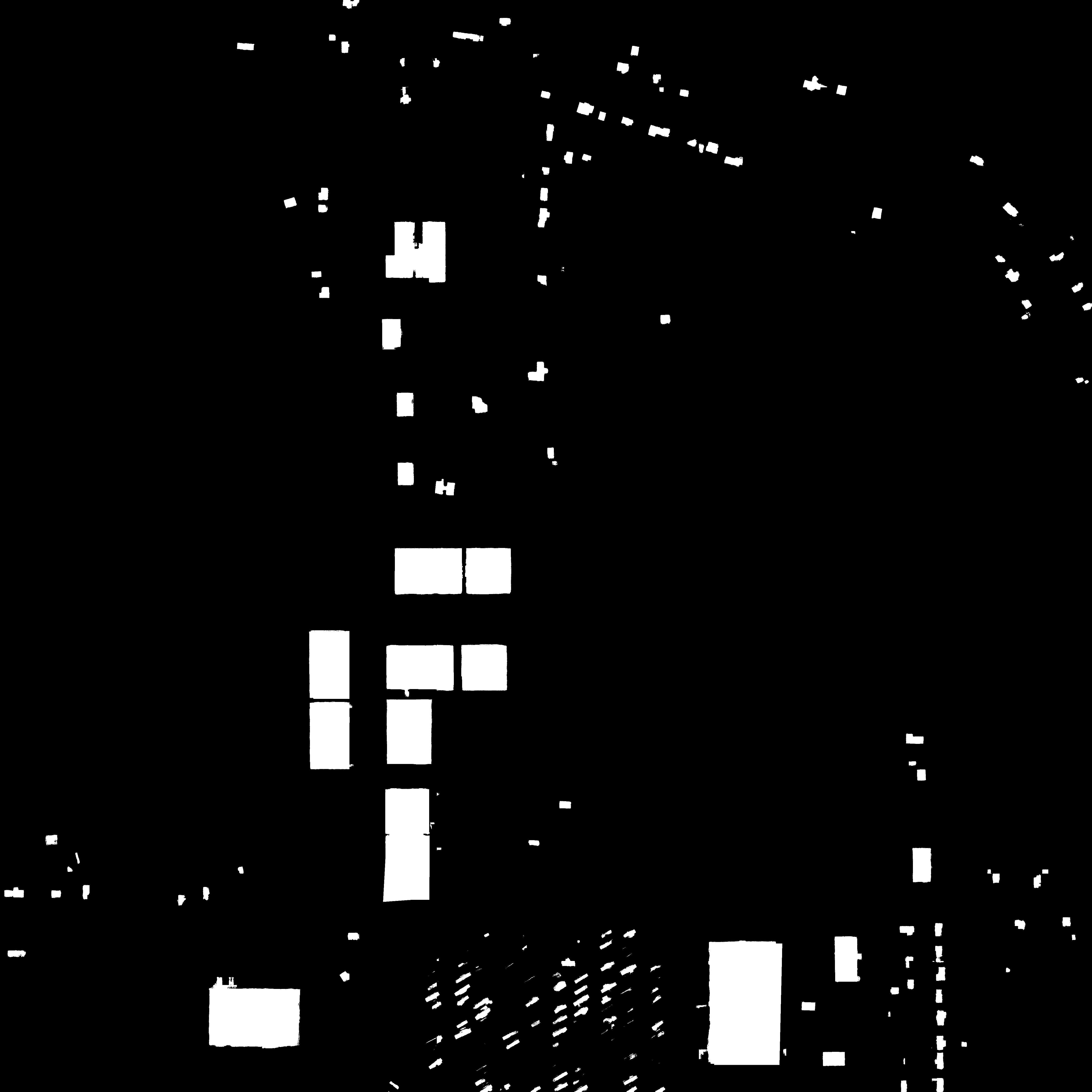} &
    \includegraphics[width=3.2cm]{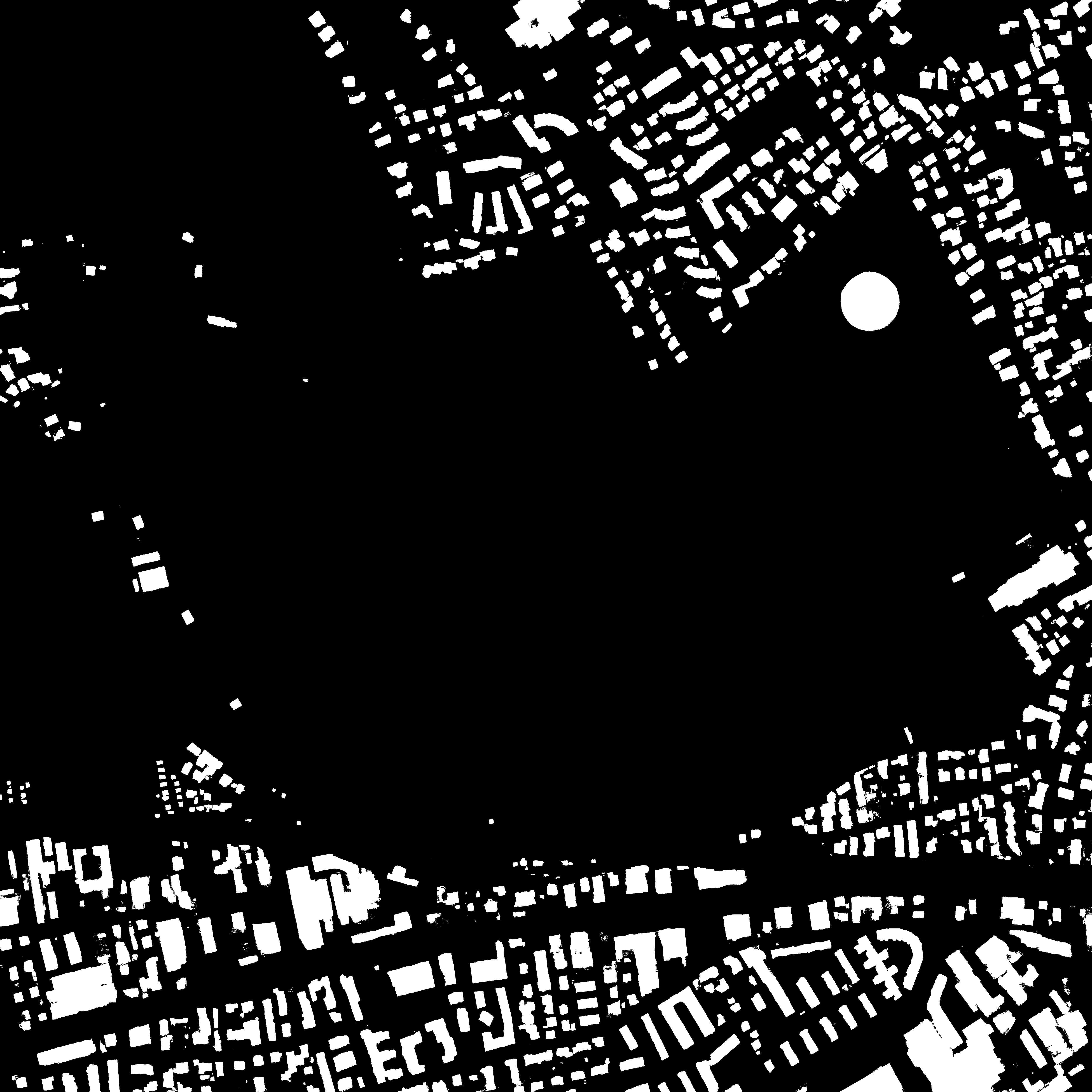} & 
    \includegraphics[width=3.2cm]{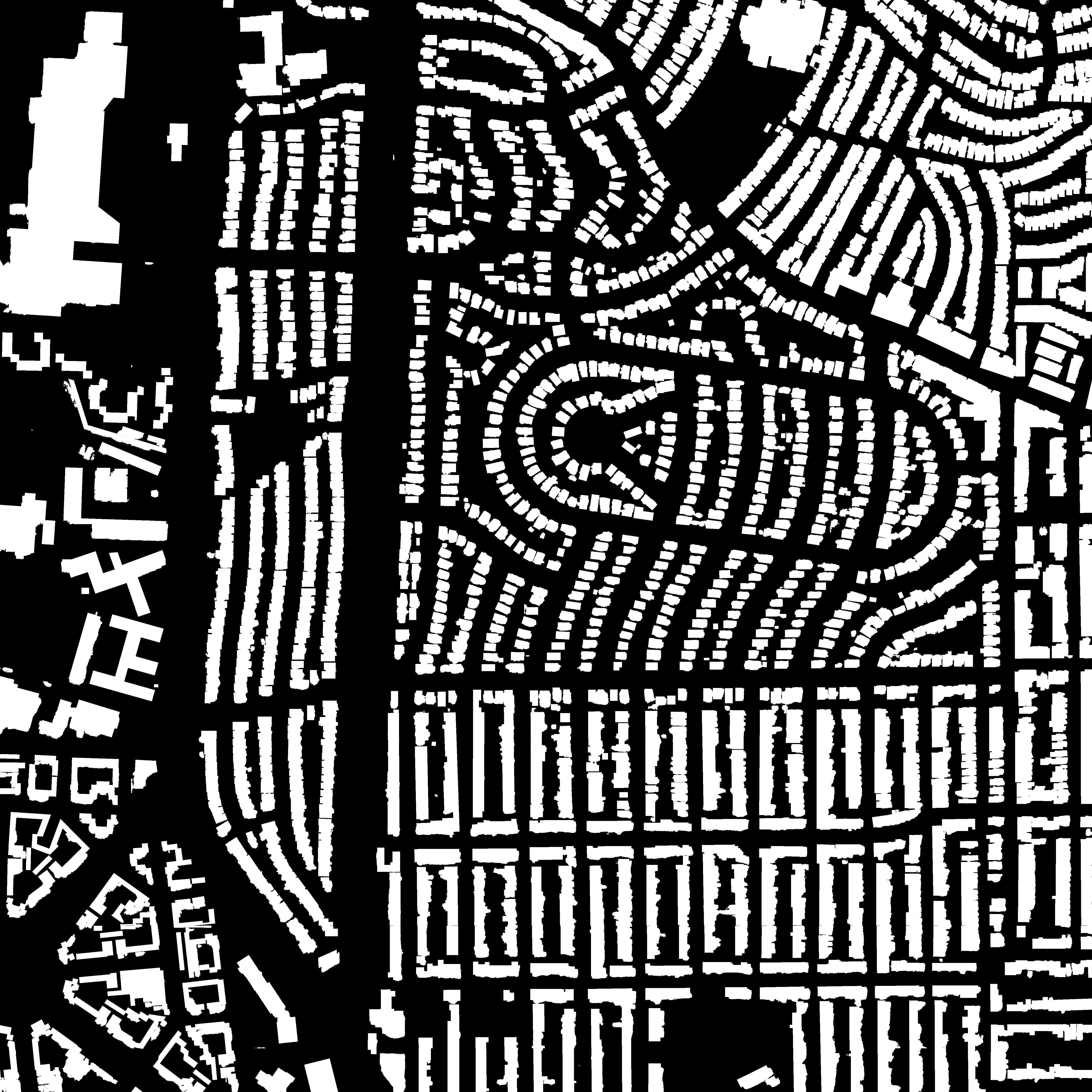} &
    \includegraphics[width=3.2cm]{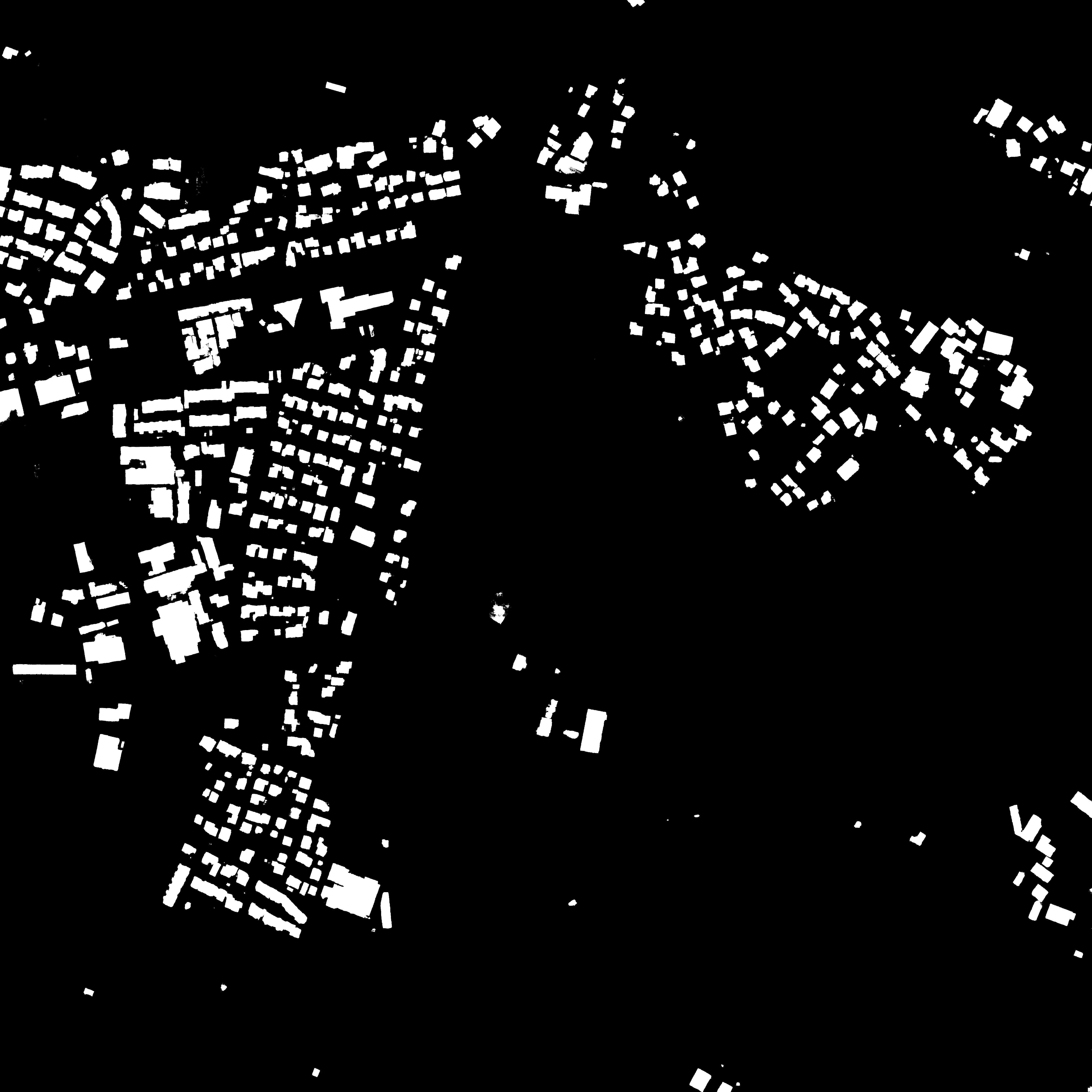}
\end{tabular}
\captionof{figure}{\centering Results on the INRIA Aerial Image Labeling Test Dataset. Row 1: Input image. Row 2: Predicted Label Map.}
\label{fig:seg_inriatest}
\end{table*}

\bgroup
\def\arraystretch{1.2}
\setlength\tabcolsep{0.05cm}
\begin{table*}[!ht]
\begin{center}
\resizebox{\linewidth}{!}{%
    \begin{tabular}{ | c | c | c | c | c| c | c | c | c| c | p{1.5cm} |}
    \hline
    Model & \multicolumn{1}{| c |}{Train Cities}& \multicolumn{1}{| c |}{Train IoU}& \multicolumn{1}{| c |}{Train Acc.}& \multicolumn{1}{| c |}{Val. City}& \multicolumn{1}{| c |}{Val. IoU} & \multicolumn{1}{| c |}{Val. Acc.} & \multicolumn{1}{| c |}{INRIA Val. Dataset IoU} & \multicolumn{1}{| c |}{INRIA Val. Dataset Acc.} \\
    \hline
    \hline
    \hline
    \hline
    Model 1 & Austin, Chicago, Kitsap, W. Tyrol & 80.26  & 96.01  & Vienna & 78.24 & 94.13 & 79.47 & 96.54 \\
    \hline
    Model 2 & Austin, Chicago, Kitsap, Vienna & 81.86 & 96.74 &  W. Tyrol  & 79.32 & 98.29 & 79.15 & 97.23\\
    \hline
    Model 3 & Austin, Chicago, W. Tyrol, Vienna & 82.93 & 94.11 & Kitsap & 70.26 & 99.22 & 81.74 & 97.14\\
    \hline
    Model 4 & Austin, Kitsap, W. Tyrol, Vienna & 82.26  & 95.03 &  Chicago  & 72.63 & 92.46 & 82.97 & 95.38\\
    \hline
    Model 5 &  Chicago, Kitsap, W. Tyrol, Vienna & 79.66 & 95.29 &  Austin & 80.29 & 96.78 & 77.45 & 96.37\\
    \hline
    \end{tabular}}
\end{center}
\caption{\centering Comparison of different models in our ensemble of k-fold training on the training and validation subsets of the INRIA Aerial Image Labeling Dataset. Val.: Validation. Acc.: Accuracy}
\label{tab:kfold}
\end{table*}
\egroup
\bgroup

\bgroup
\def\arraystretch{1.2}
\setlength\tabcolsep{0.02cm}
\begin{table*}[!ht]
\begin{center}
\resizebox{\linewidth}{!}{%
    \begin{tabular}{ | c | c | c | c | c| c | c | c | p{1.5cm} |}
    \hline
    Method & \multicolumn{1}{| c |}{Evaluation Metrics}& \multicolumn{1}{| c |}{Austin}& \multicolumn{1}{| c |}{Chicago}& \multicolumn{1}{| c |}{Kitsap}& \multicolumn{1}{| c |}{W. Tyrol}& \multicolumn{1}{| c |}{Vienna} & \multicolumn{1}{| c |}{Overall}\\
    \hline
    \hline
    \hline
    FCN (baseline) \cite{maggiori2017dataset} 
    & (IoU, Accuracy)  & (47.66, 92.22) & (53.62, 88.59) & (33.70, 98.58 ) & (46.86, 95.83) & (60.60, 88.72) & (53.82, 92.79)\\
    \hline
    MLP (baseline) \cite{maggiori2017dataset} & (IoU, Accuracy) & (61.20, 94.20) & (61.30, 90.43 ) & (51.50, 98.92 ) & (57.95, 96.66) & (72.13, 91.87 ) & (64.67, 94.42) \\
    \hline
    Mask R-CNN \cite{pan2019building} & (IoU, Accuracy) & (65.63, 94.09) &  (48.07, 85.56) & (54.38, 97.32) & (70.84, 98.14 ) & (64.40, 87.40) &  (59.53, 92.49) \\
    \hline
    MSMT-Stage-1 \cite{marcu2018multi} & (IoU, Accuracy) & (75.39, 95.99) & (67.93, 92.02) & (66.35, 99.24 ) & (74.07, 97.78) & (77.12, 92.49) & (73.31, 96.06) \\ 
    \hline
    SegNet+Multi-Task Loss \cite{bischke2019multi} & (IoU, Accuracy) & (72.43, 95.71 ) & (77.68, 95.60) & (72.28, 95.81) & (64.34, 98.76) & (76.15, 94.48) & (74.49, 96.07) \\
    \hline
    2-levels U-Nets \cite{khalel2018automatic} & (IoU, Accuracy) & (77.29, 96.69) & (68.52, 92.40) & (72.84, 99.25) & (75.38, 98.11) & (78.72, 93.79) & (74.55, 96.05) \\
    \hline
    U-Net \cite{pan2019building} & (IoU, Accuracy) & (79.95, 97.10) & (70.18, 92.67) & (68.56, 99.31) & (76.29, 98.15 ) & (79.92, 94.25 ) & (76.16, 96.31) \\
    \hline
    GMEDN \cite{ma2020building} & (IoU, Accuracy) & (80.53, 97.19) & (70.42, 92.86) & (68.47, 99.30 ) & (75.29, 98.05) & (80.72, 94.54) & (76.69, 96.43) \\ 
    \hline
    GAN-SCA \cite{pan2019building} & (IoU, Accuracy) & (81.01, 97.26) & (71.73, 93.32) & (68.54, 99.30 ) & (78.62, 98.32) & (81.62, 94.84) & (77.75, 96.61) \\ 
    \hline
    SEResNeXt101-FPN-CPA \cite{sebastian2020contextual} & (IoU, Accuracy) & (80.15, 97.18) & (69.54, 92.78) & (70.36, 99.32 ) & (80.83, 98.46) & (81.43, 94.67) & (77.29, 96.48) \\ 
    \hline
    Building-A-Nets \cite{li2018building} & (IoU, Accuracy) & (80.14, 96.91) & (\textbf{79.31}, \textbf{97.06}) & (72.77, 96.99 ) & (74.55, 93.52) & (75.71, \textbf{98.09}) & (78.73, 96.71) \\ 
    \hline
    %Our Method (no TTA) & (IoU, Accuracy)  & (82.97, 97.67) & (75.77, 94.45) & (72.96, 99.19) & (84.68, 98.82) & (82.78, 94.91) & (80.24, 96.89)
    %\\  
    %\hline
    Our Method & (IoU, Accuracy)  & (\textbf{83.78}, \textbf{97.75}) & (76.39, 94.83) & (\textbf{73.25}, \textbf{99.37}) & (\textbf{85.72}, \textbf{98.91}) & (\textbf{83.19}, 95.09) & (\textbf{81.28}, \textbf{97.03})
    \\  
    \hline
    \end{tabular}}
\end{center}
\caption{Comparison of state-of-the-art networks for the INRIA Validation Dataset. The best results are highlighted in bold.  }%TTA: Test Time Augmentation.}
\label{tab:results2}
\end{table*}
\egroup
\bgroup

\bgroup
\def\arraystretch{1.2}
\setlength\tabcolsep{0.01cm}
\begin{table*}[!ht]
\begin{center}
\resizebox{\linewidth}{!}{%
    \begin{tabular}{ | c | c | c | c | c | c| c | c | c | p{1.5cm} |}
    \hline
    Method & \multicolumn{1}{| c |}{Evaluation Metrics}& \multicolumn{1}{| c |}{Bellingham}& \multicolumn{1}{| c |}{Bloomington}& \multicolumn{1}{| c |}{Innsbruck}& \multicolumn{1}{| c |}{San Francisco}& \multicolumn{1}{| c |}{East Tyrol} & \multicolumn{1}{| c |}{Overall}\\
    \hline
    \hline
    \hline
    \hline
    Building-A-Nets \cite{li2018building} & (IoU, Accuracy) & (65.50, 96.39)	&(66.63, 96.85)	&(72.59, 96.73)	&(76.14, 91.96)	&(71.86, 97.48)	&(72.36, 95.88) \\
    \hline
    U-Net-ResNet101	 \cite{girard2021polygonal}& (IoU, Accuracy) & 	(69.75,	96.77) & (72.04, 97.13)	& (74.64, 96.83)	& (74.55, 91.14)	& (77.40, 97.92)	& (73.91, 95.96)  \\ 
    \hline
    Zorzi et al. \cite{zorzi2021machine} & (IoU, Accuracy)  & (70.36, 96.99) & (73.01, 97.36) & (73.34, 96.77) & (75.88, 91.55) & (76.15, 97.84) & (74.40, 96.10) \\
    \hline
    DS-Net \cite{liao2020learning}& (IoU, Accuracy)  & (71.74,	97.22)	&(70.55, 97.27)	&(75.44, 97.11)	&(77.26, 92.47)	&(78.54, 98.10) & (75.52, 96.43) \\
    \hline
    Zhang et al. \cite{zhang2020local}& (IoU, Accuracy) & (72.25, 97.25) &	(72.49, 97.41)	& (75.21, 97.07) &	(77.70, 92.54) & (78.06, 98.04) &	(75.94,	96.46) \\
    \hline
    Milosavljevic et al. \cite{milosavljevic2020automated}& (IoU, Accuracy) & (73.90, 97.35) &	(72.97,	97.39) &	(77.31, 97.32) & (76.46, 92.01) &	(80.41,	98.23)	& (76.27,	96.46) \\ 
    \hline
    E-D-Net \cite{zhu2021ed}& (IoU, Accuracy) & (73.12, 97.22) & (75.58, 97.64) & (77.66, 97.31) & (79.81, 93.26) & (80.61, 98.25) & (78.08, 96.73) \\
    \hline
    ICT-Net \cite{chatterjee2021semantic} & (IoU, Accuracy) & (\textbf{74.63}, \textbf{97.47}) & (\textbf{80.80}, \textbf{98.18}) & (\textbf{79.50}, \textbf{97.58}) & (\textbf{81.85}, \textbf{94.08}) & (\textbf{81.71}, \textbf{98.39}) & (\textbf{80.32}, \textbf{97.14}) \\ 
    \hline
    Our Method & (IoU, Accuracy)  & (74.41, 97.03)  & (77.29, 97.64)  & (76.93, 96.70) & (76.82, 90.49) & (80.11, 98.16) & (77.86, 96.41)
    \\  
    \hline
    \end{tabular}}
\end{center}
\caption{\centering Comparison of our framework with other
  state-of-the-art approaches on the test set of the INRIA
  Aerial Image Labeling Dataset. The best results are
  highlighted in bold.}
\label{tab:inriatest}
\end{table*}
\egroup
\bgroup

As mentioned in Section~\ref{sec:datacreate}, we adopt a
k-fold validation strategy for training our network on the
INRIA Dataset. In our experiments, k = 5. In
Table~\ref{tab:kfold}, we report the training as well as the
validation IoU and accuracy scores of these 5 models. We
also report the overall performance of each model on the
INRIA Validation Dataset.

In Table~\ref{tab:results2}, we compare the result of our
framework with some of the state-of-the-art approaches on
the INRIA Validation Dataset. Specifically, we report the
IoU and accuracy scores for the different methods.  Since
the dataset comes with a disproportionately large number of
true negatives for the background images, the accuracy
numbers achieved with this dataset are generally high, as
can be seen by the entries for accuracy in
Tables~\ref{tab:kfold}-\ref{tab:inriatest}.  On the other
hand, since the IoU metric takes into account both the false
alarms and missing detections, we believe that that is a
better metric of performance on this dataset.

For the individual cities, as shown in
Table~\ref{tab:results2}, we have highlighted the highest
valued entries for each of the two evaluation metrics. Our
network achieves performance improvement of at least 3.42\%,
0.56\%, 6.05\% and 1.92\% over Austin, Kitsap, W. Tyrol and
Vienna respectively. Our network also gives better accuracy
for Austin, Kitsap and W. Tyrol.  For Chicago, though our
IoU and accuracy are smaller than \cite{li2018building} by
3.82\% and 2.35\% respectively, overall our algorithm
outperforms \cite{li2018building} as well as other
state-of-the-art methods by at least 3.24\% and 0.33\% in
terms of IoU and accuracy respectively.

These results show that our network gives consistently good
performance over all the cities in the INRIA Validation
Dataset, while also yielding the best performance for a
subset of the cities. Figures~\ref{fig:seg_inria1} and
\ref{fig:seg_inriatest} illustrate some of our building
segmentation results on the INRIA Validation and Test
Dataset.

In Table~\ref{tab:inriatest}, we compare the performance of
our framework with some other state-of-the-art methods on
the official INRIA Test Dataset. Though we do not achieve
best scores on this subset, our performance is pretty
competitive with the state-of-the art methods. Most of the
state-of-the-art methods that perform better than us on the
INRIA Test Dataset either use pretrained feature extraction
networks \cite{krizhevsky2012imagenet, simonyan2014very} as
backbones or are significantly deeper than our proposed
network. This shows effective generalization capability of
our network. Notice the drop in both the accuracy and IoU
values when applying the trained network to a set of
different geographic areas. This is to be expected, since
each city has some unique specifics.

\subsection{Quantitative Evaluation on the WHU Building Dataset}

In Table~\ref{tab:whu}, we report the IoU, precision, recall and F1-scores obtained using our proposed algorithm on the WHU test dataset and compare these scores with some of the best performing state-of-the-art building segmentation approaches. %Our method without TTA achieves 91.68\% IoU, 96.41\% precision, 94.92\% recall and 95.66\% F1 score, respectively. As can be seen from Table~\ref{tab:whu}, our proposed approach without TTA performs very similar to the previous best scoring algorithm -- ARC-Net \cite{liu2020arc} by Liu et al. 
As can be seen from Table~\ref{tab:whu}, our proposed method outperforms the previous best scoring algorithm (ARC-Net \cite{liu2020arc}) by 0.51\%, 0.34\%, 0.15\% and 0.29\% in IoU, precision, recall and F1-score respectively. Figure~\ref{fig:seg_whu} illustrates some qualitative results of our algorithm on the WHU dataset. The last column in the figure shows the high degree of completeness (i.e. high number of true positives and true negatives, very few false positives and false negatives) in our segmentation results.
\begin{figure*}
    \centering
    \subfloat{{\includegraphics[width=3.8cm]{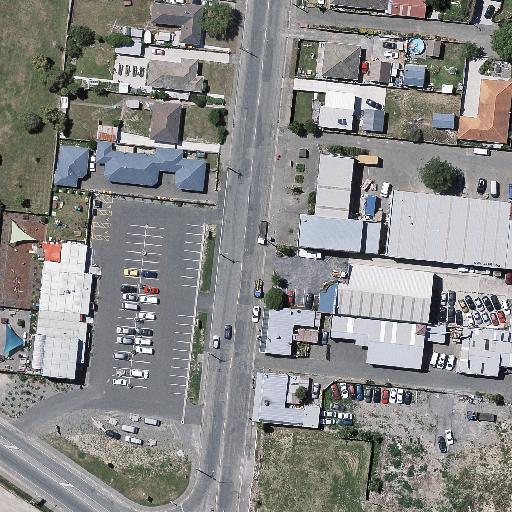} }}
    \subfloat{{\includegraphics[width=3.8cm]{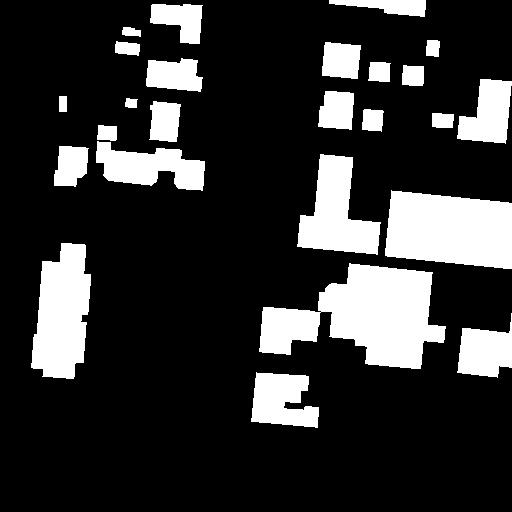} }}
    \subfloat{{\includegraphics[width=3.8cm]{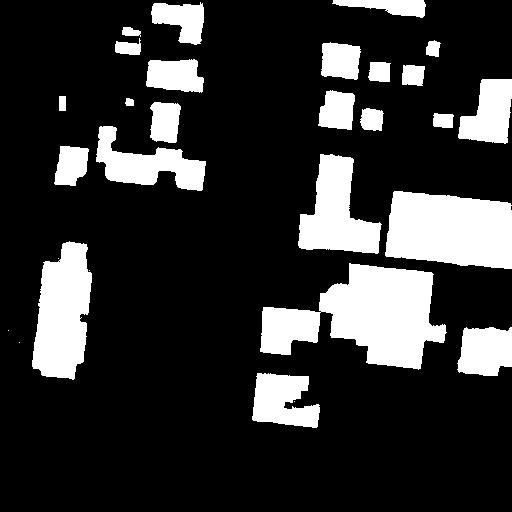} }}
    \subfloat{{\includegraphics[width=3.8cm]{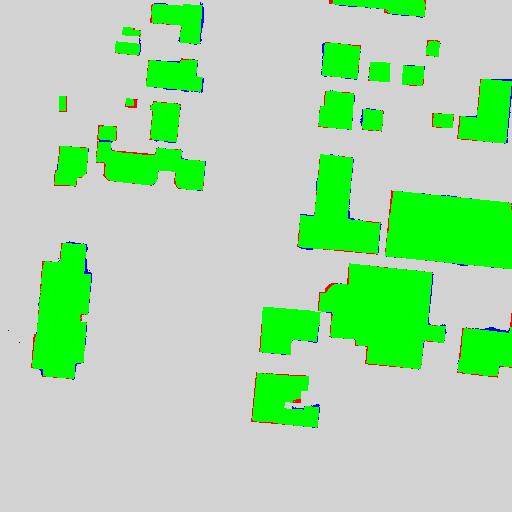} }}
    \hfill
    \subfloat{{\includegraphics[width=3.8cm]{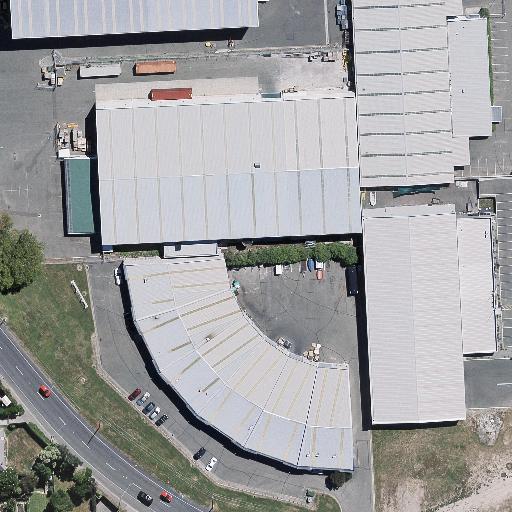} }}
    \subfloat{{\includegraphics[width=3.8cm]{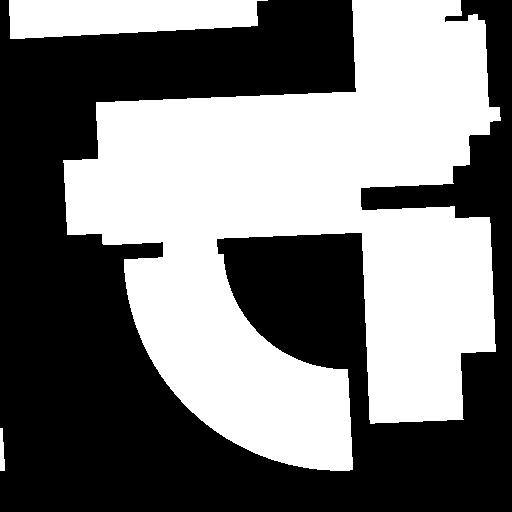} }}
    \subfloat{{\includegraphics[width=3.8cm]{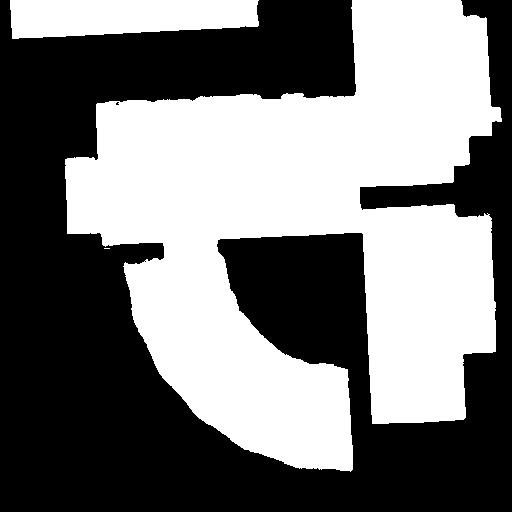} }}
    \subfloat{{\includegraphics[width=3.8cm]{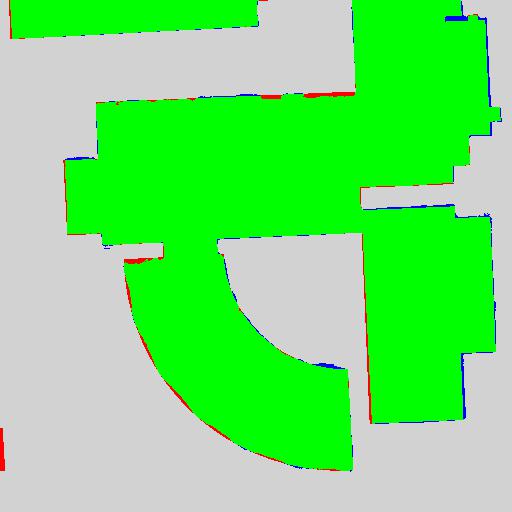} }}
    \hfill
    \subfloat{{\includegraphics[width=3.8cm]{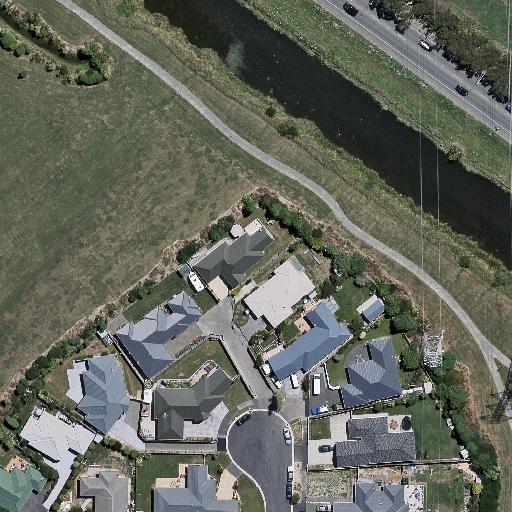} }}
    \subfloat{{\includegraphics[width=3.8cm]{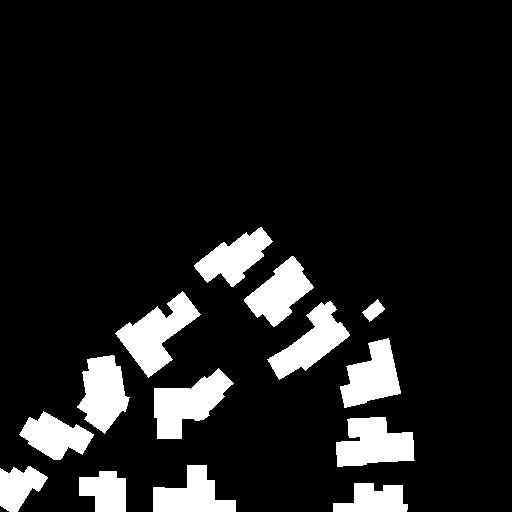} }}
    \subfloat{{\includegraphics[width=3.8cm]{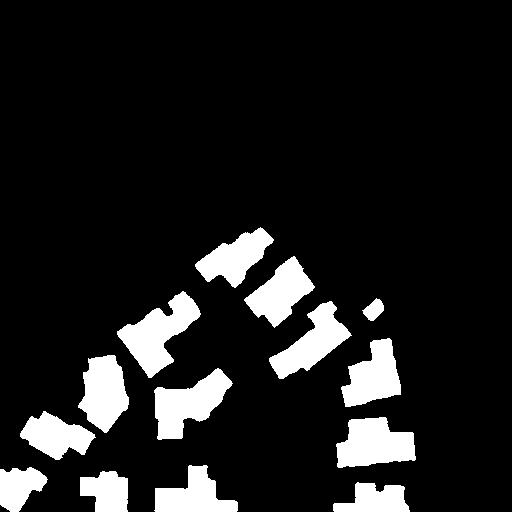} }}
    \subfloat{{\includegraphics[width=3.8cm]{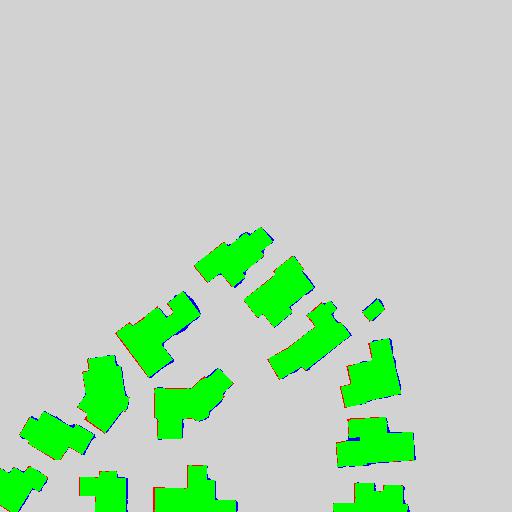} }}
    \hfill
    \subfloat{{\includegraphics[width=3.8cm]{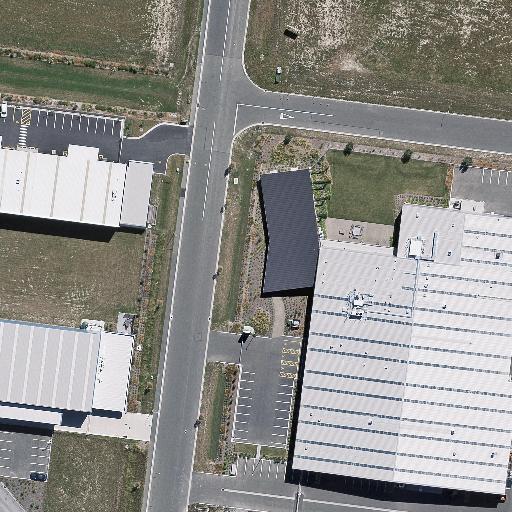} }}
    \subfloat{{\includegraphics[width=3.8cm]{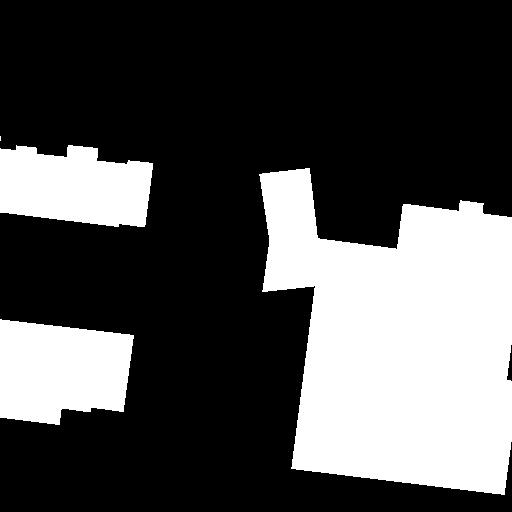} }}
    \subfloat{{\includegraphics[width=3.8cm]{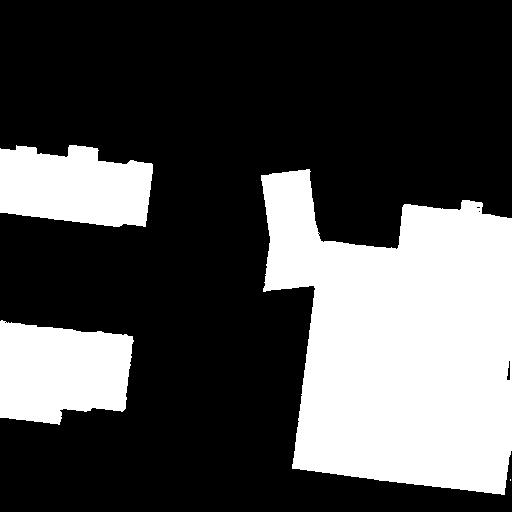} }}
    \subfloat{{\includegraphics[width=3.8cm]{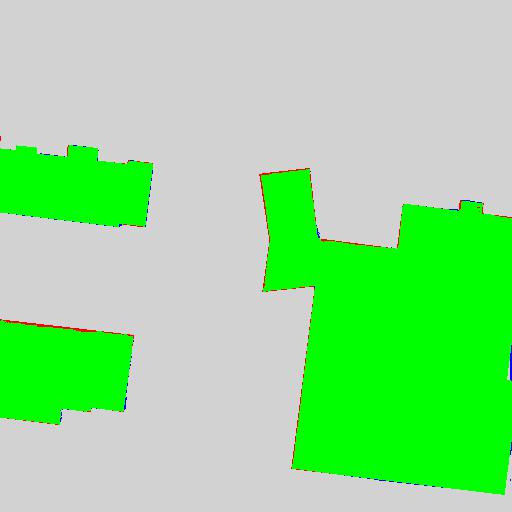} }}
    \caption{\centering Results on the WHU Building Dataset. Column 1: Input image. Column 2: Ground-truth Label Map. Column 3: Predicted Label Map. Column 4: Green: True Positives;
    Blue: False Positives; Red: False Negatives;
    Grey: True Negatives.}
  \label{fig:seg_whu}
\end{figure*}

\bgroup
\def\arraystretch{1.2}
\setlength\tabcolsep{0.65cm}
\begin{table*}[b]
%\small
\begin{center}
\resizebox{0.7\linewidth}{!}{%
    \begin{tabular}{ | c | c | c | c | c | c | p{1.5cm} |}
    \hline
    Method & \multicolumn{1}{| c |}{IoU} & {Precision} & {Recall} & {F1}\\
    \hline
    \hline
    \hline
    BRRNet \cite{chen2021dr, shao2020brrnet} & 85.9 & 93.5 & 91.3 & 92.4 \\
    \hline
    DRNet \cite{chen2021dr} & 86.0 & 92.7 & 92.2 & 92.5 \\ 
    \hline
    RefineNet \cite{lin2017refinenet, he2021boundary} & 86.9 & 93.7 & 92.3 & 93.0 \\
    \hline
    PISANet \cite{zhou2020robust} & 87.97 & 94.20 & 92.94 & 93.55 \\
    \hline
    SiU-Net \cite{ji2018fully} & 88.4 & 93.8 & 93.9 & 93.8 \\
    \hline
    SRI-Net \cite{liu2019building} & 89.09 & 95.21 & 93.28 & 94.23\\
    \hline 
    BMFR-Net \cite{ran2021building} & 89.32 & 94.31 & 94.42 & 94.36 \\
    \hline
    Chen et al. \cite{chen2021self} & 89.39 & 93.25 & 95.56 & 94.4 \\
    \hline
    Res-U-Net \cite{xu2018building} & 89.46 & 94.29 & 94.53 & 94.43 \\
    \hline
    HRLinkNetv2 \cite{wu2021hrlinknet} & 89.53 & 94.56 & 94.40 & 94.48 \\
    \hline
    DeepLab v3 + \cite{guo2021scale} & 89.61 & 94.68 & 92.36 & 94.52 \\
    \hline
    DE-Net \cite{liu2019net} & 90.12 & 95.00 & 94.60 & 94.80\\
    \hline
    DS-Net2 \cite{guo2021scale} & 90.4 & 94.85 & 95.06 & 94.96 \\
    \hline
    He et al. \cite{he2021boundary} & 90.5 & 95.1 & 94.9 & 95.0 \\
    \hline
    MA-FCN \cite{wei2019toward} & 90.7 & 95.2 & 95.1 & 95.15\\
    \hline
    ARC-Net \cite{liu2020arc} & 91.8 & 96.4 & 95.1 & 95.70\\
    \hline
    %Our Method (no TTA) & 91.68 & 96.41 & 94.92  & 95.66  \\ 
    %\hline
    Our Method  & \textbf{92.27} & \textbf{96.73} & \textbf{95.24} & \textbf{95.98}\\  
    \hline
    \end{tabular}}
\end{center}
\caption{\centering IoU, Precision, Recall and F1-scores for the state-of-the-art networks on the WHU Building Dataset. The best results are highlighted in bold.} % TTA: Test Time Augmentation. 
\label{tab:whu}
\end{table*}
\egroup
\bgroup

\subsection{Quantitative Evaluation on the DeepGlobe Building Dataset}
Table~\ref{tab:dg} illustrates the quantitative performance of our proposed algorithm on the DeepGlobe Building Dataset. Our algorithm %without TTA achieves F1-scores of 0.895, 0.780, 0.679 and 0.607 over Vegas, Paris, Shanghai and Khartoum respectively; on applying TTA, the F1-scores improves to 
achieves F1-scores of 0.896, 0.785, 0.687 and 0.613 over Vegas, Paris, Shanghai and Khartoum respectively. We outperform the previous best (published)  F1-scores obtained by TernausNetV2 \cite{iglovikov2018ternausnetv2} by 0.56\%, 0.51\%, 1.03\% and 1.65\% over Vegas, Paris, Shanghai and Khartoum respectively. Overall, our algorithm outperforms the popular TernausNetV2 network by 0.81\%. 

To this end, we emphasize the fact that most of the state-of-the-art methods reported in Table~\ref{tab:dg} use multi-spectral information; whereas our algorithm uses only RGB images for building footprint extraction. We believe incorporating additional spectral information would further improve our algorithm's segmentation performance.

In addition to the state-of-the-art methods reported in Table~\ref{tab:dg}, several other papers \cite{liu2019improved, zhang2020local, jung2021boundary} have shown experimental results on the DeepGlobe Building Dataset. However, they have either chosen their own set of {\em test} images or have reported pixel-wise performance scores. In this paper, we report only those works which have reported object-wise performance scores on the test dataset provided by the original DeepGlobe 2018 Competition organizers during the development phase. 
\begin{table*}[b]
\centering
\begin{tabular}{cccc}
    Vegas & Paris  & Shanghai & Khartoum
    \\
    \includegraphics[width=4cm]{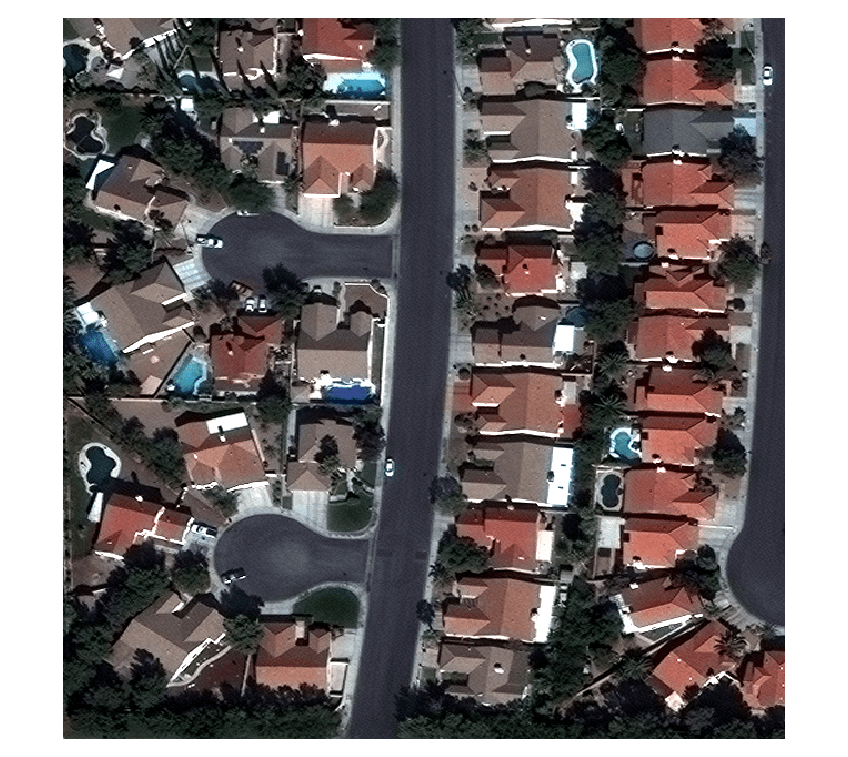} &
    \includegraphics[width=4cm]{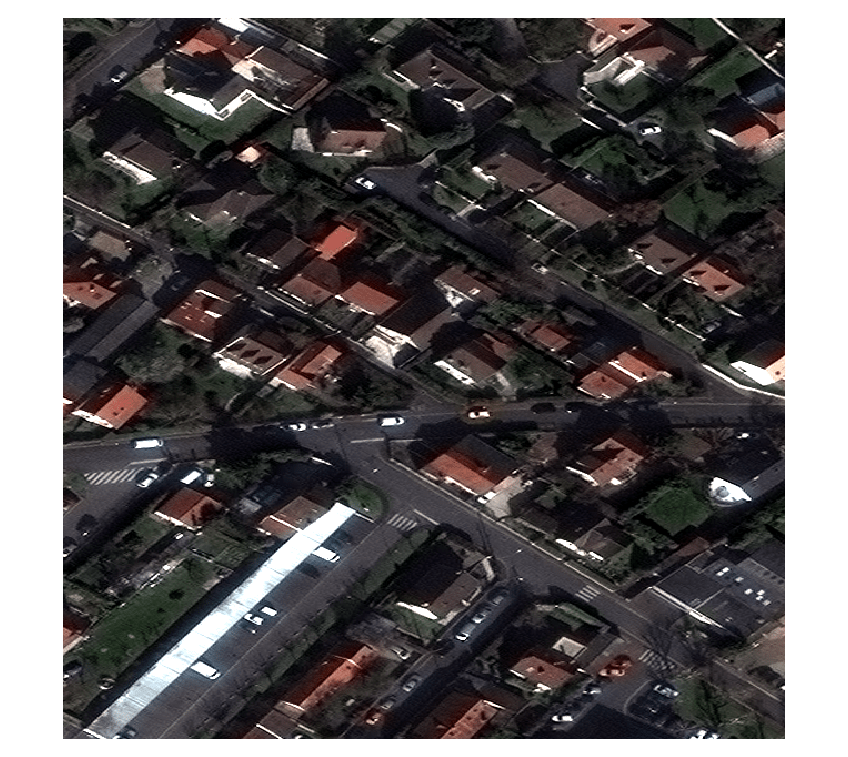} &
    \includegraphics[width=4.25cm]{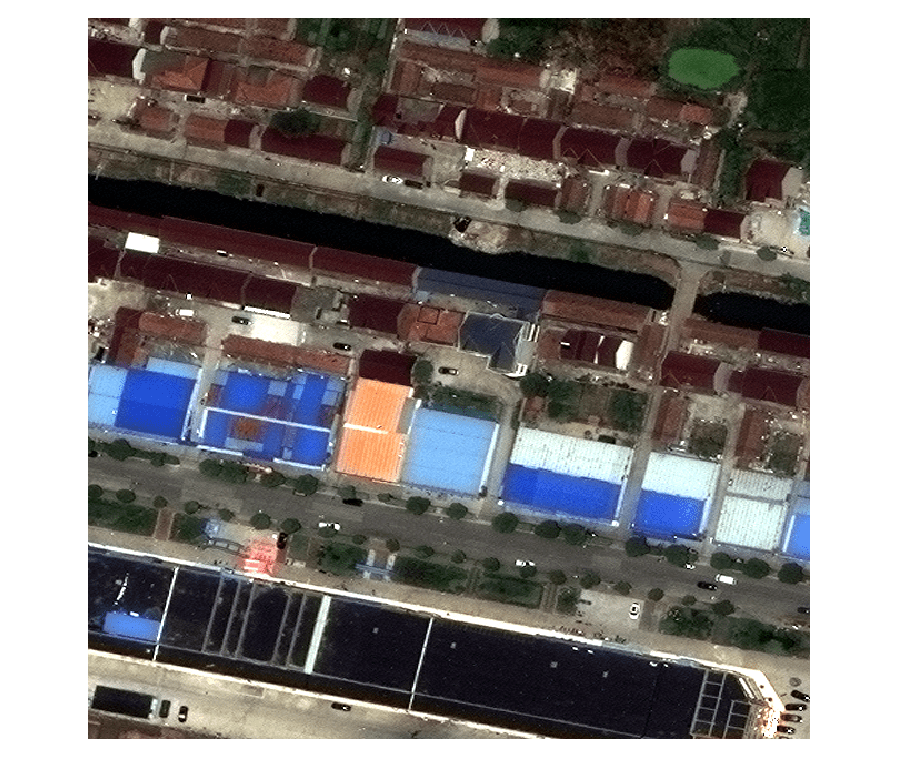} & 
    \includegraphics[width=4.22cm]{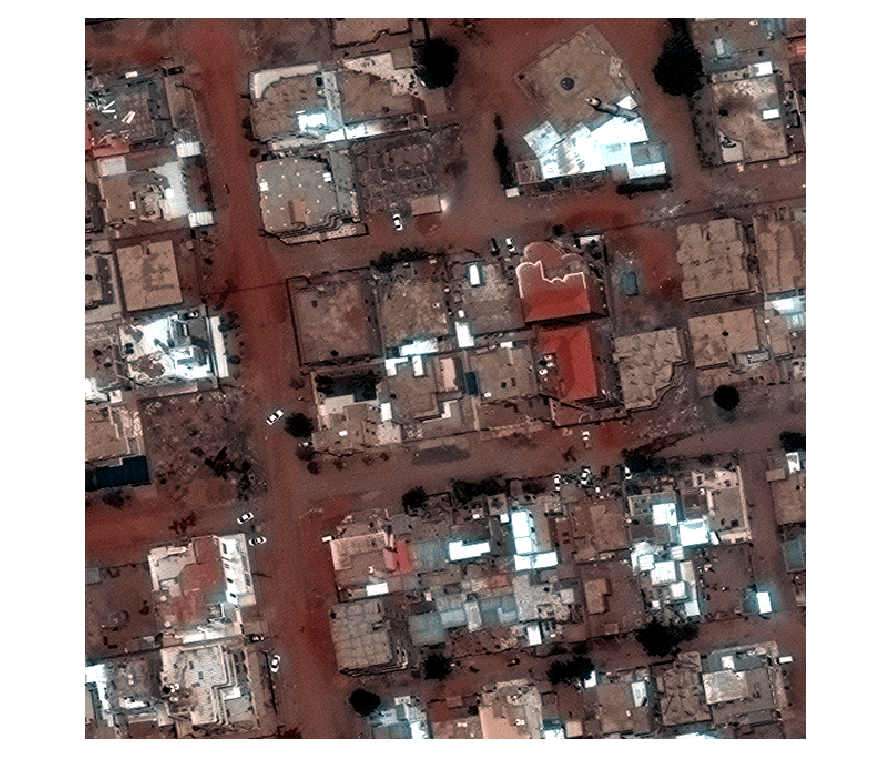} 
    \\
    \\
    \includegraphics[width=4cm]{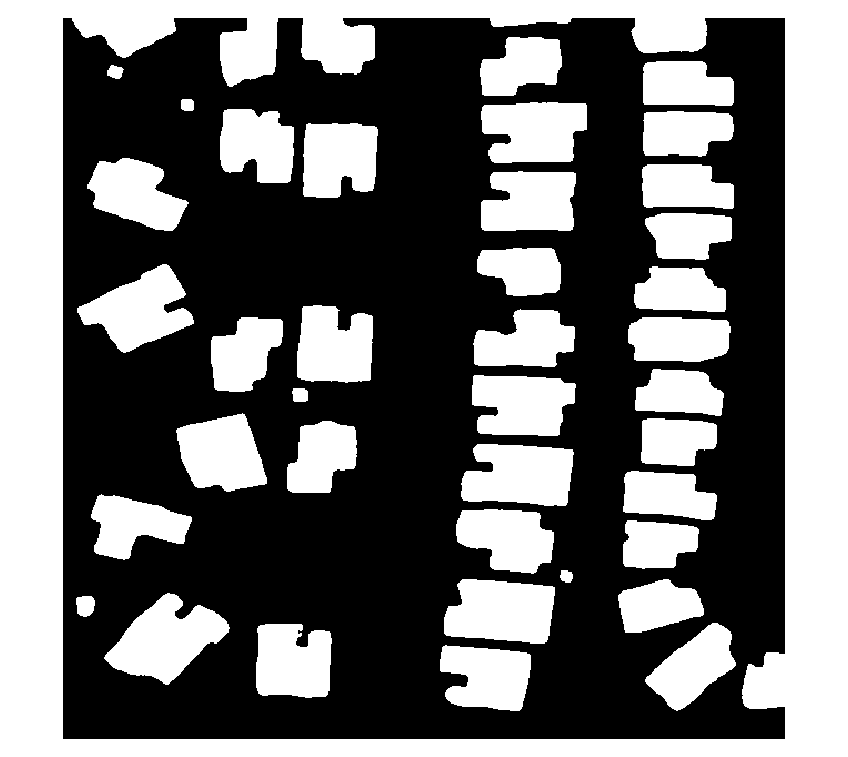} &
    \includegraphics[width=4cm]{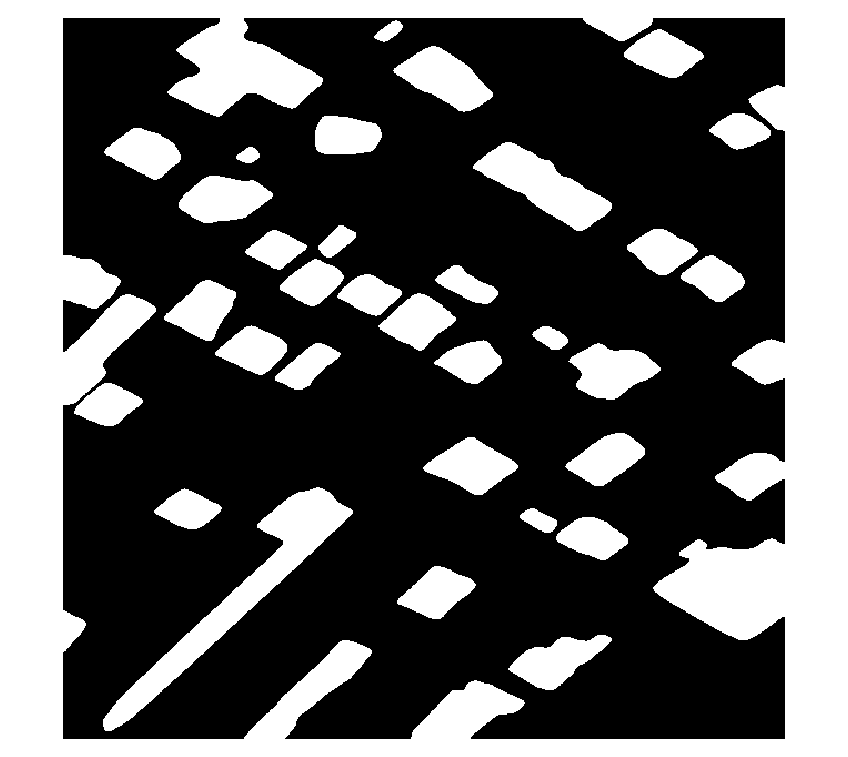} &
    \includegraphics[width=4.25cm]{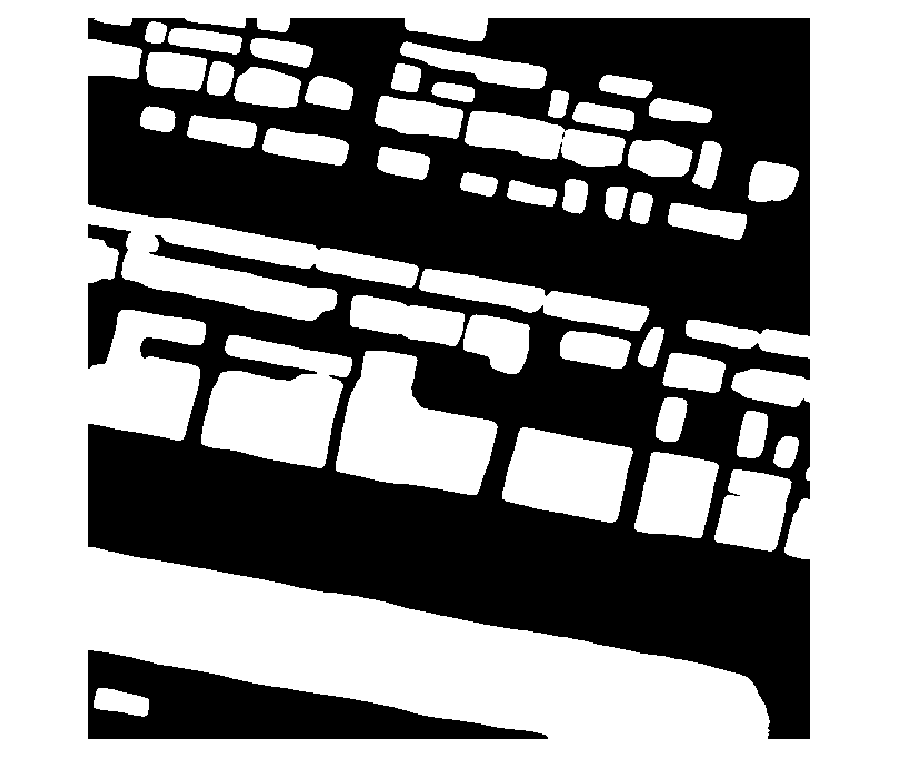} & 
    \includegraphics[width=4.2cm]{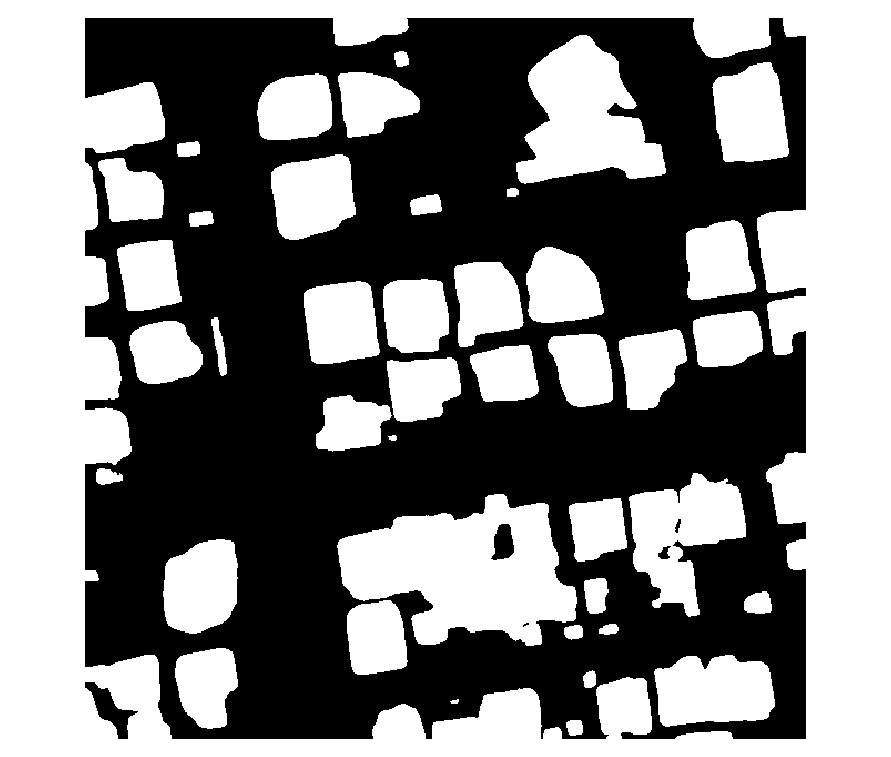} 
\end{tabular}
\captionof{figure}{\centering Qualitative results on the test subset of DeepGlobe Building Dataset. Row 1: Input image. Row 2: Predicted Label Map.}
\label{fig:seg_dg_test}
\end{table*}
\bgroup
\def\arraystretch{1.2}
\setlength\tabcolsep{0.5cm}
\begin{table*}[b]
%\small
\begin{center}
\resizebox{0.8\linewidth}{!}{%
    \begin{tabular}{ | c | c | c | c | c | c | c | p{1.5cm} |}
    \hline
    Method & \multicolumn{1}{| c |}{Vegas} & {Paris} & {Shanghai} & {Khartoum} & Overall\\
    \hline 
    \hline 
    \hline 
    %XD\_XD \cite{DeepGlobe18} & 0.885 & 0.745 & 0.597 & 0.544 & 0.693 \\
    %\hline 
    %wleitte \cite{DeepGlobe18} & 0.829 & 0.679 & 0.581 & 0.483 & 0.643 \\
    %\hline 
    %nofto \cite{DeepGlobe18} & 0.787 & 0.584 & 0.520 & 0.424 & 0.579 \\
    %\hline
    Li et al. \cite{li2018semantic} & 0.886 & 0.749 & 0.618 & 0.554 & 0.701 \\
    %\hline
    %Prathap et al. \cite{prathap2018deep} & 0.883 & 0.760 & 0.604 & 0.584 & - \\
    \hline
    Golovanov et al. \cite{golovanov2018building} & - & - & - & - & 0.707 \\
    \hline
    Zhao et al. \cite{Zhao_2018_CVPR_Workshops} & 0.879 & 0.753 & 0.642 & 0.568 & 0.713 \\
    \hline
    Hamaguchi et al \cite{Hamaguchi_2018_CVPR_Workshops} & - & - & - & - & 0.726 \\
    \hline
    TernausNetV2 \cite{iglovikov2018ternausnetv2} & 0.891 & 0.781 & 0.680 & 0.603 & 0.739 \\
    \hline
    Ali\_DI\_Deep\_Learning$^{**}$   & - & - & - & - & \textbf{0.749} \\
    \hline
    %Our Method (no TTA) & 0.895 & 0.780 & 0.679  & 0.607 &  0.740\\
    %\hline
    Our Method & \textbf{0.896} & \textbf{0.785} & \textbf{0.687}  & \textbf{0.613} &  0.745\\
    \hline
    \end{tabular}}
\end{center}
\caption{\centering F1-scores for the state-of-the-art networks on the test subset of DeepGlobe Building Dataset. The best results are highlighted in bold. $^{**}$Leading the DeepGlobe 2018 public leaderboard. Citation is unknown.}
\label{tab:dg}
\end{table*}
\egroup
\bgroup

\begin{table*}[ht]
\centering
\begin{tabular}{ccccc}
    Vegas & 
    \raisebox{-.5\height}{\includegraphics[width=3.7cm]{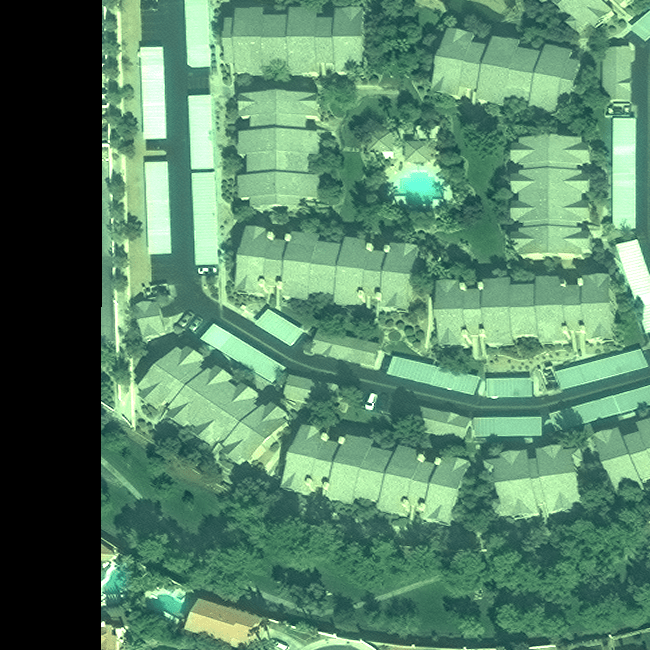}} & 
    \raisebox{-.5\height}{\includegraphics[width=3.7cm]{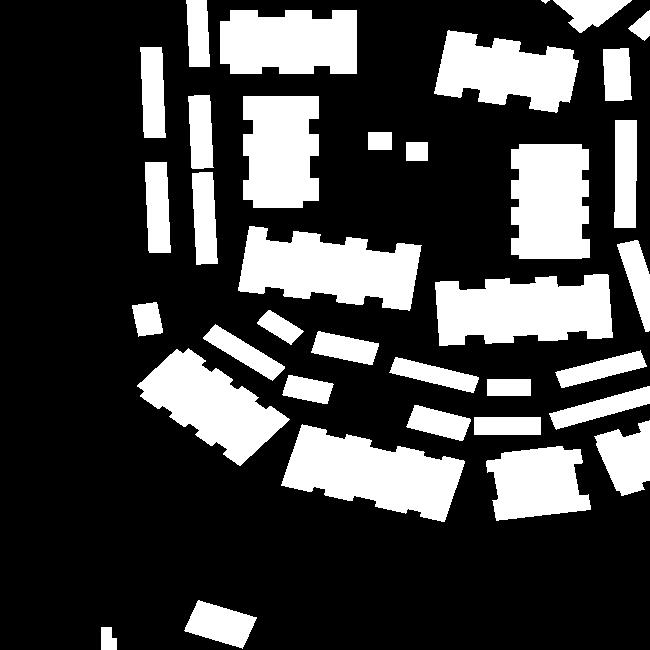}} &
    \raisebox{-.5\height}{\includegraphics[width=3.7cm]{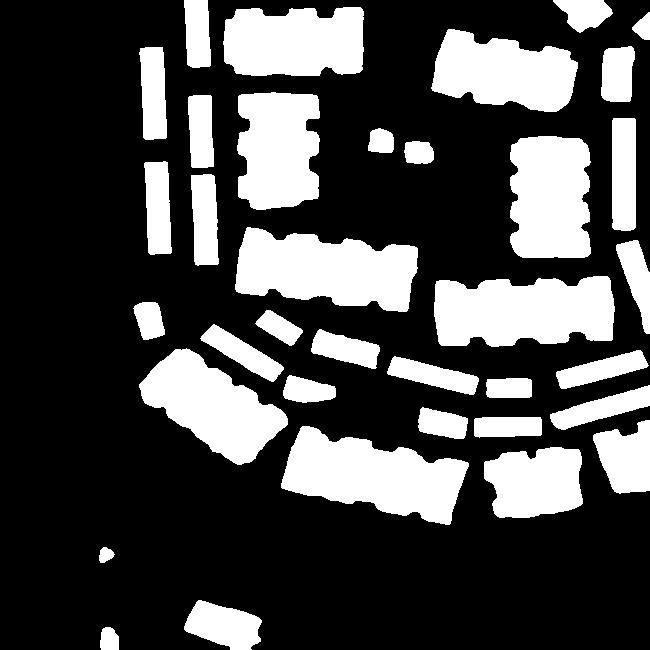}} &
    \raisebox{-.5\height}{\includegraphics[width=3.7cm]{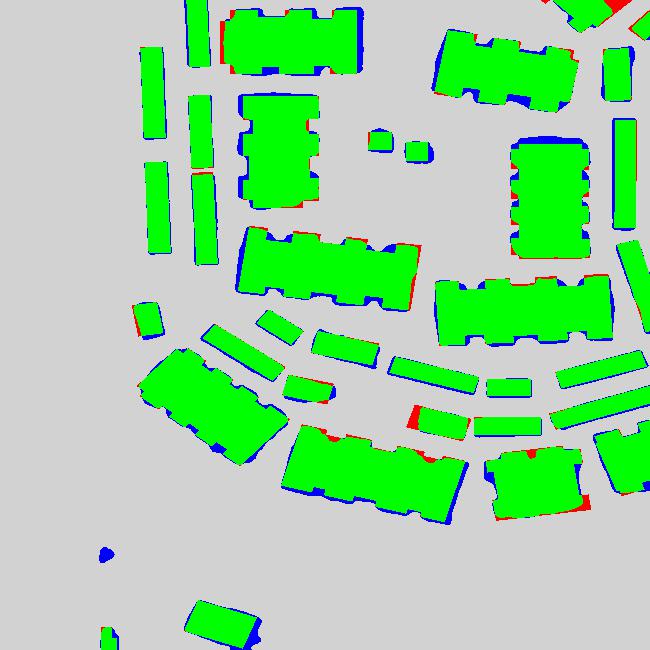}} \\
    \\
    Paris & 
    \raisebox{-.5\height}{\includegraphics[width=3.7cm]{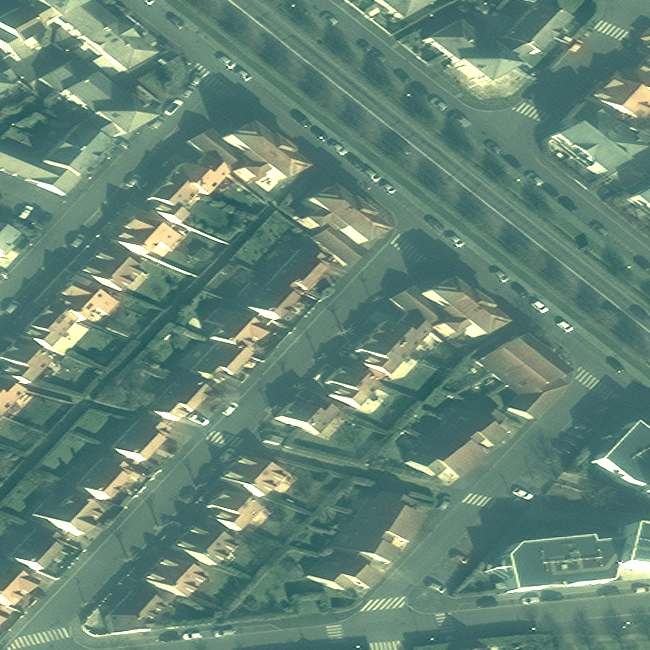}} & 
    \raisebox{-.5\height}{\includegraphics[width=3.7cm]{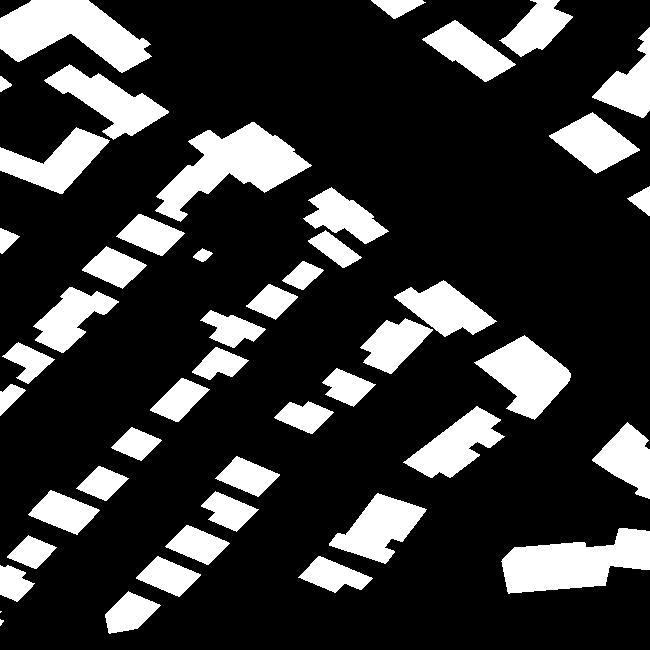}} & 
    \raisebox{-.5\height}{\includegraphics[width=3.7cm]{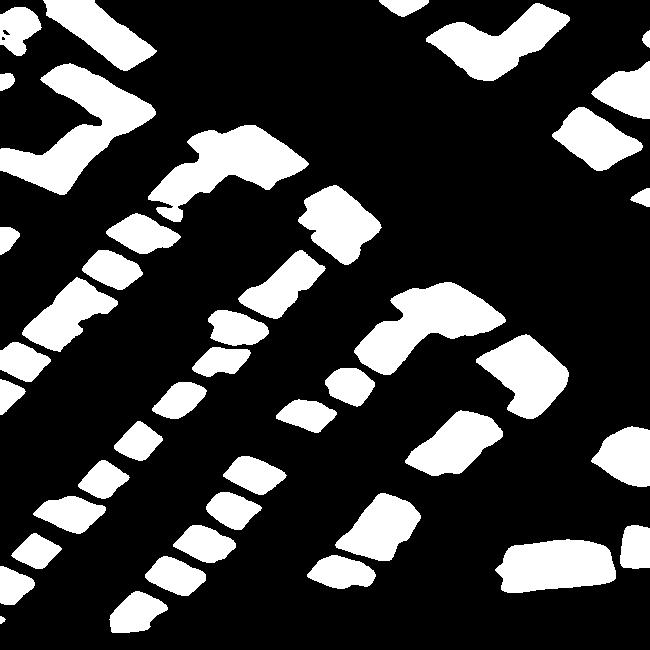}} &
    \raisebox{-.5\height}{\includegraphics[width=3.7cm]{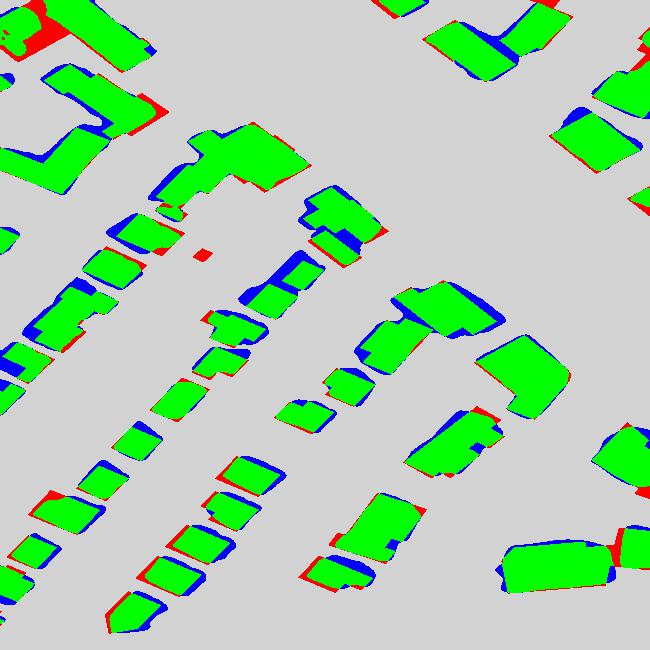}}
    \\
    \\
    Shanghai & 
    \raisebox{-.5\height}{\includegraphics[width=3.7cm]{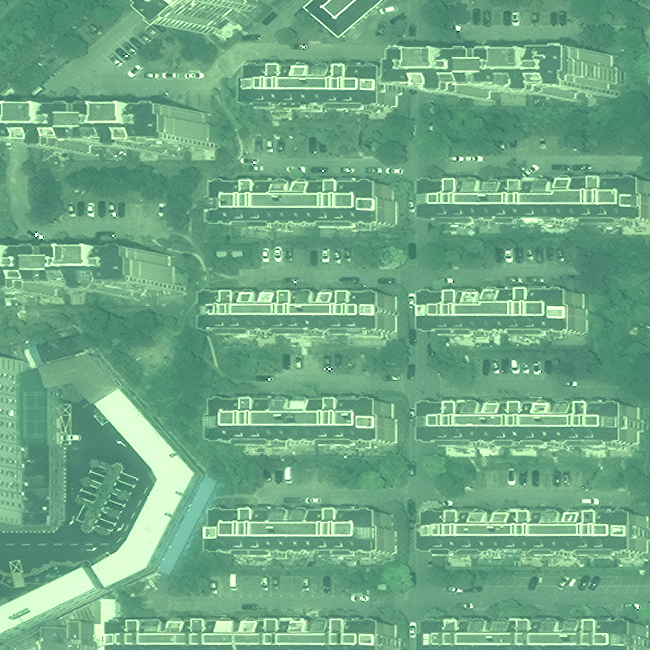}} & 
    \raisebox{-.5\height}{\includegraphics[width=3.7cm]{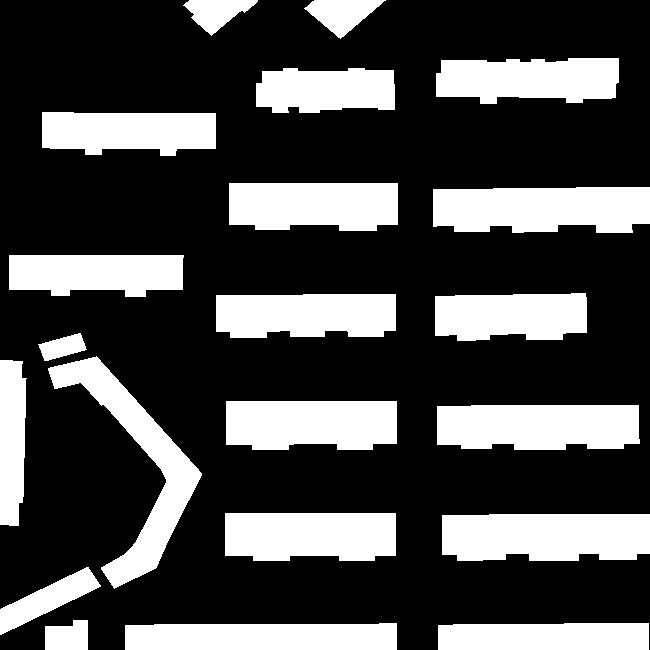}} & 
    \raisebox{-.5\height}{\includegraphics[width=3.7cm]{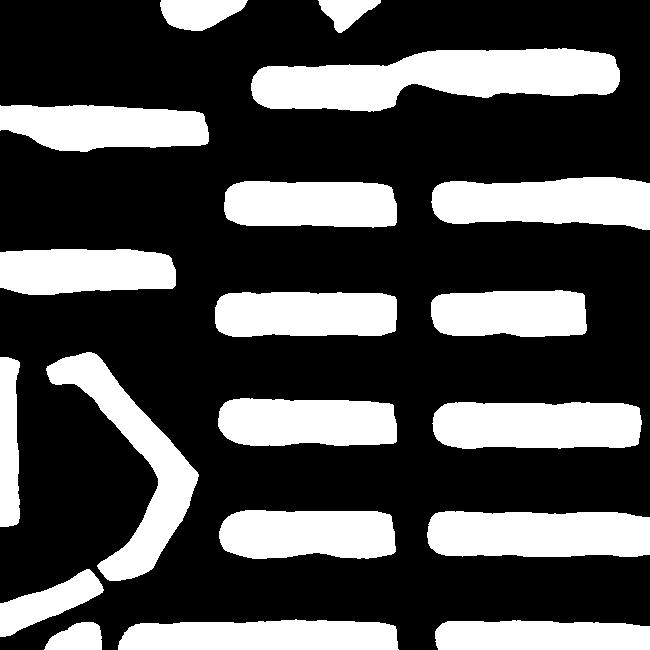}} &
    \raisebox{-.5\height}{\includegraphics[width=3.7cm]{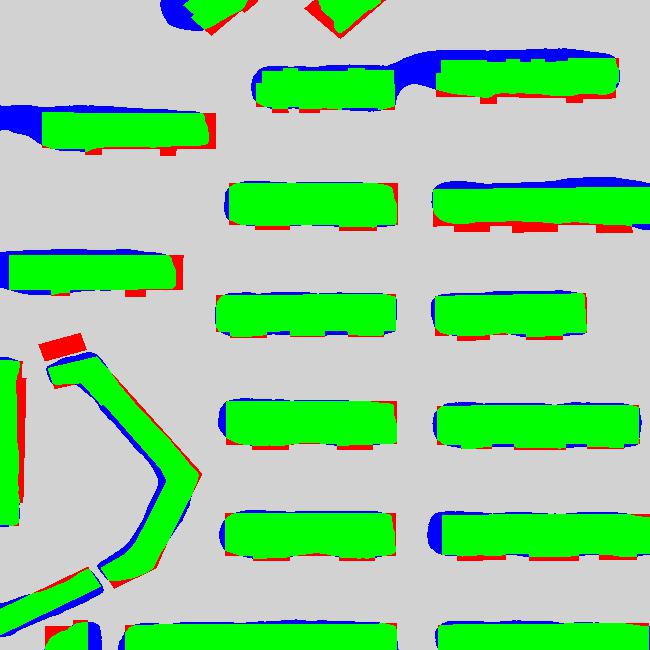}}
    \\
    \\
    Khartoum & 
    \raisebox{-.5\height}{\includegraphics[width=3.7cm]{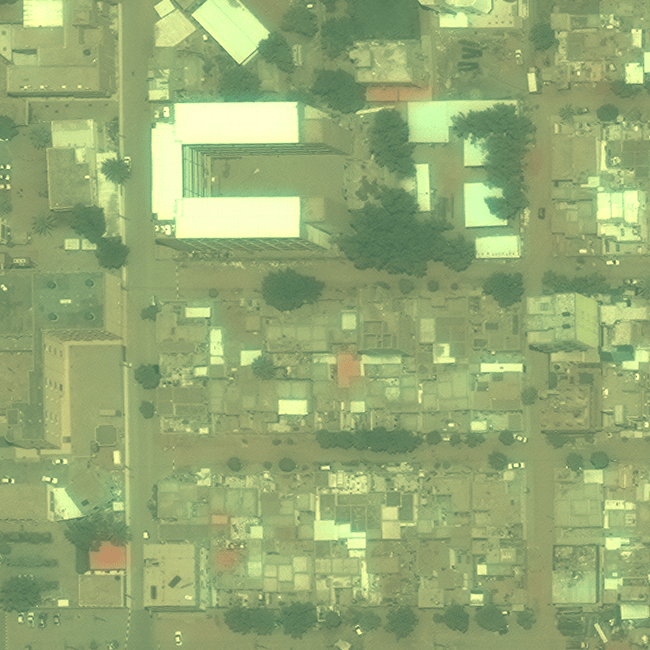}} & 
    \raisebox{-.5\height}{\includegraphics[width=3.7cm]{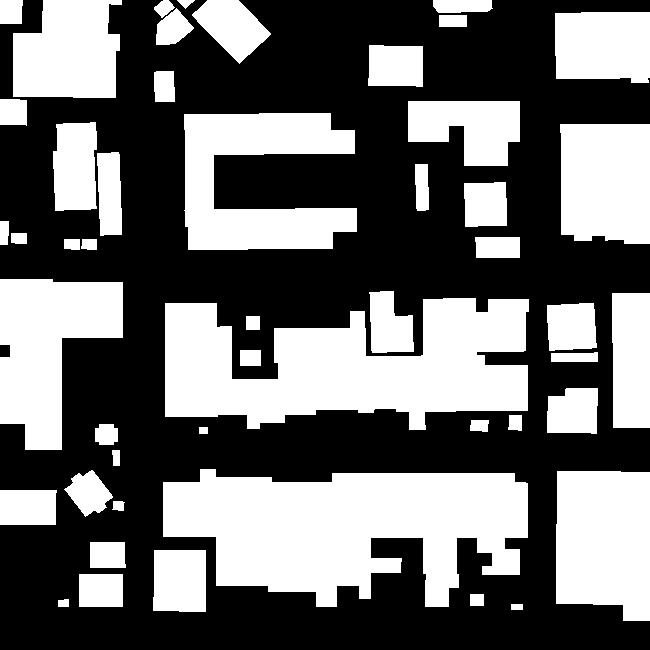}} & 
    \raisebox{-.5\height}{\includegraphics[width=3.7cm]{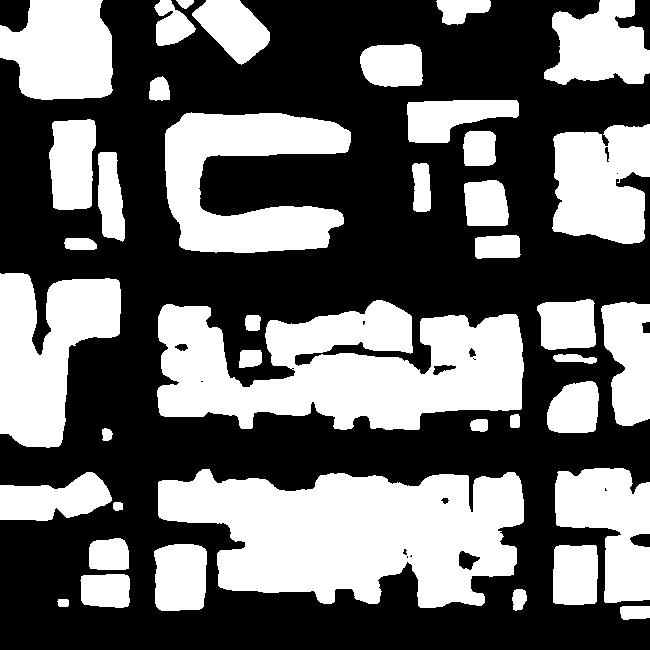}} &
    \raisebox{-.5\height}{\includegraphics[width=3.7cm]{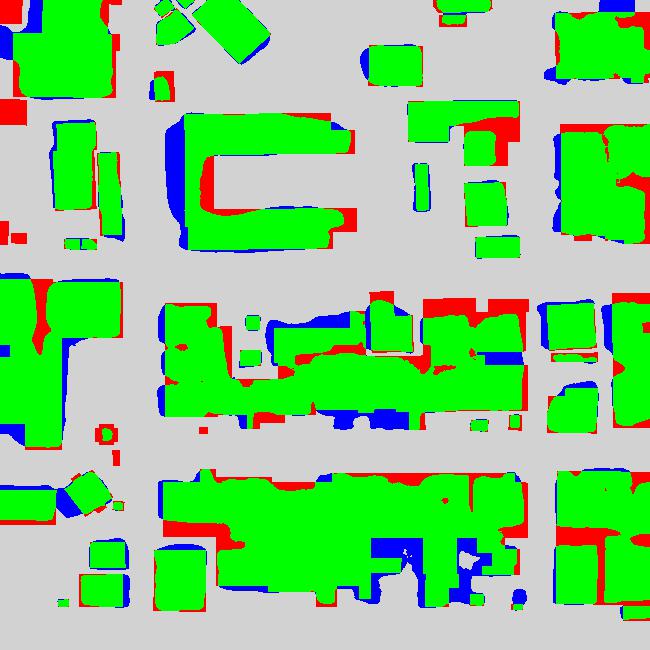}}
\end{tabular}
\captionof{figure}{\centering Qualitative results on the validation subset of DeepGlobe Building Dataset. Column 1: Input image. Column 2: Ground-truth Label Map. Column 3: Predicted Label Map. Column 4: Green: True Positives; Blue: False Positives; Red: False Negatives; Grey: True Negatives.}
\label{fig:seg_dg_val}
\end{table*}

\section{Discussion on the Results and an Ablation Study}
\label{sec:discussionandablation}

The goal of this section is to present a comprehensive
overview of the performance of our approach over all four
datasets that takes into account the characteristics of
each. Subsequently, in a separate subsection, we present an
ablation study to verify the effectiveness of the modules
for the uncertainty attention and refinement, and also of
the deep supervision that is used in our network.

\subsection{Discussion}
\label{sec:discussion}

The results reported in
Tables~\ref{tab:results1}--\ref{tab:ablation} clearly demonstrate
the effectiveness of our proposed algorithm in building
segmentation from remotely sensed images. Owing to the Edge
Attention Unit and the Hausdorff Loss used in our framework
for training, we get accurate building boundaries, as can be
seen in Figure~\ref{fig:seg_good}. The Uncertainty Attention
Module helps us to achieve high number of true positives and
avoid false alarms (See column 4 of
Figure~\ref{fig:seg_inria1}) by giving more attention to the
ambiguous regions of an aerial scene. Further, the Reverse
Attention Unit assists us to identify the missing detections
by refining the intermediate label maps in a top-down
fashion. We also observe significant improvement in the
predictive performance of our algorithm when TTA is applied. Tables~\ref{tab:results1} and \ref{tab:non} report scores for both TTA and non-TTA versions of our algorithm. Tables~\ref{tab:adv}--\ref{tab:ablation} only report our TTA applied results.

With regard to the INRIA dataset, it is evident from
Table~\ref{tab:results2} that the performance of our
algorithm for the Chicago area is not the best. The
buildings in Chicago are located very close to one another,
and the network finds it difficult to clearly separate the
building boundaries of adjacent buildings. We see the same
situation in the San Francisco region -- buildings in San
Francisco area are also densely packed.  Obviously, our
framework needs further improvements in separating the
buildings that are in close proximity to one another. We
believe this issue arises as we use a dilation operator in
our edge refinement module. Using an accurate contour
extraction algorithm should help us in alleviating this
problem.

In general, ground-truth label inconsistencies in the
datasets hinder our training process to some extent, and
also impact the overall evaluation scores. Specifically, in
addition to the building masks being not perfectly aligned
to the actual buildings, the Massachusetts Buildings Dataset
also contains false labels. Some examples of noisy labels in
the Massachusetts Dataset can be seen in column 2 of
Figure~\ref{fig:noisy_label}. Moreover, in some of the
images, the buildings encompassing playgrounds or parking
lots are labeled as a single building instance without
capturing the actual shape of the building (column 1 of
Figure~\ref{fig:noisy_label}). However, our
network identifies the building pixels accurately, as
illustrated in row 3 of columns 1 and 2 of
Figure~\ref{fig:noisy_label}. Similar noisy labels appear in
the INRIA Aerial Image Labeling Dataset. Column 3 of
Figure~\ref{fig:noisy_label} shows an image patch over
Vienna where in the ground-truth, smaller building
structures close to one-another are clubbed as a one large
building. Still, our network accurately predicts each
smaller structure. Kitsap County not only has a very sparse
distribution of buildings, but mis-labels are also prevalent
in the dataset. This severely impacts the evaluation
scores. Out of 5 images in the validation dataset, 2 of the
images have false building labels. One such example is shown
in column 4 of Figure~\ref{fig:noisy_label}. We achieve an IoU
of 86.42\% as opposed to 73.25\% when we leave out those 2
images from the validation set. This kind of mis-labels are
found through the training subset as well. However, our
network is robust to such mis-labels as evident from the
qualitative as well as quantitative results.

Our network yields across-the-board superior performance on
the WHU Building Dataset. We believe that the main reason
for that is the fact that the ground-truth building maps
provided in the WHU dataset are more accurate.  We should
also mention the relatively low complexity of this dataset
in relation to the other three datasets that cover more
difficult terrains with high buildings, diverse topography,
more occlusions and shadows.

For the DeepGlobe Dataset, our algorithm achieves the best
results for Vegas and second highest F1-score for Paris. The
images in the Vegas and Paris subsets are mostly collected
from residential regions. Unlike the other two cities in the
DeepGlobe dataset, the buildings in Vegas and Paris have
more unified architectural style. For Shanghai, our proposed
method faced difficulty in correctly extracting buildings
with green roofs or buildings that are of extremely small
size. In Khartoum, there are many building groups, and it is
hard to judge, even by the human eye, whether a group of
neighboring buildings should be extracted entirely or
separately in many regions.

\begin{figure*}
    \centering
    \subfloat{{\includegraphics[width=4.2cm]{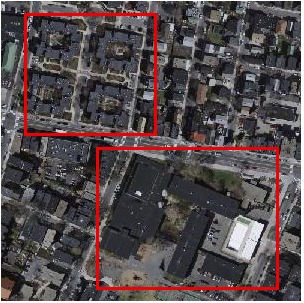} }}
    \subfloat{{\includegraphics[width=4.2cm]{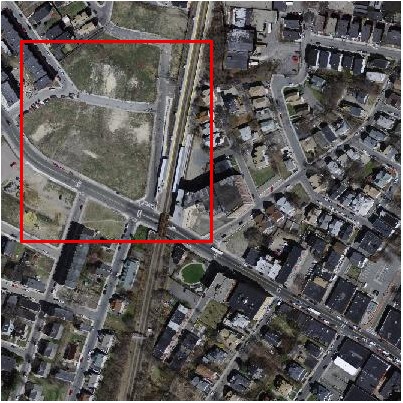} }}
    \subfloat{{\includegraphics[width=4.2cm]{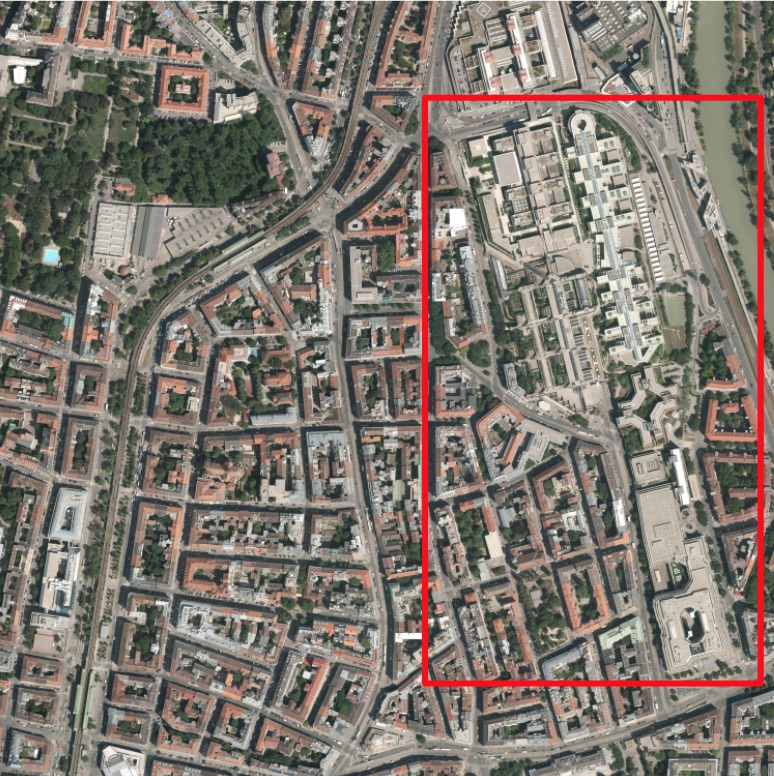} }}
    \subfloat{{\includegraphics[width=4.2cm]{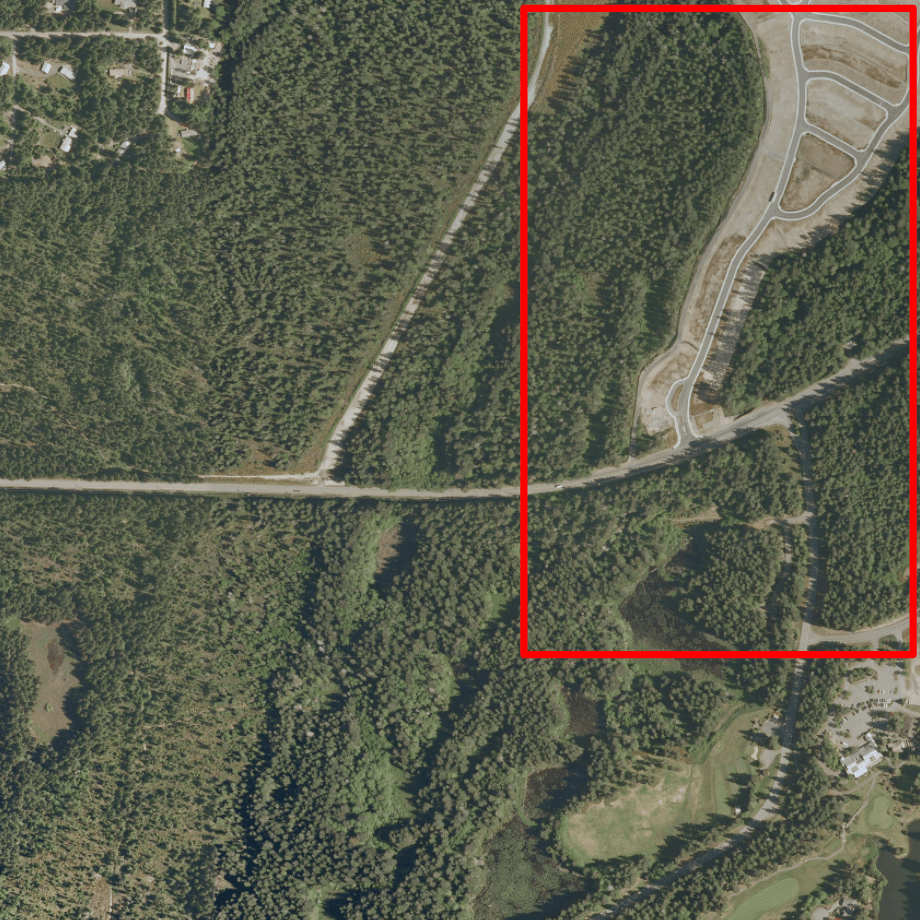} }}
    \hfill
    \subfloat{{\includegraphics[width=4.2cm]{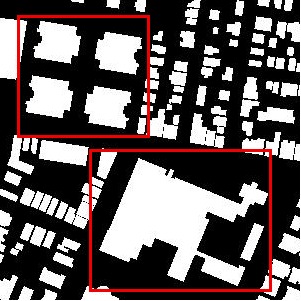} }}
    \subfloat{{\includegraphics[width=4.2cm]{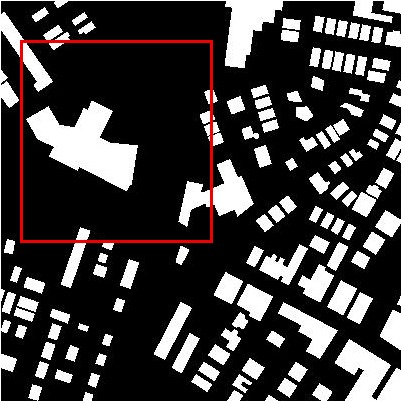} }}
    \subfloat{{\includegraphics[width=4.2cm]{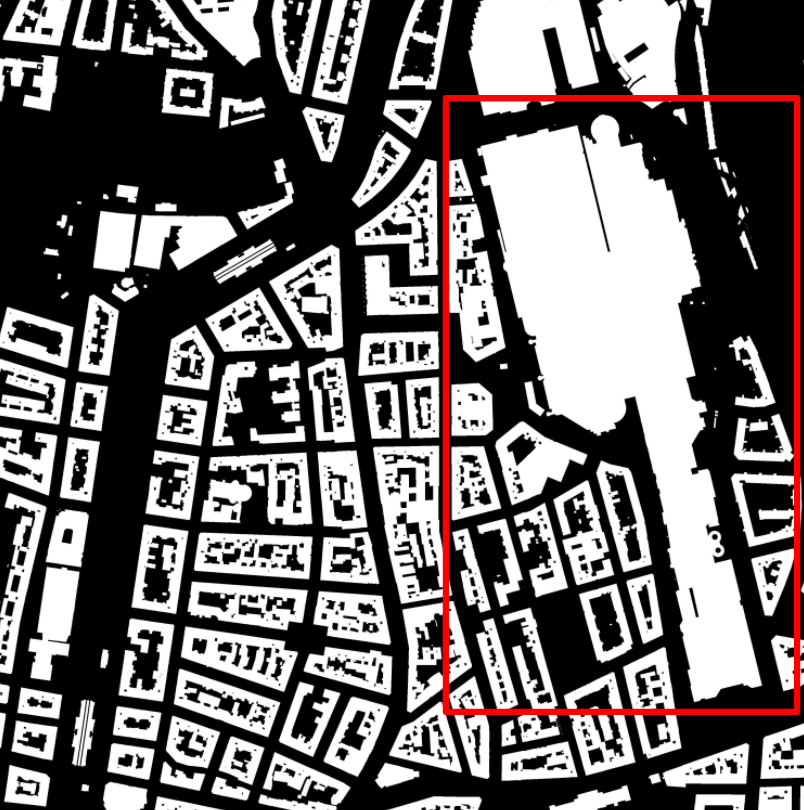} }}
    \subfloat{{\includegraphics[width=4.2cm]{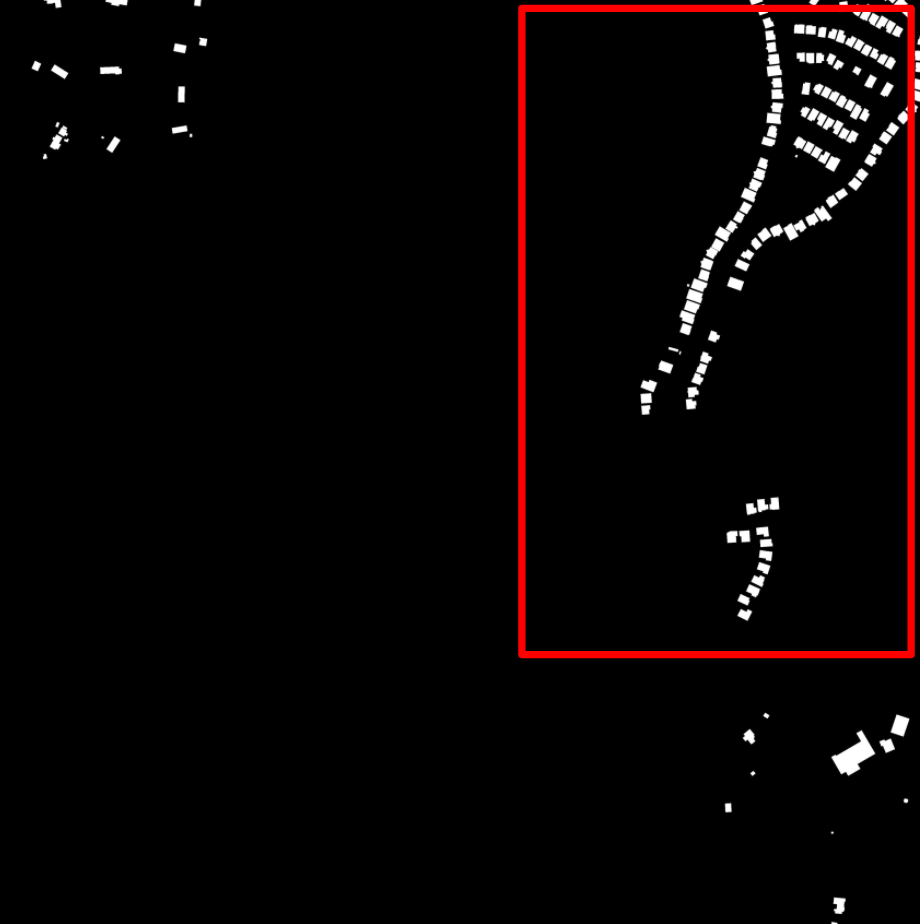} }}
    \hfill
    \subfloat{{\includegraphics[width=4.2cm]{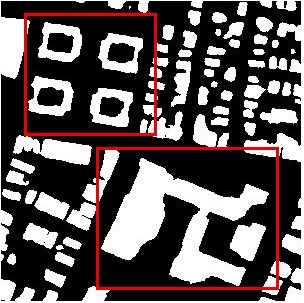} }}
    \subfloat{{\includegraphics[width=4.2cm]{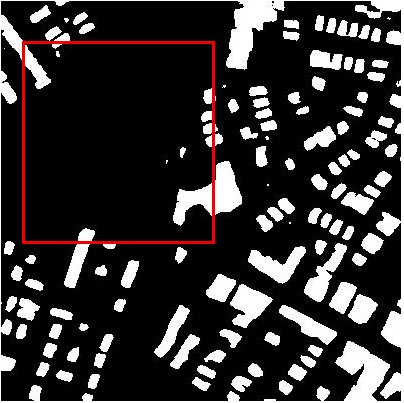} }}
    \subfloat{{\includegraphics[width=4.2cm]{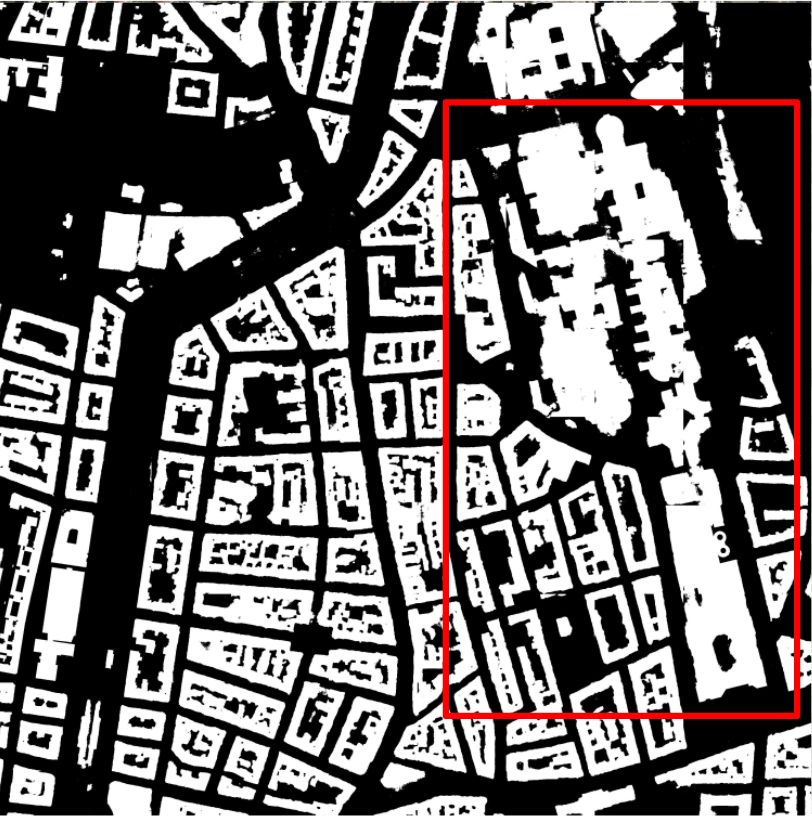} }}
    \subfloat{{\includegraphics[width=4.2cm]{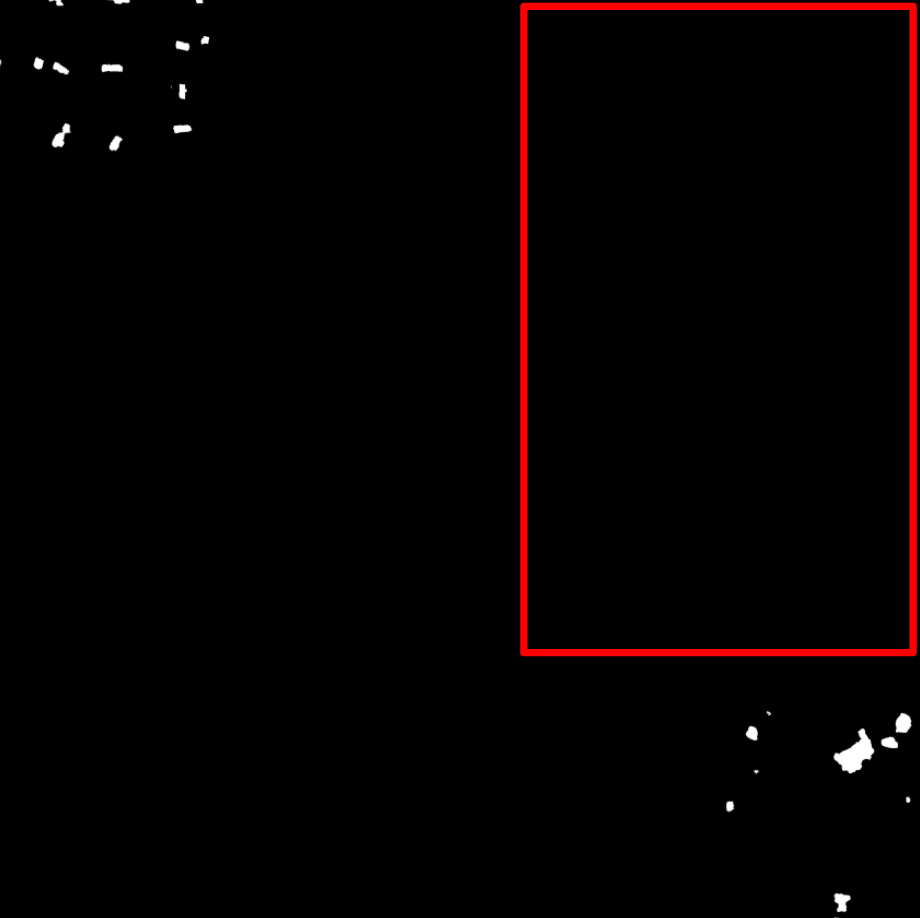} }}
    \caption{\centering Noisy labels in the Massachusetts Buildings Dataset (columns 1, 2) and the INRIA Aerial Image Labeling Dataset (columns 3, 4). Row 1: Input Image. Row 2: Ground-truth Labels. Row 3: Predicted Labels. The red boxes represent the areas where noisy labels are present in the ground-truth label maps.}
  \label{fig:noisy_label}
\end{figure*}
\begin{figure*}%[t!p]
  \centering
  \subfloat{{\includegraphics[width=4.2cm]{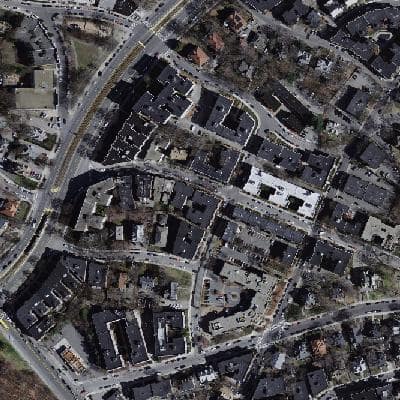} }}
  \subfloat{{\includegraphics[width=4.2cm]{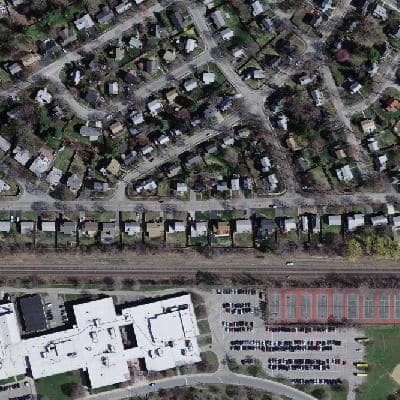} }}
  \subfloat{{\includegraphics[width=4.2cm]{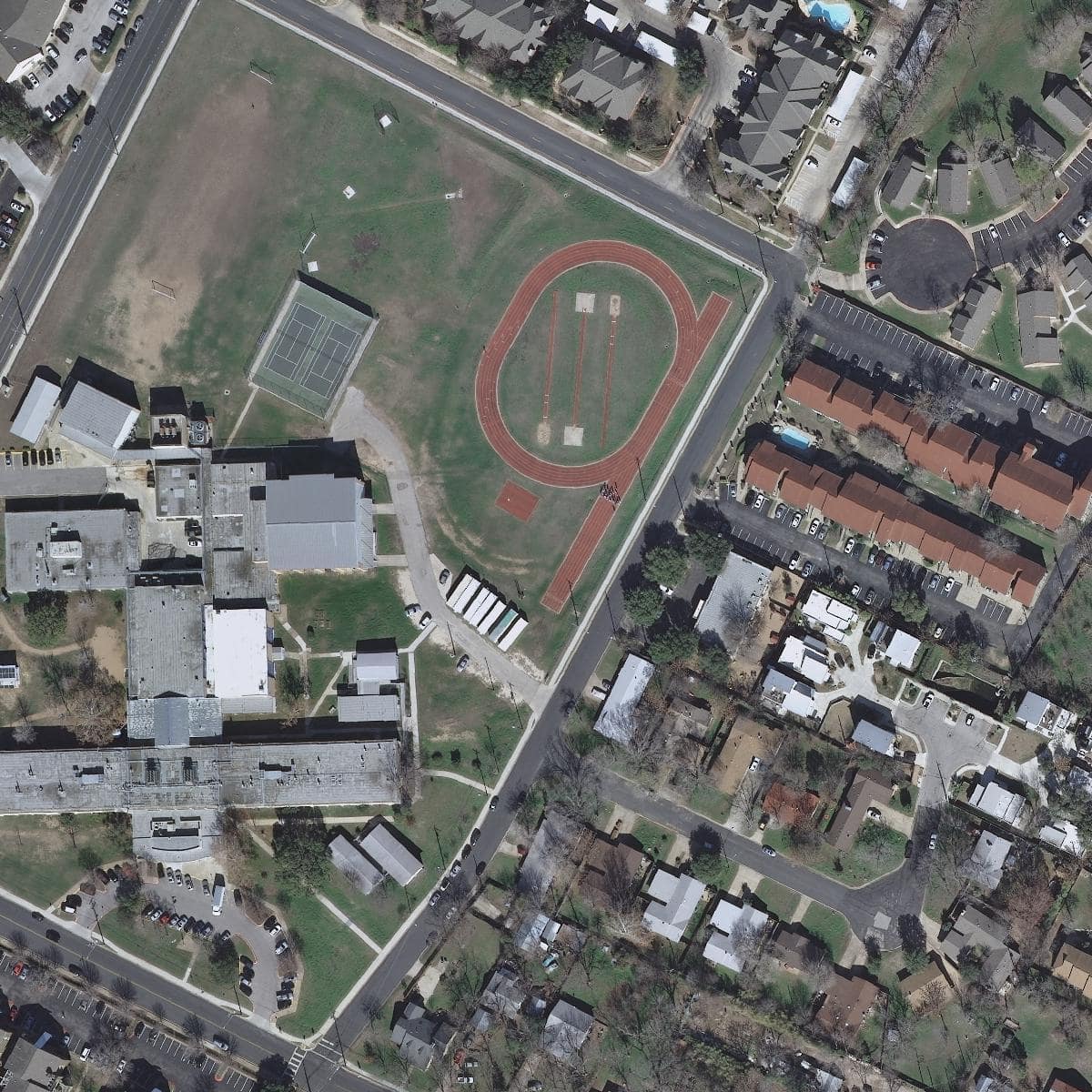} }}
  \subfloat{{\includegraphics[width=4.2cm]{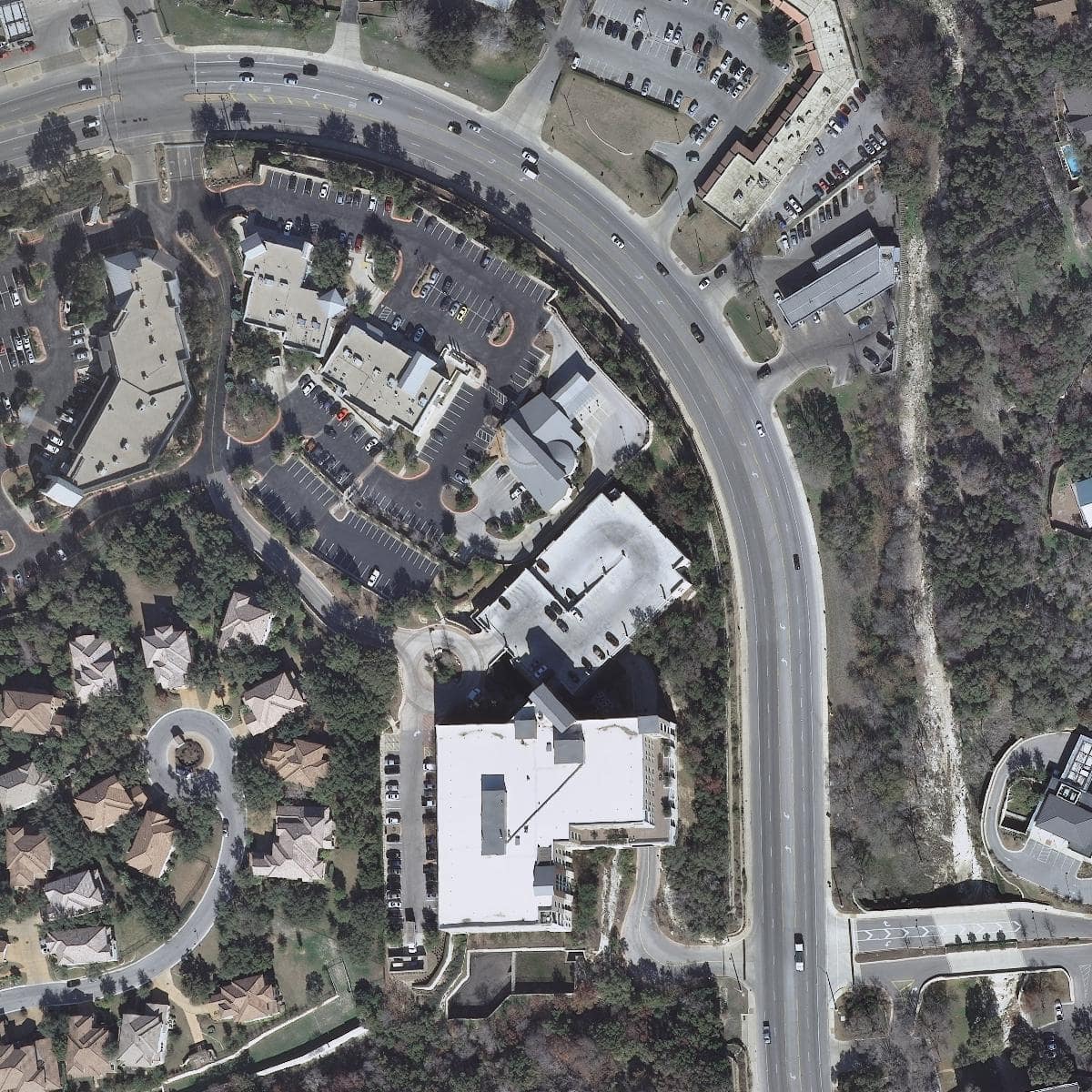} }}
  \hfill
  \subfloat{{\includegraphics[width=4.2cm]{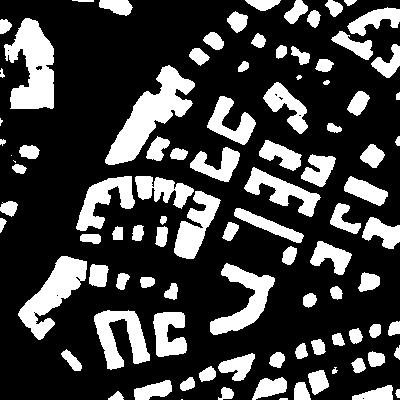} }}
  \subfloat{{\includegraphics[width=4.2cm]{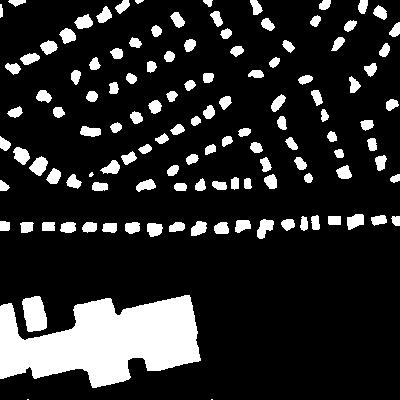} }}
  \subfloat{{\includegraphics[width=4.2cm]{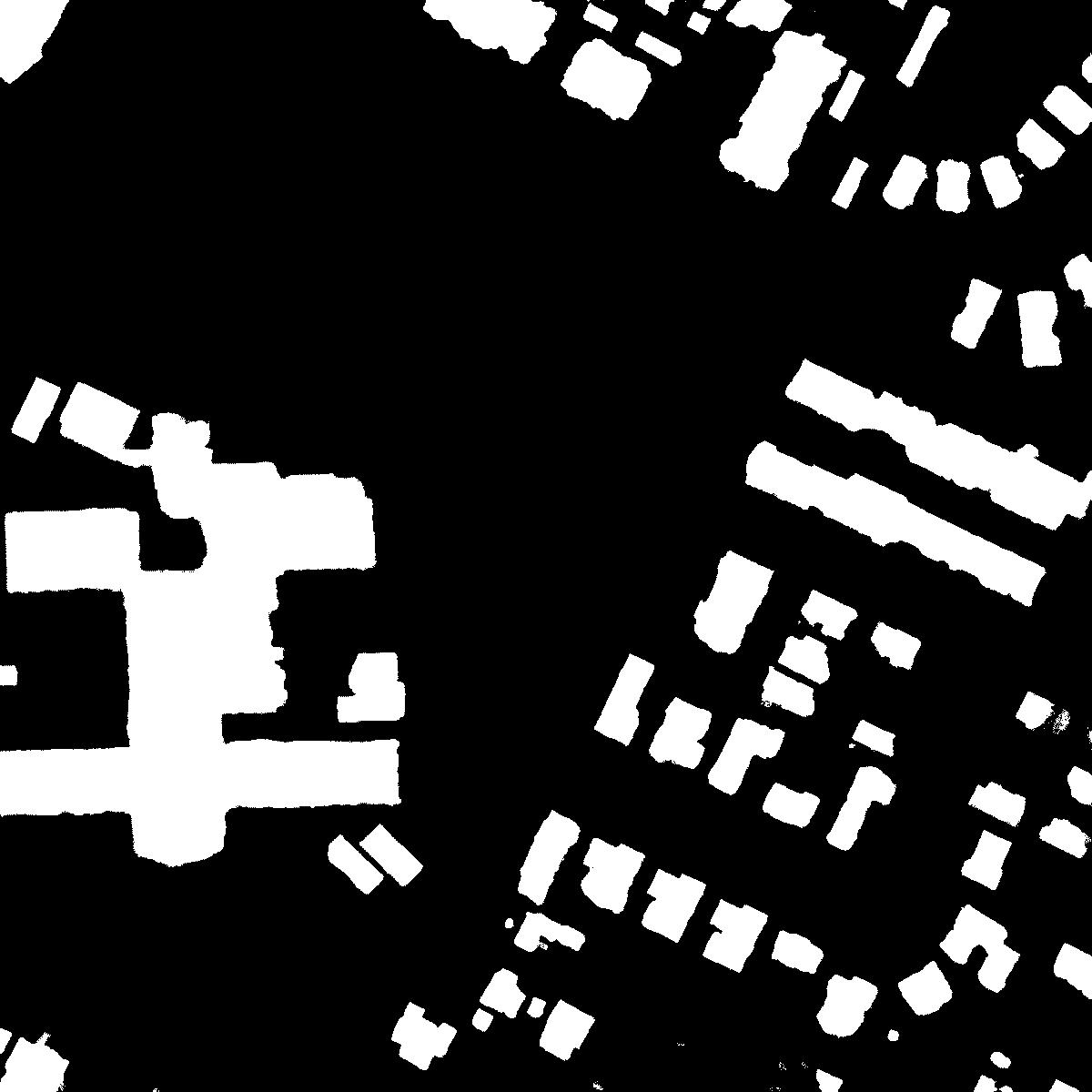} }}
  \subfloat{{\includegraphics[width=4.2cm]{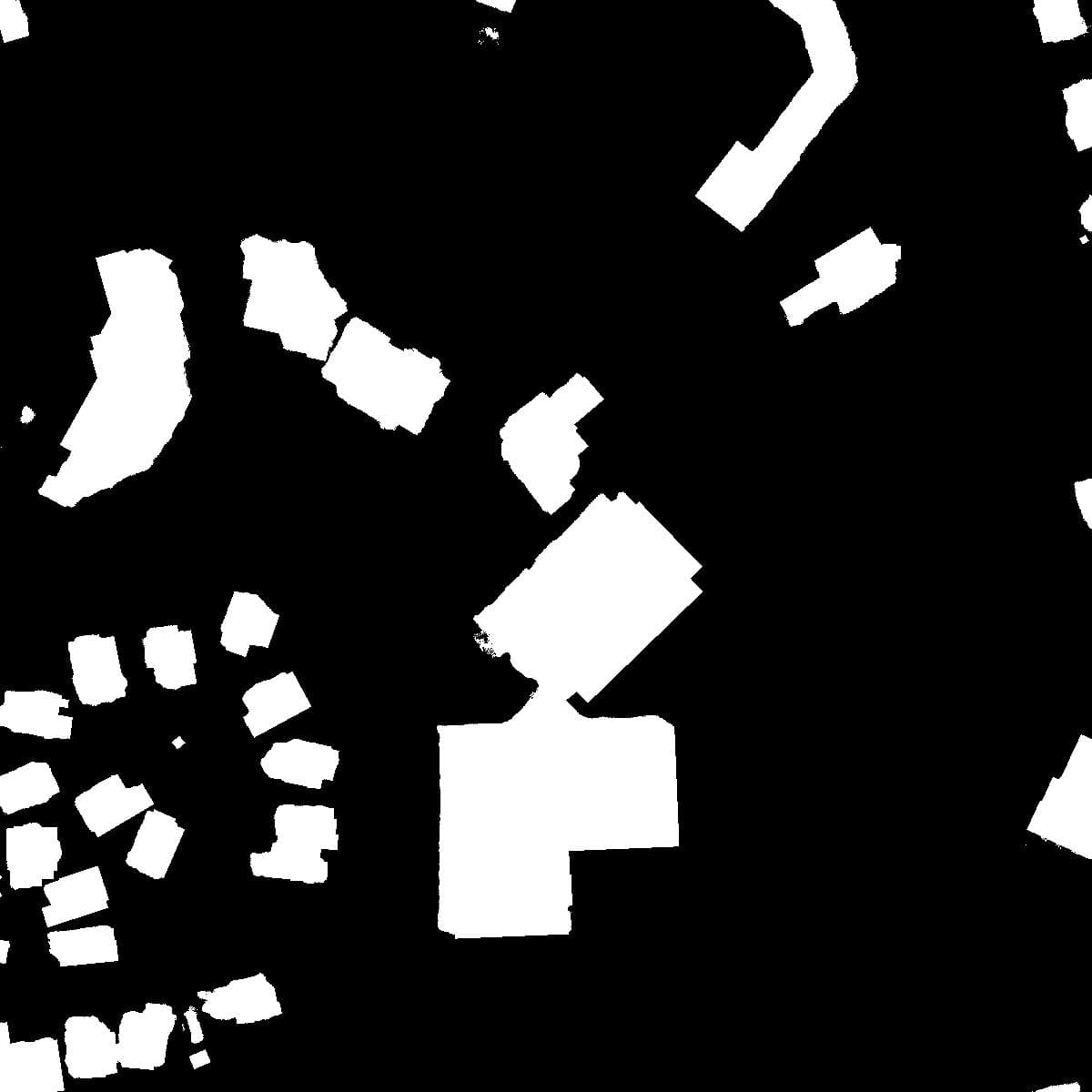} }}
  \caption{\centering Crisp building boundaries
    using our proposed approach. Row 1: Input Image. Row 2: Predicted Labels.}
  \label{fig:seg_good}
\end{figure*}
\begin{figure*}%[t!p]
  \centering
  \subfloat{{\includegraphics[width=3.3cm]{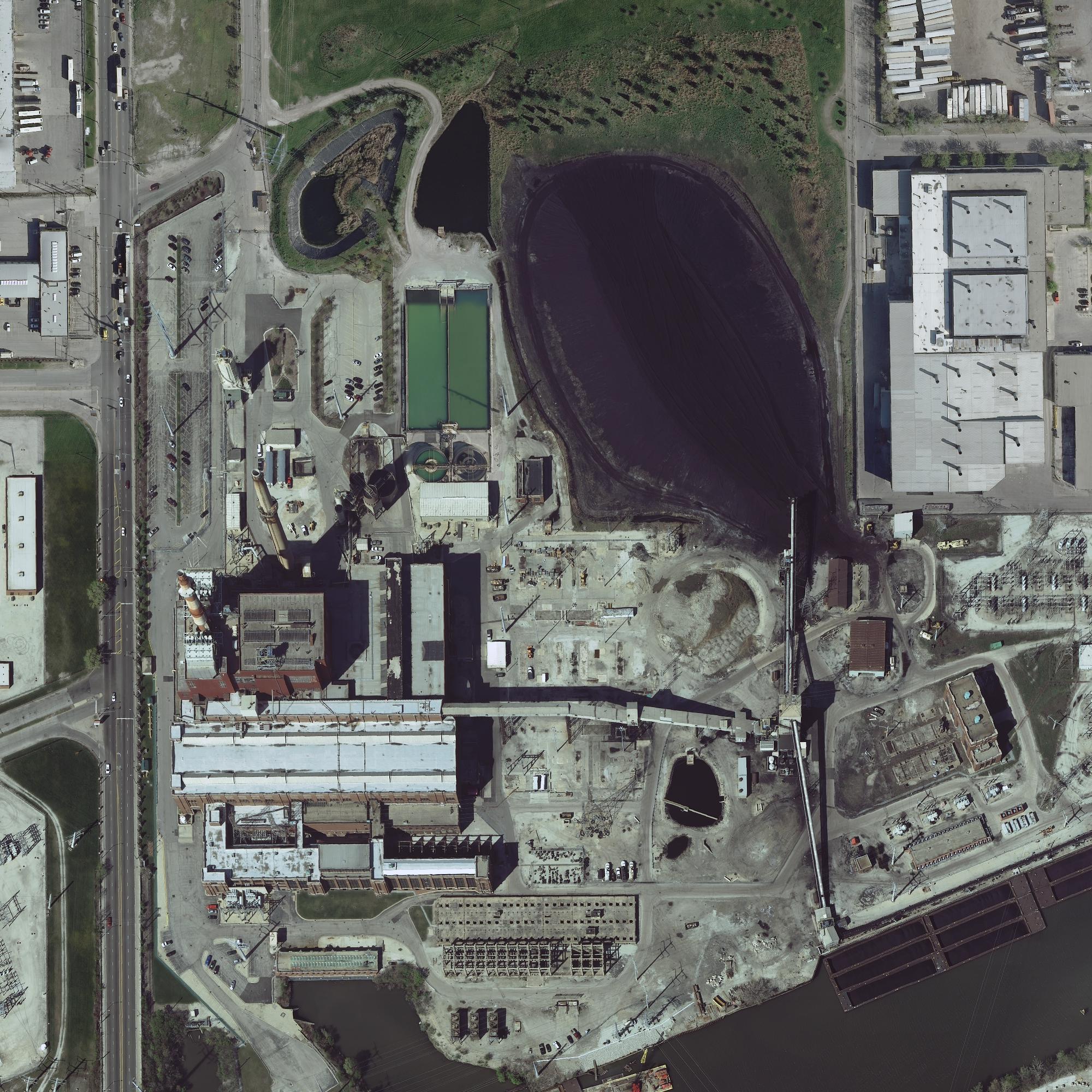} }}
  \subfloat{{\includegraphics[width=3.3cm]{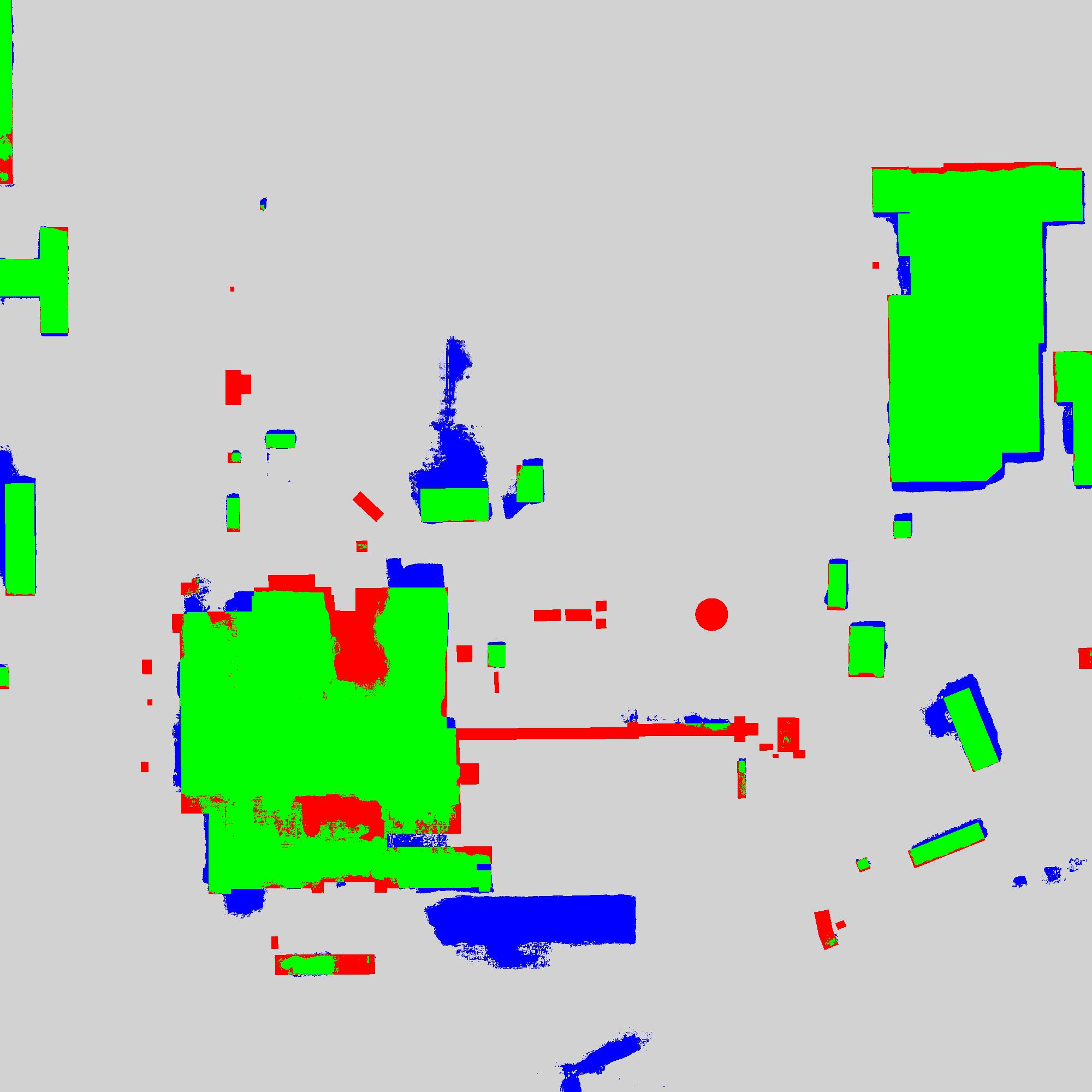} }}
  \subfloat{{\includegraphics[width=3.3cm]{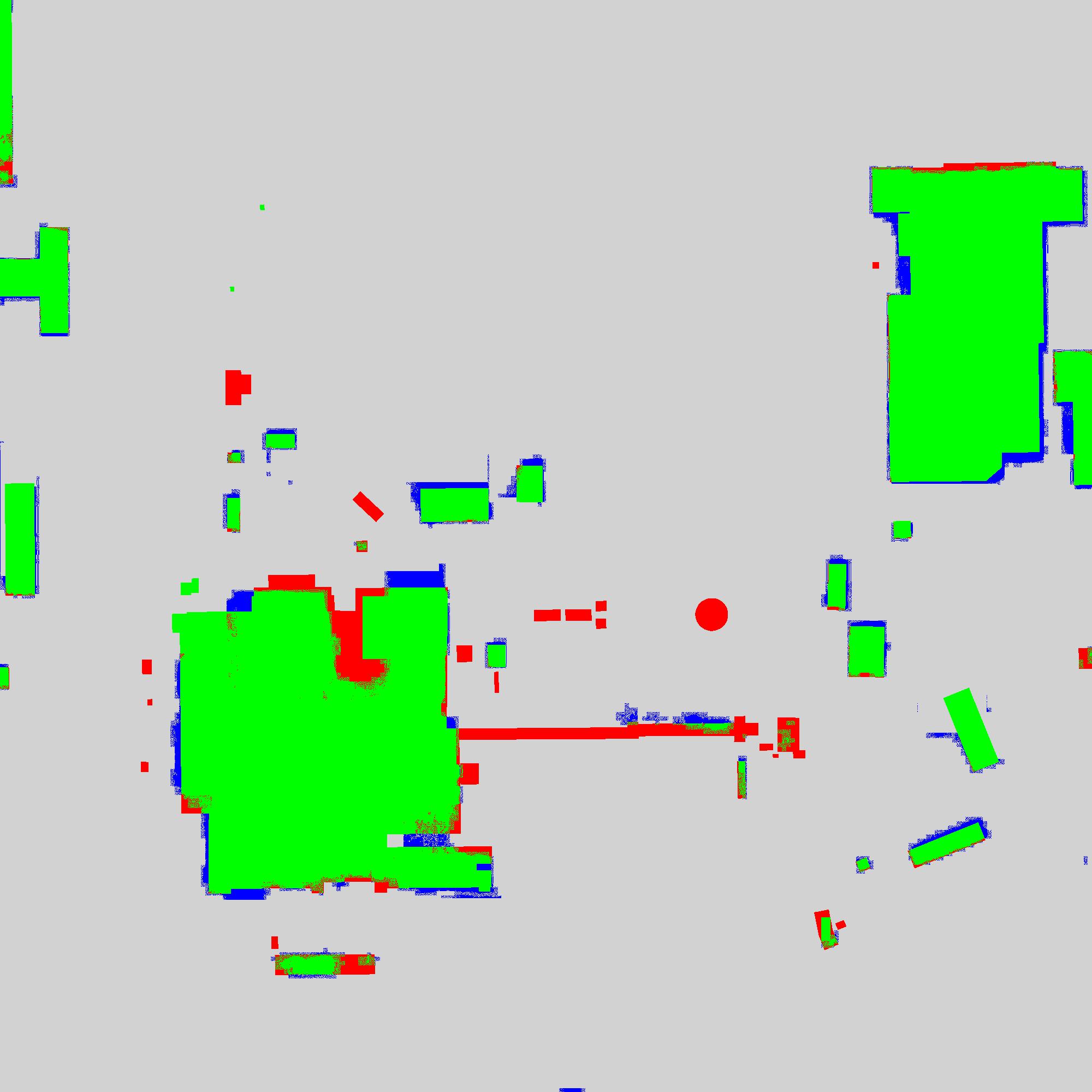} }}
  \subfloat{{\includegraphics[width=3.3cm]{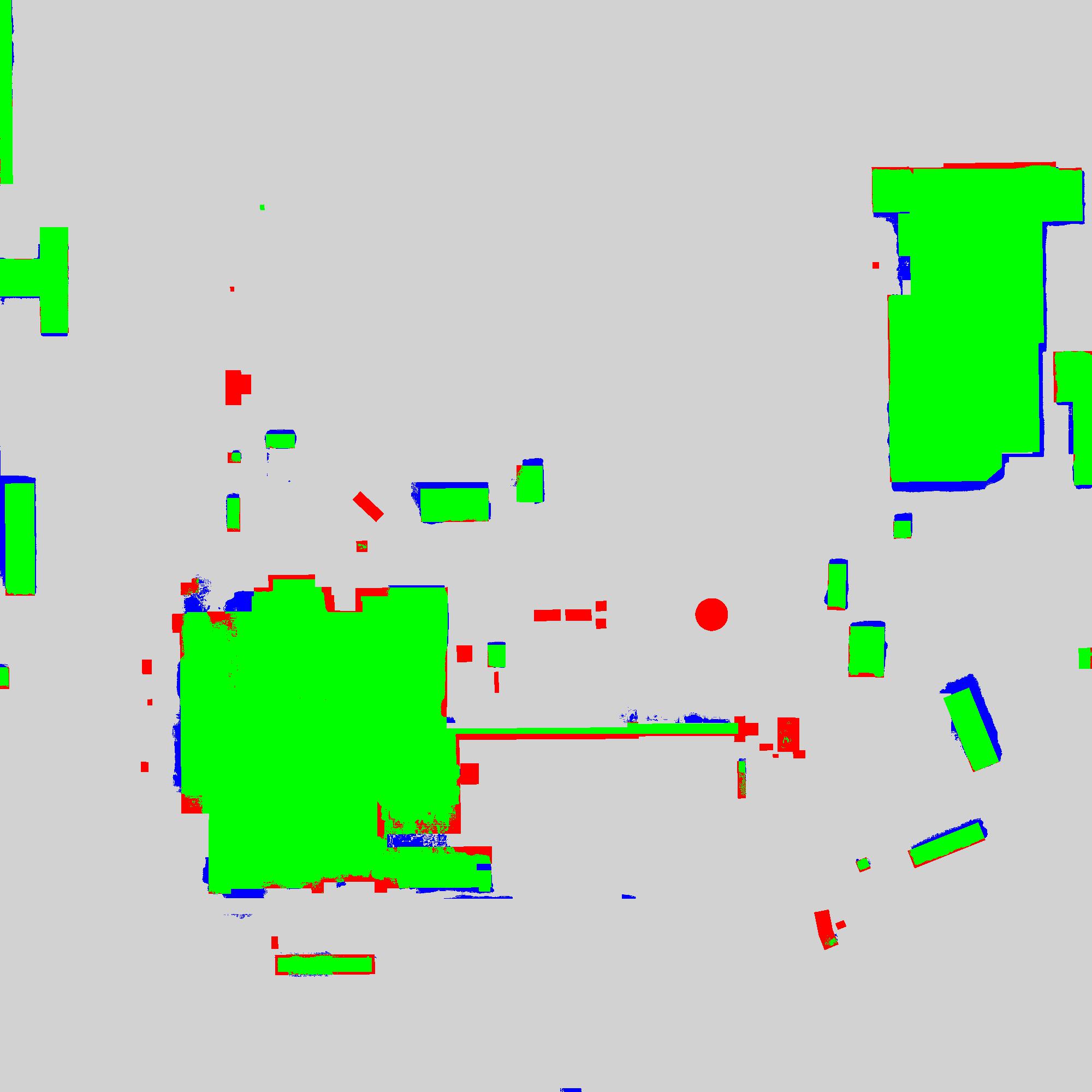} }}
  \subfloat{{\includegraphics[width=3.3cm]{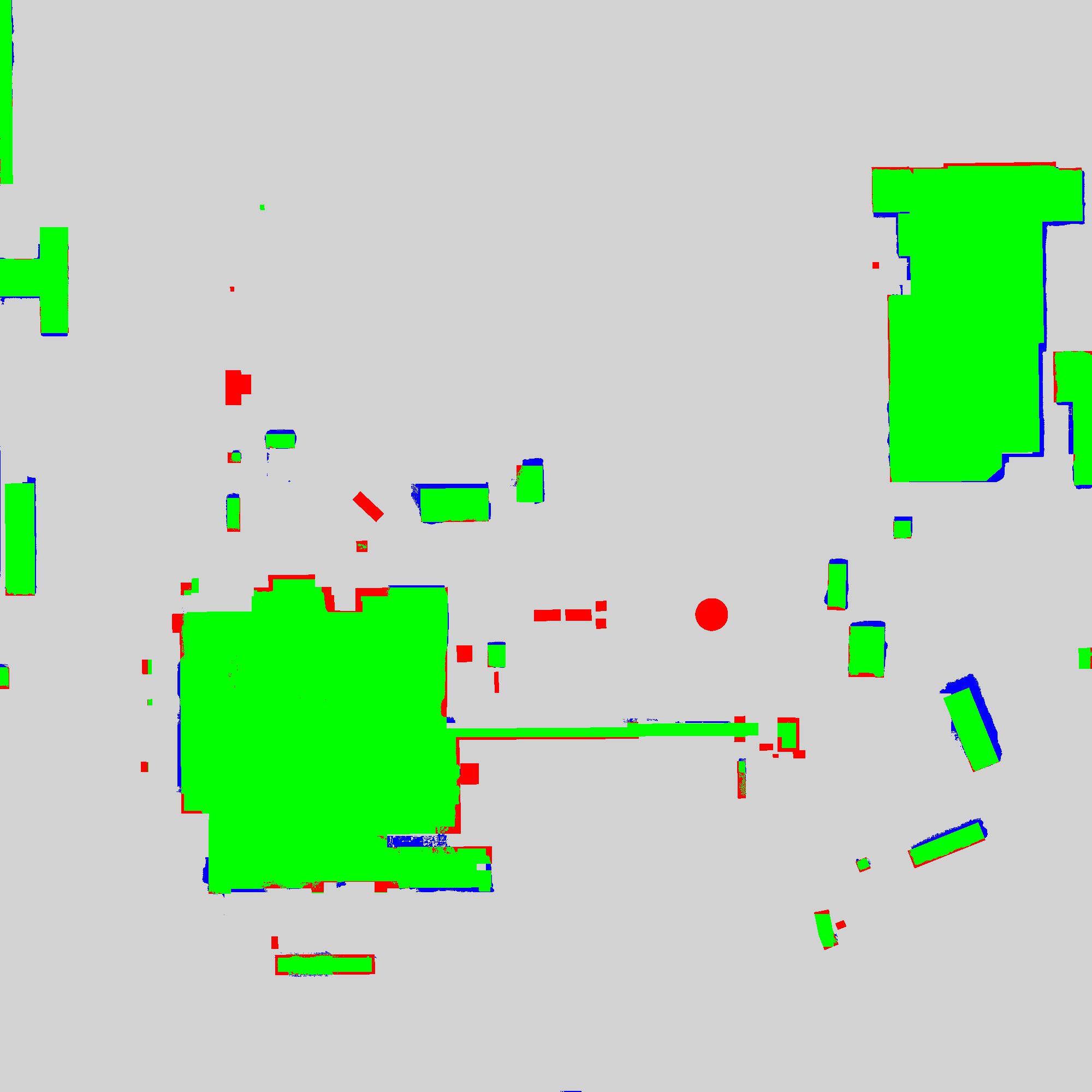} }}
  \hfill
  \subfloat{{\includegraphics[width=3.3cm]{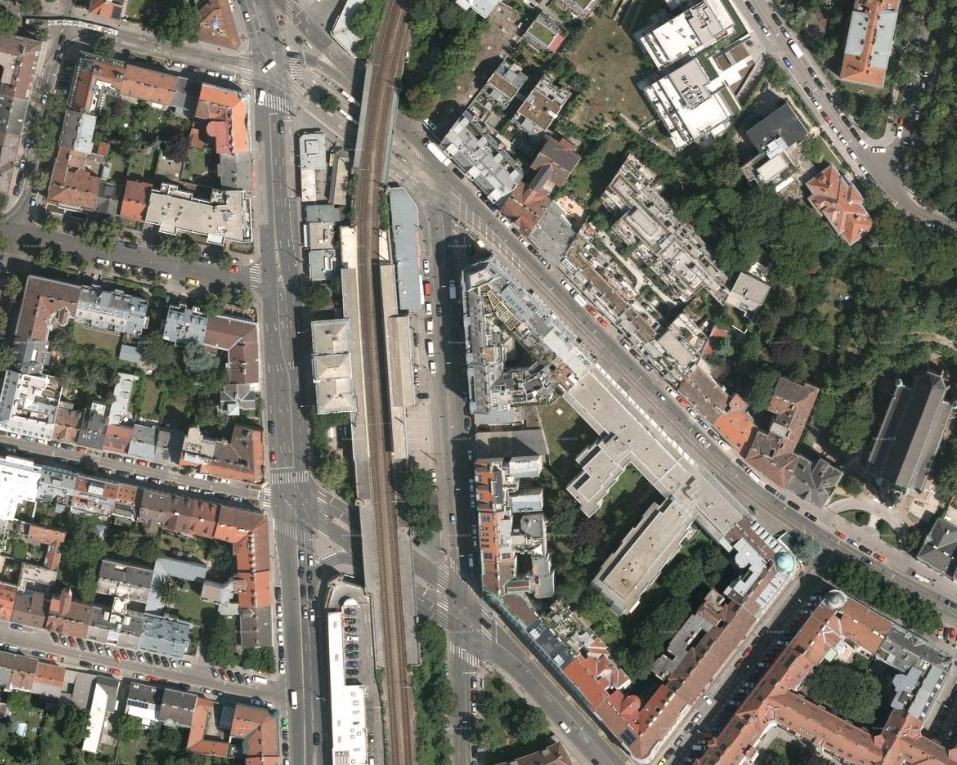} }} 
  \subfloat{{\includegraphics[width=3.3cm]{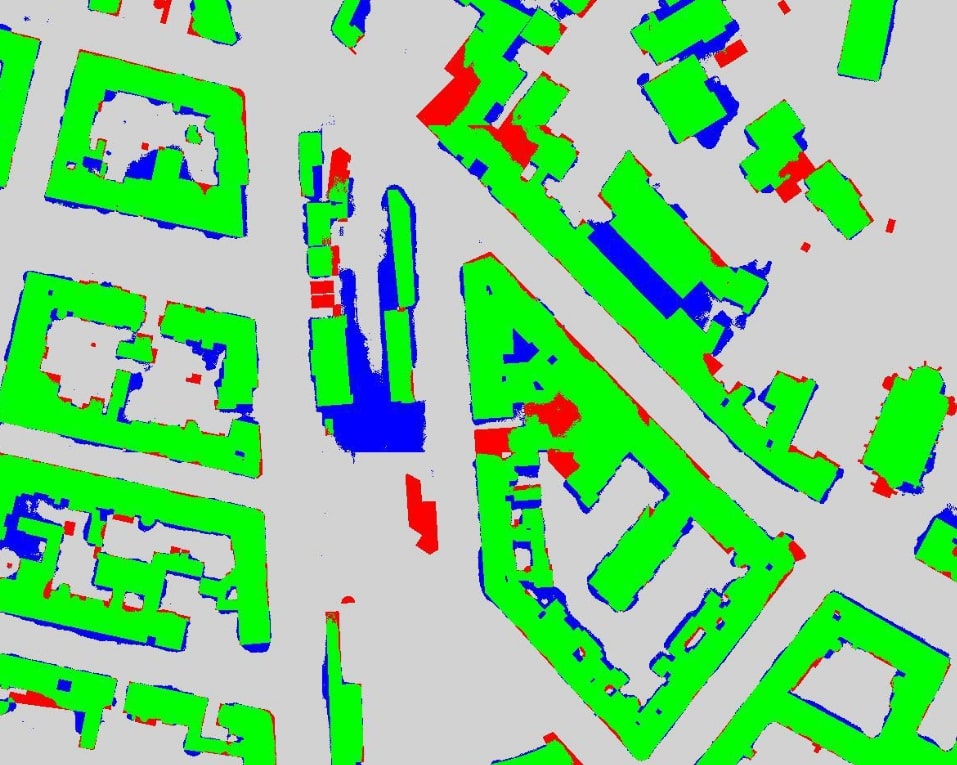} }}
  \subfloat{{\includegraphics[width=3.3cm]{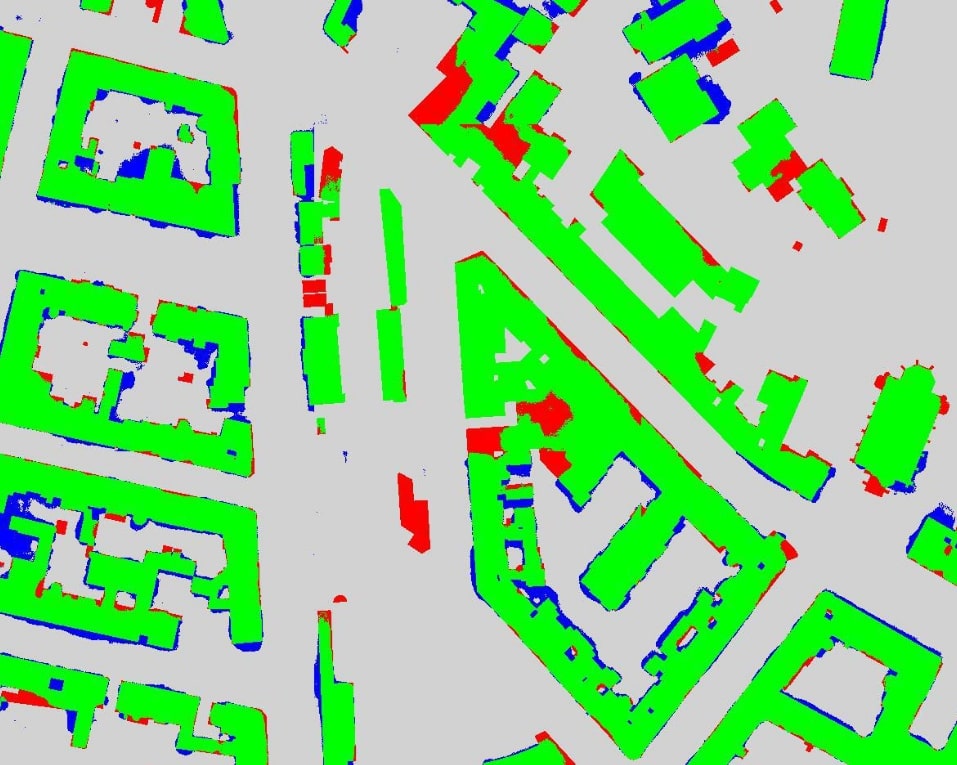} }}
  \subfloat{{\includegraphics[width=3.3cm]{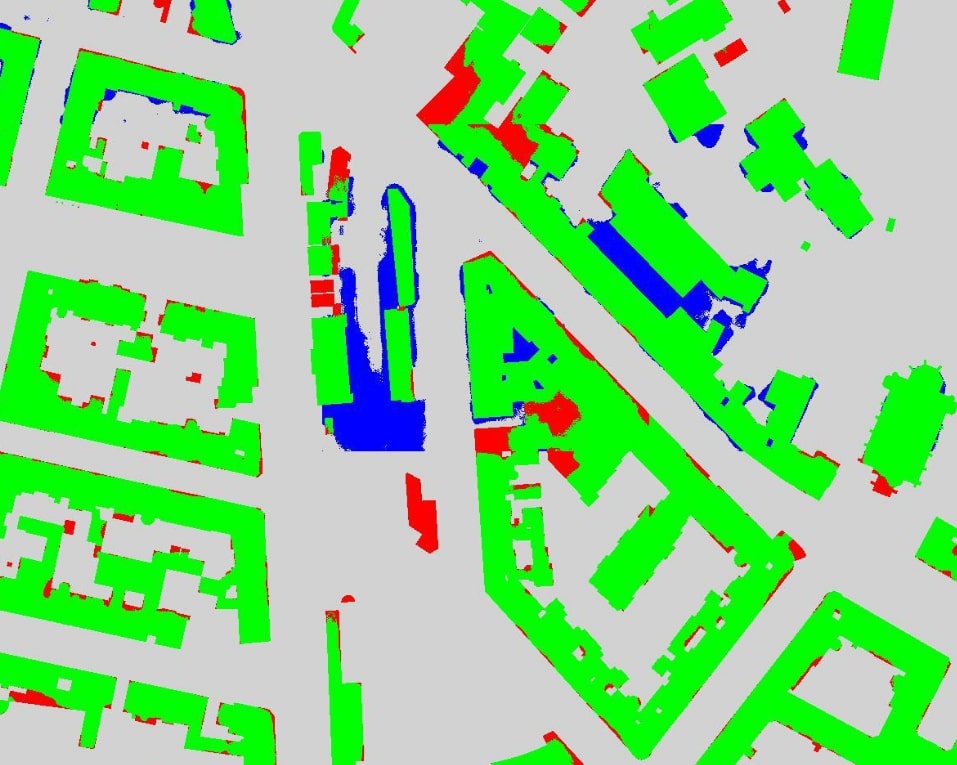} }}
  \subfloat{{\includegraphics[width=3.3cm]{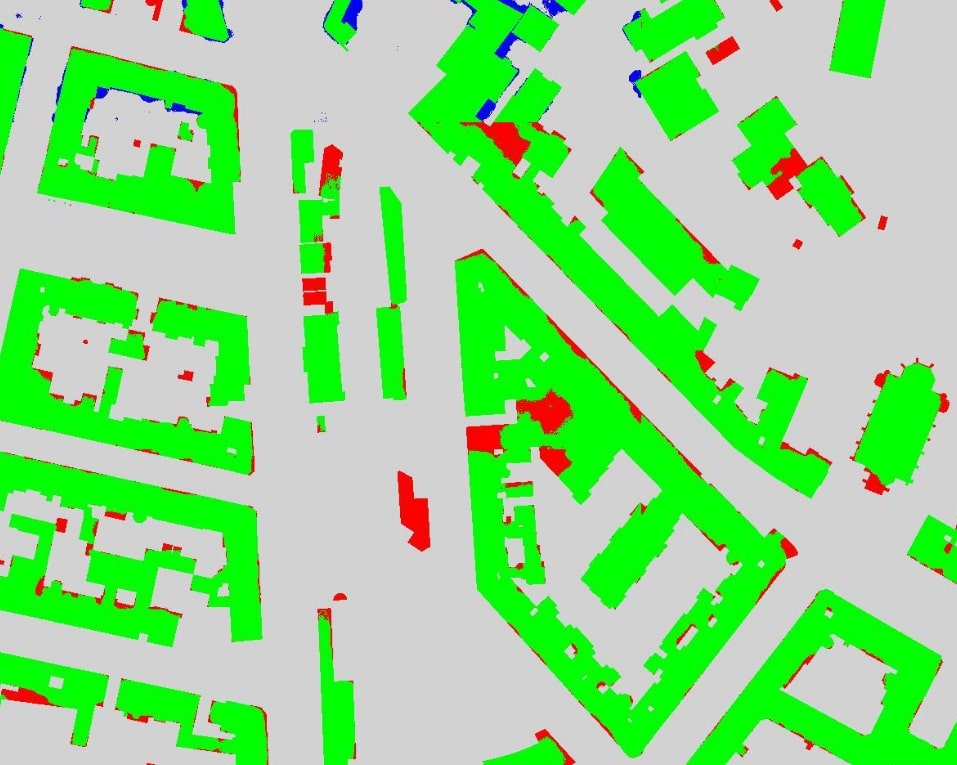} }}
  \caption{\centering Ablation study results on Chicago (row 1) and Vienna (row 2) areas of the INRIA Aerial Image Labeling Dataset. \: Column 1: Input Image. Column 2: Base GAN Architecture (BGA). Column 3: BGA + Uncertainty Attention Module (UAM). Column 4: BGA + Refinement Module (RM). Column 5: BGA + UAM + RM. All the results are from models trained with deep supervision. Test time augmentation is used for all models. Green: True Positives; Blue: False Positives; Red: False Negatives; Grey: True Negatives.}
  \label{fig:ablation}
\end{figure*}

\bgroup
\def\arraystretch{1.5}
\setlength\tabcolsep{0.3cm}
\begin{table*}[!ht]
\begin{center}
\resizebox{0.8\linewidth}{!}{%
    \begin{tabular}{ | c | c | c | c | c| c | c | p{1.5cm} |}
    \hline
    Method & \multicolumn{1}{| c |}{Austin} & \multicolumn{1}{| c |}{Chicago}& \multicolumn{1}{| c |}{Kitsap}& \multicolumn{1}{| c |}{W. Tyrol}& \multicolumn{1}{| c |}{Vienna} & \multicolumn{1}{| c |}{Overall}\\
    \hline
    \hline
    \hline
    Vanilla Generator (VG) & 77.52 &  69.08 & 65.69 & 76.89 & 79.45 & 75.31\\
    \hline
    Base GAN Architecture (BGA) (VG + C)                       & 78.97 &  70.21 & 68.07 & 77.86 & 79.98 & 75.93\\
    \hline
    BGA + DS            & 80.31 &  71.77 & 68.86 & 79.67 & 80.18 & 77.89\\ 
    \hline
    BGA + UAM + DS       & 81.56 &  73.86 & 70.64 & 81.49 & 81.87 & 79.36\\
    \hline
    BGA + RM + DS       & 80.95 &  73.12 & 72.01 & 82.73 & 81.13 & 78.84 \\
    \hline
    BGA + UAM + RM + DS  & 83.78 & 76.39 & 73.25 & 85.72 & 83.19 & 81.28\\ 
    \hline
    \hline
    \end{tabular}}
\end{center}
\caption{Mean IoU scores for the ablation studies performed on the INRIA Validation Dataset. C: Critic, DS: Deep Supervision, UAM: Uncertainty Attention Module, RM: Refinement Module.}
\label{tab:ablation}
\end{table*}
\egroup
\bgroup

\subsection{Ablation Study}
\label{sec:ablation}

To verify the effectiveness of the Uncertainty Attention
Module, the Refinement Module, and of the deep supervision
technique we have used, we conducted ablation studies using
the INRIA Aerial Validation Dataset. We trained 6 different
architectures -- (a) the vanilla Generator (VG --- no
attention, deep supervision or critic) (b) the base GAN
architecture (BGA --- VG + critic); (c) the base GAN
architecture with deep supervision (DS); (d) the base GAN
architecture with deep supervision and the Uncertainty
Attention Module; (e) the base GAN architecture with deep supervision and the
Refinement Module; and, (f) the base GAN architecture with Deep Supervision,
the Uncertainty Attention Module and the Refinement
Module. All the architectures were trained independently
with identical training hyper-parameters. Test Time
Augmentation is applied while evaluating the performance of
the trained models on the validation images. As mentioned in
Section~\ref{sec:datacreate}, for the INRIA dataset, all the
experiments are conducted using our k-fold validation
strategy.

The mean IoU scores for these 6 models are reported in
Table~\ref{tab:ablation}. On adding the critic, the overall IoU 
of the Vanilla Generator improves by 0.82\%. With deep supervision, 
we achieve an overall improvement of 2.58\% relative to the BGA. 
The Uncertainty Attention Module and the Refinement
Module further improve the mean IoU scores by 1.89\% and
1.22\% respectively. Finally when we combine all these
components, our model outperforms the baseline GAN model by 7.04\%.

Figure~\ref{fig:ablation} demonstrates the qualitative
performance improvements obtained with the Uncertainty
Attention Module and the Refinement Module. In the first row and second column
of Figure~\ref{fig:ablation}, the large building is labeled
incorrectly due to the presence of shadow and absence of
global context in the base architecture. However, adding the
Uncertainty Attention Module improves the segmentation
result, as shown in row 1 and column 3 of
Figure~\ref{fig:ablation}. Similar results can be seen in
row 2, where the base network can not distinguish between
roads and buildings since they are similar in color. On the
contrary, the model with the Uncertainty Attention Module
accurately identifies the building
pixels. Column 4 of Figure~\ref{fig:ablation} demonstrates results when
we add the Refinement Module to the base GAN architecture. We
can observe that the Refinement Module has identified
precise building boundaries compared to the base model. When
we incorporate both the Uncertainty Attention and the
Refinement Modules, we can observe the overall improvement
compared to the base module in column 5 of Figure~\ref{fig:ablation}.

\section{Conclusion}
\label{sec:conclusion}

This paper has presented an attention-enhanced residual
refining GAN framework for detecting buildings in aerial and satellite 
images.  The proposed approach uses an Uncertainty
Attention Module to resolve uncertainties in classification
and a Refinement Module to refine the building
labels. Specifically, the Refinement Module, whose main job
is to refine intermediate prediction maps, uses an Edge
Attention Unit to improve the quality of building boundaries
and a Reverse Attention Unit to seek missed detections
in the intermediate prediction maps. The results demonstrate
the effectiveness of our building detection approach even
when the buildings are present amidst complex background or
are only partly visible due to the presence of shadows. The experimental evaluations that we have conducted in this paper also shows that the proposed method performs equally well on aerial as well as satellite images.  In
the future, we plan to investigate how to utilize multi-spectral information for further improvement of our network's capability. Extensive investigations on more diverse datasets (like, roads) have been left for the future.

%\section*{Acknowledgments}
{\small
\bibliographystyle{IEEEtran}
\bibliography{egbib}
}
\newpage
\begin{comment}
\section{Biography Section}
If you have an EPS/PDF photo (graphicx package needed), extra braces are
 needed around the contents of the optional argument to biography to prevent
 the LaTeX parser from getting confused when it sees the complicated
 $\backslash${\tt{includegraphics}} command within an optional argument. (You can create
 your own custom macro containing the $\backslash${\tt{includegraphics}} command to make things
 simpler here.)
 
\vspace{11pt}

\bf{If you include a photo:}\vspace{-33pt}
\begin{IEEEbiography}[{\includegraphics[width=1in,height=1.25in,clip,keepaspectratio]{fig1}}]{Michael Shell}
Use $\backslash${\tt{begin\{IEEEbiography\}}} and then for the 1st argument use $\backslash${\tt{includegraphics}} to declare and link the author photo.
Use the author name as the 3rd argument followed by the biography text.
\end{IEEEbiography}

\vspace{11pt}

\bf{If you will not include a photo:}\vspace{-33pt}
\begin{IEEEbiographynophoto}{John Doe}
Use $\backslash${\tt{begin\{IEEEbiographynophoto\}}} and the author name as the argument followed by the biography text.
\end{IEEEbiographynophoto}

\vfill
\end{comment}
\end{document}